\documentclass[twoside,11pt]{article}

\usepackage{jmlr2e}
\usepackage{bookmark}

\usepackage{tikz}
\usetikzlibrary{arrows}
\usepackage{subfigure}
\usepackage{amsmath}
\usepackage{bm}
\usepackage{paralist}
\usepackage{comment}
\usepackage{bbm}
\usepackage{lastpage}

\newtheorem{innercustomthm}{Theorem}
\newenvironment{customthm}[1]
  {\renewcommand\theinnercustomthm{#1}\innercustomthm}
  {\endinnercustomthm}

\newtheorem{innercustomcor}{Corollary}
\newenvironment{customcor}[1]
  {\renewcommand\theinnercustomcor{#1}\innercustomcor}
  {\endinnercustomcor}

\DeclareMathOperator*{\argmin}{arg\,min}

\definecolor{blue}{rgb}{0,0,1}
\definecolor{orange}{rgb}{1,0.5,0.1}
\definecolor{brightblue}{rgb}{0.92,0.92,1}
\definecolor{brightgrey}{rgb}{0.96,0.96,1}
\definecolor{green}{rgb}{0.2,1.0,0.2}
\definecolor{red}{rgb}{1,0,0}

\DeclareMathOperator*{\CI}{{\,\perp\mkern-12mu\perp\,}}
\DeclareMathOperator*{\nCI}{{\,\not\mkern-1mu\perp\mkern-12mu\perp\,}}
\DeclareMathOperator*{\SEP}{\perp}
\DeclareMathOperator*{\nSEP}{\not\perp}
\newcommand\indep[4]{{#1} \CI_{#4} {#2} \given {#3}}
\newcommand\dep[4]{{#1} \nCI_{#4} {#2} \given {#3}}
\newcommand\sep[4]{{#1} \SEP_{#4} {#2} \given {#3}}
\newcommand\con[4]{{#1} \nSEP_{#4} {#2} \given {#3}}
\newcommand{\dsep}[4]{{#1} \SEP_{#4}^d {#2} \given {#3}}

\newcommand{\sigmasep}[4]{{#1} \SEP_{#4}^\sigma {#2} \given {#3}}

\newcommand{\Prb}{\mathbb{P}}

\newcommand\idcausal{\dashrightarrow}

\newcommand\B[1]{\bm{#1}}
\newcommand\C[1]{\mathcal{#1}}
\newcommand\BC[1]{\bm{\mathcal{#1}}}
\newcommand\mathbfsc[1]{\text{\normalfont\scshape#1}}

\newcommand\an[1]{\mathbfsc{an}(#1)}

\newcommand\pa[1]{\mathbfsc{pa}(#1)}
\newcommand\ch[1]{\mathbfsc{ch}(#1)}
\newcommand\ncol[1]{\mathbfsc{ncol}(#1)}
\newcommand\col[1]{\mathbfsc{col}(#1)}

\newcommand\ansub[2]{\mathbfsc{an}_{#1}(#2)}
\newcommand\desub[2]{\mathbfsc{de}_{#1}(#2)}
\newcommand\sccsub[2]{\mathbfsc{sc}_{#1}(#2)}
\newcommand\pasub[2]{\mathbfsc{pa}_{#1}(#2)}
\newcommand\chsub[2]{\mathbfsc{ch}_{#1}(#2)}
\newcommand\given{\,|\,}
\newcommand\causes{\idcausal}

\newcommand{\xto}[1]{\stackrel{#1}{\to}}
\newcommand\eref[1]{(\ref{#1})}
\newtheorem{assumption}{Assumption}

\newcommand{\ot}{\leftarrow}
\newcommand{\oto}{\leftrightarrow}
\newcommand{\ots}{\leftarrow\mkern-13mu\ast\,}
\newcommand{\otc}{\leftarrow\mkern-9mu\circ\,}
\newcommand{\sto}{\,\ast\mkern-13mu\to}

\newcommand{\sts}{\,\ast\mkern-10mu\relbar\mkern-10mu\ast\,}
\newcommand{\ctc}{\,\circ\mkern-9mu\relbar\mkern-9mu\circ\,}
\newcommand{\cto}{\,\circ\mkern-9mu\rightarrow}
\newcommand{\ctt}{\,\circ\mkern-9mu\relbar\!\!\!\relbar}
\newcommand{\ttt}{\relbar\!\!\!\relbar}
\newcommand{\ttc}{\relbar\!\!\!\relbar\mkern-9mu\circ\,}
\newcommand{\RN}{\mathbb{R}}
\newcommand{\I}{\mathbbm{1}}

\newcommand{\intervene}{\mathrm{do}}

\newcommand{\DMAG}{\mathrm{DMAG}}

\newcommand{\CDPAG}{\mathrm{CDPAG}}
\newcommand{\IM}{\mathrm{IM}}

\newcommand{\alg}[1]{\texttt{#1}}
\newcommand{\boldcap}[1]{\textbf{#1}}

\DeclareSymbolFont{bbold}{U}{bbold}{m}{n}
\DeclareSymbolFontAlphabet{\mathbbold}{bbold}

\newcommand{\JCIABC}{\ref{ass:uncaused}, \ref{ass:unconfounded}, \ref{ass:dependences}}
\newcommand{\JCIAB}{\ref{ass:uncaused}, \ref{ass:unconfounded}}

\jmlrheading{21}{2020}{1-108}{3/17; Revised 1/20}{3/20}{17-123}{Joris M.~Mooij, Sara Magliacane and Tom Claassen}
\ShortHeadings{Joint Causal Inference from Multiple Contexts}{Mooij, Magliacane and Claassen}
\firstpageno{1}

\tikzstyle{var}=[circle,draw=black,fill=white,thick,minimum size=20pt,inner sep=0pt]
\tikzstyle{varh}=[circle,draw=gray,fill=white,thick,minimum size=20pt,inner sep=0pt,dashed]
\tikzstyle{arr}=[->,>=stealth',draw=black,thick]
\tikzstyle{carr}=[o->,>=stealth',draw=black,thick]
\tikzstyle{carc}=[o-o,>=stealth',draw=black,thick]
\tikzstyle{arrh}=[->,>=stealth',draw=gray,fill=gray,thick,dashed]
\tikzstyle{biarr}=[<->,>=stealth',draw=black,fill=black,thick]
\tikzstyle{biarrh}=[<->,>=stealth',draw=gray,fill=gray,thick,dashed]
\tikzstyle{noarr}=[draw=black,fill=black,thick]
\tikzstyle{noarrh}=[draw=gray,fill=gray,thick,dashed]
\tikzstyle{fac}=[rectangle,draw=black!50,fill=black!20,thick,minimum size=10pt] 
\tikzstyle{varc}=[rectangle,draw=black,fill=white,thick,minimum size=20pt,inner sep=0pt]
\tikzstyle{varch}=[rectangle,draw=black,fill=white,thick,minimum size=20pt,inner sep=0pt,dashed]

\graphicspath{{fig/},{fig/simul/},{fig/sachs/}}

\begin{document}
\title{Joint Causal Inference from Multiple Contexts}

\author{\name Joris M. Mooij\thanks{Part of this work was done while the authors were with the Informatics Institute of the University of Amsterdam.} \email \url{j.m.mooij@uva.nl}\\
  \addr Korteweg-De Vries Institute, University of Amsterdam\\
  \addr Postbox 94248, 1090 GE Amsterdam, The Netherlands
  \AND 
  \name Sara Magliacane \email \url{sara.magliacane@ibm.com}\\
  \addr MIT-IBM Watson AI Lab, IBM Research\\
  \addr 75 Binney St, Cambridge, MA 02142, USA
  \AND
  \name Tom Claassen \email \url{tomc@cs.ru.nl}\\
  \addr Institute for Computing and Information Sciences, Radboud University Nijmegen\\
  \addr Postbox 9010, 6500 GL Nijmegen, The Netherlands
}

\editor{Peter Spirtes}

\maketitle

\begin{abstract}%
The gold standard for discovering causal relations is by means of experimentation.
Over the last decades, alternative methods have been proposed that can
infer causal relations between variables from certain
statistical patterns in purely observational data. 
We introduce \emph{Joint Causal
Inference (JCI)}, a novel approach to causal discovery from
multiple data sets from different contexts that elegantly unifies both approaches. JCI is a causal modeling 
framework rather than a specific algorithm, and it can be implemented using any
causal discovery algorithm that can take into account certain background knowledge.
JCI can deal with different types of interventions (e.g., perfect, imperfect, stochastic, etc.) 
in a unified fashion, and does
not require knowledge of intervention targets or types in case of interventional data.
We explain how several well-known causal discovery algorithms can be seen as addressing special
cases of the JCI framework, and we also propose novel implementations that extend
existing causal discovery methods for purely observational data to the JCI setting.
We evaluate different JCI implementations on synthetic data 
and on flow cytometry protein expression data and conclude that JCI implementations can 
considerably outperform state-of-the-art causal discovery algorithms.
\end{abstract}

\begin{keywords}
causal discovery, causal modeling, causal inference, observational and experimental data, interventions, 
randomized controlled trials
\end{keywords}

\section{Introduction}\label{sec:introduction}

The aim of causal discovery is to learn the causal relations between variables
of a system of interest from data. As a simple example, suppose a researcher
wants to find out whether playing violent computer games causes aggressive
behavior. She gathers observational data by taking a sample from pupils at several high
schools in different countries and observes a significant correlation between the daily amount of hours
spent on playing violent computer games, and aggressive behavior at
school (see also Figure~\ref{fig:example}).
This in itself does not yet imply a causal relation between the two in either direction.
Indeed, an alternative explanation of the observed correlation could be the
presence of a confounder (a latent common cause), for example, a genetic
predisposition towards violence that makes the carrier particularly enjoy such
games and also make him behave more aggressively. 
The most reliable way to establish whether playing violent computer games
causes aggressive behavior, is by means of \emph{experimentation}, for example by a randomized controlled trial \citep{Fisher1935}.
This would imply assigning each pupil to one out of two groups randomly, 
where the pupils in one group are forced to play violent computer games for
several hours a day,
while the pupils in the other group are forced to abstain from playing those games.
After several months, the aggressive behavior in both groups is measured. 
If a significant correlation between group and outcome is observed (or 
equivalently, the outcome is significantly different between the two groups),
it can then be concluded that playing violent computer
games indeed causes aggressive behavior.

Given the ethical and practical problems that such an experiment would involve, 
one might wonder whether there are alternative ways to answer this question. One such
alternative is to combine data from different
contexts. For example, in some countries the government may have decided
to forbid certain ultra-violent games from being sold. In addition, some schools may have introduced
certain measures to discourage aggressive behavior. By combining the data
from these different contexts in an appropriate way, one may be able to identify
the presence or absence of a causal effect of playing violent computer games
on aggressive behavior. For example, in the setting of Figure~\ref{fig:example}(c),
the causal relationship between the two variables of interest turns out to 
be identifiable from conditional independence relationships in pooled data 
from all the contexts.
In particular, in that case the observed correlation between playing violent
computer games and aggressive behavior could be unambiguously attributed to a 
causal effect of one on the other, just from \emph{combining} multiple readily
available data sets, \emph{without} the need for an impractical experiment.\footnote{One can show that the conditional dependence $\dep{C_\alpha}{X_2}{C_\beta}{}$ and conditional independence $\indep{C_\alpha}{X_2}{\{X_1,C_\beta\}}{}$ in the pooled data that are entailed by the causal graph, together with the assumption that neither $C_\alpha$ nor $C_\beta$ is caused by $X_1$ or $X_2$, suffice to arrive at this conclusion.}
In this paper, we propose a simple and general way to combine and analyze data sets from 
different contexts that enables one to draw such strong causal conclusions.

\begin{figure}\centering%
\begin{tikzpicture}
  \begin{scope}
  \node at (-1,2) {(a)};
  \node[var] (X1) at (0,1) {$X_1$};  
  \node[var] (X2) at (2,1) {$X_2$};  
  \draw[arr] (X1) -- (X2);
  \end{scope}
  \begin{scope}[xshift=5cm]
  \node at (-1,2) {(b)};
  \node[var] (X1) at (0,1) {$X_1$};  
  \node[var] (X2) at (2,1) {$X_2$};  
  \draw[biarr] (X1) -- (X2);
  \end{scope}
  \begin{scope}[xshift=10cm]
  \node at (-1,2) {(c)};
  \node[var] (X1) at (0,0) {$X_1$};  
  \node[var] (X2) at (2,0) {$X_2$};  
  \draw[arr] (X1) -- (X2);
  \node[var] (C1) at (0,2) {$C_\alpha$};
  \node[var] (C2) at (2,2) {$C_\beta$};
  \draw (-0.5,1) edge[dotted] (2.5,1);
  \draw[biarr] (C1) -- (C2);
  \draw[arr] (C1) -- (X1);
  \draw[arr] (C2) -- (X2);
  \end{scope}
\end{tikzpicture}
  \caption{Different causal graphs relating $X_1$, the daily amount of hours spent on playing violent computer games, and $X_2$, a measure of aggressive behavior. (a) Playing violent computer games causes aggressive behavior; (b) The observed correlation between $X_1$ and $X_2$ is explained by a latent confounder, e.g., a genetic predisposition towards violence. (c) Hypothetical causal graph also involving context variables $C_\alpha$, which indicates whether ultra-violent games have been
  banned by the government, and $C_\beta$, which represents school interventions to stimulate social behavior. Without considering contexts, it is not possible to distinguish between (a) and (b) based on conditional independences in the data. In scenario (c), JCI allows one to infer from conditional independences in the pooled data that $X_1$ causes $X_2$ and that $X_1$ and $X_2$ are not confounded (assuming that context variables $C_\alpha$ and $C_\beta$ are not caused by system variables $X_1$ and $X_2$).
\label{fig:example}}
\end{figure}

While experimentation is still the gold standard to establish causal relationships,
researchers realized in the early nineties that there are other methods that require
only \emph{purely observational} data \citep{SGS2000,Pearl2009}. Many methods for causal
discovery from purely observational data have been proposed over the last decades,
relying on different assumptions.
These can be roughly divided into \emph{constraint-based} causal discovery methods, 
such as the PC \citep{SGS2000}, IC \citep{Pearl2009} and FCI algorithms \citep{SMR1999,Zhang2008_AI},
\emph{score-based} causal discovery methods \citep[e.g.,][]{CooperHerskovits1992,HGC1995,Chickering2002,KoivistoSood2004}, 
and methods exploiting other statistical patterns in the joint distribution \citep[e.g.,][]{Mooij++_JMLR_16,PJS2017}.
Originally, these methods were designed to estimate the causal graph of the system 
from a single data set corresponding to a single (purely observational) context.

More recently, various causal discovery methods have been proposed that extend these techniques 
to deal with multiple data sets from different contexts.
As an example, the data sets may correspond with a baseline 
of purely observational data consisting of measurements concerning the ``natural'' state of the 
system, and data consisting of measurements under different perturbations of the system due to 
external interventions on the system.\footnote{In
certain parts of the causal discovery literature, the word ``intervention'' has become 
synonymous to ``perfect intervention'' (i.e., an intervention that precisely sets a variable or set of variables
to a certain value without directly affecting any other variables in the system), but in this work we use it in the more
general meaning of any external perturbation of the system.} More generally, they can correspond to measurements
of the system in different environments.
These methods can be divided into two main approaches:
\begin{enumerate}[(a)]
\item methods that obtain statistics or constraints from each context separately and then construct a single context-independent causal graph by combining these statistics, but never directly compare data from different contexts \citep{Claassen++_NIPS2010,IOD2011,Hyttinen++2012,HEJ2014,triantafillou2015constraint,Rothenhausler++2015,ForreMooij_UAI_18};
\item methods that pool all data and construct a single context-independent causal graph directly from the pooled data
\citep{Cooper1997,CooperYoo1999,TianPearl2001,SPP05,EatonMurphy07,Trigger2007,GIES2012,MooijHeskes_UAI_13,ICP2016,oates2016estimating,Zhang++_IJCAI17}.
\end{enumerate}
In this paper, we propose \emph{Joint Causal Inference (JCI)}, a framework for causal modeling 
of a system in different contexts and for causal discovery from multiple data sets consisting of 
measurements obtained in different contexts, which takes the latter approach. 
As will be discussed in more detail in Section~\ref{sec:related_work}, 
JCI is the most generally applicable of those approaches---for example, it allows for the
presence of latent confounders and cyclic causal relationships---and also offers most flexibility in terms of
its implementation. 
While the ingredients of the JCI framework are not novel, the added value of the 
framework is that on the one hand it 
arrives at a unifying description of a diverse spectrum of existing approaches, while on the 
other hand it serves to inspire new implementations, such as the adaptations of FCI that we
propose in this work. Technically, this is achieved by formulating the problem in terms of
a (standard) Structural Causal Model that considers system and environment as subsystems of one
joint system, rather than other types of representations in which the system is 
modeled conditionally on its environment \citep{Dawid2002,BareinboimPearl2013,oates2016estimating,YangKatcoffUhler2018,ForreMooij_UAI_19}. This allows us to apply the standard notion of 
statistical independence in the same ways as is commonly done in the
purely observational setting. As we observed in our experiments (that are reported in Section~\ref{sec:experiments}), 
the novel algorithms proposed in this work compare favorably with the state-of-the-art in causal discovery
on synthetic data in many settings.

The key idea of JCI is to (i) consider auxiliary context variables that describe the context of each data set, (ii) pool all the data from different contexts, including the values of the context variables, into a single data set, and finally (iii) apply standard
causal discovery methods to the pooled data, incorporating appropriate background knowledge on the
causal relationships involving the context variables. The framework is simple and very generally 
applicable as it allows
one to deal with latent confounding and cycles (if the causal discovery method supports this)
and various types of interventions in a unified way.
It does not require background knowledge on the intervention types and targets, making it very
suitable to the application on complex systems in which the effects of certain interventions are
not known \emph{a priori}, a situation that often occurs in practice. On the other hand, if such 
background knowledge is available, it can be exploited.

JCI can be implemented using any causal discovery method that can incorporate the appropriate
background knowledge on the relationships between context and system variables. 
This allows one to benefit from the availability of sophisticated 
and powerful causal discovery methods that have been primarily designed for a single data set 
from a single context by extending their application domain to the setting of multiple data sets 
from multiple contexts. For example, we will show in this work how FCI \citep{SMR1999,Zhang2008_AI} can easily be adapted to the
JCI setting. At the same time, JCI accommodates various well-known causal discovery
methods as special cases, such as the standard randomized controlled trial setting \citep{Fisher1935}, Local Causal Discovery (LCD) \citep{Cooper1997}
and Invariant Causal Prediction (ICP) \citep{ICP2016}.
By explicitly introducing the context variables and treating them analogously to the system 
variables (but with additional background knowledge about their causal relations with the system variables), JCI makes
it possible to elegantly combine the principles of 
causal discovery from experimentation with those of causal discovery from purely observational data
to achieve a causal discovery framework that is more powerful than either of the two separately.

This paper is structured as follows. In Section~\ref{sec:background} we describe the relevant causal
modeling and discovery concepts and define terminology and notation. In Section~\ref{sec:JCI} we
introduce the JCI framework and modeling assumptions. In Section~\ref{sec:causal_discovery}, 
we show how JCI can be implemented using various causal discovery methods, 
and compare it with related work. In Section~\ref{sec:experiments} we
report experimental results on synthetic and flow cytometry data. 
We conclude in Section~\ref{sec:conclusion} with some promising directions for future developments.

\section{Background}\label{sec:background}

In this section, we present the background material on which we will base our
exposition. We start in Section~\ref{sec:graphical_causal_modeling} 
with a brief subsection stating the basic definitions and
results in the field of graphical causal modeling that 
we will use in this paper. In addition to covering material that is 
standard in the field, we review more recent extensions to the cyclic 
setting \citep{Bongers++_1611.06221v3}. Because the cyclic setting is quite
similar to the acyclic one that is mostly considered in the literature, we 
decided to present both cases in parallel rather
than first explaining the acyclic setting and then explaining how everything
generalizes to the cyclic setting.\footnote{The disadvantage is that our
notation and definitions deviate somewhat from those commonly used in the acyclic 
causal discovery literature. Therefore, we recommend reading this section also
to those readers that are already familiar with the theory of acyclic structural 
causal models.}
In Section~\ref{sec:CDexp}, we discuss the key
idea of causal discovery from experimentation (in the setting of a randomized
controlled trial, or A/B-testing) in these terms. We finish with Section~\ref{sec:CDobs}
that briefly illustrates the basic idea underlying constraint-based causal discovery 
from purely observational data in a simple setting.

\subsection{Graphical Causal Modeling}\label{sec:graphical_causal_modeling}

We briefly summarize some basic definitions and results in the field of graphical causal modeling.
For more details, we refer the reader to \citet{Pearl2009} and \citet{Bongers++_1611.06221v3}.

\subsubsection{Directed Mixed Graphs}
A \emph{Directed Mixed Graph} (DMG) is a graph $\C{G} = \langle \C{V},\C{E},\C{F} \rangle$ with 
nodes $\C{V}$ and two types of edges: \emph{directed} edges $\C{E} \subseteq \C{V}^2$, and 
\emph{bidirected} edges $\C{F} \subseteq \{\{i,j\} : i, j \in \C{V}, i \ne j\}$. 
We will denote a directed edge $(i,j) \in \C{E}$ as $i \rightarrow j$ or $j \ot i$, and
call $i$ a \emph{parent} of $j$ and $j$ a \emph{child} of $i$. We denote all parents of
$j$ in the graph $\C{G}$ as $\pasub{\C{G}}{j} := \{i \in \C{V} : i \to j \in \C{E}\}$, and all children of $i$ in $\C{G}$ as
$\chsub{\C{G}}{i} := \{j \in \C{V} : i \to j \in \C{E}\}$. We allow for self-cycles $i \to i$,
so a variable can be its own parent and child.
We will denote a bidirected edge $\{i,j\} \in \C{F}$ as $i \oto j$ or $j \oto i$,
and call $i$ and $j$ \emph{spouses}.
Two nodes $i,j \in \C{V}$ are called \emph{adjacent in $\C{G}$} if they are connected by an edge (or multiple edges), i.e., 
if $i\to j \in \C{E}$ or $i \ot j \in \C{E}$ or $i \oto j \in \C{F}$.
For a subset of nodes $\C{W} \subseteq \C{V}$, we define the \emph{induced subgraph} $\C{G}_{\C{W}}
:= (\C{W},\C{E}\cap \C{W}^2,\C{F}\cap \{\{i,j\} : i,j \in \C{W}, i \ne j\})$, i.e., 
with nodes $\C{W}$ and exactly those edges of $\C{G}$ that connect nodes in $\C{W}$.

A \emph{walk between $i,j\in\C{V}$} is a tuple $\langle i_0,e_1,i_1,e_2,i_3,\dots,e_n,i_n \rangle$ 
of alternating nodes and edges in $\C{G}$ ($n \ge 0$), such that all $i_0,\dots,i_n \in \C{V}$,
all $e_1,\dots,e_n \in \C{E} \cup \C{F}$, starting with node $i_0=i$ and ending with node $i_n=j$,
and such that for all $k=1,\dots,n$, the edge $e_k$ connects the two nodes $i_{k-1}$ and
$i_k$ in $\C{G}$. If the walk contains each node at most once, it is called a \emph{path}.
A \emph{trivial walk (path)} consists just of a single node and zero edges.
A \emph{directed walk (path) from $i \in \C{V}$ to $j \in \C{V}$} is a walk (path) between $i$ and $j$ such that every edge 
$e_k$ on the walk (path) is of the form $i_{k-1} \to i_k$, i.e., every edge is directed and points away from $i$. 
By repeatedly taking parents, we obtain the \emph{ancestors} of $j$:
$\ansub{\C{G}}{j} := \{i \in \C{V} : i = i_0 \to i_1 \to i_2 \to \dots \to i_n = j \text{ in } \C{G}\}$.
Similarly, we define the \emph{descendants} of $i$: $\desub{\C{G}}{i} := \{j \in \C{V}: i = i_0 \to i_1 \to i_2 \to \dots \to i_n = j \text{ in } \C{G}\}$. In particular, each node is ancestor and descendant of itself.
A \emph{directed cycle} is a directed path from $i$ to $j$ such that in addition, $j \to i \in \C{E}$.
An \emph{almost directed cycle} is a directed path from $i$ to $j$ such that in addition, $j \oto i \in \C{F}$.
All nodes on directed cycles passing through $i \in \C{V}$ together form the \emph{strongly-connected component}
$\sccsub{\C{G}}{i}:= \ansub{\C{G}}{i} \cap \desub{\C{G}}{i}$ of $i$.
We extend the definitions to sets $I \subseteq \C{V}$ by setting $\ansub{\C{G}}{I} := \cup_{i\in I} \ansub{\C{G}}{i}$, 
and similarly for $\desub{\C{G}}{I}$ and $\sccsub{\C{G}}{I}$.  A directed mixed graph $\C{G}$ 
is \emph{acyclic} if it does not contain any directed cycle, in which case it is known as an 
\emph{Acyclic Directed Mixed Graph (ADMG)}. A directed mixed graph that does not contain 
bidirected edges is known as a \emph{Directed Graph (DG)}. If a directed mixed graph does not
contain bidirected edges and is acyclic, it is called a \emph{Directed Acyclic Graph (DAG)}.

A node $i_k$ on a walk (path) $\pi = \langle i_0,e_1,i_1,e_2,i_3,\dots,e_n,i_n \rangle$ in $\C{G}$ is said to 
form a \emph{collider on $\pi$} if it is a non-endpoint node ($1 \le k < n$) and the two edges $e_k,e_{k+1}$ 
meet head-to-head on their shared node $i_k$ (i.e., if the two
subsequent edges are of the form $i_{k-1} \to i_k \ot i_{k+1}$, $i_{k-1} \oto i_k \ot i_{k+1}$, 
$i_{k-1} \to i_k \oto i_{k+1}$, or $i_{k-1} \oto i_k \oto i_{k+1}$). Otherwise (that is, if it is an endpoint node, i.e., $k=0$ or $k=n$, or if the two subsequent edges are of the form $i_{k-1} \to i_k \to i_{k+1}$, $i_{k-1} \ot i_k \ot i_{k+1}$,
$i_{k-1} \ot i_k \to i_{k+1}$, $i_{k-1} \oto i_k \to i_{k+1}$, or $i_{k-1} \ot i_k \oto i_{k+1}$), 
$i_k$ is called a \emph{non-collider on $\pi$}.
We will denote the colliders on a walk $\pi$ as $\col{\pi}$ and the non-colliders on $\pi$
(including the endpoints of $\pi$) as $\ncol{\pi}$.
A triple of nodes $\langle i,j,k \rangle$ in $\C{G}$ is called an \emph{unshielded triple} if
$i$ is adjacent to $j$, $j$ is adjacent to $k$ and $i$ is not adjacent to $k$ in $\C{G}$.

\subsubsection{Structural Causal Models}
Directed Mixed Graphs form a convenient graphical representation for
variables (labelled by the nodes) and their functional relations (expressed
by the edges) in a \emph{Structural Causal Model (SCM)} \citep{Pearl2009},
also known as a (non-parametric) \emph{Structural Equation Model (SEM)} \citep{Wright1921}.
Several slightly different definitions of SCMs have been proposed in the literature, which 
all have their (dis)advantages. Here we use a variant of the definition in 
\citet{Bongers++_1611.06221v3} that is most convenient for our purposes.
The reason we use SCMs to formulate JCI (rather than for example the 
more well-known causal Bayesian networks) is that SCMs are expressive enough 
to model both latent common causes and cyclic causal relationships.
\begin{definition}\label{def:SCM}
A Structural Causal Model (SCM) is a tuple $\C{M} = \langle \C{I}, \C{J}, \C{H}, \BC{X}, \BC{E}, \B{f}, \Prb_{\BC{E}} \rangle$ of:
\begin{compactenum}[(i)]
\item a finite index set $\C{I}$ for the endogenous variables in the model;
\item a finite index set $\C{J}$ for the latent exogenous variables in the model (disjoint from $\C{I}$);
\item a directed graph $\C{H}$ with nodes $\C{I} \cup \C{J}$, and directed edges pointing from
  $\C{I} \cup \C{J}$ to $\C{I}$;\label{def:SCMgraph}
\item a product of Borel\footnote{A \emph{Borel space} is both a measurable and a topological space, such that
  the sigma-algebra is generated by the open sets. Most spaces that one encounters in applications as the domain of a random variable are (isomorphic to) Borel spaces.} spaces $\BC{X} = \prod_{i \in \C{I}} \C{X}_i$, which define the
    domains of the endogenous variables; 
\item a product of Borel spaces $\BC{E} = \prod_{j \in \C{J}} \C{E}_j$, which define the
    domains of the exogenous variables;
\item a product probability measure $\Prb_{\BC{E}} = \prod_{j \in \C{J}} \Prb_{\C{E}_j}$ on $\BC{E}$ specifying
    the \emph{exogenous distribution};
\item a measurable function $\B{f} : \BC{X} \times \BC{E} \to \BC{X}$, the \emph{causal
    mechanism}, such that each of its components $f_i$ only depends on a particular subset of the variables, as
specified by the directed graph $\C{H}$:
      $$f_i : \BC{X}_{\pasub{\C{H}}{i} \cap \C{I}} \times \BC{E}_{\pasub{\C{H}}{i} \cap \C{J}} \to \C{X}_i, \qquad i \in \C{I}.$$
\end{compactenum}
\end{definition}
In discussing the concepts and properties of SCMs, the graphical representation of various objects and their relations
in Figure~\ref{fig:scms} may be helpful. This shows how the SCM is the basic object containing all information, and how other
representations can be derived from the SCM. In the rest of this section, we will discuss this in more detail.

\begin{figure}\centering
\begin{tikzpicture}
  \node[draw=black] (SCM) at (-0.5,0.5) {Simple SCM};
  \node[draw=black] (iSCM) at (4,1) {Intervened SCM};
  \node[draw=black] (iP) at (9,1) {Interventional Distribution};
  \node[draw=black] (mSCM) at (4,0) {Marginal SCM};
  \node[draw=black] (afG) at (2,-1) {Augmented Graph};
  \node[draw=black] (fG) at (4,-2.25) {Graph};
  \node[draw=black] (Seps) at (3.3,-3.5) {$d$/$\sigma$-separations};
  \node[draw=black] (ccR) at (8,-1) {Latent Confounders};
  \node[draw=black] (cR) at (8,-2.25) {Direct Causes};
  \node[draw=black] (icR) at (8,-3.5) {Causal Relations};
  \node[draw=black] (P) at (-1,-3.5) {Observational Distribution};
  \node[draw=black] (CI) at (1,-6) {(Conditional) Independences};
  \draw[->,thick] (afG) -- (fG);
  \draw[->,thick] (SCM) -- (afG);
  \draw[->,thick] (SCM) -- (iSCM);
  \draw[->,thick] (SCM) -- (mSCM);
  \draw[->,thick] (SCM) -- (P);
  \draw[->,thick] (iSCM) -- (iP);
  \draw[->,thick,bend right] (P) edge (CI);
  \draw[->,thick] (fG) edge (Seps);
  \draw[->,thick,bend left=10] (Seps) edge [anchor=west] node[text width=1.5cm,yshift=-3mm] {Markov Property} (CI);
  \draw[->,thick,bend left=10,dashed] (CI) edge [anchor=east,xshift=-1.2cm] node {Faithfulness} (Seps);
  \draw[->,thick] (fG) edge (cR);
  \draw[->,thick] (fG) edge (ccR);
  \draw[->,thick] (cR) edge (icR);
\end{tikzpicture}
  \caption{Relationships between various representations of simple SCMs. Directed edges represent mappings.
  Intervened and marginal SCMs are always defined and are also simple.\label{fig:scms}}
\end{figure}

We refer to the graph $\C{H}$ in Definition~\ref{def:SCM}(\ref{def:SCMgraph}) as the \emph{augmented graph} of $\C{M}$.
In contrast, the \emph{graph} of $\C{M}$, denoted $\C{G}(\C{M})$, is the directed mixed graph
with nodes $\C{I}$, directed edges $i_1 \to i_2$ iff $i_1 \to i_2 \in \C{H}$, and bidirected edges
$i_1 \oto i_2$ iff there exists $j \in \pasub{\C{H}}{i_1} \cap \pasub{\C{H}}{i_2} \cap \C{J}$.\footnote{This
definition of graph makes a slight simplification: 
a more precise definition would leave out edges that are redundant. 
For example, if the structural equation for $X_2$ reads $X_2 = 0 \cdot X_1 + X_3$ it could be that $1 \to 2 \in \C{H}$, 
but this edge would not appear in $\C{G}(\C{M})$.
For the rigorous version of this definition, see \citet{Bongers++_1611.06221v3}.}
While the augmented graph $\C{H}$ shows in detail the functional dependence of endogenous variables
on the (independent) exogenous variables, the graph $\C{G}(\C{M})$ provides an abstraction by not including
the exogenous variables explicitly, but using bidirected edges to represent any shared dependence of
pairs of endogenous variables on a common exogenous parent.
If $\C{G}(\C{M})$ is acyclic, we call the SCM $\C{M}$ \emph{acyclic}, otherwise we call the SCM \emph{cyclic}. 
If $\C{G}(\C{M})$ contains no bidirected edges, we call the endogenous variables in the SCM $\C{M}$ \emph{causally sufficient}.

A pair of random variables $(\B{X},\B{E})$ is called a \emph{solution} of the SCM $\C{M}$ if
$\B{X} = (X_i)_{i \in \C{I}}$ with $X_i \in \C{X}_i$ for all $i \in \C{I}$,
$\B{E} = (E_j)_{j \in \C{J}}$ with $E_j \in \C{E}_j$ for all $j \in \C{J}$,
the distribution $\Prb(\B{E})$ is equal to the exogenous distribution $\Prb_{\BC{E}}$, and
the \emph{structural equations}:
$$X_i = f_i(\B{X}_{\pasub{\C{H}}{i} \cap \C{I}}, \B{E}_{\pasub{\C{H}}{i} \cap \C{J}})\quad\text{a.s.}$$
hold for all $i \in \C{I}$.
An SCM is often specified informally by specifying only the structural equations and the
density\footnote{We denote a probability measure (or distribution) of a random variable $\B{X}$ by
$\Prb(\B{X})$, and a density of $\B{X}$ with respect to some fixed product measure by $p(\B{X})$.}  of the exogenous distribution with respect to some product measure, for example:
$$\C{M}: \begin{cases}
  X_i = f_i(\B{X}_{\pasub{\C{H}}{i} \cap \C{I}}, \B{E}_{\pasub{\C{H}}{i} \cap \C{J}}), & \quad i \in \C{I},\\
  p(\B{E}) = \prod_{j\in\C{J}} p(E_j). &
\end{cases}$$

For acyclic SCMs, solutions exist and have a unique distribution that is determined by the SCM.
This is not generally the case in cyclic SCMs, as these could have no solution at all, or 
could have multiple solutions with different distributions \citep{Bongers++_1611.06221v3}. 
\begin{definition}\label{def:unique_solvability_wrt}
An SCM $\C{M}$ is said to be \emph{uniquely solvable w.r.t.\ $\C{O} \subseteq \C{I}$} if there exists 
  a measurable mapping $\B{g}_{\C{O}} : \BC{X}_{(\pasub{\C{H}}{\C{O}}\setminus\C{O})\cap\C{I}} \times \BC{E}_{\pasub{\C{H}}{\C{O}} \cap \C{J}} \to \BC{X}_{\C{O}}$ 
such that for $\Prb_{\BC{E}}$-almost every $\B{e}$ for all $\B{x} \in \BC{X}$:
$$
    \B{x}_{\C{O}} = \B{g}_{\C{O}}(\B{x}_{(\pasub{\C{H}}{\C{O}}\setminus\C{O})\cap\C{I}}, \B{e}_{\pasub{\C{H}}{\C{O}}\cap\C{J}}) 
    \quad\iff\quad \B{x}_{\C{O}} = \B{f}_{\C{O}}(\B{x},\B{e}) \,.
$$
(Loosely speaking: the structural equations for $\C{O}$ have a unique solution for $\B{X}_{\C{O}}$ in terms of the other variables appearing in those equations.)
\end{definition}
If $\C{M}$ is uniquely solvable with respect to $\C{I}$ (in particular, this holds if $\C{M}$ is acyclic), then it induces a unique \emph{observational distribution 
$\Prb_{\C{M}}(\B{X})$}.

Given an SCM that models a certain system, we can model the system after an idealized intervention in which an external
influence enforces a subset of endogenous variables to take on certain values, while leaving the rest of the system untouched.
\begin{definition}
Let $\C{M}$ be an SCM. The \emph{perfect intervention} with target $I \subseteq \C{I}$
and value $\B{\xi}_I \in \BC{X}_I$ induces the \emph{intervened SCM} $\C{M}_{\intervene(I,\B{\xi}_I)}$ obtained by copying $\C{M}$,
but letting $\tilde{\C{H}}$ be $\C{H}$ without the edges $\{j \to i \in \C{H}: j \in \C{I} \cup \C{J}, i \in I\}$, and
modifying the causal mechanism into $\tilde{\B{f}}$ such that
\begin{equation*}
  \tilde{f}_i(\B{x},\B{e}) = \begin{cases}
    \xi_i & i \in I \\
    f_i(\B{x},\B{e}) & i \notin I.
  \end{cases}
\end{equation*}
\end{definition}
The interpretation is that the causal mechanisms that normally determine the values of the components $i \in I$ are 
replaced by mechanisms that assign the values $\xi_i$. Other types of interventions are possible as well 
(see also Section~\ref{sec:modeling_interventions}). If the intervened SCM $\C{M}_{\intervene(I,\B{\xi}_I)}$ induces a 
unique observational distribution, this is denoted as $\Prb_{\C{M}}\big(\B{X} \given \intervene(I,\B{\xi}_I)\big)$ and
referred to as the \emph{interventional distribution of $\C{M}$ under the perfect intervention $\intervene(I,\B{\xi}_I)$}.
\citet{Pearl2009} derived the \emph{do-calculus} for acyclic SCMs, consisting of three rules that express relationships between 
interventional distributions of an SCM. 

\subsubsection{Simple Structural Causal Models}\label{sec:simple_scm}

The theory of general cyclic Structural Causal Models is rather involved \citep{Bongers++_1611.06221v3}.
In this work, for simplicity of exposition, we will focus on a certain subclass of SCMs that has many convenient properties
and for which the theory simplifies considerably:
\begin{definition}\label{def:simple_scm}
An SCM $\C{M}$ is called \emph{simple} if it is uniquely solvable with respect to any subset $\C{O} \subseteq \C{I}$.
\end{definition}
All acyclic SCMs are simple. 
Simple SCMs provide a special case of the more general class of \emph{modular} SCMs \citep{ForreMooij_1710.08775}.
The class of simple SCMs can be thought of as a generalization of acyclic SCMs that allows for (weak) cyclic causal relations, but preserves many of the convenient properties that acyclic SCMs have.

Indeed, a simple SCM induces a unique observational distribution. 
Its marginalizations are always defined \citep{Bongers++_1611.06221v3},
and are also simple; in other words, the class of simple SCMs is closed under marginalizations. 
The class of simple SCMs is also closed under perfect interventions, and hence, 
all perfect interventional distributions of a simple SCM are uniquely defined. 
Without loss of generality, one can assume that simple SCMs have no self-cycles. 
The causal interpretation of the graph of an SCM with cycles and/or bidirected edges can be rather subtle in general.
However, for graphs of simple SCMs there is a straightforward causal interpretation:
\begin{definition}
  Let $\C{M}$ be a simple SCM. If $i \to j \in \C{G}(\C{M})$ we call $i$ a \emph{direct cause of $j$ according to $\C{M}$}.
  If there exists a directed path $i \to \dots \to j \in \C{G}(\C{M})$, i.e., if $i \in \ansub{\C{G}(\C{M})}{j}$, then we call
  $i$ a \emph{cause of $j$ according to $\C{M}$}. If there exists a bidirected edge $i \oto j \in \C{G}(\C{M})$, then we call
  $i$ and $j$ \emph{confounded according to $\C{M}$}.
\end{definition}

We conclude that the graph $\C{G}(\C{M})$ of a simple SCM can be interpreted as its \emph{causal graph}.
In the next subsection, we will discuss how the same graph $\C{G}(\C{M})$ of a simple SCM $\C{M}$ also represents
the conditional independences that must hold in the observational distribution of $\C{M}$.

\subsubsection{Structural Causal Models: Markov Properties}\label{sec:markov}
Under certain conditions, the graph $\C{G}(\C{M})$ of an SCM $\C{M}$ can be interpreted as a statistical
graphical model, i.e., it allows one to read off
conditional independences that must hold in the observational distribution $\Prb_{\C{M}}(\B{X})$.
One of the most common formulations of such \emph{Markov properties} involves the following notion of \emph{$d$-separation},
first proposed by \citet{Pearl1986} in the context of DAGs,
and later shown to be more generally applicable:\footnote{It is also sometimes called ``$m$-separation'' in the ADMG literature.}
\begin{definition}[$d$-separation]
We say that a walk $\langle i_0 \dots i_n \rangle$ in DMG $\C{G} = \langle \C{V},\C{E},\C{F} \rangle$ is \emph{$d$-blocked by $C \subseteq \C{V}$} if:
\begin{compactenum}[(i)]
  \item its first node $i_0 \in C$ or its last node $i_n \in C$, or
  \item it contains a collider $i_k \notin \ansub{\C{G}}{C}$, or
  \item it contains a non-collider $i_k \in C$.
\end{compactenum}
If all paths in $\C{G}$ between any node in set $A \subseteq \C{V}$ and any node in set $B \subseteq \C{V}$
are $d$-blocked by a set $C \subseteq \C{V}$, we say that $A$ is \emph{$d$-separated}
from $B$ by $C$, and we write $\dsep{A}{B}{C}{\C{G}}$.
\end{definition}
In the general cyclic case, however, the notion of $d$-separation is too strong, as was already pointed out by
\citet{Spirtes94}. A solution is to replace it with a non-trivial generalization of $d$-separation, 
known as $\sigma$-separation \citep{ForreMooij_1710.08775}:
\begin{definition}[$\sigma$-separation]
We say that a walk $\langle i_0 \dots i_n \rangle$ in DMG $\C{G} = \langle \C{V},\C{E},\C{F} \rangle$ is \emph{$\sigma$-blocked by $C \subseteq \C{V}$} if:
\begin{compactenum}[(i)]
\item its first node $i_0 \in C$ or its last node $i_n \in C$, or
\item it contains a collider $i_k \notin \ansub{\C{G}}{C}$, or
\item it contains a non-collider $i_k \in C$ that points to a 
neighboring node on the walk in another strongly-connected component (i.e.,
$i_{k-1} \to i_k \to i_{k+1}$ or $i_{k-1}\oto i_k \to i_{k+1}$ with $i_{k+1} \notin \sccsub{\C{G}}{i_k}$,
$i_{k-1} \ot i_k \ot i_{k+1}$ or $i_{k-1}\ot i_k \oto i_{k+1}$ with $i_{k-1} \notin \sccsub{\C{G}}{i_k}$,
or $i_{k-1} \ot i_k \to i_{k+1}$ with $i_{k-1} \notin \sccsub{\C{G}}{i_k}$ or $i_{k+1} \notin \sccsub{\C{G}}{i_k}$).
\end{compactenum}
If all paths in $\C{G}$ between any node in set $A \subseteq \C{V}$ and any node in set $B \subseteq \C{V}$
are $\sigma$-blocked by a set $C \subseteq \C{V}$, we say that $A$ is \emph{$\sigma$-separated}
from $B$ by $C$, and we write $\sigmasep{A}{B}{C}{\C{G}}$.
\end{definition}
\citet{ForreMooij_1710.08775} proved the following fundamental result for modular SCMs, which we formulate here only for
the special case of simple SCMs:
\begin{theorem}[Generalized Directed Global Markov Property]\label{thm:sigma_separation}
Any solution $(\B{X},\B{E})$ of a simple SCM $\C{M}$ obeys the \emph{Generalized Directed Global Markov Property} 
with respect to the graph $\C{G}(\C{M})$:
$$\sigmasep{A}{B}{C}{\C{G}(\C{M})} \implies \indep{\B{X}_A}{\B{X}_B}{\B{X}_C}{\Prb_{\C{M}}(\B{X})} \qquad \forall A,B,C \subseteq \C{I}.$$
\end{theorem}
The following stronger Markov properties, in which $\sigma$-separation is replaced by the more familiar notion of
$d$-separation, have been derived for special cases by \citet{ForreMooij_1710.08775} (where
again we consider only the special case of simple SCMs):
\begin{theorem}[Directed Global Markov Property]\label{thm:d_separation}
Let $\C{M} = \langle \C{I}, \C{J}, \C{H}, \BC{X}, \BC{E}, \B{f}, \Prb_{\BC{E}} \rangle$ be a simple SCM. If $\C{M}$ satisfies at least one of the following three conditions:
  \begin{compactenum}[(i)]
    \item $\C{M}$ is acyclic;
    \item all endogenous spaces $\C{X}_i$ are discrete;
    \item $\C{M}$ is linear (i.e., $\C{X}_i = \RN$ for each $i \in\C{I}$, $\C{E}_j = \RN$ for each $j\in\C{J}$, and each causal mechanism $f_i : \BC{X}_{\pasub{\C{H}}{i} \cap \C{I}} \times \C{E}_{\pasub{\C{H}}{i} \cap \C{J}} \to \C{X}_i$ is linear), each causal mechanism $f_i$ depends non-trivially on some exogenous variable(s), and its exogenous distribution has a density $p(\B{E})$ with respect to Lebesgue measure; 
  \end{compactenum}
then any solution $(\B{X},\B{E})$ of $\C{M}$ obeys the \emph{Directed Global Markov Property}
with respect to the graph $\C{G}(\C{M})$:
$$\dsep{A}{B}{C}{\C{G}(\C{M})} \implies \indep{\B{X}_A}{\B{X}_B}{\B{X}_C}{\Prb_{\C{M}}(\B{X})} \qquad \forall A,B,C \subseteq \C{I}.$$
\end{theorem}
\begin{quote}\end{quote}
Of these cases, the acyclic and linear cases are well-known.\footnote{The acyclic case 
was first shown in the context of 
linear-Gaussian structural equation models \citep{SRMSG98,Kos99}. The discrete case fixes the
erroneous theorem by \citet{PearlDechter96}, for which a counterexample
was found by \citet{Neal2000}, by adding the unique solvability condition,
and extends it to allow for latent common causes. The linear case
extends existing results for the linear-Gaussian setting without
latent common causes \citep{Spirtes94,Spi95,Koster96} to a linear (possibly non-Gaussian) setting with
latent common causes.}

We conclude that simple SCMs also have convenient Markov
properties. A simple SCM induces a unique observational distribution that satisfies the Generalized 
Directed Global Markov Property; under additional conditions, it satisfies even the Directed Global
Markov Property. Similarly, for any perfect intervention, a simple SCM induces a unique interventional
distribution that satisfies the (Generalized) Directed Global Markov Property with respect to the 
intervened graph. We conclude that the graph $\C{G}(\C{M})$ of a simple SCM has two interpretations:
it expresses both the causal structure between the variables as well as the conditional independence
structure of the solutions. These two interpretations of the graph $\C{G}(\C{M})$ of a simple SCM can be combined 
into a causal do-calculus \citep{ForreMooij_UAI_19} that extends the acyclic do-calculus of \citet{Pearl2009}
to the class of simple (or more generally, modular) SCMs.

The starting point for constraint-based approaches to causal discovery from observational data is 
to assume that the data is modelled by an (unknown) SCM $\C{M}$, such that its observational
distribution $\Prb_{\C{M}}(\B{X})$ exists and satisfies a Markov property with respect to its graph
$\C{G}(\C{M})$. In addition, one usually assumes the \emph{faithfulness assumption} to hold \citep{SGS2000,Pearl2009}, 
i.e., that the graph explains \emph{all} conditional independences present in the 
observational distribution. For the cases in which the $d$-separation criterion Theorem~\ref{thm:d_separation} applies, this amounts to assuming
the following implication:
$$\dsep{A}{B}{C}{\C{G}(\C{M})} \impliedby \indep{\B{X}_A}{\B{X}_B}{\B{X}_C}{\Prb_{\C{M}}(\B{X})} \qquad \forall A,B,C \subseteq \C{V}.$$
\citet{Meek1995} has shown completeness properties of $d$-separation. More specifically,  \citet{Meek1995} 
showed that faithfulness holds generically for DAGs if (i) all variable domains are finite, or
(ii) if all variables are real-valued, linearly related and have a multivariate Gaussian distribution. 
This in particular provides some justification for assuming faithfulness.
On the other hand, no completeness results are known yet for the general
cyclic case in which the $\sigma$-separation criterion Theorem~\ref{thm:sigma_separation} applies. Nevertheless, we believe that such results can be
shown, and we will assume for simple SCMs a similar faithfulness assumption as for the $d$-separation case:
$$\sigmasep{A}{B}{C}{\C{G}(\C{M})} \impliedby \indep{\B{X}_A}{\B{X}_B}{\B{X}_C}{\Prb_{\C{M}}(\B{X})} \qquad \forall A,B,C \subseteq \C{V}.$$

\subsection{Causal Discovery by Experimentation}\label{sec:CDexp}

The gold standard for causal discovery is by means of experimentation. For example, 
randomized controlled trials \citep{Fisher1935} form the foundation 
of modern evidence-based medicine. In engineering, A/B-testing is a
common protocol to optimize certain causal effects of an engineered
system. Toddlers learn causal representations of the world through playful
experimentation. 

We will discuss here the simplest randomized controlled trial setting by formulating
it in terms of the graphical causal terminology introduced in the last section. 
The experimental procedure is as follows.
Consider two variables, ``treatment'' $C$ and ``outcome'' $X$. In the simplest setting, one considers a 
binary treatment variable, where $C=1$ corresponds to ``treat with drug'' and
$C=0$ corresponds to ``treat with placebo''. For example, the drug could be aspirin,
and outcome could be the severity of headache perceived two
hours later. Patients are split into two groups,
the treatment and the control group, by means of a coin flip that assigns a value
of $C$ to every patient.\footnote{Usually this is done in a double-blind way, so
that neither the patient nor the doctor knows which group a patient has been assigned
to.} Patients are treated depending on the assigned value of $C$, i.e., patients in the
treatment group are treated with the drug and patients in the control group are treated
with a placebo. Some time after treatment, the outcome $X$ is measured for each patient. 
This yields a data set $(C_n,X_n)_{n=1}^N$
with two measurements ($C_n$, $X_n$) for the $n^{\mathrm{th}}$ patient. 
If the distribution of outcome $X$ significantly differs between the two groups,
one concludes that treatment is a cause of outcome. 

The important underlying causal assumptions that ensure the validity of the
conclusion are:
\begin{enumerate}[(i)]
  \item outcome $X$ is not a cause of treatment $C$ (which is commonly deemed 
  justified if the outcome is an event that occurs later in time than the treatment event);
  \item there is no latent confounder of treatment and outcome (this is where the 
randomization comes in: if treatment is decided solely by a proper coin flip, 
then it seems reasonable to assume that there cannot be any latent common cause 
of the coin flip $C$ and the outcome $X$ that is not just a combination of two
statistically independent separate causes of $C$ and $X$), 
\item no selection bias is present in the data (in other words, no data is
missing; for example, if only those patients that did not suffer from certain 
treatment side effects are included in the data set, then this assumption will be violated).
\end{enumerate}

\begin{figure}\centering\begin{tikzpicture}
    \node at (0,2) {(a) Two separate data sets:};
    \node[anchor=east] at (-3.5,0) {\scalebox{0.8}{\small\begin{tabular}{c}
      Placebo \\
      ($C=0$): \\
      \hline
      $X$ \\
      \hline
      \color{blue!70}-0.2 \\
      \color{blue!70} 0.6 \\
      \color{blue!70}-1.7 \\
      \color{blue!70}\dots \\
      \hline
    \end{tabular}}};
    \node at (-2,0) {\includegraphics[width=3cm]{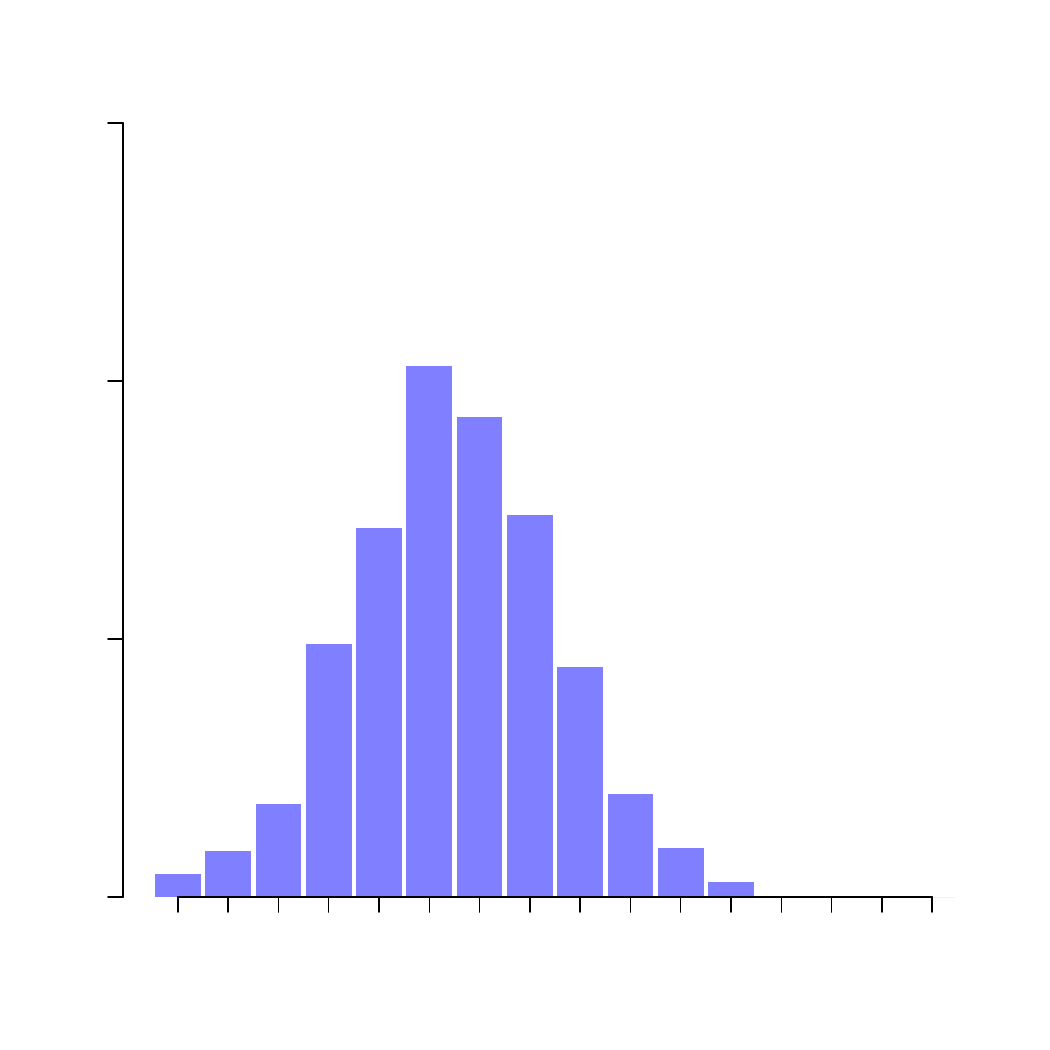}};

    \node[anchor=east] at (1.5,0) {\scalebox{0.8}{\small\begin{tabular}{c}
      Drug \\
      ($C=1$): \\
      \hline
      $X$ \\
      \hline
      \color{red!70} -0.3 \\
      \color{red!70} 1.8  \\ 
      \color{red!70} -0.1 \\ 
      \color{red!70} \dots  \\ 
      \hline
    \end{tabular}}};
    \node at (3,0) {\includegraphics[width=3cm]{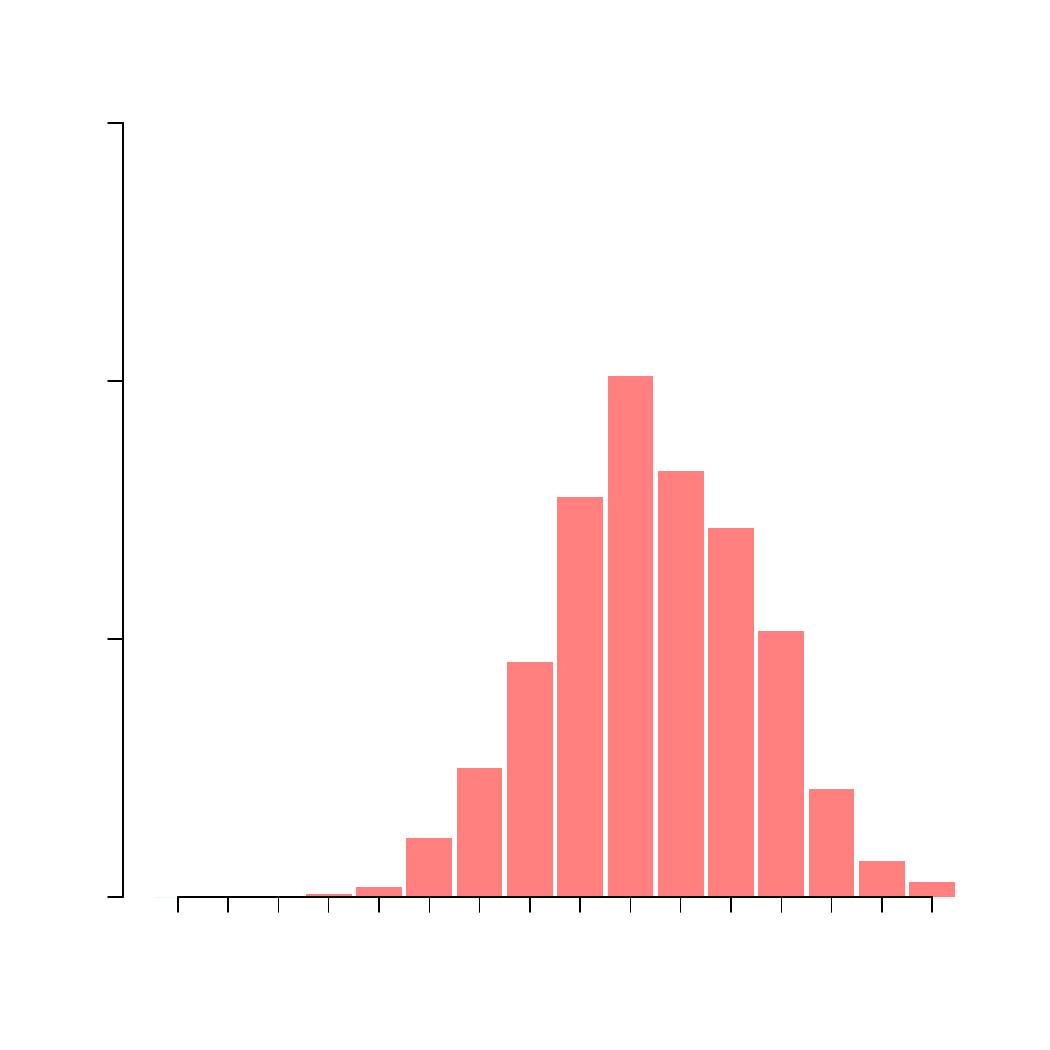}};

    \begin{scope}[xshift=6.9cm]
    \node at (0,2) {(b) Pooled data:};
    \node[anchor=east] at (0,0) {\scalebox{0.8}{\small\begin{tabular}{cc}
      $C$ & $X$ \\
      \hline
      0 & \color{blue!70}-0.2 \\
      0 & \color{blue!70} 0.6 \\
      0 & \color{blue!70}-1.7 \\
      0 & \color{blue!70}\dots \\
      1 & \color{red!70} -0.3 \\
      1 & \color{red!70} 1.8  \\ 
      1 & \color{red!70} -0.1  \\ 
      1 & \color{red!70} \dots \\ 
      \hline
    \end{tabular}}};
      \node at (1.5,0) {\includegraphics[width=3cm,trim={0cm 3mm 0cm 0cm},clip]{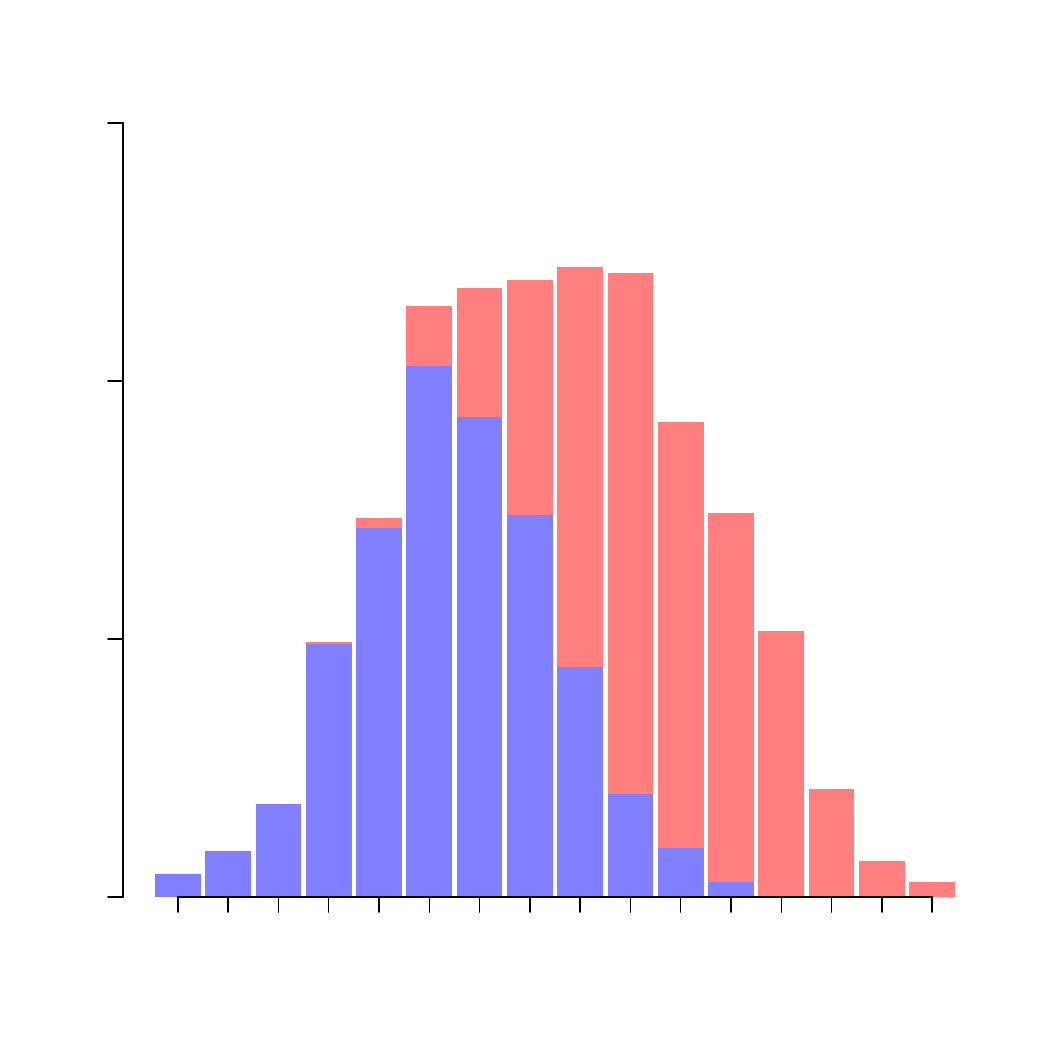}};
    \end{scope}
  \end{tikzpicture}
  \caption{Illustration of the data from an example randomized controlled trial. The data can either be interpreted
  as (a) two separate data sets, one for the treatment and one for the control group, or (b) 
  as a single data set including a context variable indicating treatment/control. Note that in this
  particular example, $C$ is dependent on $X$ in the pooled data (or equivalently, the distribution of $X$
  differs between contexts $C=0$ and $C=1$), which implies that $C$ is a cause of $X$.\label{fig:RCT_example}}
\end{figure}
Under these assumptions, one can show that
if the distribution of the outcome $X$ differs between the two groups of
patients (``treatment group'' with $C=1$ vs.\ ``control group'' with $C=0$), 
then treatment must be a cause of outcome, at least in this population of patients (see Proposition~\ref{prop:RCT}).
There are two conceptually slightly different ways of testing this in the data, depending on whether
we treat the data as a single pooled data set, or rather as two separate data sets (each one corresponding 
to a particular patient group), see also Figure~\ref{fig:RCT_example}.
If we consider the data about outcome $X$
in the two groups as two \emph{separate} data sets (corresponding to the same
variable $X$, but measured in different contexts $C$), then the question is whether the
distribution of $X$ is statistically different in the two data sets. This can 
be tested with a two-sample test, for example, a $t$-test or a Wilcoxon test. 
The other alternative is to consider the data as a single \emph{pooled} data set 
(by pooling the data for the two groups), and let the value of $C$ indicate the
context of each sample (treatment or control). The question now becomes whether the
conditional distribution of $X$ given $C=0$ differs from the conditional distribution
of $X$ given $C=1$, i.e., whether $\Prb(X \given C=0) \ne \Prb(X \given C=1)$. 
In other words, we have to test whether
there is a statistically significant \emph{dependence} $C \nCI X$ in the pooled data 
between treatment $C$ and outcome $X$; if there is, it must be due
to the treatment $C$ causing the outcome $X$, as the following proposition shows:
\begin{proposition}\label{prop:RCT}
Suppose that the data-generating process on context variable $C$ and outcome variable $X$ 
can be modeled by a simple SCM $\C{M}$ and no selection bias is present.\footnote{The 
context variable $C$ is here considered as an \emph{endogenous} variable in the SCM, as explained
in Section~\ref{sec:JCI_system_context}.}
Under the randomized controlled trial assumptions:
  \begin{compactenum}[(i)]
  \item $C \ot X \notin \C{G}(\C{M})$ (``outcome $X$ is not a cause of treatment $C$'')
  \item $C \oto X \notin \C{G}(\C{M})$ (``there is no latent confounder of treatment $C$ and outcome $X$''),
  \end{compactenum}
a dependence $C \nCI X$ in the joint distribution $\Prb(C,X)$ implies that $C$ causes $X$. Furthermore,
the causal effect of $C$ on $X$ is given by:
  \begin{equation}\label{eq:RCT_causal_effect}
    \Prb_{\C{M}}\big(X \given \intervene(C=c)\big) = \Prb_{\C{M}}(X \given C=c).
  \end{equation}
\end{proposition}
\begin{proof}
Out of the eight possible graphs $\C{G}(\C{M})$, only two satisfy the assumptions:\\
\begin{center}\begin{tikzpicture}
  \begin{scope}
    \node at (0,-0.8) {$C \CI X$};
    \node[var] (C1) at (-1,0) {$C$};
    \node[var] (X1) at (1,0) {$X$};
  \end{scope}
  \begin{scope}[xshift=4cm]
    \node[var] (C1) at (-1,0) {$C$};
    \node[var] (X1) at (1,0) {$X$};
    \draw[arr] (C1) edge (X1);
  \end{scope}
  \begin{scope}[xshift=8cm]
    \node[var] (C1) at (-1,0) {$C$};
    \node[var] (X1) at (1,0) {$X$};
    \draw[arr,bend left] (X1) edge (C1);
    \draw[red, line width=2mm, fill opacity=.30, draw opacity=.30] (-1,-0.5) edge (1,0.5);
    \draw[red, line width=2mm, fill opacity=.30, draw opacity=.30] (-1,0.5) edge (1,-0.5);
  \end{scope}
  \begin{scope}[xshift=12cm]
    \node[var] (C1) at (-1,0) {$C$};
    \node[var] (X1) at (1,0) {$X$};
    \draw[arr] (C1) edge (X1);
    \draw[arr,bend left] (X1) edge (C1);
    \draw[red, line width=2mm, fill opacity=.30, draw opacity=.30] (-1,-0.5) edge (1,0.5);
    \draw[red, line width=2mm, fill opacity=.30, draw opacity=.30] (-1,0.5) edge (1,-0.5);
  \end{scope}
  \begin{scope}[yshift=-2.5cm]
    \node[var] (C1) at (-1,0) {$C$};
    \node[var] (X1) at (1,0) {$X$};
    \draw[biarr,bend left] (C1) edge (X1);
    \draw[red, line width=2mm, fill opacity=.30, draw opacity=.30] (-1,-0.5) edge (1,0.5);
    \draw[red, line width=2mm, fill opacity=.30, draw opacity=.30] (-1,0.5) edge (1,-0.5);
  \end{scope}
  \begin{scope}[yshift=-2.5cm,xshift=4cm]
    \node[var] (C1) at (-1,0) {$C$};
    \node[var] (X1) at (1,0) {$X$};
    \draw[arr] (C1) edge (X1);
    \draw[biarr,bend left] (C1) edge (X1);
    \draw[red, line width=2mm, fill opacity=.30, draw opacity=.30] (-1,-0.5) edge (1,0.5);
    \draw[red, line width=2mm, fill opacity=.30, draw opacity=.30] (-1,0.5) edge (1,-0.5);
  \end{scope}
  \begin{scope}[yshift=-2.5cm,xshift=8cm]
    \node[var] (C1) at (-1,0) {$C$};
    \node[var] (X1) at (1,0) {$X$};
    \draw[arr,bend left] (X1) edge (C1);
    \draw[biarr,bend left] (C1) edge (X1);
    \draw[red, line width=2mm, fill opacity=.30, draw opacity=.30] (-1,-0.5) edge (1,0.5);
    \draw[red, line width=2mm, fill opacity=.30, draw opacity=.30] (-1,0.5) edge (1,-0.5);
  \end{scope}
  \begin{scope}[yshift=-2.5cm,xshift=12cm]
    \node[var] (C1) at (-1,0) {$C$};
    \node[var] (X1) at (1,0) {$X$};
    \draw[arr] (C1) edge (X1);
    \draw[arr,bend left] (X1) edge (C1);
    \draw[biarr,bend left] (C1) edge (X1);
    \draw[red, line width=2mm, fill opacity=.30, draw opacity=.30] (-1,-0.5) edge (1,0.5);
    \draw[red, line width=2mm, fill opacity=.30, draw opacity=.30] (-1,0.5) edge (1,-0.5);
  \end{scope}
  \end{tikzpicture}\end{center}
By the Markov property (Theorem~\ref{thm:sigma_separation}), if the edge $C \to X$ were absent in $\C{G}(\C{M})$, then $C$ would be independent of $X$. 
Therefore, if $C \nCI X$, the edge $C \to X$ must be in $\C{G}(\C{M})$. 
In both cases, the causal do-calculus applied to $\C{G}(\C{M})$
yields the identity \eref{eq:RCT_causal_effect}. 
\end{proof}

Of course, in this straightforward example the equivalence between the two approaches (differences between
two separate data sets vs.\ properties of a single pooled data set) is trivial, and the reader may wonder
why we emphasize it. The reason is that the key idea of our approach is precisely this: \emph{reducing
an apparently complicated causal discovery problem with multiple data sets to a more 
standard causal discovery problem involving a single pooled data set.}
The Joint Causal Inference framework that we propose in this paper can be considered as an 
extension of this randomized controlled trial setting to multiple treatment and outcome variables.

It is important to realize that the simple causal reasoning for the RCT \emph{cannot} be made when looking at the two data sets in isolation (i.e., by considering only properties of $\Prb(X \given C=0)$ and $\Prb(X \given C=1)$ separately, 
and not using in addition any other properties of the joint distribution $\Prb(X,C)$). The latter approach
is commonly used by constraint-based methods for causal discovery from multiple data sets 
\citep[e.g.,][]{ION2009,Claassen++_NIPS2010,IOD2011,HEJ2014,triantafillou2015constraint,Rothenhausler++2015,ForreMooij_UAI_18}.
Under the assumptions made, the crucial (and possibly very strong) signal in the data that allows one to 
draw the conclusion that $C$ causes $X$ is the
dependence $C \nCI X$ that \emph{can only be seen} in the pooled data. 
Methods that only test for conditional independences \emph{within} each context
and subsequently combine these into a single context-independent causal model will not yield any conclusion 
in this setting. The approach taken by JCI, on the other hand, is to analyze the pooled data jointly,
so that informative signals like these can be taken into account.

\subsection{Causal Discovery from Purely Observational Data}\label{sec:CDobs}

In the previous section, we discussed the current gold standard for discovering causal relations.
Over the last two decades, alternative methods have been proposed to perform causal discovery
from \emph{purely observational} data. This is intriguing and of high relevance, 
since experiments may be impossible, infeasible, impractical, unethical or too expensive to perform. 
These causal discovery methods can be divided into \emph{constraint-based} causal discovery methods, 
such as the PC \citep{SGS2000}, IC \citep{Pearl2009} and FCI algorithms 
\citep{SMR1999,Zhang2008_AI}, 
and \emph{score-based} causal discovery methods \citep[e.g.,][]{HGC1995,Chickering2002,KoivistoSood2004}.
The PC and IC algorithms and most score-based methods assume causal sufficiency (i.e., the absence of latent confounders), while
the FCI algorithm and other modern constraint-based algorithms allow for latent confounders and selection bias.
Originally, these methods have been designed to estimate the causal graph of the system 
from a single data set corresponding to a single (purely observational) context.

All these methods try to infer causal relationships on the basis of subtle statistical patterns
in the data. The most important of these patterns are conditional independences between variables.
These are exploited by most constraint-based methods, and implicitly, by score-based methods.
Other patterns, such as ``Verma constraints'' \citep{Shpitser++2014}, algebraic constraints
in the linear-Gaussian case \citep{VanOmmenMooij_UAI_17}, non-Gaussianity in 
linear models \citep{KanoShimizu2003}, and non-additivity of noise in nonlinear models
\citep{PetersMooijJanzingSchoelkopf_JMLR_14} can also be exploited. Another class of methods that has 
become popular more recently are methods that try to infer the causal direction ($A \to B$ 
vs.\ $B \to A$) from purely observational data of variable pairs \citep[see e.g.,][]{Mooij++_JMLR_16}.

Since our main goal is to enable constraint-based causal discovery from multiple contexts, we
will focus on this approach here, while noting that the JCI framework that we propose in the next
section is compatible with all approaches to causal discovery from purely observational data
that allow for multiple variables and can handle certain background knowledge (to be made
precise in Section~\ref{sec:JCI_assumptions}).

\begin{figure}
  \begin{tikzpicture}
    \begin{scope}
    \node[var] (X1) at (-1,0) {$X_1$};
    \node[var] (X2) at (1,0) {$X_2$};
    \node[var] (X3) at (0,-1) {$X_3$};
    \node[var] (X4) at (0,-2.5) {$X_4$};
    \draw[arr,dashed] (X1) edge (X3);
    \draw[biarr,bend right,dashed] (X1) edge (X3);
    \draw[arr,dashed] (X2) edge (X3);
    \draw[biarr,bend left,dashed] (X2) edge (X3);
    \draw[arr] (X3) edge (X4);
    \end{scope}
  \end{tikzpicture}\hfill
  \includegraphics[width=0.26\textwidth]{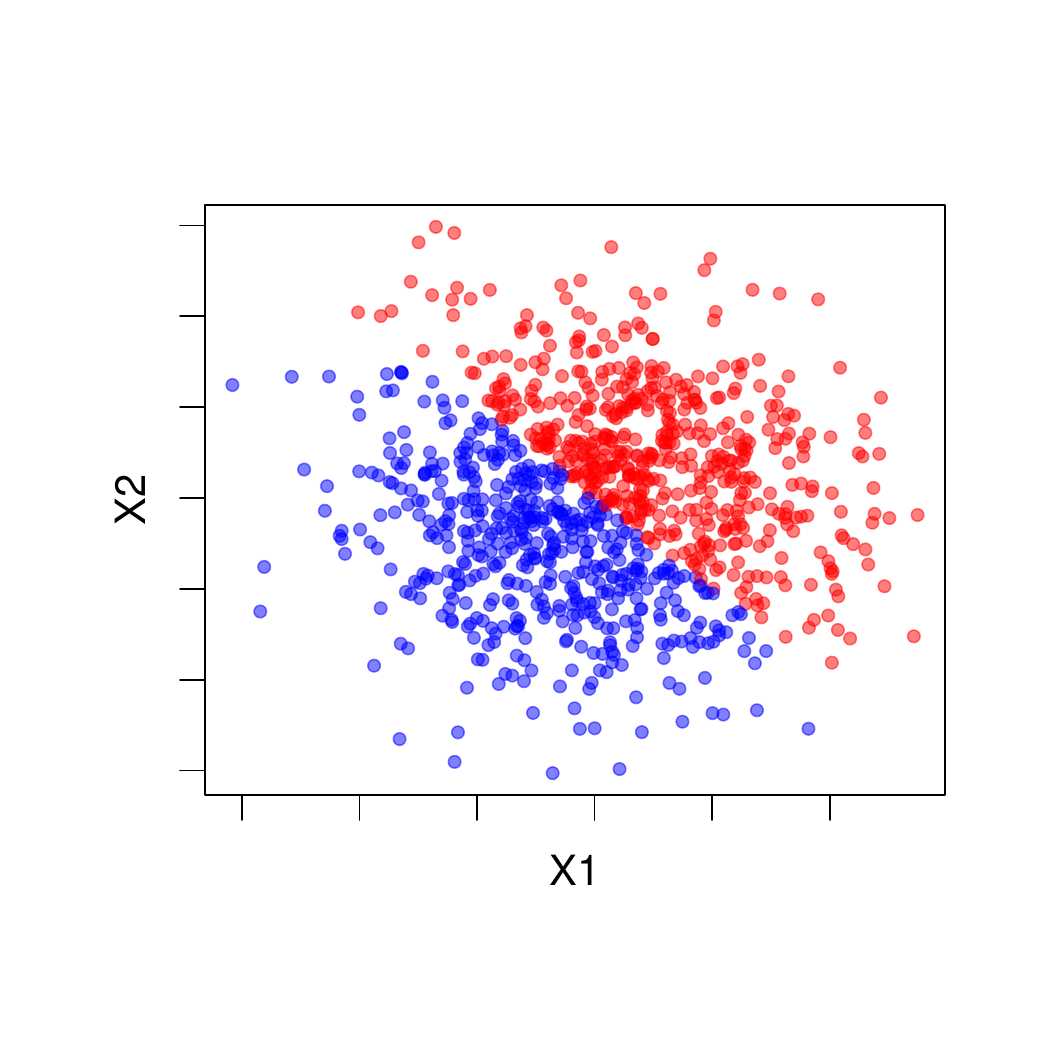}
  \includegraphics[width=0.26\textwidth]{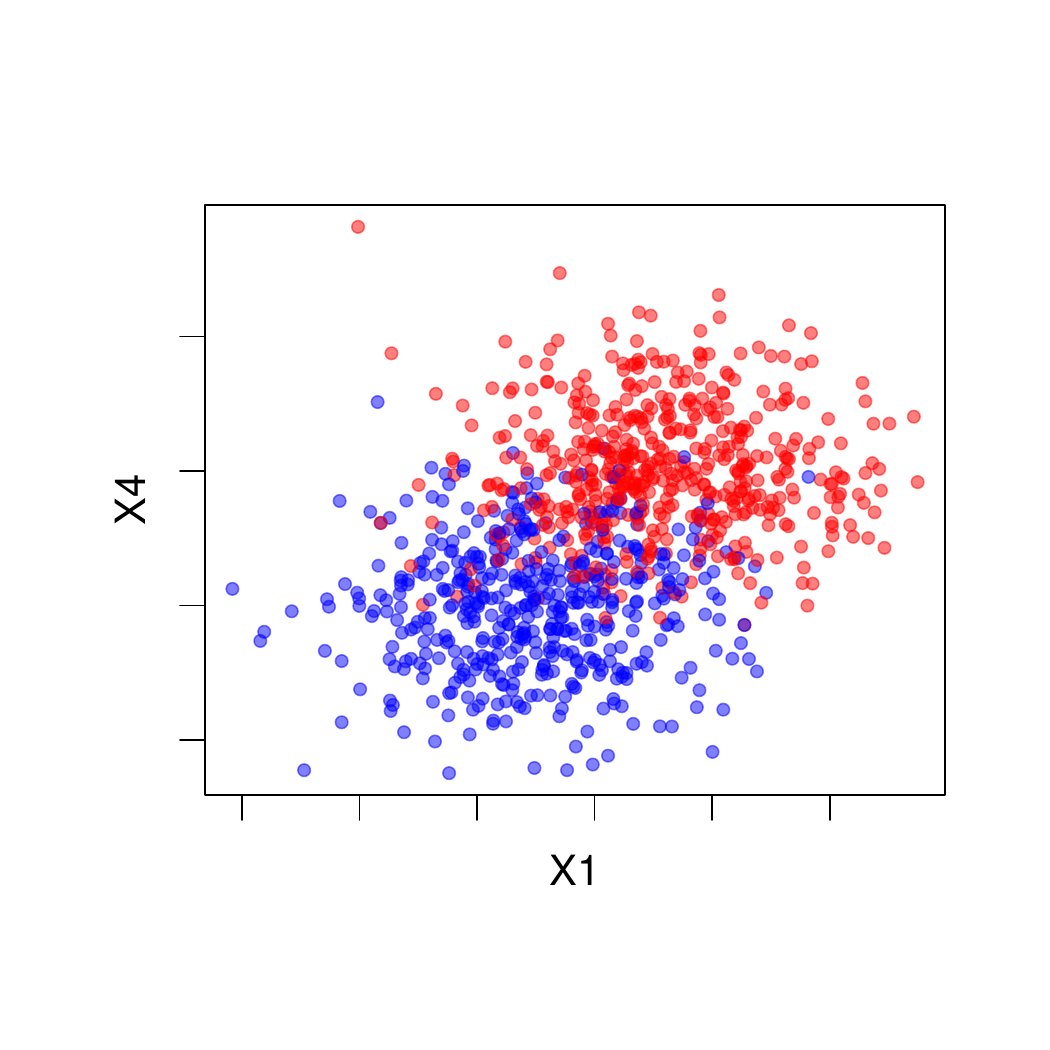}
  \includegraphics[width=0.26\textwidth]{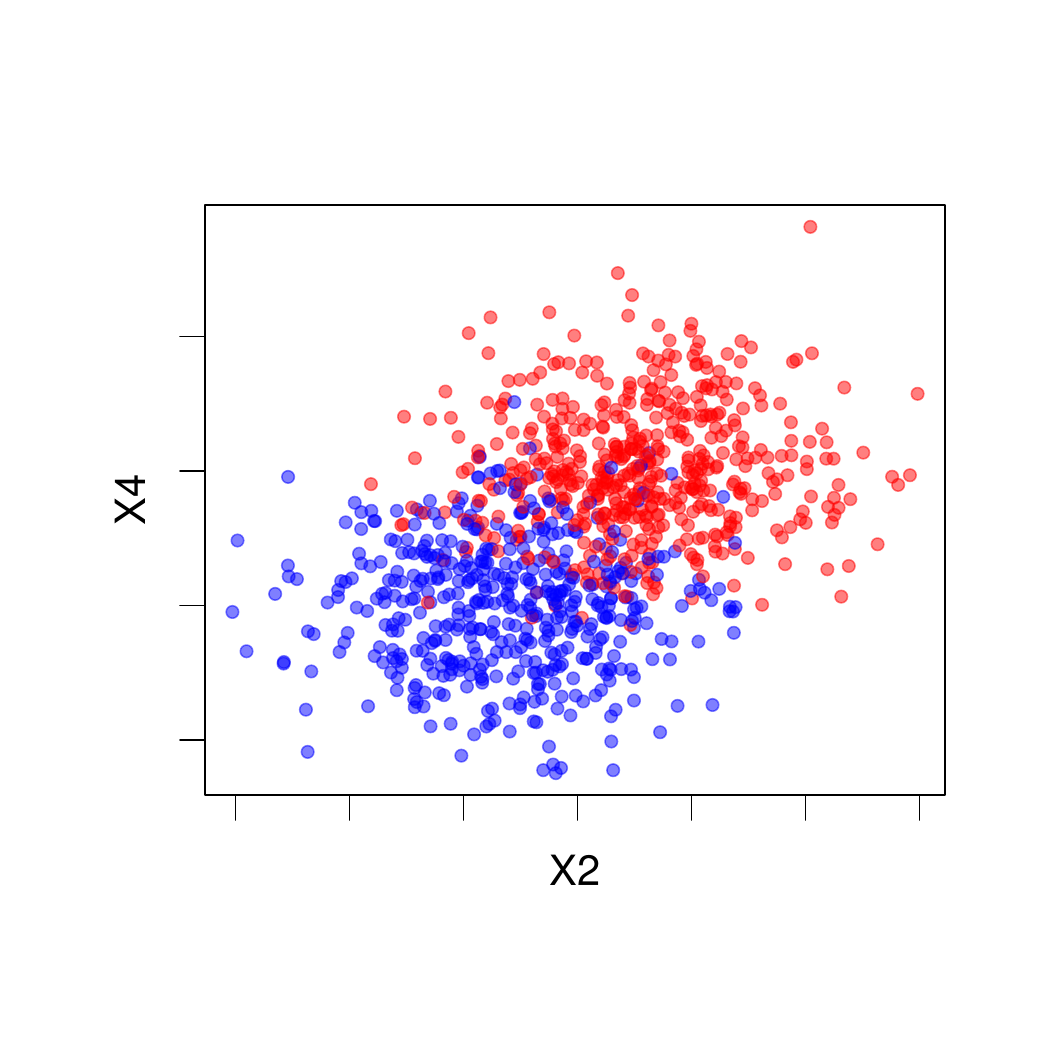}
  \caption{Left: Causal graphs satisfying the ``Y-structure'' pattern on four variables ($X_1$,$X_2$,$X_3$,$X_4$). Right: Scatter plots illustrating the Y-structure pattern in purely observational data, where $X_3$ is discrete-valued and its value is indicated by color (red/blue).\label{fig:Ystruct_data}}
\end{figure}

As discussed in detail by \citet{SGS2000},
causal discovery from conditional independence patterns in purely observational data becomes possible under strong assumptions. 
The simplest example of how certain patterns of conditional independences in the observational distribution can lead to conclusions
about the causal relations of the variables is given by the ``Y-structure'' pattern \citep{ManiPhD2006}, which is illustrated in 
Figure~\ref{fig:Ystruct_data}. We show here that the Y-structure pattern also generalizes to the cyclic case.
\begin{proposition}
  Suppose that the data-generating process on four variables $X_1,X_2,X_3,X_4$ can be modeled by a simple SCM $\C{M}$. Assume that the sampling procedure is not subject to selection bias, and that faithfulness holds. 
  If the following conditional (in)dependencies hold in the observational distribution $\Prb_{\C{M}}(\B{X})$:
  $$\begin{array}{lll}
    X_1 \nCI X_4,  & X_2 \nCI X_4, & X_1 \CI X_2, \\
    X_1 \CI X_4 \given X_3, & X_2 \CI X_4 \given X_3, & X_1 \nCI X_2 \given X_3,
  \end{array}$$
  then $X_3$ is a direct cause of $X_4$ according to $\C{M}$. Furthermore, 
  $X_3$ and $X_4$ are unconfounded according to $\C{M}$ and
  the causal effect of $X_3$ on $X_4$ is given by:
  \begin{equation}\label{eq:Y_causal_effect}
    \Prb_{\C{M}}\big(X_4 \given \intervene(X_3=x_3)\big) = \Prb_{\C{M}}(X_4 \given X_3=x_3).
  \end{equation}
\end{proposition}
\begin{proof}
By the assumed Markov and faithfulness properties, one
can check that the only (cyclic or acyclic) graphs that are compatible with the observed
conditional independences are the ones in Figure~\ref{fig:Ystruct_data} (left),
where $X_1$ must be adjacent to $X_3$ via at least one of the two dashed edges, and similarly,
$X_2$ must be adjacent to $X_3$ via at least one of the two dashed edges. Hence, $X_3$ is a 
direct cause of $X_4$ according to $\C{M}$, but $X_4$ is not a direct
cause of $X_3$ according to $\C{M}$. Also, $X_3$ and $X_4$ cannot be confounded
according to $\C{M}$.
By applying the causal do-calculus, we arrive at \eref{eq:Y_causal_effect}.
\end{proof}
This example illustrates how conditional independence patterns in the observational
distribution allow one to infer certain features of the underlying causal model. This
principle is exploited more generally by constraint-based methods, and implicitly,
by score-based methods that optimize a penalized likelihood over (equivalence classes of) causal graphs.

Typically, the graph cannot be completely identified from purely observational data.
For example, in the Y-structure case, the conditional independences in the observational data
do not allow to conclude whether the dependence between $X_1$ and $X_3$ is explained by
$X_1$ being a cause of $X_3$, or by $X_1$ and $X_3$ having a latent confounder,
or both.
However, under the assumption of faithfulness, one can deduce the Markov
equivalence class of the graph from the conditional independences in the observational data, 
i.e., the class of all DMGs that induce the same separations.
Another disadvantage of causal discovery methods from purely observational data is that
they typically need very large sample sizes and strong assumptions in order to work reliably.
These are some of the motivations to combine these ideas with those of causal discovery by
experimentation, as we will do in the next section.

\section{Joint Causal Inference}\label{sec:JCI}

In this section we present Joint Causal Inference (JCI), a novel framework for
causal discovery from multiple data sets corresponding to measurements that
have been performed in different contexts. JCI combines the existing approaches
towards causal discovery that we discussed in Sections~\ref{sec:CDexp} and \ref{sec:CDobs}.

\subsection{The Distinction between System and Context}\label{sec:JCI_system_context}

Henceforth, we will distinguish \emph{system variables} $(X_i)_{i \in \C{I}}$ describing the system of interest, and \emph{context variables} $(C_k)_{k\in\C{K}}$  describing the context in which the system has been observed.
An observation that will turn out to be crucial in what follows is that the decision of what to consider part of the ``system'' 
and what to consider part of its ``context'' does not reflect an objective property of nature, but is a choice of the modeler.

While the system variables are treated as \emph{endogenous} variables of the system of interest,
we usually (but not necessarily) think of the context variables as observed \emph{exogenous} 
variables for the system of interest. In particular, context variables could describe which interventions have been performed on the system (or more specifically, how these interventions have been performed), in which case we will also refer to them as \emph{intervention variables}. The possible interventions are not limited to the perfect interventions modeled by the do-operator of \citet{Pearl2009}, but can also be more general types of interventions
that appear in practice, like mechanism changes \citep{TianPearl2001}, soft interventions \citep{Markowetz++2005}, fat-hand interventions \citep{EatonMurphy07}, activity interventions \citep{MooijHeskes_UAI_13}, and stochastic versions of all these.
This will be discussed in more detail in Section~\ref{sec:modeling_interventions}. 
Even more generally,
a context variable could describe \emph{any} property of the environment of the system, including those
properties that one would not normally think about as an intervention. Examples are the lab in 
which measurements have been done, the time of the day, the patient population, variables like
``gender'' or ``age'', etc.\ Like system variables, context variables can be discrete or continuous (or
more generally, take values in some Borel space).

The idea of explicitly considering context variables is not novel: they have been discussed in the literature
under various names, such as ``policy variables'' \citep{SGS2000},
``force variables'' \citep{Pearl1993b}, 
``decision variables'' in influence diagrams \citep{Dawid2002}, 
``regime indicators'' \citep{Didelez++2006},
``selection variables'' in selection diagrams \citep{BareinboimPearl2013}, 
and ``environment variable'' \citep{ICP2016}.
Their use for causal discovery was already suggested by \citet{CooperYoo1999}.
Formal aspects in how these variables are treated vary across accounts, however.
For example, \citet{Dawid2002} treats system variables as random variables and chooses
to not treat context (``decision'') variables as random variables.
In this work we simply consider context variables as random variables with added background
knowledge on their causal relations, which expresses their assumed exogeneity with respect to the system.

Conceptually, context variables provide a more general notion than intervention variables,
since every intervention can be seen as a \emph{change of} context, but not every change of 
context is naturally thought of as an intervention. For example, the causal effect of some
drug on a certain health outcome may differ for males and females. Taking ``gender'' as 
a context variable that just encodes the specific subpopulation of patients we are considering
is more natural than considering it to be an intervention variable that encodes the
result of a gender-changing operation on the patient. Furthermore, interventions
usually come with an ``observational baseline'' of ``doing nothing'', but this is not always naturally available
for more general context variables (e.g., ``male'' and ``female'' could both qualify as a baseline,
while neither of the two would provide a more natural ``observational'' baseline than the other). 
When considering context variables, we do not
have to specify such a baseline, whereas if we consider them as intervention variables, one can 
always ask ``which value of the variable corresponds with no intervention?''.
Ultimately, though, both interpretations can be treated equally from a mathematical modeling perspective. 
Henceforth, we will use the term ``context variable'' in general, but ``intervention variable'' specifically
for context variables that model an external intervention on the system.

\begin{figure}\centering
\begin{tikzpicture}
  \begin{scope}[xshift=0cm]
  \draw[rounded corners] (0.5,0.5) rectangle (2.5,1.7);
  \node at (1.5,1) {system};
  \node at (1.5,2.4) {context};
  \draw[arr] (1,2) -- (1,1.4);
  \draw[arr] (1.5,2) -- (1.5,1.4);
  \draw[arr] (2,2) -- (2,1.4);
  \end{scope}
  \begin{scope}
  \draw[rounded corners,thick] (0,0) rectangle (3,3);
  \node[anchor=west] at (3.1,1.5) {meta-system};
  \end{scope}
\end{tikzpicture}
  \caption{JCI reduces modeling a system in its environment to modeling the meta-system consisting of the system \emph{and} its environment.\label{fig:JCI_key_idea}}
\end{figure}

That being said, the approach we take in JCI is simple (see also Figure~\ref{fig:JCI_key_idea}): rather than considering a causal model of the system alone (i.e., modeling only the
endogenous system variables), we broaden its
scope to include relevant parts of the environment of the system (i.e., we include the context variables as
additional endogenous variables).
Thereby, we ``internalize'' parts of the environment of the system, 
which makes the meta-system (consisting of both system and its environment) amenable to formal causal modeling.
The meta-system can now formally be considered as occurring in just a single (meta)-context, and thereby we
have reduced the problem of how to deal with multiple contexts to one of dealing with a single context only.
We will formalise this idea in the next subsection.

\subsection{Joint Causal Modeling of Multiple Contexts}\label{sec:JCI_modeling}

Different approaches to modeling multiple contexts can be taken, e.g., using
influence diagrams \citep{Dawid2002}, using selection diagrams \citep{BareinboimPearl2013},
considering only conditional models (i.e., for the conditional probability of the system
given the context) \citep{EatonMurphy07,MooijHeskes_UAI_13}, or using ioSCMs 
\citep{ForreMooij_UAI_19}. Here, we will take what is perhaps the simplest
approach: we treat both context and system variables as endogenous variables in an SCM.

We will use a simple SCM to model the meta-system (i.e., the system and its contexts) causally.
The endogenous variables of the SCM consist of the system variables $\B{X} = (X_i)_{i\in\C{I}}$ with values 
$\B{x} \in \BC{X} = \prod_{i\in\C{I}} \C{X}_i$ and the context variables $\B{C} = (C_k)_{k\in\C{K}}$ with values $\B{c} \in \BC{C} = \prod_{k\in\C{K}} \C{C}_k$. The latent exogenous variables of the SCM are denoted
$\B{E} = (E_j)_{j\in\C{J}}$ with values $\B{e} \in \BC{E} = \prod_{j\in\C{J}} \C{E}_j$.
The SCM modeling the meta-system is then assumed to be of the following form:
\begin{equation}\label{eq:SCM_JCI}
  \C{M}:
  \begin{cases}
    C_k = f_k(\B{X}_{\pasub{\C{H}}{k} \cap \C{I}}, \B{C}_{\pasub{\C{H}}{k} \cap \C{K}}, \B{E}_{\pasub{\C{H}}{k} \cap \C{J}}), & \qquad k \in \C{K}, \\
    X_i = f_i(\B{X}_{\pasub{\C{H}}{i} \cap \C{I}}, \B{C}_{\pasub{\C{H}}{i} \cap \C{K}}, \B{E}_{\pasub{\C{H}}{i} \cap \C{J}}), & \qquad i \in \C{I},\\
    \Prb(\B{E}) = \prod_{j\in\C{J}} \Prb(E_j). &
  \end{cases}
\end{equation}
The system variables $\B{X}$ and context variables $\B{C}$ are all treated as endogenous variables of the meta-system, and
the exogenous variables $\B{E}$ are independent latent variables that are assumed 
not to be caused by the system variables $\B{X}$ or the context variables $\B{C}$.\footnote{At this
stage, we have not yet incorporated the assumption that context variables are exogenous to the system,
and they are still treated equally to system variables in \eref{eq:SCM_JCI}.}
The augmented graph $\C{H}$ has nodes 
$\C{I} \cup \C{J} \cup \C{K}$ and directed edges corresponding to the functional dependencies of the causal
mechanisms on the variables.
The graph $\C{G}(\C{M})$ has only nodes $\C{I} \cup \C{K}$,
and may contain both directed and bidirected edges between the nodes, expressing direct
causal relations and latent confounders.

Note that the most general way to use SCMs to model multiple contexts would be to use
separate SCMs, one for each context.  In that approach, we could have a different 
graph for each context. Representing the contexts jointly, as in \eref{eq:SCM_JCI}, 
we simply obtain the union of those graphs. In particular, even if within each context, the system is acyclic,
it could be that the mixture of systems in different contexts has a cyclic graph.
As a simple example, consider a system with two system variables $X_1$ and $X_2$, and 
consider two different contexts, where in the first context $X_1$ causes $X_2$ (but not vice
versa), and in the second context, $X_2$ causes $X_1$ (but not vice versa); see also Figure~\ref{fig:example_cyclic}.
As a more concrete
example, the engine drives the wheels of a car when going uphill, but when going downhill, 
the rotation of the wheels drives the engine. Modeling this in
a joint SCM as in \eref{eq:SCM_JCI} requires a cyclic graph.

\begin{figure}\centering%
\begin{tikzpicture}
  \begin{scope}
  \node at (-1,2) {(a)};
  \node[var] (X1) at (0,0) {$X_1$};  
  \node[var] (X2) at (2,0) {$X_2$};  
  \draw[arr] (X1) -- (X2);
  \node (C) at (1,1.5) {$C=0$:};
  \end{scope}
  \begin{scope}[xshift=5cm]
  \node at (-1,2) {(b)};
  \node[var] (X1) at (0,0) {$X_1$};  
  \node[var] (X2) at (2,0) {$X_2$};  
  \draw[arr] (X2) -- (X1);
  \node (C) at (1,1.5) {$C=1$:};
  \end{scope}
  \begin{scope}[xshift=10cm]
  \node at (-1,2) {(c)};
  \node[var] (X1) at (0,0) {$X_1$};  
  \node[var] (X2) at (2,0) {$X_2$};  
  \draw[arr,bend left] (X1) edge (X2);
  \draw[arr,bend left] (X2) edge (X1);
  \node[var] (C) at (1,1.5) {$C$};
  \draw[arr] (C) -- (X1);
  \draw[arr] (C) -- (X2);
  \end{scope}
\end{tikzpicture}
  \caption{The graph of a mixture of two acyclic SCMs can be cyclic. (a) $X_1$ causes $X_2$ in context $C=0$; (b) $X_2$ causes $X_1$ in context $C=1$; (c) $X_1$ and $X_2$ cause each other in the joint model.\label{fig:example_cyclic}}
\end{figure}

The model \eref{eq:SCM_JCI} imposes a probability distribution $\Prb(\B{C})$ on
the context variables, the \emph{context distribution}. The context distribution
will reflect the empirical distribution of the context variables
in the \emph{pooled} data $\hat\Prb(\B{C})$, by using as the probability of a context the
fraction of the total number of samples that have been measured in that
context. In case the context variables are used to model interventions, for example, 
the context distribution is determined by the experimental design.
One might object that this makes the model very specific to the
particular setting, since it also specifies the relative numbers of samples in
each data set, but as it turns out, the conclusions of the causal discovery 
procedure do not depend on these details under reasonable assumptions, and 
therefore generalize to other context distributions. In other words, the 
behavior of the system is invariant of the context distribution.

Because the context variables are treated as endogenous variables (similarly to the system variables),
we have ``internalized'' them. 
The main advantage of our modeling approach over alternative approaches is that in \eref{eq:SCM_JCI}, context variables
are formally treated in exactly the same way as the system variables.
This implies in particular that all standard definitions and terminology of Section~\ref{sec:graphical_causal_modeling}, and all
causal discovery methods that are applicable in that setting, can be directly applied.

\subsection{Modeling Interventions as Context Changes}\label{sec:modeling_interventions}

The causal model in \eref{eq:SCM_JCI} allows one to model a perfect  
intervention in the usual way \citep{Pearl2009}. Specifically, the perfect intervention that forces $\B{X}_{I}$ to take on the value $\B{\xi}_I$ (``$\mathrm{do}(\B{X}_I=\B{\xi}_I)$'') for some subset $I\subseteq \C{I}$ and some value $\B{\xi}_I \in \prod_{i\in I} \C{X}_i$ can be modeled by replacing the structural equations for the system variables in \eref{eq:SCM_JCI} by:
$$X_i = \begin{cases}
    \xi_i & i \in I\\
    f_i( \B{X}_{\pa{i} \cap \C{I}}, \B{C}_{\pa{i} \cap \C{K}}, \B{E}_{\pa{i} \cap \C{J}}) & i \in \C{I}\setminus I,\\
  \end{cases}$$
while leaving the rest of the model invariant.\footnote{For brevity, we dropped the subscript $\C{H}$ of $\pasub{\C{H}}{\cdot}$.}

Alternatively, the context variables can be used to model interventions. For example, the same perfect intervention could
be modeled by introducing a context variable $C_k$ that has $\ch{k} = I$, no parents or spouses, and domain $\C{C}_k = \{\emptyset\} \cup \prod_{i\in I} \C{X}_i$, by taking $\B{f}_I$ to be of the following form:
\begin{equation}\label{eq:perfect_intv}
  f_i( \B{X}_{\pa{i} \cap \C{I}}, \B{C}_{\pa{i} \cap \C{K}}, \B{E}_{\pa{i} \cap \C{J}}) =
  \begin{cases}
    \tilde f_i( \B{X}_{\pa{i} \cap \C{I}}, \B{C}_{\pa{i} \cap \C{K} \setminus \{k\}}, \B{E}_{\pa{i} \cap \C{J}}) & C_k = \emptyset \\
    (C_k)_i & C_k \in \prod_{i\in I} \C{X}_i
  \end{cases}
\end{equation}
for $i \in I$. Here, $C_k = \emptyset$ corresponds to no intervention (i.e., the observational baseline).
Modeling a perfect intervention in this way is similar to the concept of ``force variables'' introduced by \citet{Pearl1993b}.
The observational distribution of the system variables is then given by
the conditional distribution $\Prb(\B{X} \given C_k = \emptyset)$, the interventional
distribution corresponding to the perfect intervention $\mathrm{do}(\B{X}_I = \B{\xi}_I)$ is given by
the conditional distribution $\Prb(\B{X} \given C_k = \B{\xi}_I)$, and the marginal distribution 
$\Prb(\B{X})$ represents a mixture of those. This is illustrated in Figure~\ref{fig:JCI_example}.

More general types of interventions such as mechanism changes \citep{TianPearl2001} can be modeled in a similar way, 
simply by not enforcing the dependence on $C_k$ to be of the form \eref{eq:perfect_intv}, but allowing more general
forms of functional dependence. For example, switching the causal mechanism of system
variable $X_i$ from mechanism $A$ to mechanism $B$ can be modeled as follows by introducing a context variable $C_k$ with $\ch{k} = \{i\}$ and domain $\C{C}_k = \{A,B\}$:
\begin{equation*}
  f_i( \B{X}_{\pa{i} \cap \C{I}}, \B{C}_{\pa{i} \cap \C{K}}, \B{E}_{\pa{i} \cap \C{J}}) =
  \begin{cases}
    \tilde f_i^{A}( \B{X}_{\pa{i} \cap \C{I}}, \B{C}_{\pa{i} \cap \C{K} \setminus \{k\}}, \B{E}_{\pa{i} \cap \C{J}}) & C_k = A \\
    \tilde f_i^{B}( \B{X}_{\pa{i} \cap \C{I}}, \B{C}_{\pa{i} \cap \C{K} \setminus \{k\}}, \B{E}_{\pa{i} \cap \C{J}}) & C_k = B. \\
  \end{cases}
\end{equation*}
As another example, a stochastic perfect intervention on $X_i$ that is only successful with a certain probability can be 
modeled by having one of the latent exogenous variables $E_j$ with $j \in \pa{i}$ determine whether the intervention was 
successful:
\begin{equation*}\begin{split}
  & f_i( \B{X}_{\pa{i} \cap \C{I}}, \B{C}_{\pa{i} \cap \C{K}}, \B{E}_{\pa{i} \cap \C{J}}) \\
  & \qquad = \begin{cases}
    \tilde f_i( \B{X}_{\pa{i} \cap \C{I}}, \B{C}_{\pa{i} \cap \C{K} \setminus \{k\}}, \B{E}_{\pa{i} \cap \C{J} \setminus \{j\}}) & C_k = \emptyset \text{ or } E_j = 0 \\
    C_k & C_k \in \C{X}_i \text{ and } E_j = 1.
  \end{cases}
\end{split}\end{equation*}

\begin{figure}[t]
  \centering\begin{tikzpicture}
    \begin{scope}[yshift=-3.5cm]
      \node at (-4,0.5) {(a)};
      \node[var,fill=blue!20!white] (X1) at (0,0) {$X_1$};
      \node[var,fill=blue!20!white] (X2) at (1.5,0) {$X_2$};
      \node[var,fill=blue!20!white] (X3) at (3,0) {$X_3$};
      \draw[arr] (X1) edge (X2);
      \draw[arr] (X2) edge (X3);
      \draw[biarr,bend left] (X1) edge (X2);
      \node at (-2,0.3) {Observational:};
      \node at (-2,-0.2) {($C_\alpha=0$)};
      \begin{scope}[yshift=-1.5cm]
        \node[var,fill=red!20!white] (X1) at (0,0) {$X_1$};
        \node[var,fill=red!20!white] (X2) at (1.5,0) {$X_2$};
        \node[var,fill=red!20!white] (X3) at (3,0) {$X_3$};
        \draw[arr] (X2) edge (X3);
        \node at (-2,0.3) {Interventional:};
        \node at (-2,-0.2) {($C_\alpha=1$)};
      \end{scope}
      \begin{scope}[xshift=7cm,yshift=-1.5cm]
        \node[var,left color=red!20!white,right color=blue!20!white,shading=axis] (I1) at (1.5,1.5) {$C_\alpha$};
        \node[var,left color=red!20!white,right color=blue!20!white,shading=axis] (X1) at (0,0) {$X_1$};
        \node[var,left color=red!20!white,right color=blue!20!white,shading=axis] (X2) at (1.5,0) {$X_2$};
        \node[var,left color=red!20!white,right color=blue!20!white,shading=axis] (X3) at (3,0) {$X_3$};
        \draw[arr] (I1) edge (X2);
        \draw[arr] (X1) edge (X2);
        \draw[arr] (X2) edge (X3);
        \draw[biarr,bend left] (X1) edge (X2);
        \node at (-1.5,1.5) {Jointly:};
      \end{scope}
    \end{scope}
    \begin{scope}[yshift=-7cm]
      \node at (-4,0.5) {(b)};
      \node[text width=4cm,anchor=north] at (-1.0,0.8) {\small\begin{tabular}{ccc}
        \multicolumn{3}{c}{Observational:} \\
        \multicolumn{3}{c}{($C_\alpha = 0$)} \\
        \hline
        $X_1$ & $X_2$ & $X_3$ \\
        \hline
        \color{blue!70}-0.2 & \color{blue!70}-0.4 & \color{blue!70} 0.6 \\
        \color{blue!70} 0.6 & \color{blue!70} 0.8 & \color{blue!70} 1.3 \\
        \color{blue!70}-1.7 & \color{blue!70} 0.1 & \color{blue!70} 0.3 \\
        \hline
      \end{tabular}};
      \node[text width=4cm,anchor=north] at (2.5,0.8) {\small\begin{tabular}{ccc}
        \multicolumn{3}{c}{Interventional:} \\
        \multicolumn{3}{c}{($C_\alpha = 1$)} \\
        \hline
        $X_1$ & $X_2$ & $X_3$ \\
        \hline
        \color{red!70} -0.3 & \color{red!70}  1.8 & \color{red!70} -0.1 \\
        \color{red!70} 1.8  & \color{red!70} -2.2 & \color{red!70} -0.2 \\ 
        \hline
      \end{tabular}};
      \node[text width=4cm,anchor=north] at (8.5,0.8) {\small\begin{tabular}{cccc}
        \multicolumn{4}{c}{Pooled:} \\
        $C_\alpha$ & $X_1$ & $X_2$ & $X_3$ \\
        \hline
        0 & \color{blue!70}-0.2 & \color{blue!70}-0.4 & \color{blue!70} 0.6 \\
        0 & \color{blue!70} 0.6 & \color{blue!70} 0.8 & \color{blue!70} 1.3 \\
        0 & \color{blue!70}-1.7 & \color{blue!70} 0.1 & \color{blue!70} 0.3 \\
        1 & \color{red!70} -0.3 & \color{red!70}  1.8 & \color{red!70} -0.1 \\
        1 & \color{red!70} 1.8  & \color{red!70} -2.2 & \color{red!70} -0.2 \\ 
        \hline
      \end{tabular}};
    \end{scope}
  \end{tikzpicture}
  \caption{Two ways of representing interventions, either through modeling contexts separately (left), or by modeling system and context jointly (right). In this example, we consider a perfect intervention on $X_2$, though the same idea applies to other types of interventions. (a) shows the corresponding causal graphs, as separate graphs for each context (left), or as a single joint graph that includes a context variable (right); (b) shows different ways of grouping the data: as separate data sets for each context (left), or as a single joint data set after pooling (right).\label{fig:JCI_example}}
\end{figure}

This approach of modeling interventions by means of context variables is very general,
as it allows to treat various types of interventions in a unified way. For example, it can deal with perfect interventions \citep{Pearl2009}, mechanism changes \citep{TianPearl2001}, soft interventions \citep{Markowetz++2005}, fat-hand interventions \citep{EatonMurphy07}, activity interventions \citep{MooijHeskes_UAI_13}, and stochastic versions of all these. 
In case the context variables are used to model interventions in this way, we 
also refer to the context distribution $\Prb(\B{C})$ (the probability for each context to occur)
as the \emph{experimental design}.

\subsection{JCI Assumptions}\label{sec:JCI_assumptions}

In this subsection, we discuss additional background knowledge on the causal relationships
of context variables that one may often have in practice, and that can be very helpful for
causal discovery.

\subsubsection{JCI Assumption~\ref{ass:simple_scm}}

First, we restate formally our basic modeling assumption:
\setcounter{assumption}{-1}
\begin{assumption}(``Joint SCM'')\label{ass:simple_scm}
The data-generating mechanism is described by a simple SCM $\C{M}$ of the form:
\begin{equation}\label{eq:SCM_JCI_ass}
  \C{M}:
  \begin{cases}
    C_k = f_k(\B{X}_{\pasub{\C{H}}{k} \cap \C{I}}, \B{C}_{\pasub{\C{H}}{k} \cap \C{K}}, \B{E}_{\pasub{\C{H}}{k} \cap \C{J}}), & \qquad k \in \C{K}, \\
    X_i = f_i(\B{X}_{\pasub{\C{H}}{i} \cap \C{I}}, \B{C}_{\pasub{\C{H}}{i} \cap \C{K}}, \B{E}_{\pasub{\C{H}}{i} \cap \C{J}}), & \qquad i \in \C{I},\\
    \Prb(\B{E}) = \prod_{j\in\C{J}} \Prb(E_j), &
  \end{cases}
\end{equation}
that jointly models the system and the context.
Its graph $\C{G}(\C{M})$ has nodes
$\C{I} \cup \C{K}$ (corresponding to system variables $\{X_i\}_{i\in\C{I}}$ and context variables $\{C_k\}_{k\in\C{K}}$).
\end{assumption}
Whereas we will always make this assumption in order to facilitate the formulation of JCI, 
the following three assumptions that we discuss are optional,
and their applicability has to be decided based on a case-by-case basis.

\subsubsection{JCI Assumption~\ref{ass:uncaused}}
Typically, when a modeler decides to distinguish a \emph{system} from its \emph{context}, the modeler possesses
background knowledge that expresses that the context is \emph{exogenous} to the system:
\begin{assumption}(``Exogeneity'', optional)\label{ass:uncaused}
  No system variable causes any context variable, i.e.,
$$\forall k \in \C{K}, \forall i\in\C{I}: \quad i \to k \notin \C{G}(\C{M}).$$
\end{assumption}
This exogeneity assumption is often easy to justify, for example if context is gender or age. 
Another common case is that the context encodes interventions
on the system that have been decided and performed on the system \emph{before} measurements on the system are performed: this already rules out any causal influence of system variables on the intervention (context) variables if time travel is not deemed possible. Of course, one can imagine settings in which a system variable was measured before an
intervention was performed on the system. For example, a doctor typically first diagnoses a patient 
\emph{before} deciding on treatment. For system variables containing the results of the medical examination
used for the diagnosis and intervention variables describing the treatment that was decided \emph{after}---and based upon---the
medical examination, JCI Assumption~\ref{ass:uncaused} would not apply.

\subsubsection{JCI Assumption~\ref{ass:unconfounded}}\label{sec:ass_unconfounded}
The second JCI assumption generalizes the randomization assumption for randomized controlled trials:
\begin{assumption}(``Complete randomized context'', optional)\label{ass:unconfounded}
No context variable is confounded with a system variable, i.e.,
$$\forall k\in\C{K}, \forall i\in\C{I}: \quad i \oto k \notin \C{G}(\C{M}).$$
\end{assumption}
This assumption is often harder to justify in practice. 
It is justifiable in experimental protocols in which the decision of which intervention to perform on the system does not depend on 
anything else that might also affect the system of interest, and in which the observed context variables provide a complete description of the context. This is ensured for example in case of proper randomization in a double-blind randomized trial setting, i.e., in which neither the patient nor the physician knows whether the patient was assigned a drug or a placebo.

Many experimental protocols that do not involve explicit coin flips or random number generators are implicitly performing randomization. For example, in the experimental procedure described by \citet{SPP05} (see also Section~\ref{sec:exp_sachs}), one starts with a collection of human immune system cells. These are divided into batches randomly, without taking into account any property of the cells. When done carefully, the experimenter tries to ensure that for example the size of a cell cannot influence the batch it ends up in, by stirring the liquid that contains the cells before pipetting. Then, after randomly assigning cells to batches, interventions are performed on each batch separately, by adding some chemical compound to the batch of cells. Finally, properties of each individual cell within each batch are measured. If the system variables reflect the measured properties of the individual cells, and the context variables encode the batch ID, this experimental procedure justifies JCI Assumption~\ref{ass:unconfounded}. 

However, one should be careful not to jump to the conclusion that the chemical compound administered to the batch is what actually causes the observed system behavior, as there may be other factors that vary across batches due to unintentional side effects of the experimental procedure. For example, the lab assistant that carries out the experiment for a particular batch of cells might influence the outcome, because slightly different experimental procedures are used by different lab assistants. Another example is that the time of the day may affect the measurements, and also correlate with batch ID. In situations like those, \emph{identifying} the batch ID with the chemical compound administered to that batch could be misleading, and could lead one to incorrectly attribute the inferred causal relation between batch ID and a certain system variable to the causal effect of the intended intervention corresponding to that batch on the system variable. This is a subtle type of error that the causal modeler should beware of. Even though we have good reasons to assume that proper randomization was performed for batch ID in the \citet{SPP05} experiment, it is questionable whether the interpretation of the context variables as concerning solely the addition of certain chemical compounds (and not any other factors that actually varied across batches) is appropriate.

The issue can also be understood by noting that JCI Assumption~\ref{ass:unconfounded} may not be preserved when marginalizing out context variables, as illustrated in Figure~\ref{fig:unconfounded_invalid}. The following example describes a situation in which this phenomenon may occur.
\begin{example}
Consider a randomized trial setup for establishing whether sugar causes plants to grow. Context variable $C_\alpha$ denotes the coin flip result, $C_\beta$ indicates whether sugar is administered to the plant, and $C_\gamma$ indicates whether water is administered to the plant. The experimenter decided to use an experimental design with two groups, and assigning plants to groups with a coin flip. One group of plants was administered a solution consisting of sugar dissolved in water on a daily basis, the other (control) group was not treated in any way. The growth rate $X_1$ of the plants was measured for both groups. Suppose the following experimental design was used:
\begin{center}\begin{tabular}{l|lll}
  $\Prb(\B{C}=\B{c})$ & $C_\alpha$ (coin flip) & $C_\beta$ (sugar) & $C_\gamma$ (water) \\
  \hline
  \small $\tfrac{1}{2}$ & 0      & 0       & 0 \\
  \small $\tfrac{1}{2}$ & 1      & 1       & 1 \\
\end{tabular}\end{center}
If one would only take context variable $C_\beta$ (did the plant get sugar?) into account and would treat $C_\alpha$ and $C_\gamma$ as latent, as in Figure~\ref{fig:unconfounded_invalid}(b), and would make JCI Assumptions~\ref{ass:uncaused} and \ref{ass:unconfounded}, one would arrive at the (wrong) conclusion that sugar causes plants to grow. However, if one would take all three context variables into account, and make JCI Assumptions~\ref{ass:uncaused} and \ref{ass:unconfounded}, one would obtain the right conclusion that at least one of the three context variables must cause plants to grow. 
\end{example}
A simple remedy to avoid the wrong conclusion if only $C_\beta$ is observed would be to drop JCI Assumption~\ref{ass:unconfounded}: then it is no longer identifiable whether $C_\beta$ causes $X_1$, or whether $C_\beta$ and $X_1$ are just confounded.

\begin{figure}\centering%
\begin{tikzpicture}
    \begin{scope}
      \node at (-2,3) {(a)};
      \draw (-2,0.75) edge[dotted] (0.5,0.75);
      \node[var] (C0) at (-0.75,2.75) {$C_\alpha$};
      \node[var] (C1) at (-1.5,1.5) {$C_\beta$};
      \node[var] (C2) at (0,1.5) {$C_\gamma$};
      \node[var] (X0) at (-0.75,0) {$X_1$};
      \draw[arr] (C0) edge (C1);
      \draw[arr] (C0) edge (C2);
      \draw[arr] (C2) edge (X0);
    \end{scope}
    \begin{scope}[xshift=5cm]
      \node at (-2,3) {(b)};
      \draw (-2,0.75) edge[dotted] (0.5,0.75);
      \node[var] (C1) at (-1.5,1.5) {$C_\beta$};
      \node[var] (X0) at (-0.75,0) {$X_1$};
      \draw[biarr] (C1) edge (X0);
    \end{scope}
  \end{tikzpicture}
  \caption{Confounding between system and context variables due to unobserved context variables. 
  (a) If all three context variables $C_\alpha, C_\beta, C_\gamma$ are observed, JCI Assumption~\ref{ass:unconfounded} would be valid. (b) After marginalizing out $C_\alpha$ and $C_\gamma$, leaving only context variable $C_\beta$ as observed, JCI Assumption~\ref{ass:unconfounded} is no longer valid.\label{fig:unconfounded_invalid}}
\end{figure}

\subsubsection{JCI Assumption~\ref{ass:dependences}}

We have seen that JCI Assumption~\ref{ass:uncaused} is often easily
justifiable, but the applicability of JCI Assumption~\ref{ass:unconfounded} may
be less obvious in practice. We will now state JCI Assumption~\ref{ass:dependences}, which
can be useful whenever both JCI Assumptions \ref{ass:uncaused} and 
\ref{ass:unconfounded} have been made as well.
\begin{assumption}(``Generic context model'', optional)\label{ass:dependences}
The context graph\footnote{Remember that $\C{G}(\C{M})_{\C{K}}$ denotes the subgraph
on the context variables $\C{K}$ induced by the causal graph $\C{G}(\C{M})$.}
$\C{G}(\C{M})_{\C{K}}$ is of the following special form: 
$$\forall k \ne k' \in \C{K}: \quad k \oto k' \in \C{G}(\C{M}) \quad \land \quad k \to k' \notin \C{G}(\C{M}).$$
\end{assumption}
In Figure~\ref{fig:JCI_ass_3}(b), this assumption is satisfied, while in Figure~\ref{fig:JCI_ass_3}(a), it is not.
We will show that JCI Assumption~\ref{ass:dependences} seems stronger than it is, since it can be made without loss of generality in many cases occurring in practice.

In order to precisely formulate and prove that claim, the following definition is needed.
\begin{definition}
Given an SCM $\C{M}$ satisfying JCI Assumption~\ref{ass:simple_scm}, define the \emph{conditional system graph}
$\C{G}(\C{M})_{\intervene(\C{K})}$ as the DMG with context nodes $\C{K}$ and system nodes $\C{I}$, 
and as directed and bidirected edges those edges in $\C{G}(\C{M})$ that
contain at least one system node in $\C{I}$ (i.e., excluding edges between context nodes).
We will graphically represent the system nodes $\C{I}$ of $\C{G}(\C{M})_{\intervene(\C{K})}$
by ellipses and the context nodes $\C{K}$ of $\C{G}(\C{M})_{\intervene(\C{K})}$ by squares.
\end{definition}
Figure~\ref{fig:JCI_ass_3}(c) shows the common conditional system graph for SCMs with graph as
given in Figure~\ref{fig:JCI_ass_3}(a) and for SCMs with graph as given in Figure~\ref{fig:JCI_ass_3}(b).
The conditional system graph provides a particular graphical representation
for an SCM with context and system variables that is less expressive than its
graph. This representation is useful when we are not interested in describing
relationships between context variables, but only in describing the
relationships between system variables and how the context affects the
system.

The following key result essentially states that when one is only interested
in modeling the causal relations involving the system variables (under JCI Assumptions~\ref{ass:uncaused} and \ref{ass:unconfounded}), one does not
need to care about the \emph{causal relations} between the context variables, as long
as one correctly models the context \emph{distribution}.
\newcommand\cmdTheoremReplaceContext{%
Assume that JCI Assumptions \ref{ass:simple_scm}, \ref{ass:uncaused} and \ref{ass:unconfounded} hold for SCM $\C{M}$:
\begin{equation*}
  \C{M}:
  \begin{cases}
    C_k = f_k(\B{C}_{\pasub{\C{H}}{k} \cap \C{K}}, \B{E}_{\pasub{\C{H}}{k} \cap \C{J}}), & \qquad k \in \C{K}, \\
    X_i = f_i(\B{X}_{\pasub{\C{H}}{i} \cap \C{I}}, \B{C}_{\pasub{\C{H}}{i} \cap \C{K}}, \B{E}_{\pasub{\C{H}}{i} \cap \C{J}}), & \qquad i \in \C{I},\\
    \Prb(\B{E}) = \prod_{j\in\C{J}} \Prb(E_j), &
  \end{cases}
\end{equation*}
For any other SCM $\tilde{\C{M}}$ satisfying JCI Assumptions \ref{ass:simple_scm}, \ref{ass:uncaused} and \ref{ass:unconfounded}
that is the same as $\C{M}$ except that it models the context differently, i.e., of the form
\begin{equation*}
  \tilde{\C{M}}:
  \begin{cases}
    C_k = \tilde{f}_k(\B{C}_{\pasub{\tilde{\C{H}}}{k} \cap \C{K}}, \B{E}_{\pasub{\tilde{\C{H}}}{k} \cap \tilde{\C{J}}}), & \qquad k \in \C{K}, \\
    X_i = f_i(\B{X}_{\pasub{\C{H}}{i} \cap \C{I}}, \B{C}_{\pasub{\C{H}}{i} \cap \C{K}}, \B{E}_{\pasub{\C{H}}{i} \cap \C{J}}), & \qquad i \in \C{I},\\
    \Prb(\B{E}) = \prod_{j\in\tilde{\C{J}}} \Prb(E_j), &
  \end{cases}
\end{equation*}
  with $\C{J} \subseteq \tilde{\C{J}}$ and $\pasub{\C{H}}{i} = \pasub{\tilde{\C{H}}}{i}$ for all $i \in \C{I}$, 
  we have that
  \begin{enumerate}[(i)]
    \item the conditional system graphs coincide: $\C{G}(\C{M})_{\intervene(\C{K})} = \C{G}(\tilde{\C{M}})_{\intervene(\C{K})}$;
    \item if $\tilde{\C{M}}$ and $\C{M}$ induce the same context distribution, i.e., $\Prb_{\C{M}}(\B{C}) = \Prb_{\tilde{\C{M}}}(\B{C})$, then for any perfect intervention on the system variables $\intervene(I,\B{\xi}_I)$ with $I \subseteq \C{I}$ (including the non-intervention $I = \emptyset$), $\tilde{\C{M}}_{\intervene(I,\B{\xi}_I)}$ is observationally equivalent to $\C{M}_{\intervene(I,\B{\xi}_I)}$.
    \item if the context graphs
      $\C{G}(\tilde{\C{M}})_{\C{K}}$ and $\C{G}(\C{M})_{\C{K}}$ induce the same separations, then also $\C{G}(\tilde{\C{M}})$ and $\C{G}(\C{M})$ induce the same separations (where ``separations'' can refer to either $d$-separations or $\sigma$-separations).
  \end{enumerate}}
\begin{theorem}\label{thm:replace_context}
\cmdTheoremReplaceContext
\end{theorem}
\begin{proof}
  See Appendix~\ref{sec:app:proofs}.
\end{proof}

The following corollary of Theorem~\ref{thm:replace_context} states that JCI Assumption~\ref{ass:dependences} can be made without loss of generality for the purposes of constraint-based causal discovery if the context distribution contains no conditional independences:
\newcommand\cmdCorollaryJCIAssC{%
Assume that JCI Assumptions \ref{ass:simple_scm}, \ref{ass:uncaused} and \ref{ass:unconfounded} hold for SCM $\C{M}$.
Then there exists an SCM $\tilde{\C{M}}$ that satisfies JCI Assumptions \ref{ass:simple_scm}, \ref{ass:uncaused} and \ref{ass:unconfounded} and \ref{ass:dependences}, such that
  \begin{enumerate}[(i)]
    \item the conditional system graphs coincide: $\C{G}(\C{M})_{\intervene(\C{K})} = \C{G}(\tilde{\C{M}})_{\intervene(\C{K})}$;
    \item for any perfect intervention on the system variables $\intervene(I,\B{\xi}_I)$ with $I \subseteq \C{I}$ (including the non-intervention $I = \emptyset$), $\tilde{\C{M}}_{\intervene(I,\B{\xi}_I)}$ is observationally equivalent to $\C{M}_{\intervene(I,\B{\xi}_I)}$;
    \item if the context distribution $\Prb_{\C{M}}(\B{C})$ contains no conditional or marginal independences, then the same $\sigma$-separations hold in $\C{G}(\tilde{\C{M}})$ as in $\C{G}(\C{M})$; if in addition, the Directed Global Markov Property holds for $\C{M}$, then also the same $d$-separations hold in $\C{G}(\tilde{\C{M}})$ as in $\C{G}(\C{M})$.\label{coro:JCI_ass_3_dependences}
  \end{enumerate}
}
\begin{corollary}\label{coro:JCI_ass_3}
\cmdCorollaryJCIAssC
\end{corollary}
\begin{proof}
  This follows from Theorem~\ref{thm:replace_context} by showing that there exists an $\tilde{\C{M}}$ that satisfies all requirements in Theorem~\ref{thm:replace_context} and JCI Assumption~\ref{ass:dependences} by construction, and that induces the same context distribution as $\C{M}$ does. For a detailed proof, see Appendix~\ref{sec:app:proofs}.
\end{proof}
An example illustrating this corollary is provided in Figure~\ref{fig:JCI_ass_3}. 

\begin{figure}\centering
  \begin{tikzpicture}
    \begin{scope}[xshift=0cm]
      \node at (-1.5,2) {(a)};
      \draw (-1.5,0.6) edge[dotted] (3,0.6);
      \node[var] (C0) at (-0.75,1.2) {$C_\alpha$};
      \node[var] (C1) at (0.75,1.2) {$C_\beta$};
      \node[var] (C2) at (2.25,1.2) {$C_\gamma$};
      \node[var] (X0) at (-0.75,0) {$X_1$};
      \node[var] (X1) at (0.75,0) {$X_2$};
      \node[var] (X2) at (-0.75,-1.2) {$X_3$};
      \draw[arr] (C0) edge (C1);
      \draw[arr] (C1) edge (C2);
      \draw[arr,bend left] (C2) edge (C1);
      \draw[arr,bend left] (C0) edge (C2);
      \draw[arr] (C0) edge (X0);
      \draw[arr] (C1) edge (X1);
      \draw[arr,bend left] (C2) edge (X2);
      \draw[arr] (X0) edge (X2);
    \end{scope}
    \begin{scope}[xshift=5cm]
      \node at (-1.5,2) {(b)};
      \draw (-1.5,0.6) edge[dotted] (3,0.6);
      \node[var] (C0) at (-0.75,1.2) {$C_\alpha$};
      \node[var] (C1) at (0.75,1.2) {$C_\beta$};
      \node[var] (C2) at (2.25,1.2) {$C_\gamma$};
      \draw[biarr] (C0) edge (C1);
      \draw[biarr] (C1) edge (C2);
      \draw[biarr,bend left] (C0) edge (C2);
      \node[var] (X0) at (-0.75,0) {$X_1$};
      \node[var] (X1) at (0.75,0) {$X_2$};
      \node[var] (X2) at (-0.75,-1.2) {$X_3$};
      \draw[arr] (C0) edge (X0);
      \draw[arr] (C1) edge (X1);
      \draw[arr,bend left] (C2) edge (X2);
      \draw[arr] (X0) edge (X2);
    \end{scope}
    \begin{scope}[xshift=10cm]
      \node at (-1.5,2) {(c)};
      \draw (-1.5,0.6) edge[dotted] (3,0.6);
      \node[varc] (C0) at (-0.75,1.2) {$C_\alpha$};
      \node[varc] (C1) at (0.75,1.2) {$C_\beta$};
      \node[varc] (C2) at (2.25,1.2) {$C_\gamma$};
      \node[var] (X0) at (-0.75,0) {$X_1$};
      \node[var] (X1) at (0.75,0) {$X_2$};
      \node[var] (X2) at (-0.75,-1.2) {$X_3$};
      \draw[arr] (C0) edge (X0);
      \draw[arr] (C1) edge (X1);
      \draw[arr,bend left] (C2) edge (X2);
      \draw[arr] (X0) edge (X2);
    \end{scope}
  \end{tikzpicture}
  \caption{Example graphs of (a) a true SCM $\C{M}$ and (b) the modified SCM $\tilde{\C{M}}$ constructed in the proof of Corollary~\ref{coro:JCI_ass_3} that satisfies JCI Assumption~\ref{ass:dependences}, and (c) their corresponding conditional system graph $\C{G}(\C{M})_{\intervene(\C{K})}$. Corollary~\ref{coro:JCI_ass_3} gives sufficient conditions for $\tilde{\C{M}}$ and $\C{M}$ to be equivalent for our purposes.\label{fig:JCI_ass_3}}
\end{figure}

JCI Assumption~\ref{ass:dependences} is typically made for convenience.
When our aim is not to model the causal relations \emph{between} the context variables, but just to use the context variables as an aid to model the causal relations between system variables and between context and system variables, Corollary~\ref{coro:JCI_ass_3} shows that we may assume JCI Assumption~\ref{ass:dependences} without loss of generality if JCI Assumptions \ref{ass:uncaused} and \ref{ass:unconfounded} are made and the context distribution contains no (conditional) independences. The causal discovery algorithm then does not need to waste time on learning the causal relations between context variables but can focus directly on learning the causal relations involving the system variables.

\begin{figure}\centering
  \begin{tikzpicture}
    \begin{scope}[xshift=0cm]
      \node at (-2,2) {(a)};
      \draw (-1.5,0.6) edge[dotted] (1.5,0.6);
      \node[var] (C0) at (-0.75,1.2) {$C_\alpha$};
      \node[var] (C1) at (0.75,1.2) {$C_\beta$};
      \node[var] (X0) at (-0.75,0) {$X_1$};
      \node[var] (X1) at (0.75,0) {$X_2$};
      \draw[arr] (C0) edge (X0);
      \draw[arr] (C1) edge (X1);
    \end{scope}
    \begin{scope}[xshift=5cm]
      \node at (-2,2) {(b)};
      \draw (-1.5,0.6) edge[dotted] (1.5,0.6);
      \node[var] (C0) at (-0.75,1.2) {$C_\alpha$};
      \node[var] (C1) at (0.75,1.2) {$C_\beta$};
      \draw[biarr,bend left] (C0) edge (C1);
      \node[var] (X0) at (-0.75,0) {$X_1$};
      \node[var] (X1) at (0.75,0) {$X_2$};
      \draw[arr] (C0) edge (X0);
      \draw[arr] (C1) edge (X1);
    \end{scope}
  \end{tikzpicture}
  \caption{Example that illustrates that
  the genericity assumption in statement (\ref{coro:JCI_ass_3_dependences}) of Corollary~\ref{coro:JCI_ass_3} is necessary.
  Graphs of (a) the true SCM $\C{M}$ and (b) the modified SCM $\tilde{\C{M}}$ constructed in the
  proof of Corollary~\ref{coro:JCI_ass_3} that are not Markov equivalent.
  The graph in (a) is identifiable under JCI Assumptions \ref{ass:simple_scm}--\ref{ass:unconfounded}. The joint distribution
  $\Prb_{\C{M}}(\B{X},\B{C})$ is not faithful with respect to the graph in (b), which is the minimal one that also satisfies JCI Assumption \ref{ass:dependences} such that $\Prb_{\C{M}}(\B{X},\B{C})$ is Markov with respect to it.\label{fig:JCI_independent_contexts}}
\end{figure}

Note that the genericity assumption in statement (\ref{coro:JCI_ass_3_dependences}) of Corollary~\ref{coro:JCI_ass_3}
(i.e., $\Prb_{\C{M}}(\B{C})$ containing no conditional independences) is necessary,
as the simple counterexample in Figure~\ref{fig:JCI_independent_contexts} shows.
Depending on how well the causal discovery algorithm can handle faithfulness violations, 
model misspecification due to incorrectly
assuming JCI Assumption~\ref{ass:dependences} even though $\Prb(\B{C})$ contains conditional 
independences might prevent successful identification of the causal relationships between system variables.
Therefore, it is prudent to check that the empirical context distribution $\hat\Prb(\B{C})$ indeed contains
no conditional independences before making JCI Assumption~\ref{ass:dependences}.

\begin{table}
\centering\begin{tabular}{lllll|l}
  $C_\alpha$ & $C_\beta$ & $C_\gamma$ & $C_\delta$ & $C_\epsilon$ & possible interpretation \\
  \hline
  0 & 0 & 0 & 0 & 0 & observational \\
  1 & 0 & 0 & 0 & 0 & intervention $\alpha$ \\
  0 & 1 & 0 & 0 & 0 & intervention $\beta$ \\
  0 & 0 & 1 & 0 & 0 & intervention $\gamma$ \\
  0 & 0 & 0 & 1 & 0 & intervention $\delta$ \\
  0 & 0 & 0 & 0 & 1 & intervention $\epsilon$ \\
\end{tabular}
  \caption{Example of a diagonal design with 5 context variables. If the context variables are indicators of interventions, the context with $C_k = 0$ for all $k \in \C{K}$ corresponds with the purely observational setting, and the other contexts in which one $C_k = 1$ and the other $C_l = 0$ for $l \ne k$ correspond with a particular intervention each.\label{tab:diagonal_design}}
\end{table}

An example of a common situation in which the context distribution contains no conditional independences is what we refer to as a \emph{diagonal design} (see also Table~\ref{tab:diagonal_design}). 
This is a simple experimental design that is often used to discover the effects of single interventions when one is not interested in understanding the interactions that multiple interventions might have. Note that two non-constant binary variables $X$, $Y$ can only be independent if $\Prb(X=1,Y=1) > 0$. Even more, they can only be conditionally independent given a third discrete variable $\B{Z}$ if $\Prb(X=1,Y=1 \given \B{Z}=\B{z}) > 0$ for all $\B{z}$ with $\Prb(\B{Z}=\B{z})>0$. Therefore, each pair of context variables is dependent in a diagonal design (as there is no context in which a pair of context variables simultaneously obtains the value 1), even conditionally on any subset of the other context variables. In other words, the context distribution $\Prb(\B{C})$ corresponding to any such diagonal design (with non-zero probability for each context) contains no conditional independences.

JCI Assumption~\ref{ass:dependences} can easily be modified for situations in which the context distribution \emph{does} contain
conditional independences. For example, in the extreme case in which all context variables are jointly independent,
one would simply assume that $\C{G}(\C{M})$ contains no directed and no bidirected edges between context variables. 
Such situations may occur for symmetric experimental designs in which all context variables are jointly independent by design (for example, factorial designs with equal sample sizes in each experimental context). However, we believe that this occurs less often in practice than the generic case in which all context variables are (conditionally) dependent, because resource constraints often lead experimenters to deviate from completely symmetric experimental designs. Therefore, rather than assuming the context variables to be jointly independent as a default, we have opted here for the more generic default of assuming that no conditional independences hold between context variables in the context distribution. 

More generally, one could replace JCI Assumption~\ref{ass:dependences} by assuming that $\C{G}(\C{M})_\C{K}$ equals
a certain graph that expresses the known conditional independences in the experimental design.
Theorem~\ref{thm:replace_context} can be applied to these more general situations as well and shows
that for the purpose of constraint-based causal discovery, any context graph 
that implies the observed conditional independences (i.e., any graph that is Markov 
equivalent to the true context graph) works.

Another alternative is to omit JCI Assumption~\ref{ass:dependences}
and instead try to infer the context subgraph $\C{G}(\C{M})_{\C{K}}$ from the data. This would
typically be computationally more expensive, but in our experience does not seem to make much of a difference
in terms of accuracy in our experiments (as we report in Section~\ref{sec:experiments}).

JCI Assumption~\ref{ass:dependences} only makes sense when both JCI Assumptions~\ref{ass:uncaused}
and \ref{ass:unconfounded} are made. If we would not make JCI Assumption~\ref{ass:uncaused}
or \ref{ass:unconfounded}, the causal relations between the observed context variables will have testable
consequences in the joint distribution in general. For an example of this,\footnote{We are grateful to
Thijs van Ommen for pointing this out.} see Figure~\ref{fig:JCI_ass3_without_ass2}.
Here, $C_\alpha$ could be ``lab'', and $C_\beta$ could be the ``temperature'' at which an experiment is performed. 
In this case, we get different conditional independences in the joint distribution $\Prb(X_1,X_2,C_\alpha,C_\beta)$
if lab causes temperature than when they are confounded (for example, by geographical location). 
Something similar can happen if context variables are caused by system variables.

\begin{figure}\centering
  \begin{tikzpicture}
    \begin{scope}[xshift=0cm]
      \node at (-2,2) {(a)};
      \draw (-1.5,0.6) edge[dotted] (1.5,0.6);
      \node[var] (C0) at (-0.75,1.2) {$C_\alpha$};
      \node[var] (C1) at (0.75,1.2) {$C_\beta$};
      \draw[biarr,bend left] (C0) edge (C1);
      \node[var] (X0) at (-0.75,0) {$X_1$};
      \node[var] (X1) at (0.75,0) {$X_2$};
      \draw[biarr] (C0) edge (X0);
      \draw[biarr] (C1) edge (X1);
    \end{scope}
    \begin{scope}[xshift=5cm]
      \node at (-2,2) {(b)};
      \draw (-1.5,0.6) edge[dotted] (1.5,0.6);
      \node[var] (C0) at (-0.75,1.2) {$C_\alpha$};
      \node[var] (C1) at (0.75,1.2) {$C_\beta$};
      \draw[arr] (C0) edge (C1);
      \node[var] (X0) at (-0.75,0) {$X_1$};
      \node[var] (X1) at (0.75,0) {$X_2$};
      \draw[biarr] (C0) edge (X0);
      \draw[biarr] (C1) edge (X1);
    \end{scope}
  \end{tikzpicture}
  \caption{If JCI Assumption~\ref{ass:unconfounded} does not apply, the causal relations between context variables have testable consequences for the conditional independences in the joint distribution. (a) $X_1 \nCI X_2 \given \{C_\alpha,C_\beta\}$; (b) $X_1 \CI X_2 \given \{C_\alpha,C_\beta\}$.\label{fig:JCI_ass3_without_ass2}}
\end{figure}

\subsubsection{Summary of JCI Assumptions and Other Background Knowledge}

Summarizing, the JCI framework rests on different assumptions, one of which is required, whereas the others are all optional.
The basic assumption that is required is JCI Assumption~\ref{ass:simple_scm}, which states that the meta-system consisting of context and system can be described by a simple SCM.
This is just the standard assumption made throughout the causal discovery literature, but now applied to the meta-system rather than to the system only.
In addition, assumptions about the causal relationships of the context variables can be made, which are all optional and can be decided on a case-by-case basis.
In most cases, we would expect JCI Assumption~\ref{ass:uncaused} (no system variable causes any context
variable) to apply. 
In some cases, also JCI Assumption~\ref{ass:unconfounded} (no system variable is confounded with any context variables) applies.
If both apply, one can assume JCI Assumption~\ref{ass:dependences} for convenience if the context distribution contains no (conditional) independences.
More generally, we only need to model the observed conditional independences in the context distribution,
not necessarily their causal relations, when our interest is in modeling the causal relations involving 
system variables only.

The reader may wonder when one can ever be sure in practice that JCI Assumption~\ref{ass:unconfounded} applies.
There is one very common scenario in which JCI Assumption~\ref{ass:unconfounded} holds. This is in a scientific
experiment in which, in chronological order:
\begin{enumerate}[(i)]
  \item an ensemble of systems is prepared in an initial state;
  \item the systems are randomly permuted and randomly divided into batches;
  \item for each batch, all systems in the batch are intervened upon simultaneously in the same way
    (following an experimental protocol determined in advance);
  \item measurements of the system variables are performed.
\end{enumerate}
The experimental protocol specifies the \emph{intended interventions} for each batch, 
which should be completely encoded as context variables.
Since the intended interventions have been decided \emph{before}
system variables are measured, the intended interventions cannot be caused by
the system variables.
Because the systems were randomly
permuted and divided into batches, the assigned batch cannot be caused by 
prior values of the system variables, or by anything else that may also have an effect on the system variables.
Because the intended interventions for each system in each batch are 
determined completely by the batch, this implies that intended interventions and system variables cannot be confounded.
As long as the context variables
provide a complete encoding of the intended interventions (i.e., the intended interventions
are in one-to-one correspondence to values of the context variables), JCI Assumptions~\ref{ass:uncaused}
and \ref{ass:unconfounded} then apply to the context variables.\footnote{If the experimenter
sticks to the experimental protocol that was fixed before the experiment was performed, any possible
influence of the system variables on the \emph{performed interventions} is excluded, and therefore the
\emph{performed interventions} will equal the \emph{intended interventions}. This means that 
the JCI modeling framework (with JCI Assumptions~\ref{ass:uncaused} and \ref{ass:unconfounded}) applies 
also when interpreting the context variables as the performed interventions. This may explain why it
is considered good scientific practice to perform an experiment according to an experimental protocol 
that was fixed beforehand, and not deviate from it in case of unexpected measurement outcomes, for example.}
If additionally, no (conditional) independences hold in the empirical context distribution,
we can also make use of JCI Assumption~\ref{ass:dependences} to simplify and speed up 
the causal discovery procedure.

In more general scenarios, such as the example in the introduction
(concerning the question whether playing violent computer games causes
aggressive behavior), the validity of JCI Assumption~\ref{ass:unconfounded}
(no confounding between context and system) should not be taken for granted. 
For example, it could be that precisely the 
schools with a more violent population of pupils see themselves forced to
actively take measures to promote social behavior. In that case, the level
of violence in the past would confound $C_\beta$ (does a school take measures
to stimulate social behavior) and $X_2$ (how violently do the pupils of the
school behave). 
Thus, in scenarios like these, it seems safer
not to rely on JCI Assumption~\ref{ass:unconfounded} as incorrectly assuming
it might lead to wrong conclusions (although we do not currently understand
the precise impact of such model misspecification).

For causal discovery in the JCI framework, knowledge of the intervention \emph{targets} (or more generally, which system variables are affected directly by which context variables) is not necessary, but it is certainly helpful and can be exploited similarly to other available background knowledge, depending on the algorithm used to implement JCI. When applying JCI on a combination of different interventional data sets, intervention targets can be learnt from data when they are not known (as the direct effects of intervention variables), similarly to how the effects of system variables can be learnt. 
One main advantage of the JCI framework is that it offers a unified way to deal with different types of interventions,
as discussed in Section~\ref{sec:modeling_interventions}. Therefore, knowledge of intervention \emph{types} (e.g., is it a perfect intervention, or a mechanism change?) is also not necessary, but can still be helpful as it provides additional background knowledge that may be exploited for causal discovery. 

In concluding this subsection, we observe that the JCI framework 
generalizes and combines the ideas of causal discovery from purely observational
data and of causal discovery by means of randomized controlled trials. Indeed,
note that if JCI is applied to a single context (i.e., 0 context variables), 
it reduces to the standard setting of causal discovery from purely observational data
described in Section~\ref{sec:CDobs}. 
If JCI is applied to a setting with a single context variable and a single system variable, 
JCI (with Assumptions \ref{ass:uncaused} and \ref{ass:unconfounded}) reduces to the randomized
controlled trial setting described in Section~\ref{sec:CDexp}. Therefore, the Joint Causal Inference framework 
truly generalizes both these special cases.

\section{Causal Discovery from Multiple Contexts with JCI}\label{sec:causal_discovery}

In this section, we discuss how causal discovery from multiple contexts
can be performed in the Joint Causal Inference framework.
Our starting point is the assumption that some model of the form \eref{eq:SCM_JCI_ass}
is an appropriate causal model for the system and its context, and we have 
obtained samples of all system variables in multiple contexts.\footnote{An interesting
problem setting considered by several researchers \citep{Claassen++_NIPS2010,IOD2011,triantafillou2015constraint,HEJ2014,ForreMooij_UAI_18} that we do not consider here would be to allow for each context a (possibly context-dependent) subset of system variables to remain unobserved.}
Suppose that the exact model $\C{M}$ and in particular, its causal graph
$\C{G}(\C{M})$, are unknown to us. The goal of \emph{causal
discovery} is to infer as much as possible about the causal graph $\C{G}(\C{M})$
from the available data and from available background knowledge about context and system.

Let us denote the data set for context $\B{c} \in \BC{C}$ as
$\C{D}^{(\B{c})} = \big((x_{in}^{(\B{c})})_{i\in\C{I}}\big)_{n=1}^{N_{\B{c}}}$, and for simplicity, assume that no values are missing.
The number of samples in each context, given by $N_{\B{c}}$, is allowed to depend on the context. 
As a first step, we \emph{pool} the data, thereby representing it as a single data set $\C{D} = (\B{x}_n,\B{c}_n)_{n=1}^N$ where $N = \sum_{\B{c}\in\BC{C}} N_{\B{c}}$. We then assume that $\C{D}$ is an i.i.d.\ sample of $\Prb_{\C{M}}(\B{X},\B{C})$, where $(\B{X},\B{C},\B{E})$ is a solution of the SCM $\C{M}$ of the form \eref{eq:SCM_JCI_ass}.\footnote{Although
this may sound as an innocuous assumption, it is not necessarily satisfied by the data generating process. For example, suppose 
that in a randomized controlled trial, it is decided \emph{a priori} that a certain number $N_0$ of patients will be assigned to the control group, and a number $N_1$ of patients to the treatment group, but which patients end up in which group is completely randomized. The resulting pooled data is not i.i.d.; indeed, if we repeat this procedure, we will always end up with the same number of patients in each group, whereas for an i.i.d.\ sample, the numbers would fluctuate around their expected values. Nevertheless, this assumption can be made here without losing much generality. In particular, for the case of binary treatment and binary outcome, \citet[Section~15.5]{Wasserman2004} shows that for independence tests based on the (log) odds ratio the i.i.d.\ assumption can be weakened accordingly. Alternatively, in a bootstrapping procedure (which we will use in practice for most implementations of JCI in Section~\ref{sec:experiments}), the resampled pooled data is i.i.d.\ by construction.\label{footnote:iid}}

In setting up the problem, we have made the simplifying assumptions that the measurement procedure is not subject to selection 
bias, nor to (independent) measurement error \citep{Blom++_UAI_18}.
We will assume that the data has been generated by an SCM in accordance with JCI Assumption~\ref{ass:simple_scm}, and optionally, a subset of JCI Assumptions \ref{ass:uncaused}, \ref{ass:unconfounded} and \ref{ass:dependences}.
To enable constraint-based causal discovery, we will assume that the joint distribution $\Prb_{\C{M}}(\B{X},\B{C})$ is faithful with respect to the graph $\C{G}(\C{M})$, using the appropriate separation criterion ($\sigma$-separation in general, or $d$-separation for specific cases, as discussed in Section~\ref{sec:faithfulness}). We will discuss the ramifications of the faithfulness assumption in more detail in Section~\ref{sec:faithfulness}. 

\begin{definition}
  We say that a particular feature of $\C{G}(\C{M})$ is \emph{identifiable} from $\Prb_{\C{M}}(\B{X},\B{C})$ and background knowledge if the feature is present in the graph $\C{G}(\tilde{\C{M}})$ of any SCM $\tilde{\C{M}}$ with $\Prb_{\tilde{\C{M}}}(\B{X},\B{C}) = \Prb_{\C{M}}(\B{X},\B{C})$ that incorporates the background knowledge.
\end{definition}
``Feature'' could refer to the presence or absence of a direct edge, a directed path, a bidirected edge, arbitrary subgraphs, or even the complete graph. The task of causal discovery is then to identify as many features of $\C{G}(\C{M})$ as possible from the data, the i.i.d.\ sample $\C{D}$ of $\Prb_{\C{M}}(\B{X},\B{C})$, and the available background knowledge.

The key insight of the Joint Causal Inference framework that allows one to deal with data from multiple contexts $(\C{D}^{(\B{c})})_{\B{c}\in\BC{C}}$ is that by
incorporating the context variables explicitly, and pooling the data, we have now reduced the causal discovery problem
to one that is mathematically equivalent to causal discovery from purely observational data $\C{D}$ and applicable 
background knowledge on the causal relations between context and system variables (a subset of JCI Assumptions~\ref{ass:uncaused} 
and ~\ref{ass:unconfounded}). If applicable, JCI Assumption~\ref{ass:dependences} can be made to reduce the computational effort.

This trick also allows us to easily learn intervention targets from data in a similar way as
we usually learn causal effects between variables from data: the intervention targets are
simply encoded as the direct effects of the intervention variables.

After discussing the faithfulness assumption in more detail, we will give a few suggestions of how JCI can be implemented in Section~\ref{sec:JCI_implementations}.

\subsection{Faithfulness Assumption}\label{sec:faithfulness}

In this subsection we will discuss the subtleties of the faithfulness assumption in the JCI setting and compare
it with alternative faithfulness assumptions that have been made in the literature.

Given a simple SCM $\C{M}$ of the form \eref{eq:SCM_JCI_ass} with graph $\C{G} := \C{G}(\C{M})$, the joint distribution $\Prb_{\C{M}}(\B{X},\B{C})$ induced by the SCM satisfies the Generalized Directed Global Markov Property (Theorem~\ref{thm:sigma_separation}) with respect to the graph $\C{G}$ of the SCM, i.e., any $\sigma$-separation $\sigmasep{U}{V}{W}{\C{G}}$ between sets of nodes $U,V,W \subseteq \C{I}\cup\C{K}$ in the graph $\C{G}$ implies a conditional independence $\indep{\tilde{\B{X}}_U}{\tilde{\B{X}}_V}{\tilde{\B{X}}_W}{\Prb_{\C{M}}(\B{X},\B{C})}$, where we write $\tilde{\B{X}} := (\B{X},\B{C})$. Under the additional assumptions of
Theorem~\ref{thm:d_separation}, the stronger Directed Global Markov Property holds (i.e., the $d$-separation criterion).

For constraint-based causal discovery, some type of faithfulness assumption is usually made. For simplicity, the faithfulness assumption that we make in this work is the standard one, but we apply it to the combination of system and its environment: we assume that the joint distribution $\Prb_{\C{M}}(\B{X},\B{C})$ is faithful with respect to the graph $\C{G}(\C{M})$ of $\C{M}$. In other words, any conditional independence $\indep{\tilde{\B{X}}_U}{\tilde{\B{X}}_V}{\tilde{\B{X}}_W}{\Prb_{\C{M}}(\B{X},\B{C})}$ for sets of nodes $U,V,W \subseteq \C{I}\cup\C{K}$ is due to the $\sigma$-separation $\sigmasep{U}{V}{W}{\C{G}}$ in $\C{G}$ (or $d$-separation, if applicable), and no other conditional independences in $\Prb_{\C{M}}(\B{X},\B{C})$ exist.
In particular, this assumption rules out any conditional independence between context variables in case JCI Assumption~\ref{ass:dependences} is made. If JCI Assumption~\ref{ass:dependences} is not made, conditional independences in the context distribution that are faithfully described by any DMG are allowed.

This faithfulness assumption allows us to deal with different types of
interventions, including perfect interventions.  For example, for
the perfect intervention on $X_2$ illustrated in
Figure~\ref{fig:JCI_example} the causal graphs $\C{G}_{\C{I}}^{(\B{c})}$
(restricted to the system variables $\{X_i\}_{i\in\C{I}}$) depend on the context
$\B{c}\in\BC{C}$: in the observational
context ($C_\alpha=0$), $X_1\to X_2$, whereas in the interventional context
($C_\alpha=1$), this direct causal relation is no longer present (as it has been
overruled by the perfect intervention). This does not invalidate the
faithfulness of the joint distribution $\Prb(C_\alpha,X_1,X_2,X_3)$ with respect to
the joint causal graph. Indeed, even though $X_1 \CI X_2 \given C_\alpha=0$, we
still have $X_1 \nCI X_2 \given C_\alpha$ because $X_1 \nCI X_2 \given
C_\alpha=1$.\footnote{Note that for a discrete context domain $\BC{C}$, we have that
$A \CI B \given \B{C}$ if and only if $A \CI B \given \B{C}=\B{c}$ for all
$\B{c}$ with $p(\B{c}) > 0$. More generally, $A \CI B \given \B{C}$ if and only if
$A \CI B \given \B{C}=\B{c}$ for almost all $\B{c}$.} In other words, the fact that $\Prb(\B{X} \given
C_\alpha=1)$ is \emph{not} faithful to the system subgraph $\C{G}_{\C{I}}$ (i.e.,
the induced subgraph of the causal graph $\C{G}$ on the system nodes $\C{I}$)
does not lead to any problem as long as we are not going to test for
independences in the subset of data corresponding to context $C_\alpha=1$ separately, but restrict
ourselves to testing independences only in the \emph{pooled} data set that combines all 
contexts.

Causal discovery methods that analyze data from each context separately \citep[e.g.,][]{GIES2012,triantafillou2015constraint,HEJ2014} typically make another faithfulness assumption.
In our notation, such approaches assume that $\Prb(\B{X} \given \B{C}=\B{c})$ is faithful w.r.t.\ a causal subgraph $\C{G}_{\C{I}}^{(\B{c})}$ that may be context-dependent, and must then reason about how these context-dependent subgraphs are related, explicitly relying on knowledge about the type of interventions (typically assuming that the interventions are perfect interventions with known targets).
This faithfulness assumption is to a certain extent stronger than ours because it requires faithfulness of the system within \emph{each} context.
On the other hand, it is to a certain extent weaker than ours because ours implies restrictions on the context distribution ($\Prb_{\C{M}}(\B{C})$ must be faithful to some DMG, at the very least) that the alternative does not have.
We consider the extension of the applicability of JCI (because no knowledge of intervention types or targets is needed) due to the faithfulness assumption we chose here to outweigh the limitations in applicability (because not \emph{every} context distribution can be handled). In the rest of this section, we will discuss some simple workarounds that can be applied in practice when dealing with faithfulness violations in the context distribution $\Prb_{\C{M}}(\B{C})$.

Under JCI Assumption~\ref{ass:dependences}, the faithfulness assumption implies that the
context distribution $\Prb(\B{C})$ does not contain any conditional
independences. In case the empirical context distribution $\hat\Prb(\B{C})$
\emph{does} contain conditional independences, one has several options. The
first (assuming also JCI Assumptions \ref{ass:uncaused} and \ref{ass:unconfounded} are made) 
is to modify the assumed graph of the context variables in JCI
Assumption~\ref{ass:dependences} such that the context distribution is faithful
to it. For example, if all context variables are jointly independent, one could simply
assume that the context graph $\C{G}(\C{M})_{\C{K}}$ has no (directed or bidirected)
edges at all. The second option is to omit JCI Assumption~\ref{ass:dependences} or some
analogue of it completely.
In that case, the faithfulness assumption still imposes the restriction that the
context distribution can be faithfully modeled by \emph{some} DMG. The third option 
is to use only data corresponding to a certain subset of
the contexts and to ignore data from other contexts. In addition, one can
sometimes work around conditional independences in the context distribution by partitioning the set of
context variables into groups of context variables, and using combined context
variables instead of the original context variables. This will be
illustrated in the next paragraph. Finally, one could ignore the faithfulness
violations in the context distribution and hope that the causal discovery
algorithm will handle them well. This last approach usually means that it 
will be harder to guarantee consistency of the approach.
Note that the faithfulness assumption for the context variables is actually testable, 
since the empirical context distribution is available, and can be directly tested for conditional independences.

The faithfulness assumption also rules out deterministic 
relations between the variables that lead to faithfulness violations. In particular, there could be 
deterministic relations between context variables.
For example, in the experimental design of the experiments in \citet{SPP05}
described in Tables~\ref{tab:sachs_experimental_design} and \ref{tab:sachs_contexts}, $C_\alpha$ is a (deterministic) function of $C_\theta$ and $C_\iota$:
$C_\alpha = \lnot (C_\theta \lor C_\iota)$. One might na\"ively believe that this could be dealt with by simply removing
context variable $C_\alpha$ from consideration, leaving only context variables $C_\beta,\dots,C_\iota$ as observed context variables, 
none of which is a (deterministic) function of the others. However, marginalizing out context variables may give
rise to violations of JCI Assumption~\ref{ass:unconfounded}, as we have seen in Section~\ref{sec:ass_unconfounded}.
An operation that is generally allowed is \emph{grouping} context variables together. In the case of the \citet{SPP05} experimental
design, we can combine
$C_\alpha$, $C_\theta$ and $C_\iota$ together into a single context variable given by the triple $(C_\alpha,C_\theta,C_\iota)$. Accidentally,
in this case this would be mathematically equivalent to the pair $(C_\theta,C_\iota)$, 
but the interpretation of $(C_\alpha,C_\theta,C_\iota)$ is different from that of $(C_\theta,C_\iota)$. Another option would be
to ignore a subset of the contexts. In this case, one could exclude the two contexts with $C_\theta=1$ or $C_\iota=1$. Then, 
$C_\alpha$ becomes a constant, and constants can be safely ignored (or, trivially combined with any other context variable).\footnote{In an earlier draft of this work \citep{Magliacane++_1611.10351v1}, we proposed to handle deterministic relations between context variables by using 
the notion of $D$-separation, first presented in \citet{Geiger1990} and later extended in \citet{SGS2000}. However,
this notion does not provide a complete characterization of conditional independences due to a combination of graph
structure and deterministic relations. Therefore, in this work we use the simpler techniques of grouping context variables
and ignoring certain contexts to deal with faithfulness violations due to deterministic relations between context variables.}

To wrap up: under JCI Assumption~\ref{ass:dependences}, it is advisable to check
that there are indeed no conditional independences between context variables in the
empirical context distribution. More generally, one should check whether the
conditional independences in the context distribution can be faithfully described
by a DMG. If not, one can try to work around by applying the tricks discussed in 
the last paragraph (grouping context variables, and omitting certain contexts). 
When grouping context variables, one should note that the inferred causal relations from
context variables to system variables may no longer be easily interpretable. A
simple example that illustrates this is to consider two interventions that are
always performed together: when drug A is prescribed, also drug B is
prescribed, and vice versa. In that case we cannot be sure whether the effect
on outcome is due to drug A or to drug B (or to both combined). Nonetheless, we can still use the
inferred causal relations between system variables.

\begin{table}[t!]
  \centering
\tiny
  \begin{tabular}{|ccccccccc|c|}
\hline
  $C_\alpha$ & $C_\beta$ & $C_\gamma$ & $C_\delta$ & $C_\epsilon$ & $C_\zeta$ & $C_\eta$ & $C_\theta$ & $C_\iota$ & $N_{\B{c}}$ \\
\hline
  1 & 0 & 0 & 0 & 0 & 0 & 0 & 0 & 0 & 853 \\
  1 & 1 & 0 & 0 & 0 & 0 & 0 & 0 & 0 & 902 \\
  1 & 0 & 1 & 0 & 0 & 0 & 0 & 0 & 0 & 911 \\
  1 & 0 & 0 & 1 & 0 & 0 & 0 & 0 & 0 & 723 \\
  1 & 0 & 0 & 0 & 1 & 0 & 0 & 0 & 0 & 810 \\
  1 & 0 & 0 & 0 & 0 & 1 & 0 & 0 & 0 & 799 \\
  1 & 0 & 0 & 0 & 0 & 0 & 1 & 0 & 0 & 848 \\
  0 & 0 & 0 & 0 & 0 & 0 & 0 & 1 & 0 & 913 \\
  0 & 0 & 0 & 0 & 0 & 0 & 0 & 0 & 1 & 707 \\
  1 & 1 & 1 & 0 & 0 & 0 & 0 & 0 & 0 & 899 \\
  1 & 1 & 0 & 1 & 0 & 0 & 0 & 0 & 0 & 753 \\
  1 & 1 & 0 & 0 & 1 & 0 & 0 & 0 & 0 & 868 \\
  1 & 1 & 0 & 0 & 0 & 1 & 0 & 0 & 0 & 759 \\
  1 & 1 & 0 & 0 & 0 & 0 & 1 & 0 & 0 & 927 \\
\hline
\end{tabular}\hfill
\begin{tabular}{|ccccccc|c|}
\hline
  $C_\beta$ & $C_\gamma$ & $C_\delta$ & $C_\epsilon$ & $C_\zeta$ & $C_\eta$ & $(C_\alpha,C_\theta,C_\iota)$ & $N_{\B{C}}$ \\
\hline
  0 & 0 & 0 & 0 & 0 & 0 & (1,0,0) & 853 \\
  1 & 0 & 0 & 0 & 0 & 0 & (1,0,0) & 902 \\
  0 & 1 & 0 & 0 & 0 & 0 & (1,0,0) & 911 \\
  0 & 0 & 1 & 0 & 0 & 0 & (1,0,0) & 723 \\
  0 & 0 & 0 & 1 & 0 & 0 & (1,0,0) & 810 \\
  0 & 0 & 0 & 0 & 1 & 0 & (1,0,0) & 799 \\
  0 & 0 & 0 & 0 & 0 & 1 & (1,0,0) & 848 \\
  0 & 0 & 0 & 0 & 0 & 0 & (0,1,0) & 913 \\
  0 & 0 & 0 & 0 & 0 & 0 & (0,0,1) & 707 \\
  1 & 1 & 0 & 0 & 0 & 0 & (1,0,0) & 899 \\
  1 & 0 & 1 & 0 & 0 & 0 & (1,0,0) & 753 \\
  1 & 0 & 0 & 1 & 0 & 0 & (1,0,0) & 868 \\
  1 & 0 & 0 & 0 & 1 & 0 & (1,0,0) & 759 \\
  1 & 0 & 0 & 0 & 0 & 1 & (1,0,0) & 927 \\
\hline
\end{tabular}
\caption{Left: Experimental design used by \citet{SPP05}. $N_{\B{c}}$ is the number of data samples in context $\B{C}$. Interpretation of context variables is provided in Table~\ref{tab:sachs_contexts}. Right: Different choice of context variables: $C_\alpha$, $C_\theta$ and $C_\iota$ have been grouped together into a single combined context variable $(C_\alpha,C_\theta,C_\iota)$ in order to deal with the deterministic relation $C_\alpha = \lnot (C_\theta \lor C_\iota)$ (see main text for details).
\label{tab:sachs_experimental_design}}
\end{table}
\begin{table}[t!]
  \centering
  \begin{tabular}{l|ll}
     & Reagent & Intervention \\
    \hline
    $C_\alpha$ & $\alpha$-CD3, $\alpha$-CD28 & global activator \\
    $C_\beta$ & ICAM-2 & global activator \\
    $C_\gamma$ & AKT inhibitor & activity of AKT \\
    $C_\delta$ & G0076 & activity of PKC \\
    $C_\epsilon$ & Psitectorigenin & abundance of PIP2 \\
    $C_\zeta$ & U0126 & MEK activity \\
    $C_\eta$ & LY294002 & PIP2, PIP3 mechanism change \\
    $C_\theta$ & PMA & PKC activity \\
    $C_\iota$ & $\beta$2CAMP & PKA activity \\
  \end{tabular}
  \caption{For each context variable in Table~\ref{tab:sachs_experimental_design}: reagents used in this experimental setting, and expected intervention type and targets as based on (our interpretation of) biological background knowledge described in \citet{SPP05}.\label{tab:sachs_contexts}}
\end{table}

\subsection{Implementing JCI}\label{sec:JCI_implementations}

Any causal discovery method that is applicable under the assumptions described in Section~\ref{sec:causal_discovery}
can be used for Joint Causal Inference. In this section we will describe some concrete examples of JCI implementations.
Identifiability may greatly benefit from taking into account the available background knowledge on the
causal graph stemming from the applicable JCI assumptions as discussed in Section~\ref{sec:JCI_assumptions}. 
In addition, taking into account background
knowledge on targets of intervention variables may help considerably.
Some logic-based causal discovery methods \citep[e.g.,][]{HEJ2014,triantafillou2015constraint,ForreMooij_UAI_18}, are ideally suited to exploit such background knowledge.
For other methods, e.g., FCI \citep{SMR1999,Zhang2008_AI}, or methods that focus on ancestral relations, e.g., ACI \citep{MagliacaneClaassenMooij_NIPS_16}, incorporating all background knowledge is less straightforward and as far as we know cannot be done with off-the-shelf implementations. Often, simple adaptations of off-the-shelf implementations can be made that do allow one to benefit from all JCI background knowledge. For example, in Section~\ref{sec:FCI-JCI} we will propose a simple adaptation of FCI that does so.

Given a causal discovery algorithm for purely observational data that can exploit the JCI background
knowledge, we can implement JCI in a straightforward fashion:
\begin{enumerate}[(i)]
  \item introduce context variables, if not already provided;
  \item pool all data sets, including the values of the context variables; 
  \item handle faithfulness violations between context variables by grouping context variables and/or leaving out certain contexts, if necessary;
  \item apply the causal discovery algorithm on the pooled data, taking into account the appropriate JCI background knowledge. 
\end{enumerate}
Any soundness, completeness and consistency results for the
causal discovery algorithm that hold for the algorithm in the purely observational setting, but
including the background knowledge, directly apply to the JCI setting, as long as there is no model 
misspecification (i.e., if the assumed JCI assumptions do hold for the true model). 
If we use JCI Assumption 3 (or something similar), we can use Theorem~\ref{thm:replace_context}
to show that what the algorithm concludes about the causal relations concerning system variables
is still correct.

In the remainder of this subsection, we discuss four JCI implementations. We will first describe
two existing algorithms (LCD and ICP) that can be used off-the-shelf for causal discovery 
within the JCI framework. Then we propose two adaptations of existing algorithms (ASD and FCI)
so that they can be used for causal discovery within the JCI framework.
In Section~\ref{sec:experiments}, we will provide empirical results on the properties of those
four algorithms.

\subsubsection{Local Causal Discovery (LCD)}

Perhaps the first implementation of JCI (apart from randomized controlled trials) is provided by the LCD algorithm by \citet{Cooper1997}.
LCD is a very simple constraint-based causal discovery algorithm that can be used
for the purely observational causal discovery setting where certain background knowledge is available,
and in particular, in the JCI setting. The basic idea behind the LCD algorithm is the following result
(which we generalized to allow for cycles):
\begin{proposition}\label{prop:LCD}
  Suppose that the data-generating process on three variables $X_1,X_2,X_3$ can be represented by a faithful,
  simple SCM $\C{M}$ with $\C{I} = \{1,2,3\}$ and that the sampling procedure is not subject to selection bias.
  If $X_2$ is not a cause of $X_1$ according to $\C{M}$, 
  the following conditional (in)dependencies in the observational distribution $\Prb_{\C{M}}(X_1,X_2,X_3)$
  $$X_1 \nCI X_2, \quad X_2 \nCI X_3, \quad X_1 \CI X_3 \given X_2$$
  imply that the graph $\C{G}(\C{M})$ must be one of the three DMGs in Figure~\ref{fig:LCD}. In particular,
  \begin{compactenum}[(i)]
  \item $X_3$ is not a cause of $X_2$ according to $\C{M}$;
  \item $X_2$ is a direct cause of $X_3$ according to $\C{M}$;
  \item $X_2$ and $X_3$ are not confounded according to $\C{M}$;
  \item the causal effect of $X_2$ on $X_3$ is given by:
    \begin{equation}\label{eq:LCD}
      \Prb_{\C{M}}\big(X_3 \given \intervene(X_2=x_2)\big) = \Prb_{\C{M}}(X_3 \given X_2=x_2).
    \end{equation}
  \end{compactenum}
\end{proposition}
\begin{proof}
  The proof proceeds by enumerating all (possibly cyclic) DMGs on three variables and ruling out the ones that do not
satisfy the assumptions.
The assumption that $X_2$ is not a cause of $X_1$ implies that there is no directed edge
$X_2 \to X_1$ in the graph $\C{G}(\C{M})$. If there were an edge between $X_1$ and $X_3$,
$X_1 \CI X_3 \given X_2$ would not hold (faithfulness). Also, since $X_1 \nCI X_2$, $X_1$ and
$X_2$ must be adjacent (Markov property). Similarly, $X_2$ and $X_3$ must be adjacent. $X_2$ cannot be
a collider on any path between $X_1$ and $X_3$ (faithfulness). Since the only possible edges between
$X_1$ and $X_2$ are $X_1 \to X_2$ and $X_1 \oto X_2$ (both of which have an arrowhead at $X_2$), 
this means that there must be a
directed edge $X_2 \to X_3$, but there cannot be a  bidirected edge $X_2 \oto X_3$ or directed edge $X_2 \ot X_3$. 
In other words, the only three possible graphs are the ones in Figure~\ref{fig:LCD}.
The causal do-calculus applied to $\C{G}(\C{M})$ yields \eref{eq:LCD}.
\end{proof}

In a JCI setting where JCI Assumption~\ref{ass:uncaused} is made, we can directly apply LCD
for causal discovery on tuples $\langle C_k, X_i, X_{i'} \rangle$ with $k \in \C{K}$,
$i \ne i' \in \C{I}$ on the pooled data.

A conservative version of LCD has been applied by \citet{Triantafillou++2017} to the task of
inferring signaling networks from mass-cytometry data. 
A high-dimensional version of LCD has been shown to be successful in predicting the
effects of gene knockouts on gene expression levels \citep{VersteegMooij_1910.02505v2}
from large-scale interventional yeast gene expression data \citep{Kemmeren2014}.
An algorithm closely related to LCD, named ``Trigger'', has been applied on genomics data \citep{Trigger2007}.
\citet{Trigger2007} motivate the JCI assumptions in the setting of learning the causal relations between gene expression levels
using single nucleotide polymorphisms (SNPs) as context variables. Since the DNA content cannot be caused by gene expression
levels, JCI Assumption~\ref{ass:uncaused} is satisfied. \citet{Trigger2007} then argue that Mendelian randomization 
justifies JCI Assumption~\ref{ass:unconfounded}. Finally, a single conditional independence in the pooled
data (as in LCD) provides the desired evidence for an unconfounded causal relation between
two gene expression levels. 

\begin{figure}
  \centering
  \begin{tikzpicture}
    \begin{scope}
      \node[var] (X1) at (-3,0) {$X_1$};
      \node[var] (X2) at (-1.5,0) {$X_2$};
      \node[var] (X3) at (0,0) {$X_3$};
      \draw[arr] (X1) edge (X2);
      \draw[arr] (X2) edge (X3);
    \end{scope}
    \begin{scope}[xshift=5cm]
      \node[var] (X1) at (-3,0) {$X_1$};
      \node[var] (X2) at (-1.5,0) {$X_2$};
      \node[var] (X3) at (0,0) {$X_3$};
      \draw[biarr,bend left] (X1) edge (X2);
      \draw[arr] (X2) edge (X3);
    \end{scope}
    \begin{scope}[xshift=10cm]
      \node[var] (X1) at (-3,0) {$X_1$};
      \node[var] (X2) at (-1.5,0) {$X_2$};
      \node[var] (X3) at (0,0) {$X_3$};
      \draw[arr] (X1) edge (X2);
      \draw[biarr,bend left] (X1) edge (X2);
      \draw[arr] (X2) edge (X3);
    \end{scope}
  \end{tikzpicture}
  \caption{All possible DMGs detected by LCD.\label{fig:LCD}}
\end{figure}

\subsubsection{Invariant Causal Prediction (ICP)}
ICP exploits invariance of the
conditional distribution of a target variable given its direct causes across multiple contexts,
assuming that none of the contexts corresponds with an intervention that targets the target
variable \citep{ICP2016}.
The implementation described in \citet{ICP2016} handles linear relationships, arbitrary interventions
(as long as they do not change the conditional distribution of the effect variable given its direct
causes), assumes the absence of latent confounders between target variable and its direct causes, and
the absence of cycles involving the target variable.
One of the main advantages of this method over others is that it provides (conservative) confidence intervals on direct causal relationships that do not require the faithfulness assumption to be made; however, that only works under
the assumption of causal sufficiency and acyclicity.
The authors discuss several possible extensions to broaden the scope of the method, but do not address this in all generality. 
A nonlinear extension of the method has been proposed recently \citep{Heinze-Deml++2018}.
ICP has been successfully applied to predict the effects of gene knockouts on gene expression levels \citep{Meinshausen++_PNAS_16} from large-scale interventional yeast gene expression data \citep{Kemmeren2014}.

ICP can be interpreted as a particular implementation of the JCI framework,
even in the general setting with nonlinear relations between variables and with latent confounders
and cycles present (although faithfulness is then required).
The following result broadens the conditions under which ICP identifies (possibly indirect) causal relations,
strengthening results of \citet{ICP2016}:
\begin{corollary}\label{cor:ICP_as_JCI}
Consider the JCI setting with a single context variable $C$ and multiple system variables
$\{X_i\}_{i\in\C{I}}$. Under JCI Assumptions~\ref{ass:simple_scm} and \ref{ass:uncaused} and faithfulness,
the ICP estimator for target $i \in \C{I}$:
$$J_i^* := \bigcap \{I \subseteq \C{I} \setminus \{i\}: \indep{C}{X_i}{\B{X}_I}{\Prb_{\C{M}}(C,\B{X})} \}$$
satisfies $J_i^* \subseteq \ansub{\C{G}(\C{M})}{i}$, i.e., the set $J_i^*$ consists only of (possibly indirect) causes of $i$.
\end{corollary}
\begin{proof}
  Follows immediately from Proposition~\ref{prop:intersection_sepsets} in Appendix~\ref{app:mci}.
\end{proof}
Note that this means that asymptotically, ICP outputs a subset of ancestors of the target variable, 
even in the presence of confounders and linear or nonlinear cycles.
Hence, ICP can be interpreted as a particular causal discovery algorithm implementing the JCI framework.

\subsubsection{ASD-JCI}

Here we introduce a novel JCI implementation that builds on the algorithm by \citet{HEJ2014} and its generalization to $\sigma$-separation \citep{ForreMooij_UAI_18} and some of the extensions proposed by \citet{MagliacaneClaassenMooij_NIPS_16}. 
Since adapting this algorithm to the JCI setting is straightforward, and the algorithm itself has been described in detail in the cited papers, we here only provide a brief description of how it works. 
More details can be found in Appendix~\ref{sec:app:asd}.

\citet{HEJ2014} proposed formulating causal discovery as an optimization problem over possible causal graphs, where
the loss function sums the weights of all the conditional (in)dependencies present in the data that would
be violated for a certain underlying causal graph, assuming Markov and faithfulness properties.
The input consists of a list of weighted conditional independence statements. The weights $\lambda$ encode the
confidence in the conditional (in)dependence, where a weight of $\lambda=\infty$ corresponds to a ``hard
constraint'' (absolute certainty) and a weight of $\lambda=0$ corresponds to ``no evidence at all''.
\citet{HEJ2014} provide an encoding of the notion of $d$-separation in Answer Set Programming (ASP), a declarative 
programming language that can be used amongst others for solving discrete optimization problems.
\citet{ForreMooij_UAI_18} generalize the encoding to $\sigma$-separation.
The optimization problem is solved by making use of an off-the-shelf ASP solver.

There may be multiple optimal solutions to the optimization problem, because the underlying causal graph may not be identifiable from the inputs. 
Nonetheless, some of the features of the causal graph (e.g., the presence or absence of a certain directed edge) may still be identifiable.
We employ the method proposed by \citet{MagliacaneClaassenMooij_NIPS_16} for scoring the confidence that a certain feature is present or absent by calculating the difference between the optimal losses under the additional hard constraints that the feature is present vs.\ that the feature is absent.
\citet{MagliacaneClaassenMooij_NIPS_16} showed that this algorithm for scoring features is sound for oracle inputs and asymptotically consistent under reasonable assumptions.

We will make use of the weights proposed in \citet{MagliacaneClaassenMooij_NIPS_16}:
$\lambda_j = \log p_j - \log \alpha$, where $p_j$ is the $p$-value of a statistical test for the
$j^\mathrm{th}$ conditional independence statement, with independence as null hypothesis, 
and $\alpha$ is a significance level (e.g., 1\%) that should decrease with sample size at a suitable rate.
These weights have the desirable property that independences get a lower weight
than strong dependencies.

As we will need an acronym for this algorithm later, we will henceforth refer to it as ASD (Accounting for Strong Dependencies),
as it essentially tries to explain the observed dependencies in the data, taking into account the statistical strength of these dependencies.
This is fundamentally different from other constraint-based algorithms such as PC or FCI, which give priority to
observed \emph{independences} and do not take into account the strength of dependencies.

Taking into account the JCI background knowledge (a subset of JCI Assumptions~\ref{ass:uncaused}, \ref{ass:unconfounded}
and \ref{ass:dependences}), and possible background knowledge on intervention targets, 
is trivial thanks to the expressive power of ASP, and can be 
done with a few lines of ASP code. The resulting algorithm is very accurate but scales only
up to a few variables due to the combinatoric explosion. 
Incorporating JCI Assumption~\ref{ass:dependences} considerably reduces computation time, as
it removes the need to learn the causal relations of the context variables. 

Since the JCI background knowledge can be completely and exactly encoded as constraints on the possible
causal graphs, we can directly extend the known soundness, completeness and consistency results for ASD:
\begin{theorem}\label{eq:ASDJCI_sound_complete_consistent}
  ASD-JCI is sound and complete for oracle inputs. 
  It is asymptotically consistent if the weights are asymptotically consistent.
\end{theorem}
\begin{proof}
  For the precise meaning of these statements, we refer the reader to Appendix~\ref{sec:app:asd}, and in particular, to Theorem~\ref{theo:ASD_sound_complete} and \ref{theo:ASD_consistent}.
\end{proof}

\subsubsection{FCI-JCI}\label{sec:FCI-JCI}

Here we introduce an adaptation of the constraint-based causal discovery algorithm FCI \citep{SMR1999,Zhang2008_AI} that can be used in a JCI setting. The FCI algorithm was designed to work under the assumption that the data was generated by an acyclic SCM.
While FCI can deal with selection bias, we will assume here for simplicity that no selection bias is present.

The FCI algorithm consists of two main phases: an adjacency search phase leading to the skeleton, followed by an edge orientation phase. In the adjacency search the algorithm searches for conditional independences to eliminate edges from the graph. The subsequent orientation stage consists of a set of graphical rules that allow invariant edge marks, signifying either causal (tail marks) or non-causal (arrowhead marks) relations, to be added to the skeleton. For a single observational data set the final result is a so-called Partial Ancestral Graph (PAG) that is a concise representation of ancestral relations and conditional independences. The PAG represents a set of Maximal Ancestral Graphs (MAGs) \citep{RichardsonSpirtes02}, and each MAG represents a set of ADMGs \citep{triantafillou2015constraint}, and each ADMG represents an infinite set of DAGs (with arbitrary number of latent variables). In the case of purely observational data, the PAG output by FCI provides a complete description of the Markov equivalence class \citep{Zhang2008_AI,Ali++2009}. For a more detailed discussion of MAGs and PAGs, we refer the reader to Appendix~\ref{sec:app:preliminaries}.

Extending FCI such that it can take into account the additional JCI background knowledge on the adjacency and causal relations between the combined set of context and system variables (see also Section~\ref{sec:JCI_assumptions}) is straightforward:
\begin{itemize}
\item
If JCI Assumption \ref{ass:dependences} is made, all context variables are connected by bidirected edges, and the adjacency phase of FCI is adapted accordingly by not removing any edges between context variables; afterwards, all edges between context variables are oriented as $k \oto k'$ for $k \ne k' \in \C{K}$. In the subsequent phase of orienting unshielded triples, only system variables can take on the role of the collider. 
\item
If JCI Assumption \ref{ass:uncaused} is made, then (since we are assuming no selection bias) any adjacent pair of a context variable $k\in\C{K}$ and a system variable $i\in\C{I}$ must be connected in the MAG by an edge with an arrowhead at $i$. Therefore, after the adjacency phase, all edges between a context and a system variable are oriented as $k \, \ast \!\!\! \rightarrow i$, with an arrowhead at the system variable $i \in \C{I}$.
\item
If both JCI Assumptions \ref{ass:uncaused} and \ref{ass:unconfounded} are made, any adjacent pair of a context variable $k\in\C{K}$ and a system variable $i\in\C{I}$ must be connected by a directed edge $k\to i$ (since we are assuming no selection bias) in the MAG. Hence, after the adjacency phase, all edges between a context and a system variable are oriented as $k \to i$, pointing from the context variable $k \in \C{K}$ to the system variable $i \in \C{I}$.
\end{itemize}
The subsequent orientation phase of the FCI algorithm does not need to be adapted. 

We will refer to this adaptation of the FCI algorithm as \alg{FCI-JCI}. In particular, 
we distinguish three variants: \alg{FCI-JCI0} (which only makes JCI Assumption~\ref{ass:simple_scm}),
\alg{FCI-JCI1} (also JCI Assumption~\ref{ass:uncaused}), and \alg{FCI-JCI123} (also JCI Assumptions~\JCIABC).
In Appendix~\ref{sec:app:dpag_ancestors} we discuss how one can read off the identified causal and non-causal relations 
from the PAG output by FCI or FCI-JCI. 
Furthermore, in Appendix~\ref{sec:app:fcijci123_intervention_targets} we discuss how one can read off the direct targets and
direct non-targets of interventions represented by context nodes from the PAG output by \alg{FCI-JCI123}.

\subsubsection{Speeding up \alg{FCI-JCI123}\label{sec:fci-jci123r}}

Adding the context nodes may make FCI-JCI considerably slower than FCI on a single context. In this subsection,
we propose a further adaptation of \alg{FCI-JCI123}. By exploiting the following observation, we can achieve a considerable speedup:
\begin{lemma}\label{lemm:restricted_separations_jci12}
  Let $\C{M}$ be an SCM that satisfies JCI Assumptions \ref{ass:simple_scm}, \ref{ass:uncaused}, \ref{ass:unconfounded}. 
  Then for $X \subseteq \C{I}$ and $Y, Z \subseteq \C{I} \cup \C{K}$:
  $$\sep{X}{Y}{Z}{\C{G}(\C{M})} \implies \sep{X}{Y}{Z \cup (\C{K} \setminus Y)}{\C{G}(\C{M})}$$
  (for both d-separation as well as for $\sigma$-separation).
\end{lemma}
\begin{proof}
  By contradiction: Suppose there exists a path $\langle v_0,e_1,v_1,e_2,v_3,\dots,e_{n-1},v_n \rangle$ between $X$ and $Y$ that is open given $Z \cup (\C{K} \setminus Y)$ and contains no non-endpoint nodes in $X \cup Y$, but is closed given $Z$.
  That can only happen if the path contains a collider $v_i$ ($0 < i < n$) that is $\C{G}(\C{M})$-ancestor of $\C{K} \setminus (Y \cup Z)$.
  Then $v_i \in \C{K}$ (because no system node is ancestor of $\C{K}$ by JCI Assumption~\ref{ass:uncaused}).
  Let $1 \le j < i$ be the lowest index such that $v_l \in \C{K}$ for all $l \in \{j, j+1, \dots, i\}$.
  Then $v_{j-1} \ot v_j$ on the path (by JCI Assumptions~\JCIAB), where $v_{j-1}$ is in another strongly-connected component than $v_j$, and therefore $v_j \in \C{K} \setminus Y$ blocks the path, which is a contradiction.
\end{proof}
This implies that in the skeleton search of \alg{FCI-JCI12} and \alg{FCI-JCI123}, the search spaces for finding separating sets between pairs of nodes can be reduced.
Indeed, instead of testing for each subset $B$ of $A \subseteq \C{I} \cup \C{K}$ whether $B$ d-separates node $v$ from node $w$, one can test for each subset $B$ of $A \setminus \C{K}$ whether $B \cup \C{K} \setminus \{v,w\}$ d-separates $v$ from $w$.\footnote{Of course, the power of the conditional independence test may be reduced when conditioning on many variables. However, if the number of contexts in the pooled data is small, one can design conditional independence tests that do not suffer from this problem.}
For the next stages of the FCI algorithm, it does not matter \emph{which} separating set is found (as long as \emph{any} separating set is found if there is one), as follows from \citet[Lemma 3.2.1 and 3.2.2]{Zhang2006}.
This modification to the skeleton phase reduces the worst-case number of conditional independence tests by a factor that is exponential in the number of context variables. 
We will refer to the adapted version of \alg{FCI-JCI123} that implements this modified version of the skeleton search as \alg{FCI-JCI123r}. 

\subsubsection{Soundness, Completeness and Consistency Results for FCI-JCI}

The FCI algorithm was shown to be sound and complete \citep{Zhang2008_AI} for oracle inputs.
In Appendix~\ref{sec:app:fcijci}, we prove:
\begin{theorem}\label{theo:fcijci}
  Let $\C{M}$ be an acyclic SCM that satisfies JCI Assumption~\ref{ass:simple_scm}.
  Assume that its distribution $\Prb_{\C{M}}(\B{X},\B{C})$ is faithful w.r.t.\ the graph $\C{G}(\C{M})$. 
  Then, with input $\Prb_{\C{M}}(\B{X},\B{C})$:
  \begin{itemize}
    \item \alg{FCI-JCI0} is sound and complete;
    \item \alg{FCI-JCI1} is sound if $\C{M}$ also satisfies JCI Assumption~\ref{ass:uncaused};
    \item \alg{FCI-JCI12} is sound if $\C{M}$ also satisfies JCI Assumptions~\ref{ass:uncaused} and \ref{ass:unconfounded};
    \item \alg{FCI-JCI123} is sound and complete if $\C{M}$ also satisfies JCI Assumptions~\JCIABC.
  \end{itemize}
  Here, \emph{sound} means that the output of the algorithm is a DPAG that contains the true $\DMAG(\C{M})$, and \emph{complete} means that all edge marks of the true $\DMAG(\C{M})$ that can be identified from the (conditional) independences in $\Prb_{\C{M}}(\B{X},\B{C})$ and the JCI background knowledge have been oriented in the DPAG output by the algorithm.
\end{theorem}
\begin{proof}
  We refer the reader to Appendix~\ref{sec:app:fcijci}, and in particular to Theorems~\ref{theo:fcijci_sound}, \ref{theo:fcijci0_complete}, and \ref{theo:fcijci123_complete}.
\end{proof}
In practice, one often does not have access to the joint distribution $\Prb_{\C{M}}(\B{X},\B{C})$, but only to a finite sample of it.
In that case we have:
\begin{corollary}
  The FCI variants mentioned in Theorem~\ref{theo:fcijci} are also asymptotically consistent under the assumptions stated in Theorem~\ref{theo:fcijci} if the conditional independence test (including the choice of the threshold to decide between independence and dependence) is consistent.
\end{corollary}
\begin{proof}
  Direct application of Lemma~\ref{lemm:consistency_constraint_based} in Appendix~\ref{sec:app:fcijci}.
\end{proof}

\subsection{Related Work}\label{sec:related_work}

In this section we provide a more detailed comparison with related work.
Since the pioneering work by \citet{Fisher1935}, many different causal discovery methods that can deal with data from different
contexts have been proposed. Table~\ref{tab:related_work} provides an overview of some of these methods and the features 
they offer. Note that JCI offers most features of all methods. 
By implementing the JCI framework using sophisticated causal discovery methods for observational data (plus background knowledge)
one obtains versatile and powerful causal discovery algorithms for multiple contexts. We will now discuss in detail some of
the aspects of the related work.

\begin{table}\small
\begin{tabular}{lcccccccccccccccc}
  & \rotatebox[origin=l]{90}{Latent confounders} & \rotatebox[origin=l]{90}{Nonlinear mechanisms} & \rotatebox[origin=l]{90}{Cycles} & \rotatebox[origin=l]{90}{Perfect interventions} & \rotatebox[origin=l]{90}{Mechanism changes} & \rotatebox[origin=l]{90}{Activity interventions} & \rotatebox[origin=l]{90}{Other context changes} & \rotatebox[origin=l]{90}{Unknown intervention/context targets} & \rotatebox[origin=l]{90}{Learns intervention/context targets} & \rotatebox[origin=l]{90}{Global causal discovery} & \rotatebox[origin=l]{90}{Different variables in each context} & \rotatebox[origin=l]{90}{Combination strategy} \\ 
  \hline
  \citep{Fisher1935}                      & +     &     + & +     & + & + & + & +     & + & + & - & -     & b \\
  LCD \citep{Cooper1997}                  & +     &     + & +     & + & + & + & +     & + & - & - & -     & b \\
  \citep{CooperYoo1999}                   & -     &     + & -     & + & - & - & -     & - & - & + & -     & b \\
  \citep{TianPearl2001}                   & -     &     + & -     & - & + & - & +     & - & - & + & -     & b \\
  \citep{SPP05}                           & -     &     + & -     & + & - & - & -     & - & - & + & -     & b \\
  \citep{EatonMurphy07}                   & -     &     + & -     & + & + & + & +     & + & + & + & -     & b \\
  Trigger \citep{Trigger2007}             & +     &     + & +     & + & + & + & +     & + & - & - & -     & b \\
  \citep{Claassen++_NIPS2010}             & +     &     + & -     & - & + & + & +     & + & - & + & +     & a \\
  \citep{IOD2011}                         & +     &     + & -     & + & + & + & +     & + & - & + & +     & a \\
  \citep{GIES2012}                        & -     &     + & -     & + & - & - & -     & - & - & + & -     & b \\
  \citep{Hyttinen++2012}                  & +     &     - & +     & + & - & - & -     & - & - & + & -     & a \\
  \citep{MooijHeskes_UAI_13}              & -     & $\pm$ & $\pm$ & + & + & + & +     & - & - & + & -     & b \\
  \citep{HEJ2014}                         & +     &     + & $\pm$ & + & - & - & -     & - & - & + & +     & a \\
  \citep{triantafillou2015constraint}     & +     &     + & -     & + & - & - & -     & - & - & + & +     & a \\
  \citep{Rothenhausler++2015}             & +     &     - & $\pm$ & - & - & - & +     & + & + & + & -     & a \\
  \citep{oates2016estimating}             & -     &     - & -     & - & - & - & +     & - & - & + & -     & b \\
  ICP \citep{ICP2016}                     & +     &     + & +     & + & + & + & +     & + & - & - & -     & b \\
  \citep{Zhang++_IJCAI17}                 & -     &     + & -     & + & + & + & +     & + & + & + & -     & b \\
  \citep{YangKatcoffUhler2018}            & -     &     + & -     & + & + & - & -     & - & - & + & -     & a/b \\
  \citep{ForreMooij_UAI_18}               & +     &     + & +     & + & - & - & -     & - & - & + & +     & a \\
  \hline
  Joint Causal Inference (this work)      & +     &     + & +     & + & + & + & +     & + & + & + & -     & b \\
  FCI-JCI (this work)                     & +     &     + & ?     & + & + & + & +     & + & + & + & -     & b \\
  ASD-JCI (this work)                     & +     &     + & +     & + & + & + & +     & + & + & + & -     & b \\
  \hline
\end{tabular}
  \caption{Overview of causal discovery methods that can combine data from multiple contexts. Features offered by the original implementations of these methods are indicated. When a feature is offered only under additional restrictive assumptions, it is indicated with a $\pm$ sign. Combination strategies (right-most column) are: (a) obtain statistics or constraints from each context separately and then construct a single causal graph based on the combined statistics, (b) pool all data and construct a single causal graph directly from the pooled data.\label{tab:related_work}}
\end{table}

\subsubsection{Latent Confounders}
Most score-based methods that combine multiple contexts 
\citep[like the ones by][]{CooperYoo1999,TianPearl2001,SPP05,EatonMurphy07,GIES2012,MooijHeskes_UAI_13,oates2016estimating} 
and some constraint-based methods \citep{Zhang++_IJCAI17,YangKatcoffUhler2018} assume \emph{causal sufficiency}, i.e., 
that no latent confounders are present. This simplifies the causal discovery 
problem considerably, but the assumption is likely violated in practice and may lead to wrong conclusions.
This is well-known for causal discovery from a single observational data set, but also applies to the JCI setting.

\subsubsection{Cycles}

As we have seen in Proposition~\ref{prop:RCT}, the method by \citet{Fisher1935} can handle cycles. Less well-known is that also
LCD \citep{Cooper1997} and Trigger \citep{Trigger2007} can handle cycles (see Proposition~\ref{prop:LCD}). 
\citet{Hyttinen++2012} provide an 
algorithm for linear SCMs with cycles and confounders that deals with perfect interventions. The methods by
\citet{HEJ2014} and \citet{MooijHeskes_UAI_13} can deal with cycles in a linear (or approximately linear) setting. The method
by \citet{HEJ2014} relies on $d$-separation, which only applies in certain settings (see Theorem~\ref{thm:d_separation}). 
The method can be modified to use $\sigma$-separation instead \citep{ForreMooij_UAI_18}.
The way \citet{MooijHeskes_UAI_13} handle cycles is not as
straightforward. Generally, their method could handle nonlinear cyclic models, but for computational reasons, their implementation
linearizes the SCMs around each (context-dependent) equilibrium, thereby basically assuming that $d$-separation holds 
\emph{within each context}. The method by \citet{Rothenhausler++2015} assumes linearity and can deal with cycles in that case,
under a certain condition that suffices to prove identifiability of the method. The method by \citet{ICP2016} can handle cycles,
as our Corollary~\ref{cor:ICP_as_JCI} shows. The JCI framework in general allows for cycles, but requires its implementation 
to support this.

\subsubsection{Selection Bias}
The only causal discovery method for multiple data sets that is explicitly claimed to be able to deal with selection bias (i.e., conditioning on a latent variable that is a common effect of one or more of the observed variables), at least to some extent, is the IOD algorithm \citep{IOD2011}. It allows for different sets of observed (system) variables in each context and for different distributions in each context, while assuming that each context can be described by a MAG that is the marginal of a common MAG defined on the union of all system variables. 
It performs conditional independence tests in each data set separately, and merges the $p$-values of the test results using Fisher's method. It then constructs the PAG that represents simultaneously all contexts. 
Since it does not assume invariance of the distribution across contexts, it can deal with a single (latent) context variable that models mechanism changes or other ``soft'' interventions that do not change the conditional independences in the distribution. It can also deal with perfect interventions since Fisher's method is used to test for independence in \emph{all} contexts (see also Figure~\ref{fig:JCI_example}). 

\subsubsection{Imperfect Interventions and Other Context Changes}

\citet{CooperYoo1999} provided the first score-based causal discovery algorithm
that could deal with data from multiple contexts, focusing on perfect
interventions with known targets. They describe in detail how to handle perfect
interventions and introduced the idea of adding explicit context variables to
deal with mechanism changes, which was later refined by \citet{EatonMurphy07},
who provide an algorithm that can handle (stochastic) perfect interventions
with unknown targets, soft interventions, and mechanism changes. Also \citet{SPP05}
use a score-based causal discovery algorithm based on the ideas of
\citet{CooperYoo1999} that uses a greedy search strategy through the space of
DAGs. 

Another recent approach to constraint-based causal discovery in a JCI setting
is the one by \citet{YangKatcoffUhler2018}; these authors propose the algorithm
IGSP that can be seen as an implementation of JCI for causally sufficient, 
acyclic models with a diagonal experimental design under JCI Assumptions~\JCIAB, 
for mechanism changes with intervention targets assumed to be known. An advantage 
of IGSP over our JCI approach is that essentially no assumptions on the context 
distribution need to be made (apart from positivity) since it relies on a weaker
faithfulness assumption.

\citet{TianPearl2001} were the first to consider \emph{mechanism changes}. They deal with
sequences of mechanism changes, exploiting changes in the distribution
to infer descendants of the changed mechanism. This is followed by a constraint-based
approach from observational data that also takes into account the background knowledge
on the causal ordering of the system variables inferred from analysing the interventional
data. A similar approach
using the differences between data from experimental conditions and an observational 
baseline as background knowledge for a constraint-based approach was applied by 
\citet{MagliacaneClaassenMooij_NIPS_16} on the data of \citet{SPP05}.

\citet{Claassen++_NIPS2010} handle certain \emph{environment changes}: direct causal
relations between system variables are assumed to be invariant across contexts, but
latent confounding (and more generally, the exogenous distribution) may differ between contexts.
\citet{Rothenhausler++2015} assume stochastic \emph{shift interventions} in which the 
mean of a target variable is shifted by an (independent) random amount.
Various multi-task ``structure learning'' (i.e., Bayesian network learning) approaches 
that put a prior on the similarity of the DAGs in multiple contexts which encourages
them to be similar have been proposed \citep[e.g.,][]{oates2014,oates2016exact}. 

Some methods which allow for a single context variable have been applied in settings
on time-series data, by using time as the context variable
\citep{Friedman++2000,Zhang++_IJCAI17}. This extends the more usual approach
of treating time-series data by assuming \emph{invariance} of the causal structure across
time as in dynamic Bayesian networks (DBNs) \citep{Murphy2002}, methods based on
Granger causality \citep{Gra69}, or constraint-based approaches \citep{EntnerHoyer2010}.

JCI allows one to handle all interventions and context changes discussed above in a unified way. 

\subsubsection{Multiple Context Variables}
Some causal discovery methods for combining data from different contexts that explicitly consider a context variable,
allow for a single context variable only, for example, LCD, ICP, and the method by \citet{Zhang++_IJCAI17}. 
There is an important advantage to allowing multiple context variables, as JCI does generally.
One might argue that the case of multiple context variables can always be reduced to a case with a single context variable, by simply combining all context variables $\{C_k\}_{k\in\C{K}}$ into a single tuple $\B{C} = (C_k)_{k\in\C{K}}$. 
However, this reduction to a single context variable typically loses information.
This is illustrated in Figure~\ref{fig:multiple_contexts_advantage}.
When using only a single context variable in that case, the DMG cannot be identified from conditional independences in the data.
On the other hand, when using all three context variables with JCI, the complete DMG can be identified, even when the causal relations between context and system variables are unknown.

\begin{figure}
  \centering
  \begin{tikzpicture}
    \begin{scope}
      \draw (-2,0.6) edge[dotted] (2,0.6);
      \node[var] (C) at (0,1.2) {$\B{C}$};
      \node[var] (X0) at (-1.5,0) {$X_1$};
      \node[var] (X1) at (0,0) {$X_2$};
      \node[var] (X2) at (1.5,0) {$X_3$};
      \draw[arr] (X0) edge (X1);
      \draw[arr] (X1) edge (X2);
      \draw[arr] (C) edge (X0);
      \draw[arr] (C) edge (X1);
      \draw[arr] (C) edge (X2);
    \end{scope}
    \begin{scope}[xshift=5cm]
      \draw (-2,0.6) edge[dotted] (2,0.6);
      \node[var] (C0) at (-1.5,1.2) {$C_\alpha$};
      \node[var] (C1) at (0,1.2) {$C_\beta$};
      \node[var] (C2) at (1.5,1.2) {$C_\gamma$};
      \node[var] (X0) at (-1.5,0) {$X_1$};
      \node[var] (X1) at (0,0) {$X_2$};
      \node[var] (X2) at (1.5,0) {$X_3$};
      \draw[arr] (X0) edge (X1);
      \draw[arr] (X1) edge (X2);
      \draw[arr] (C0) edge (X0);
      \draw[arr] (C1) edge (X1);
      \draw[arr] (C2) edge (X2);
      \draw[biarr] (C0) edge (C1);
      \draw[biarr,bend left] (C0) edge (C2);
      \draw[biarr] (C1) edge (C2);
    \end{scope}
  \end{tikzpicture}
  \caption{Example that shows that allowing multiple context variables (right) has advantages over considering a single context variable only (left). The causal graph on the right can be identified by JCI (with JCI Assumptions~\JCIABC) from conditional independences in the pooled data, whereas the causal graph on the left is not identifiable.\label{fig:multiple_contexts_advantage}}
\end{figure}

\subsubsection{Dependent Context Variables}
If one allows for multiple context variables and considers the joint distribution on context and system variables, as we do in JCI, one should account for possible dependencies between the context variables. Indeed, incorrectly assuming the context variables to be independent \emph{a priori} may lead to wrong conclusions. An example is provided in Figure~\ref{fig:JCI_dependent_contexts}. In that example, incorrectly assuming the contexts to be independent leads to the wrong conclusion that context variable $C_\beta$ causes the system variable $X_0$, at least for causal discovery algorithms that are tolerant to faithfulness violations.

This issue was recognized and addressed in recent work \citep{oates2016estimating} by introducing a novel graphical modeling framework, Conditional DAGs (CDAGs), which bears some similarity with our approach. However, a disadvantage of the CDAG framework is that existing causal discovery methods cannot be directly applied to learn a CDAG from data, and the wealth of results on causal modeling with SCMs cannot be used directly. One of the key advantages of the JCI framework is that it utilizes existing theory and methods, as it reduces a causal discovery problem from multiple contexts to a purely observational one with background knowledge. This is one of the reasons why JCI offers many more features than the approach by \citet{oates2016estimating}. We also note that CDAGs can be dealt with as a special case of the JCI framework. 


\begin{figure}
  \centering
  \begin{tikzpicture}
    \begin{scope}
      \draw (-2,0.6) edge[dotted] (0.5,0.6);
      \node[var] (C0) at (-1.5,1.2) {$C_\alpha$};
      \node[var] (C1) at (0,1.2) {$C_\beta$};
      \node[var] (X0) at (-1.5,0) {$X_1$};
      \draw[arr] (C0) edge (X0);
      \draw[biarr] (C0) edge (C1);
    \end{scope}
    \begin{scope}[xshift=5cm]
      \draw (-2,0.6) edge[dotted] (0.5,0.6);
      \node[var] (C0) at (-1.5,1.2) {$C_\alpha$};
      \node[var] (C1) at (0,1.2) {$C_\beta$};
      \node[var] (X0) at (-1.5,0) {$X_1$};
      \draw[arr] (C0) edge (X0);
      \draw[arr] (C1) edge (X0);
    \end{scope}
  \end{tikzpicture}
  \caption{Example that shows that incorrectly assuming independent context variables can lead to wrong conclusions when using causal discovery algorithms that are tolerant to faithfulness violations. Left: true causal graph, with dependent context variables, which is identifiable by JCI. Right: causal graph that reproduces all conditional dependencies (except $C_\alpha \nCI C_\beta$) and minimizes the number of faithfulness violations (faithfulness here implies $C_\beta \nCI X_1 \given C_\alpha$), when (incorrectly) assuming that context variables are independent.\label{fig:JCI_dependent_contexts}}
\end{figure}

\subsubsection{Partially Overlapping Sets of Variables}

A particular case of missing data that has been addressed by some of the methods
is when the set of observed variables differ between data sets, while still having some overlap.
The first one to address this using constraint-based causal discovery was \citet{ION2009},
and several other methods have been proposed over the years
\citep{Claassen++_NIPS2010,IOD2011,HEJ2014,triantafillou2015constraint}.
JCI can only deal with this when strengthening its faithfulness assumption: one would need to
assume that the context variables are discrete, and that every conditional distribution 
$\Prb(\B{X} \given \B{C}=\B{c})$ for $\B{c} \in \B{C}$ with $\Prb(\B{C}=\B{c})>0$ is 
faithful with respect to the marginalization of the same DMG $\C{G}_{\C{I}}$ on the
system variables that were observed in that context (or, in case of perfect interventions
with known targets, the corresponding marginal intervened DMG).

\subsubsection{Influence Diagrams}

Our representation of a system within a context imposed by its environment bears strong similarities with influence diagrams \citep{Dawid2002}. 
A formal difference is that we consider the context variables to be random variables that reflect the empirical distribution of the experimental design, whereas in influence diagrams they are interpreted as non-random decision variables. 
The advantage of treating context variables as random variables is that this allows one to apply standard causal discovery techniques (designed for random variables) \emph{jointly} on system and context variables. 
In particular, the standard notion of statistical conditional independence \citep{Dawid1979} suffices. 
If one would like to treat the context variables as decision (i.e., non-random) variables, extended notions of conditional independence would be necessary \citep{ForreMooij_UAI_19}. 
Since we can always view the context variables as random variables in the \emph{empirical distribution} of the experimental design (see also Footnote~\ref{footnote:iid}), this allows us to make use of the standard notion of conditional independence for the purposes of causal discovery.

\subsubsection{Selection Diagrams}
Our representation of a system within a context imposed by its environment also bears some similarities with selection diagrams \citep{BareinboimPearl2013}.
Selection diagrams have also been used for causal modeling in different contexts, but one crucial difference is that we are modeling the \emph{joint} distribution on the intervention and system variables, whereas a selection diagram represents the \emph{conditional} distribution of the system variables given the intervention (``selection'') variables. 
Because we are modeling the joint distribution and not only the conditional one, we can apply standard causal discovery techniques directly on pooled data, something that would not be as trivial when using selection diagrams instead.

\citet{BareinboimPearl2013} define a selection diagram as follows:
\begin{definition}
Let $\C{M} = \langle \C{I}, \C{J}, \C{H}, \BC{X}, \BC{E}, \B{f}, \Prb_{\BC{E}} \rangle$ and
$\C{M}^* = \langle \C{I}, \C{J}, \C{H}, \BC{X}, \BC{E}, \B{f}^*, \Prb^*_{\BC{E}} \rangle$ be
two acyclic SCMs corresponding to two different contexts, that only differ with respect to their
causal mechanisms and exogenous distributions. In particular, they share the same 
augmented graph $\C{H}$ and hence also their graphs are identical, $\C{G}(\C{M}) = \C{G}(\C{M}^*)$. 
The \emph{selection diagram} $\C{S}$ induced by $\langle \C{M}, \C{M}^*\rangle$ 
is the acyclic directed mixed graph with nodes $\C{I} \dot\cup \overline{\C{I}}$, where 
$\overline{\C{I}} := \{\bar{i} : i \in \C{I}\}$ is a copy of $\C{I}$ of \emph{selection variable} indices, such that
  \begin{enumerate}
    \item the induced subgraph of $\C{S}$ on $\C{I}$ equals the common graph of $\C{M}$ and $\C{M}^*$, i.e., $\C{S}_{\C{I}} = \C{G}(\C{M}) = \C{G}(\C{M}^*)$, and
    \item for each $i\in \C{I}$ such that $f_i \ne f_i^*$ or $\Prb_{\BC{E}_{\pasub{\C{G}(\C{M})}{i}}} \ne \Prb^*_{\BC{E}_{\pasub{\C{G}(\C{M})}{i}}}$ there is an edge $\bar{i} \to i$ in $\C{S}$.
  \end{enumerate}
\end{definition}

From the definition, it is apparent that a selection diagram essentially models \emph{two} contexts, and
that the selection variables in the selection diagram correspond to the \emph{children of the context variable}
in our representation.
Indeed, consider a JCI model of the form \eref{eq:SCM_JCI_ass} (p.\ \pageref{eq:SCM_JCI_ass}) with a single binary context variable $C$. The
joint SCM can be split into two context-specific SCMs, $\C{M}^0$ and $\C{M}^1$, and the induced 
selection diagram $\C{D}$ can be obtained from the causal graph $\C{G}(\C{M})$ as follows: (i) 
each edge $i_1 \to i_2$ or $i_1 \oto i_2$ in $\C{G}(\C{M})$ between system variables $i_1,i_2 \in \C{I}$ is also
in $\C{D}$; (ii) if $C \to i$ in $\C{G}(\C{M})$ for $i \in \C{I}$ then $\bar{i} \to i$ is in $\C{D}$. Since the JCI
framework can be used to learn (features of the) causal graph $\C{G}(\C{M})$ from data, this means that
we can thereby learn (features of) the selection diagram from data. 

It is not clear how a selection diagram could be used to represent the same information that an SCM with multiple context variables can represent.
Indeed, even though the selection diagram has multiple selection variables, it is still modeling only two contexts, corresponding with just a single binary context variable in the JCI framework.

\section{Experiments}\label{sec:experiments}

In this section we report on the experiments we performed with JCI, comparing
various implementations of the framework with several baselines and
state-of-the-art causal discovery methods. We experimented both with simulated data
with perfectly known ground truth and with real-world data where the ground
truth is only known approximately. The source code that we used for producing
the results and plots in this section is provided under a free and open source
license as Online Appendix 1.

\subsection{Methods and Baselines}

In our experiments we study different implementations of JCI, based on two 
existing causal discovery algorithms: ASD \citep{HEJ2014,MagliacaneClaassenMooij_NIPS_16,ForreMooij_UAI_18} and FCI \citep{SMR1999,Zhang2008_AI}.
The ASD algorithm is accurate but slow, while FCI is faster but less accurate 
due to its ``greedy'' approach. Another difference between both methods is that 
ASD can deal with partial inputs, while for FCI it is necessary to provide 
all independence test results it asks for. Although FCI (and with some small
extensions, also ASD)
can deal with selection bias, we ignore this additional complication here
and use simplified implementations that assume that there is no selection bias.
For the cyclic case, we used the adaptation of the ASD algorithm proposed by
\citet{ForreMooij_UAI_18} that replaces $d$-separation with
its general cyclic generalization, $\sigma$-separation. Adapting FCI to the cyclic case
seems less straightforward and is beyond the scope of this paper.\footnote{The 
$d$-separation case has recently been addressed by \citet{Strobl2018}.}

Table~\ref{tab:algorithms} provides an overview of all implementations that we have 
studied here. The methods will be discussed in more detail in the next few subsections.
The ``CI Tests'' column in the table describes what conditional
independence test are performed, and how, and can have the following values:
\begin{description}
\item[A]: use all variables, including context variables; the conditional independence
  tests performed are of the form $\tilde{X}_a \CI \tilde{X}_b \given \tilde{\B{X}}_S$ with $\{a\} \cup \{b\} \cup S \subseteq \C{I} \cup \C{K}$
  and $\{a\}, \{b\}, S$ mutually disjoint.
\item[S]: use only system variables; the conditional independence tests performed are 
  of the form $X_a \CI X_b \given \B{X}_S$ with $\{a\} \cup \{b\} \cup S \subseteq \C{I}$
  and $\{a\}, \{b\}, S$ mutually disjoint.
\item[SS]: system variables only, separately for each context; the conditional independence tests performed
  are of the form $X_a \CI X_b \given \B{X}_S, \B{C}=\B{c}$ for $\{a\} \cup \{b\} \cup S \subseteq \C{I}$
  and $\{a\}, \{b\}, S$ mutually disjoint, for all values $\B{c} \in \BC{C}$ with $\Prb(\B{c}) > 0$.
  Note that this method assumes that the context domain $\BC{C}$ is discrete.
\item[SF]: system variables only, separately for each context, using Fisher's method;\footnote{Fisher's method \citep{Fisher1925}
  aggregates $N$ independent $p$-values $\{p_i\}_{i=1}^N$ by computing the $p$-value 
  for the statistic $F := -2 \sum_{i=1}^N \log p_i$, which has a $\chi^2$ distribution with $2 N$ degrees of freedom if all $N$ $p$-values $p_i$ are independent.}
  the conditional independence tests performed are of the form $X_a \CI X_b \given \B{X}_S, \B{C}=\B{c}$ for 
  $\{a\} \cup \{b\} \cup S \subseteq \C{I}$ and $\{a\}, \{b\}, S$ mutually disjoint, for all values $\B{c} \in \BC{C}$ with $\Prb(\B{c}) > 0$. Note that this method assumes that the context domain $\BC{C}$ is discrete.
\item[NC]: test all variables, except for conditional independences between context variables; the conditional independence
  tests performed are of the form $X_a \CI \tilde{X}_b \given \tilde{\B{X}}_S$ with $a \in \C{I}$, $\{b\} \cup S \subseteq \C{I} \cup \C{K}$ and $\{a\}, \{b\}, S$ mutually disjoint.
\item[PF]: test all pairs of context and system variables, conditioning on the remaining context variables; the conditional
  independence tests that are performed are of the form $X_i \CI C_k \given \B{C}_{\C{K} \setminus \{k\}}$ for $i \in \C{I}$, $k \in \C{K}$.
\end{description}

\subsubsection{ASD Variants}
For ASD, we implemented different variants as described in Table~\ref{tab:algorithms}.
The variants \alg{ASD-obs}, \alg{ASD-pooled} and \alg{ASD-pikt} use 
the original implementation of \citet{HEJ2014} in the acyclic case
and the $\sigma$-separation adaptation of \citet{ForreMooij_UAI_18} in the
cyclic case, with the weights and query method
of \citet{MagliacaneClaassenMooij_NIPS_16}. \alg{ASD-obs} only uses the 
observational context and ignores data from the other contexts, \alg{ASD-pooled} 
pools data from all contexts but does \emph{not} add context variables, and
\alg{ASD-pikt} uses data from all contexts and assumes that the contexts correspond
to perfect interventions with known targets. Inspired by the approach of
\citet{ION2009} and \citet{IOD2011}, we also implemented a variant \alg{ASD-meta}
that uses Fisher's method as a ``meta-analysis'' method to combine the $p$-values of conditional
independence tests performed with data from each context separately into a single 
overall $p$-value, which was then used as input for ASD. 
We use these implementations as state-of-the-art causal discovery methods for comparison.

For the JCI approach, we study several variants that differ in terms of which JCI assumptions
they make, whether they also test conditional independences between context variables, 
whether all context variables are used or whether they are first merged into a single
context variable, and whether the intervention targets of the context variables are 
considered to be known or not. The different implementations are described in detail
in Table~\ref{tab:algorithms}. 
\alg{ASD-JCI123sc} and \alg{ASD-JCI1sc} both use a single merged context variable $\B{C} = (C_k)_{k\in\C{K}}$, whereas
the other ASD-JCI variants use all context variables $\{C_k\}_{k\in\C{K}}$ as separate variables. The background knowledge for
all ASD-JCI variants is a subset of the three JCI Assumptions \ref{ass:uncaused}, \ref{ass:unconfounded} and \ref{ass:dependences}. The only ASD-JCI variant that uses background knowledge on intervention targets is \alg{ASD-JCI123kt} (but it does not make any assumptions on the \emph{type} of the intervention).

\subsubsection{FCI Variants}
We implemented different variants of the FCI algorithm by adapting the implementation in the
\texttt{R} package \texttt{pcalg} \citep{pcalg}. We used the default configuration, i.e., an order-independent
(``stable'') skeleton phase, and no conservative or majority rule modifications \citep{colombo2014order}.
For simplicity, we assumed that no selection bias is present, which 
means that the rules $\mathcal{R}$5--$\mathcal{R}$7 in \citet{Zhang2008_AI} can
be ignored in the FCI algorithm, and only PAGs without (possibly) undirected edges need to be considered.
We consider two variants of FCI as the current state-of-the-art:
\alg{FCI-obs}, which uses only the observational context, and \alg{FCI-pooled}, which uses 
pooled data from all contexts but does \emph{not} add context variables. We also implemented
a ``meta-analysis'' approach \alg{FCI-meta} that uses Fisher's method to combine the $p$-values
from separate contexts into overall $p$-values that are used as input for the FCI algorithm.
Finally, we have three variants (\alg{FCI-JCI123}, \alg{FCI-JCI1} and \alg{FCI-JCI0}) 
referring to the JCI adaptation of the FCI algorithm as described in Section~\ref{sec:FCI-JCI},
for three different combinations of JCI Assumptions (we have not yet implemented and evaluated the speedup for
\alg{FCI-JCI123r} but leave that for future work).

The output of FCI is a PAG. Here we will not evaluate the PAG itself, since estimating the PAG
is often not the ultimate task in causal discovery, but instead we will evaluate 
presence and absence of ancestral relations that can be identified from the estimated PAG. 
The procedure we use for this task is explained in Appendix~\ref{sec:app:dpag_ancestors}.
For the special case of \alg{FCI-JCI123}, we also read off the direct intervention
targets (and non-targets) from the estimated PAG, as explained in Appendix~\ref{sec:app:fcijci123_intervention_targets}.
We encode identified presence of a feature by $+1$, identified absence by $-1$, and an unidentifiable
feature by $0$. These predictions can then also easily be bootstrapped (or more precisely,
bagged). The bagged feature scores can then be used as a score for the confidence that
the feature is present.

\subsubsection{LCD Variants}

The \alg{LCD} implementation simply iterates over all context variables and
ordered pairs of system variables and tests for the LCD pattern. As
conditional independence test we test whether the partial correlation vanishes.
As confidence measure for an LCD pattern $\langle C,X,Y\rangle$ (see also Figure~\ref{fig:LCD}, where $\langle X_1,X_2,X_3 \rangle$ corresponds with $\langle C,X,Y\rangle$), we use $-\log p_{C\CI Y}$.
Note that LCD predicts the presence of an ancestral relation $X \in \an{Y}$, the absence of
a confounder between $X$ and $Y$, and the absence of a direct causal effect of $C$ on $Y$.\footnote{The absence of a bidirected edge $X \oto Y$ in the marginalization of the graph $\C{G}$ on $\{C,X,Y\}$
implies also the absence of the bidirected edge $X \oto Y$ in the full graph $\C{G}$. Furthermore, the
absence of the direct edge $C \to Y$ in this marginalization implies the absence of the direct edge $C \to Y$ in the full
graph $\C{G}$.}

\subsubsection{ICP Variants}
We also compare with the \texttt{ICP} function in the \texttt{R} package \texttt{InvariantCausalPrediction} \citep{ICP2016}.\footnote{We used the default arguments, except that we set \texttt{stopIfEmpty} to \texttt{TRUE}.}
Under the assumption of causal sufficiency, ICP returns direct causal relations. However,
in our setting, in which causal sufficiency cannot be assumed, ICP will generally return
ancestral causal relations, as shown in Corollary~\ref{cor:ICP_as_JCI}. Note that ICP
assumes a single context variable, whereas one typically may have multiple context variables
in the data. One way to deal with this is to merge all context variables before the pooled
data is input to \texttt{ICP}, which we do in \alg{ICP-sc}. Another way is to run ICP 
on each context variable separately, hiding the other context variables when feeding the
pooled data to \texttt{ICP}, and finally merging all predictions. This is done in \alg{ICP-mc}. 

The conditional independence tests that ICP performs are of the form 
$\B{C} \CI X_i \given \B{X}_S$ with $S \subseteq \C{I} \setminus \{i\}$, for $i \in \C{I}$.
The default conditional independence test used in ICP first linearly regresses $X_i$ on $\B{X}_S$ for
each context $\B{C}=\B{c}$ individually, and once globally. It then tests whether there exists
a context $\B{c}$ in which the mean or the variance of the regression residuals is different from
the global mean or variance of the residuals. All $p$-values are then combined using a Bonferroni correction.
Note that this test can detect more conditional dependencies than a simple partial correlation test, 
as it also considers the variation between the variances across contexts. 

As a confidence score for the ancestral causal relation $i \in \an{j}$, we use $-\log p_{i\causes j}$, 
where $p_{i\causes j}$ is the $p$-value returned by ICP for system variable $i$ being ancestor of system variable $j$.

\begin{table}\centering
  \small\scalebox{0.95}{\begin{tabular}{llllllll}
  Name                   & Data          & Context variables & \multicolumn{3}{c}{JCI Assumptions} & Interventions & CI Tests \\
                         &               &                   & \ref{ass:uncaused} & \ref{ass:unconfounded} & \ref{ass:dependences} &                         &          \\
  \hline
  \multicolumn{8}{l}{\textbf{Baselines}} \\
  \hline
  \alg{ASD-obs}       & observational & none                & $-$ & $-$ & $-$ & none        & S \\
  \alg{ASD-pooled}    & pooled        & none                & $-$ & $-$ & $-$ & any         & S  \\
  \alg{ASD-meta}      & all           & none                & $-$ & $-$ & $-$ & any         & SF \\
  \alg{ASD-pikt}      & all           & none                & $-$ & $-$ & $-$ & perfect, KT & SS \\
  \hline
  \alg{FCI-obs}       & observational & none                & $-$ & $-$ & $-$ & none        & S  \\
  \alg{FCI-pooled}    & pooled        & none                & $-$ & $-$ & $-$ & any         & S  \\
  \alg{FCI-meta}      & all           & none                & $-$ & $-$ & $-$ & any         & SF \\
  \hline
  \multicolumn{8}{l}{\textbf{JCI Implementations}} \\
  \hline
  \alg{ASD-JCI0}      & pooled        & all                 & $-$ & $-$ & $-$ & any         & A  \\
  \alg{ASD-JCI1}      & pooled        & all                 & $+$ & $-$ & $-$ & any         & A  \\
  \alg{ASD-JCI12}     & pooled        & all                 & $+$ & $+$ & $-$ & any         & A  \\
  \alg{ASD-JCI123}    & pooled        & all                 & $+$ & $+$ & $+$ & any         & NC \\
  \alg{ASD-JCI123kt}  & pooled        & all                 & $+$ & $+$ & $+$ & any, KT     & NC \\
  \hline
  \alg{FCI-JCI123}    & pooled        & all                 & $+$ & $+$ & $+$ & any         & NC \\
  \alg{FCI-JCI1}      & pooled        & all                 & $+$ & $-$ & $-$ & any         & A  \\
  \alg{FCI-JCI0}      & pooled        & all                 & $-$ & $-$ & $-$ & any         & A  \\
  \hline
  \alg{LCD-sc}        & pooled        & single (merged)     & $+$ & $-$ & $-$ & any         & A$^*$  \\
  \alg{ICP-sc}        & pooled        & single (merged)     & $+$ & $-$ & $-$ & any         & A$^*$  \\
  \alg{ASD-JCI1-sc}   & pooled        & single (merged)     & $+$ & $-$ & $-$ & any         & A$^*$  \\
  \alg{ASD-JCI123-sc} & pooled        & single (merged)     & $+$ & $+$ & $+$ & any         & A$^*$  \\
  \hline
  \alg{LCD-mc}        & pooled        & all (one-by-one)    & $+$ & $-$ & $-$ & any         & A  \\
  \alg{ICP-mc}        & pooled        & all (one-by-one)    & $+$ & $-$ & $-$ & any         & A$^*$  \\
  \hline
  \alg{Fisher}        & pooled        & all (one-by-one)    & $+$ & $+$ & $-$ & any         & PF \\
  \hline
  \end{tabular}}
  \caption{Variants of implemented JCI algorithms and baselines used in our experiments.
  JCI Assumption~\ref{ass:simple_scm} is always assumed. ``KT'' is an abbreviation of ``known targets''.
  The meaning of ``CI Tests'' is (more detailed explanation in main text): A: use all variables, including context variables; S: use only system variables; SS: system variables only, separately for each context; SF: system variables only, separately for each context, using Fisher's method to combine them into a single $p$-value; NC: test all variables, except for conditional independences between context variables; PF: test all pairs of context and system variables, conditioning on the remaining context variables. 
  The meaning of the superscript $^*$ is explained in Section~\ref{sec:CI}. Bootstrapped versions of methods will be indicated with a suffix ``\alg{-bs}'' (and have been omitted from this table for clarity).\label{tab:algorithms}}
\end{table}

\subsubsection{Fisher's Test for Causality}\label{sec:Fisher}

This is a very simple and immensely popular baseline in which we simply go through
all pairs $(i,k)$ of a system variable $i \in \C{I}$ and a context variable $k
\in \C{K}$, perform the conditional independence test $X_i \CI C_k \given
\B{C}_{\C{K} \setminus \{k\}}$ on the pooled data resulting in $p$-value
$p_{ik}$. It is limited to discovery of ancestral causal relations from
context to system variables.
As confidence value for the ancestral causal relation $k \in
\an{i}$ we report $-\log p_{ik} + \log \alpha$, where $\alpha$ is the
threshold for the independence test.

\subsubsection{Bootstrapping}

A simple way to improve the stability of causal discovery algorithms is
bootstrapping. For a method that outputs a confidence measure for a certain
prediction we simply average the confidence measures over bootstrap samples.
For FCI, as confidence measure we simply take a $\{-1,0,1\}$-valued variable encoding the
identifiable absence/unidentifiability/identifiable presence of an ancestral
relation (or direct intervention target, for \alg{FCI-JCI123}). 
For LCD, we average $-\log p_{C\CI Y}$ over bootstrap samples. 
For ICP, we similarly average the negative logarithm of the $p$-values for the
discovered ancestral causal relations. We do not bootstrap ASD variants because
of the high computational complexity.
In our experiments, we use 100 bootstrap samples. Bootstrapped methods are indicated
with a suffix ``\alg{-bs}''.

\subsubsection{Conditional Independence Tests\label{sec:CI}}

Using an appropriate conditional independence test is important to obtain good causal discovery results. In this work we will
use two different conditional independence tests, both relying on the assumption that the context variables are
discrete and the system variables have a multivariate Gaussian distribution given the context. However,
the JCI framework imposes no principled restrictions on the conditional independence tests used, so one
could also use non-parametric tests instead, for example.

The default conditional independence test that we used is the following. For testing $\tilde{X}_a \CI \tilde{X}_b \given \tilde{\B{X}}_S$, we distinguish two cases:
\begin{itemize}
  \item $S \cap \C{K} = \emptyset$: the test then reduces to a standard partial correlations test.
  \item Otherwise, we go through all observed values $\B{c}_{S \cap \C{K}}$ of $\B{C}_{S \cap \C{K}}$, 
    and use a standard partial correlations test
    to calculate a $p$-value for $\tilde{X}_a \CI \tilde{X}_b \given \tilde{\B{X}}_{S\setminus\C{K}}, \B{C}_{S \cap \C{K}}=\B{c}_{S \cap \C{K}}$. We then aggregate the
    $p$-values corresponding to observed values of $\B{C}_{S \cap \C{K}}$ into one overall $p$-value for $\tilde{X}_a \CI \tilde{X}_b \given \tilde{\B{X}}_S$ using
    Fisher's method for aggregating $p$-values.\footnote{Not to be confused with Fisher's test for causality that we
    described in Section~\ref{sec:Fisher}.}
\end{itemize}
Note that for the case with zero context variables (i.e., a
single context), this test reduces to a standard partial correlations test. 


For the ICP implementations, we make use of the implementation in the \texttt{R} package \texttt{InvariantCausalPrediction}. 
This by default makes use of a conditional independence test that uses more than just partial correlations. We extended this
conditional independence test to allow for conditioning on context variables, and make use of this extended test in all the 
algorithms that assume a single merged context, i.e., the ones
marked with a $^*$ in Table~\ref{tab:algorithms}. It assumes that there is a single context variable, i.e., $|\C{K}| = 1$.
For testing $\tilde{X}_a \CI \tilde{X}_b \given \tilde{\B{X}}_S$, it distinguishes the following cases:
\begin{itemize}
  \item If $(\{a\}\cup\{b\}\cup S) \cap \C{K} = \emptyset$, it reduces to a standard partial correlations test.
  \item If $\tilde{X}_a = \B{C}$ is the context variable, it uses linear regression to fit $\tilde{X}_b$ as a linear function of $\B{X}_S$, using data pooled over all contexts, and calculates the corresponding residuals. It then goes through all observed values $\B{c}$ of $\B{C}$, and tests whether the residuals in context $\B{c}$ have a different distribution than the residuals in the other contexts (i.e., with $\B{C} \ne \B{c}$). This two-sample test is performed by comparing the means by a $t$-test, and the variances by an $F$-test. The
    two resulting $p$-values are combined with a Bonferroni correction. The resulting $p$-values, one for each context, are then also combined with a Bonferroni correction.\footnote{This is the default test  for continuous data in the \texttt{ICP} function of the \texttt{R} package \texttt{InvariantCausalPrediction}.}
  \item If $\tilde{X}_b = \B{C}$ is the context variable, proceed similarly as in the previous case.
  \item If the context variable $\B{C}$ is part of $\tilde{X}_S$, 
   we go through all observed values $\B{c}$ of $\B{C}$, 
   and use a standard partial correlations test
    to calculate a $p$-value for $X_a \CI X_b \given \B{X}_{S\setminus\C{K}}, \B{C}=\B{c}$. We then aggregate the
    $p$-values corresponding to observed values of $\B{C}$ into one overall $p$-value for $X_a \CI X_b \given \tilde{\B{X}}_S$ using
   Fisher's method for aggregating $p$-values.
\end{itemize}

As a final note regarding the conditional independence tests, we state that
\alg{ASD-pikt} uses a standard partial correlation test to calculate a
$p$-value for each context separately.  It subsequently combines all
these $p$-values by adding the $\log p$-values, but taking into account how the
graph structure is changed through perfect interventions.

The choice of the $p$-value threshold $\alpha$ for rejecting the null hypothesis
of independence is an important one. To obtain consistent results, one should let
$\alpha$ decrease to $0$ with increasing sample size. In our experiments, we used
fixed sample size and simply used a global threshold $\alpha = 0.01$.

\subsection{Simulations}\label{sec:simulations}

We simulated random linear-Gaussian SCMs with $p$ system variables and $q$
context variables. We considered both the acyclic setting and the cyclic one.
We simulated stochastic interventions of two different intervention types:
mechanism changes, and perfect interventions.

Random causal graphs
were simulated by drawing directed edges independently between system variables
with probability $\epsilon$. For the acyclic models, we only allowed directed
edges $i_1\to i_2$ for $i_1<i_2$ with $i_1,i_2 \in \C{I}$. For cyclic models, we allowed
directed edges $i_1\to i_2$ for $i_1\ne i_2$ with $i_1,i_2 \in \C{I}$, and subsequently
selected only the graphs in which at least one cycle exists. We drew bidirected
edges independently between all unordered pairs of system variables with
probability $\eta$, and associated each bidirected edge with a separate latent
confounding variable. For each context variable, we randomly selected a single
system variable as its target, while ensuring that each system variable has at
most one context variable as its direct cause. We sampled all linear
coefficients between system variables, context variables and confounders from
the uniform distribution on $[-1.5,-0.5] \cup [0.5,1.5]$. The exogenous variables
(``error terms'') were sampled independently from the standard-normal
distribution. To ensure that system variables have comparable scales, we
rescaled the weight matrix such that each system variable would have variance 1
if all its direct causes would be i.i.d.\ standard-normal. 

We used binary context variables in a ``diagonal'' design. This means that for
each random SCM, we simulated $q+1$ contexts, with the first context
being purely observational (i.e., $C_k = 0$ for all $k\in\{1,\dots,q\}$), and the 
other $q$ contexts corresponding with one of the context variables turned on
(say $C_{k'} = 1$ for some $k' \in \{1,\dots,q\}$) and the others turned off ($C_k = 0$
for the other $k \in \{1,\dots,q\} \setminus \{k'\}$). We either took all interventions
to be mechanism changes, or all interventions to be perfect.
For mechanism changes, we simply add the value of the parent context variable
to the structural equation (i.e., this corresponds with adding a constant offset
of 1 to the intervention target variable when the intervention is turned on). 
For perfect interventions, we additionally set the linear coefficients of incoming 
edges on the intervention target to zero. 
Finally, we sampled $N$ observed values of system variables from each context and
combined all samples into one pooled data set. This was done for each random SCM
separately.

\subsection{Evaluation}

In evaluating the results, we consider different prediction tasks: establishing 
the absence or presence of ancestral causal relations between system variables,
the absence or presence of direct causal relations between system variables, and
the absence or presence of confounders between system variables. 
In addition, we consider predicting the
absence or presence of indirect intervention targets (i.e., whether or not some 
context variable is ancestor of some system variable) and of direct intervention
targets (i.e., whether or not some context variable is parent of some system variable).

Each method outputs a confidence score for each feature of interest, where
positive scores mean that it is more likely that a feature is present in the
causal graph $\C{G}$, whereas negative scores mean that it is more likely that
a feature is absent. The higher the absolute value of the score, the more likely
its presence or absence is. The predictions are pooled both within model instances
(e.g., all possible ancestral relations $i \in \ansub{\C{G}}{j}$ for all ordered
pairs of system variables $i,j \in \C{I}$) and across model instances to gather 
more statistics. The scores are
then ranked and turned into ROC curves and PR curves (one PR curve for the
presence, and one for the absence of the features) by comparing with the true
features. In the ROC curves, we use solid lines for positive scores (feature present)
and negative scores (feature absent), and dotted lines for vanishing scores (feature presence/absence is unknown).

\subsection{Results: Small Simulated Models}

We first present results for small models with $p=4$ system variables and $0 \le q \le 4$ (as a default, $q=2$) context variables.
We used $\epsilon = 0.5$, $\eta = 0.5$, and sampled $N_{\B{c}} = 500$ samples for each context, i.e.,
$N = 500 (q+1)$ samples in total.

\subsubsection{ASD-JCI vs.\ Baselines (Causal Mechanism Changes)\label{sec:results_asd-jci_mc}}

\begin{figure}[t]
\centerline{%
\includegraphics[width=0.24\textwidth]{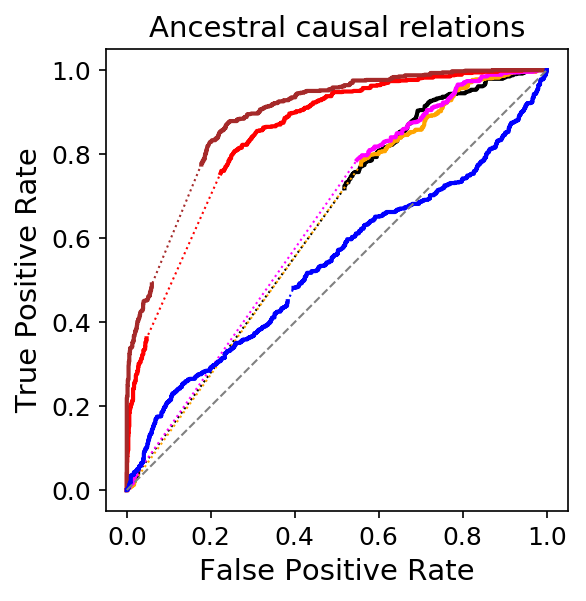}
\includegraphics[width=0.24\textwidth]{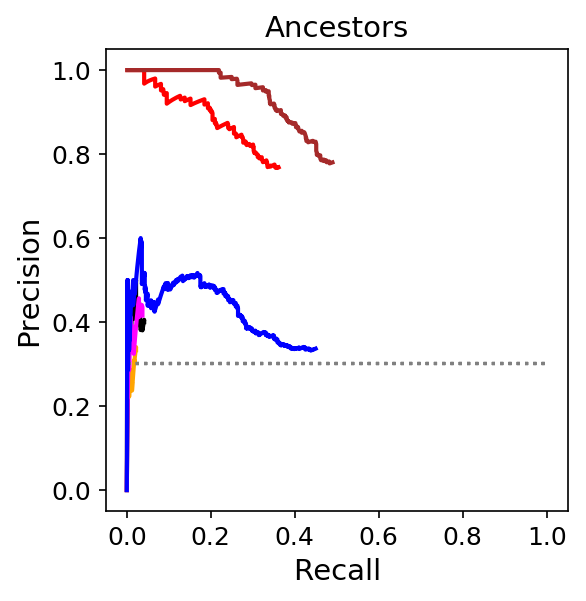}
\includegraphics[width=0.24\textwidth]{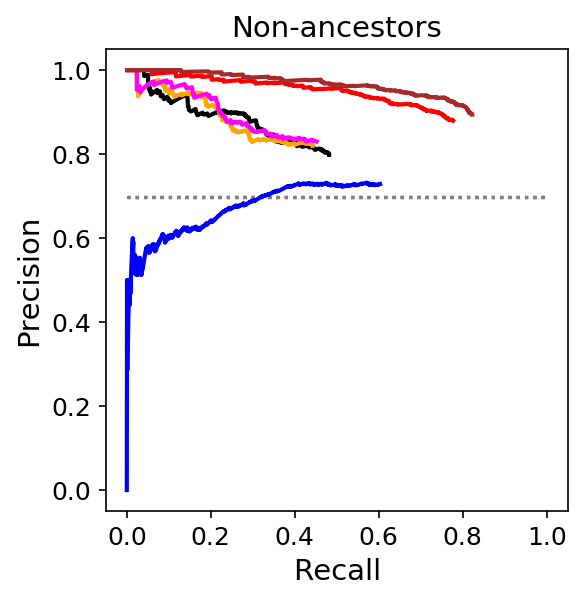}
\includegraphics[width=0.24\textwidth]{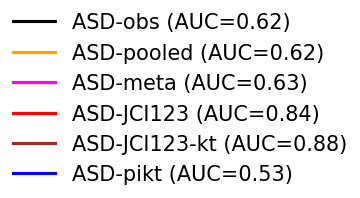}
}
\centerline{%
\includegraphics[width=0.24\textwidth]{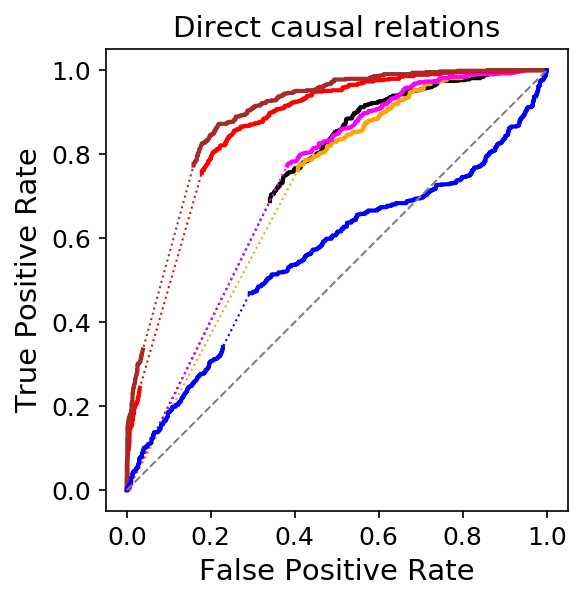}
\includegraphics[width=0.24\textwidth]{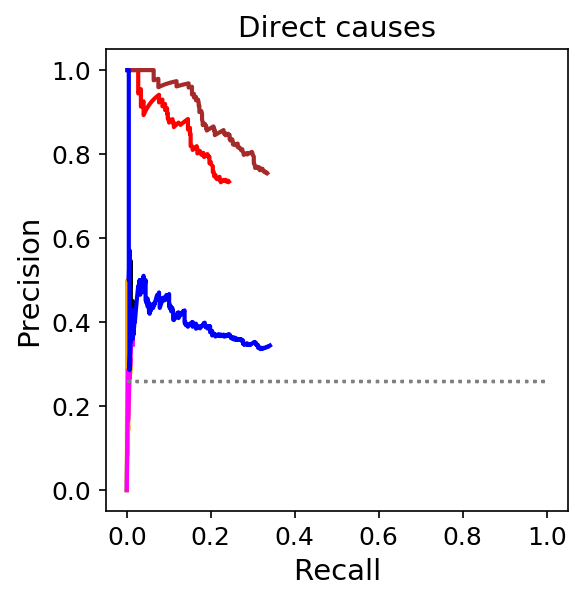}
\includegraphics[width=0.24\textwidth]{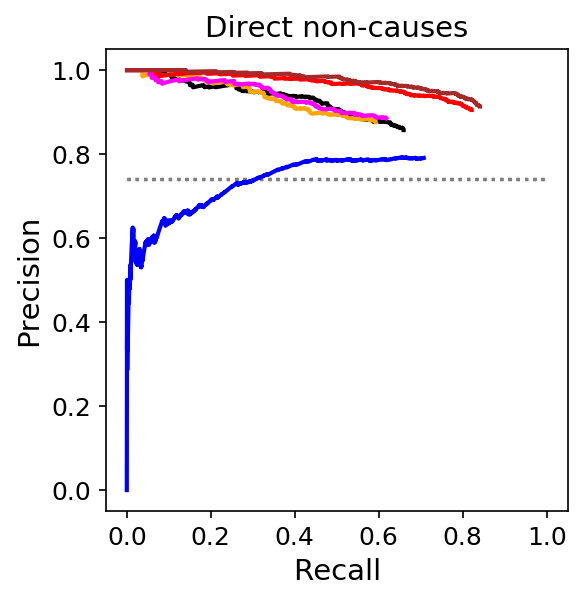}
\includegraphics[width=0.24\textwidth]{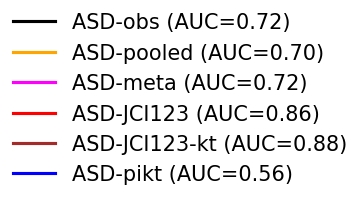}
}
\centerline{%
\includegraphics[width=0.24\textwidth]{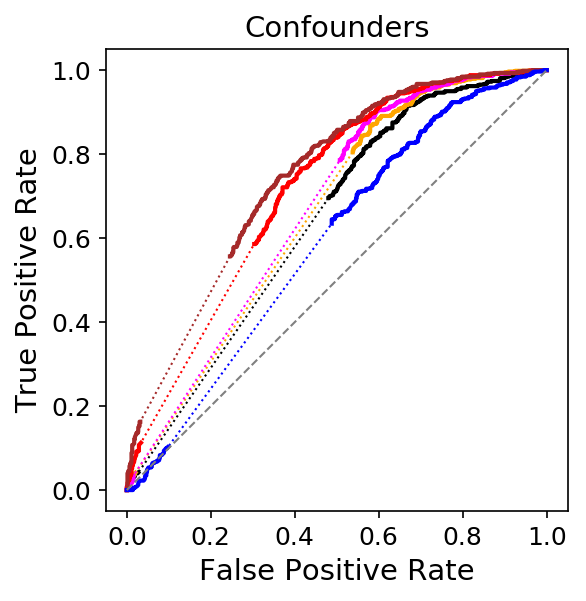}
\includegraphics[width=0.24\textwidth]{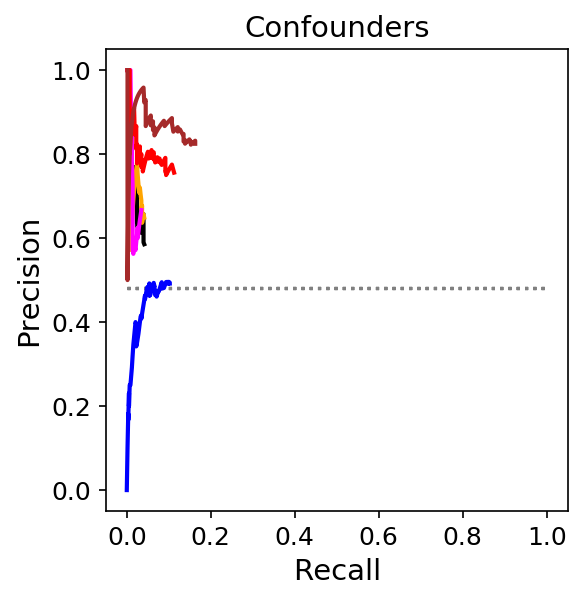}
\includegraphics[width=0.24\textwidth]{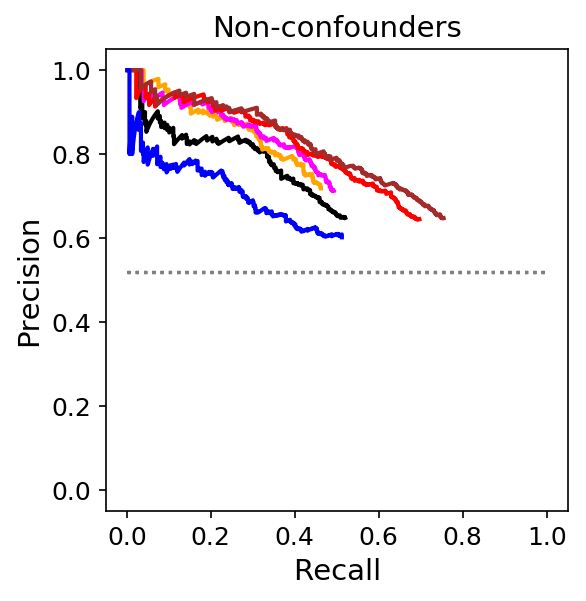}
\includegraphics[width=0.24\textwidth]{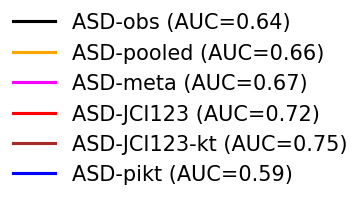}
}
\caption{\boldcap{Results of some ASD variants (acyclic, causal mechanism changes)} for small models. The two JCI variants (with unknown/known intervention targets) strongly outperform the baselines in this setting. From top to bottom: ancestral relations, direct causes, confounders. From left to right: ROC curves, PR curves for presence of feature, PR curves for absence of feature.\label{fig:simul_p4_q2_acyclic_mc_asdjci}}
\end{figure}
\begin{figure}[t]
\centerline{%
\includegraphics[width=0.24\textwidth]{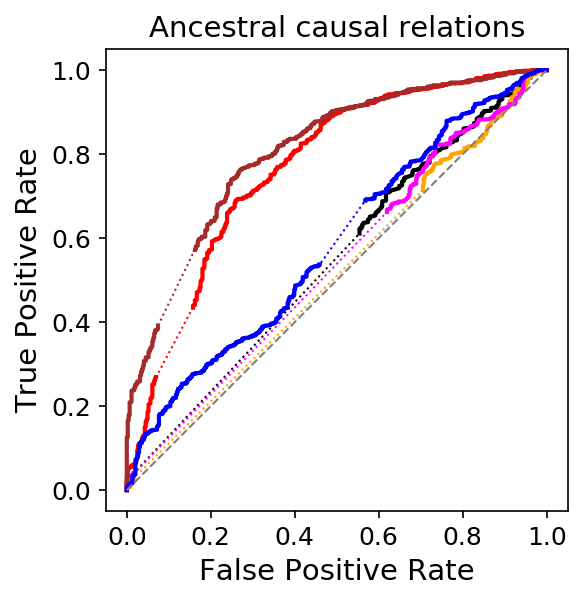}
\includegraphics[width=0.24\textwidth]{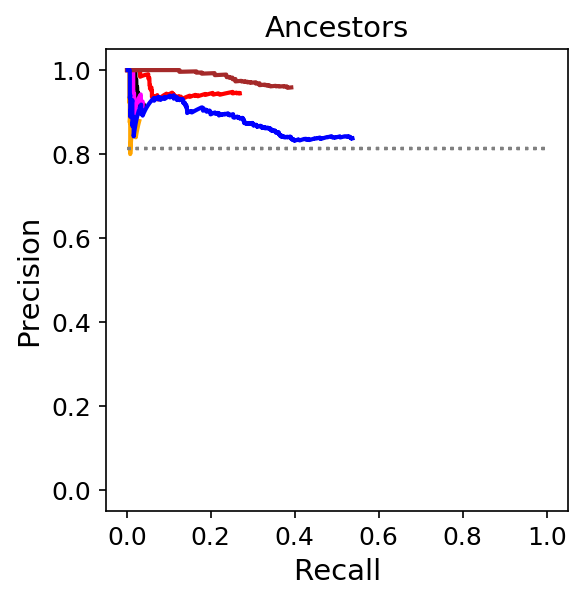}
\includegraphics[width=0.24\textwidth]{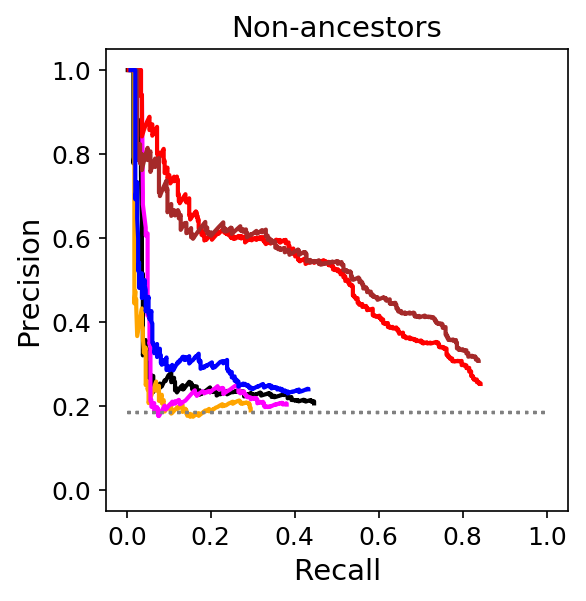}
\includegraphics[width=0.24\textwidth]{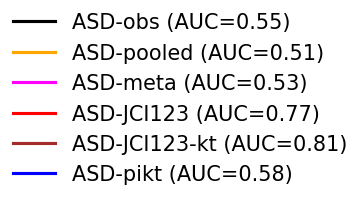}
}
\centerline{%
\includegraphics[width=0.24\textwidth]{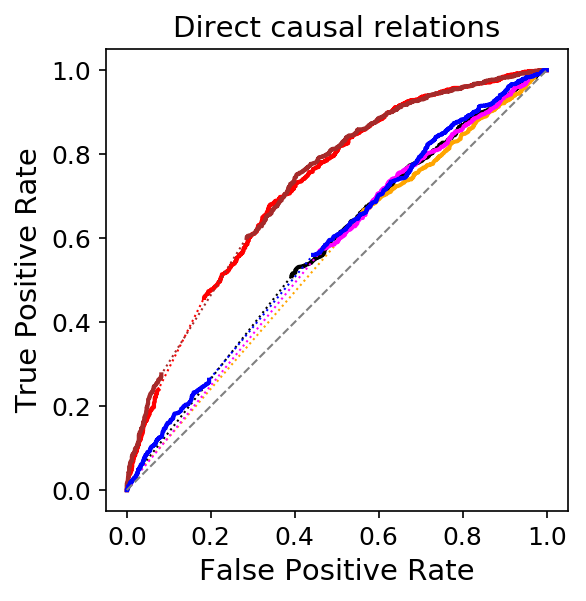}
\includegraphics[width=0.24\textwidth]{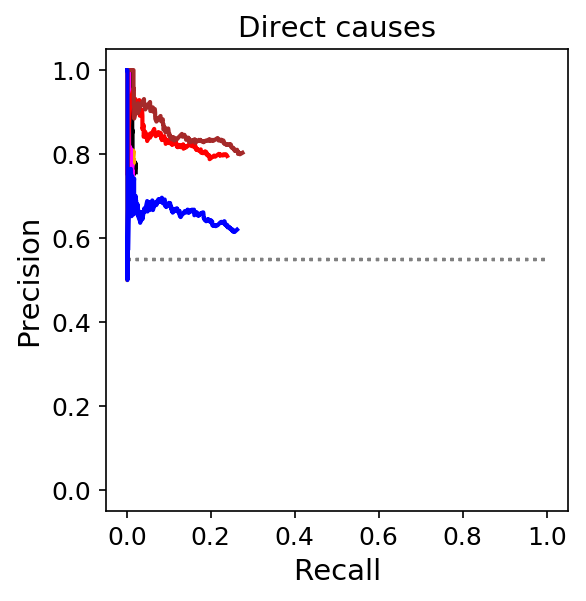}
\includegraphics[width=0.24\textwidth]{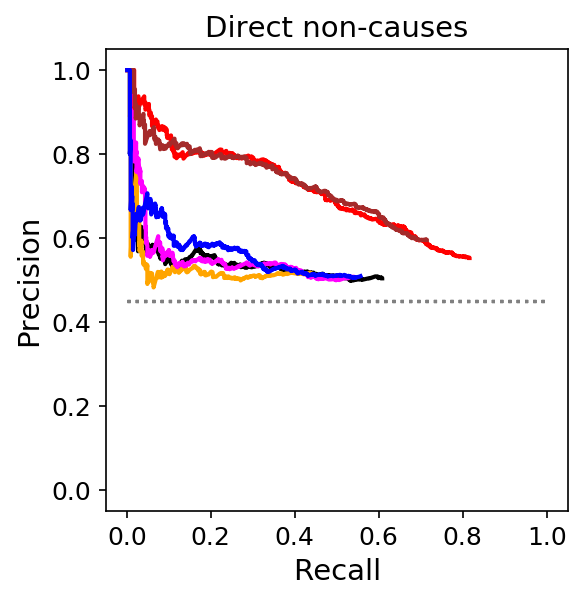}
\includegraphics[width=0.24\textwidth]{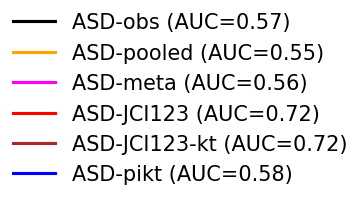}
}
\centerline{%
\includegraphics[width=0.24\textwidth]{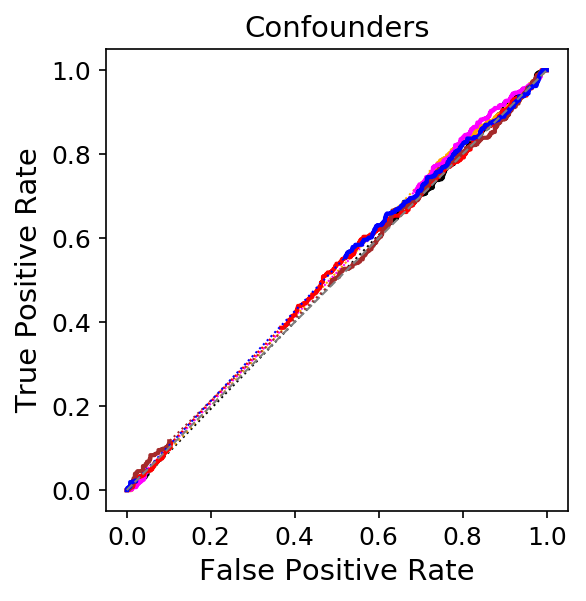}
\includegraphics[width=0.24\textwidth]{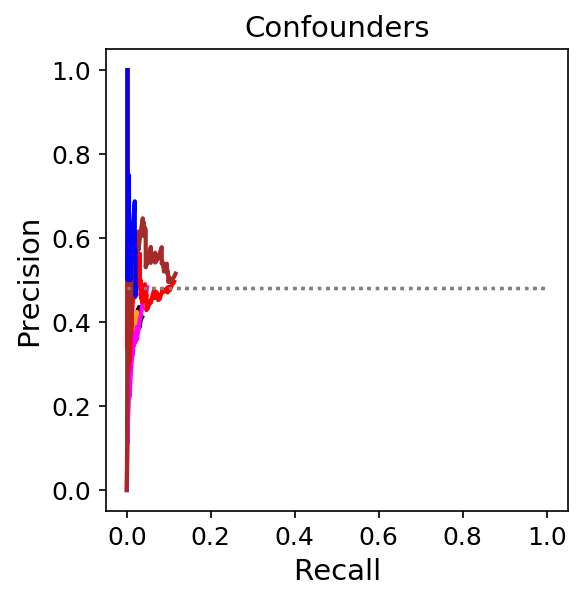}
\includegraphics[width=0.24\textwidth]{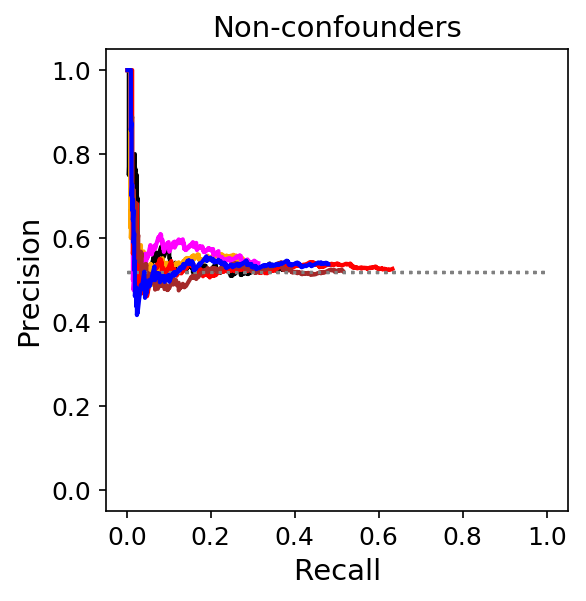}
\includegraphics[width=0.24\textwidth]{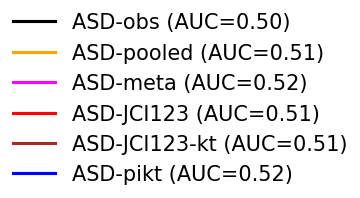}
}
  \caption{\boldcap{Results of some ASD variants (cyclic, causal mechanism changes)} for small models. The two JCI variants substantially outperform
the baselines in this setting.\label{fig:simul_p4_q2_cyclic_mc_asdjci}}
\end{figure}

We start by showing off the advantage that JCI can offer over existing methods. We first consider only ASD variants
because this most clearly shows the impact of how one merges data from different contexts and how one treats
the context variables, since the other aspects of the causal discovery algorithm are the same for all ASD variants. 
In Figure~\ref{fig:simul_p4_q2_acyclic_mc_asdjci} we present results for several ASD variants
for acyclic models with causal mechanism changes. We compare the JCI variants \alg{ASD-JCI123} (unknown
intervention targets) and \alg{ASD-JCI123kt} (known intervention targets) with the available baselines,
\alg{ASD-obs} (observational data only), \alg{ASD-pooled} (pooled data from all context treated as if
they were all observational, context variables not included), \alg{ASD-meta} (using Fisher's method to
combine $p$-values from conditional independence tests in separate contexts), and \alg{ASD-pikt} (which
assumes that interventions are perfect and uses knowledge of intervention targets).
The tasks of predicting ancestral causal relations and direct causal relations show relatively similar 
ROC and PR curves. Predicting the absence or presence of confounders is a more challenging task.

The three baselines \alg{ASD-obs}, \alg{ASD-pooled} and \alg{ASD-meta} show very similar performance behaviors.
In particular, for the tasks of predicting the presence of the features, these baselines perform poorly, not
much better than random guessing. This is partially due to the small sample size, but also to the fact that
many relationships are simply not identifiable from purely observational data alone.
\alg{ASD-pikt} even performs poorly on nearly all prediction tasks in this simulation setting because it 
incorrectly assumes that the interventions are perfect.
The two JCI variants, on the other hand, strongly outperform the baselines and obtain very high precisions. 
In particular, even without knowing the intervention targets, \alg{ASD-JCI123} manages to predict
the presence of (direct and indirect) causal relations at maximum precision for low recall. Exploiting knowledge of
the intervention targets, \alg{ASD-JCI123-kt} obtains an even more impressive precision 
for predicting ancestral causal relations for a large range of recalls. 
This illustrates the significant improvement in precision that JCI can yield.

Figure~\ref{fig:simul_p4_q2_cyclic_mc_asdjci} shows a similar picture in the cyclic setting, where all ASD variants
make use of $\sigma$-separation \citep{ForreMooij_UAI_18}. The task of predicting the presence of 
ancestral relations is easier than in the acyclic setting, because for most pairs of system variables, one is
ancestor of the other due to the cycles. The task of predicting the absence of ancestral
relations, on the other hand, is more challenging. Detecting the presence or absence of confounders in this setting
has become nearly impossible, for any method. For the other tasks, the JCI approach again
shows substantially improved precisions compared to the baselines. 

\subsubsection{ASD-JCI vs.\ Baselines (Perfect Interventions)}

\begin{figure}[t]
\centerline{%
\includegraphics[width=0.24\textwidth]{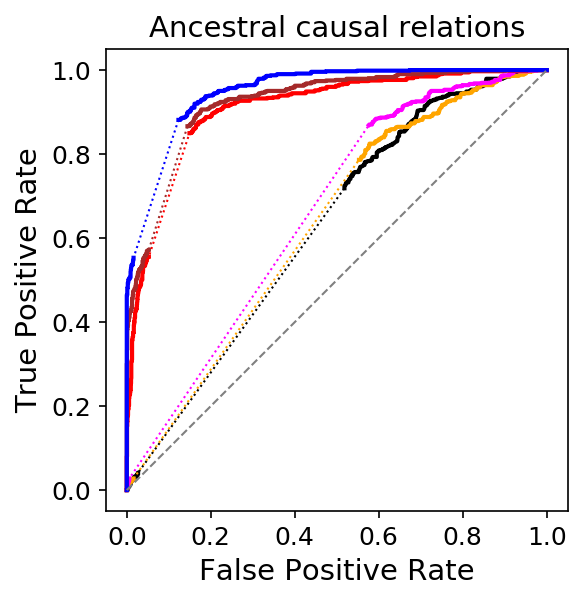}
\includegraphics[width=0.24\textwidth]{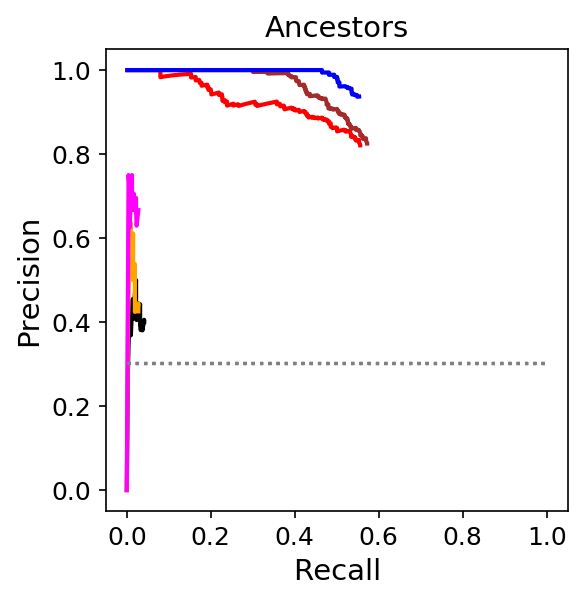}
\includegraphics[width=0.24\textwidth]{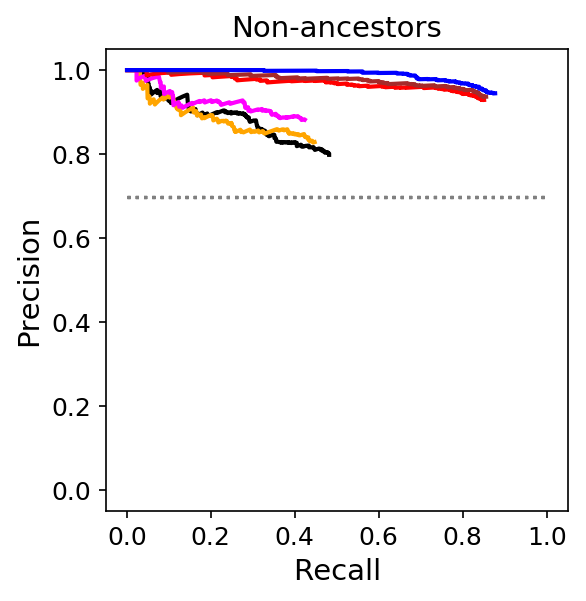}
\includegraphics[width=0.24\textwidth]{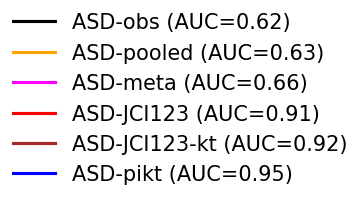}
}
\centerline{%
\includegraphics[width=0.24\textwidth]{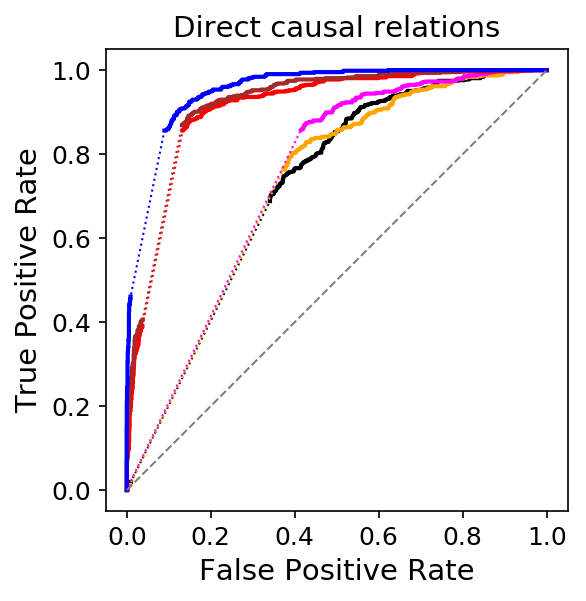}
\includegraphics[width=0.24\textwidth]{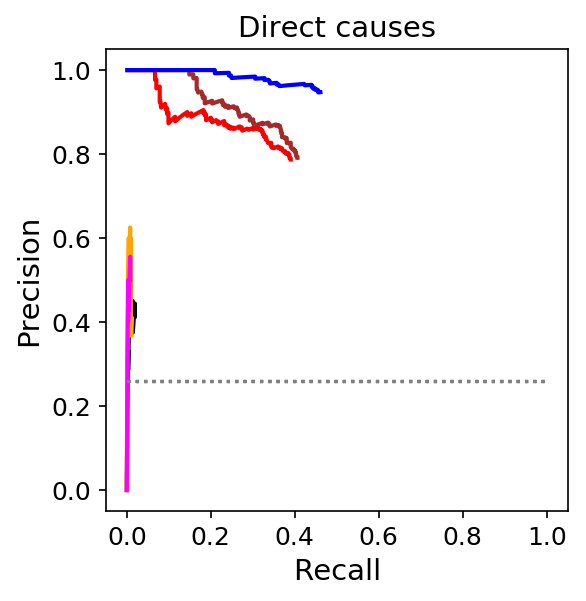}
\includegraphics[width=0.24\textwidth]{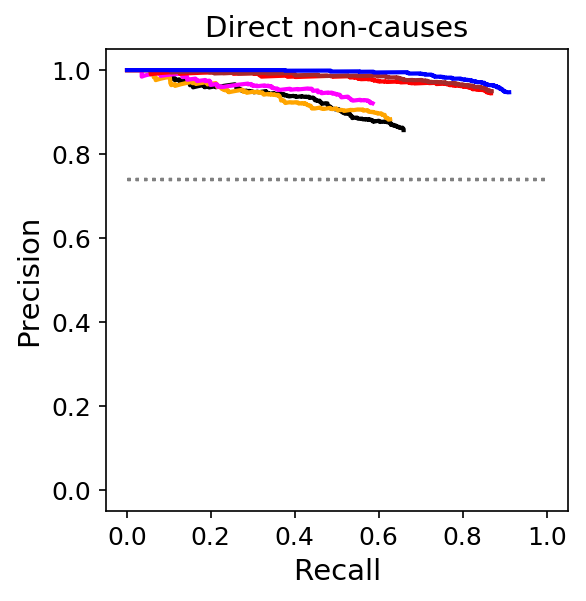}
\includegraphics[width=0.24\textwidth]{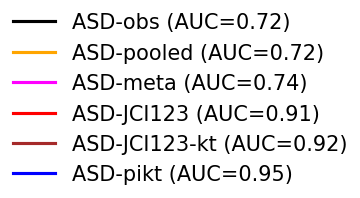}
}
\centerline{%
\includegraphics[width=0.24\textwidth]{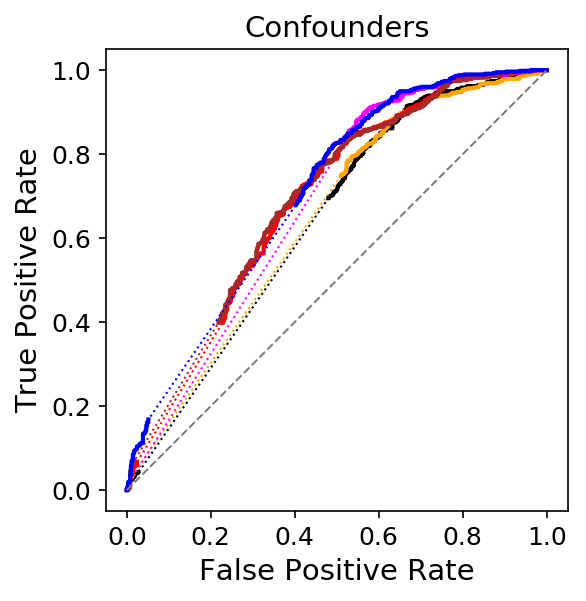}
\includegraphics[width=0.24\textwidth]{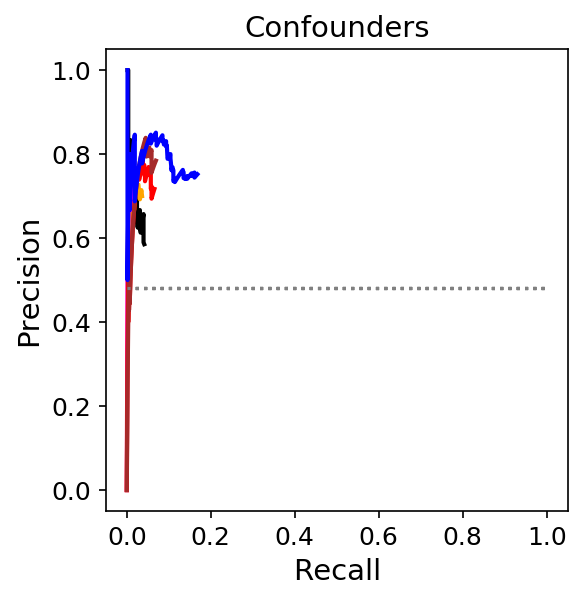}
\includegraphics[width=0.24\textwidth]{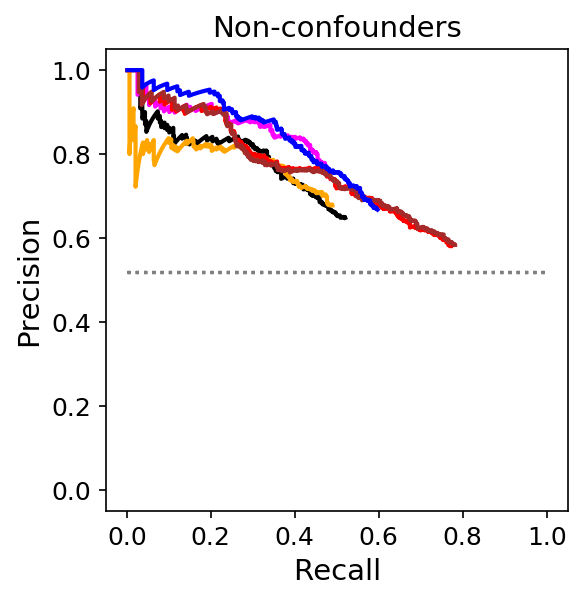}
\includegraphics[width=0.24\textwidth]{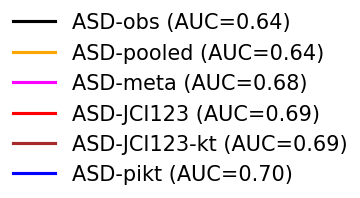}
}
  \caption{\boldcap{Results of some ASD variants (acyclic, perfect interventions)} for small models. \alg{ASD-pikt}, which takes
into account the perfect nature of the interventions, is the best performing method in this setting. The JCI
variants do not assume perfect interventions, but still yield a vast improvement of the precision of the
predicted features with respect to the other baselines.\label{fig:simul_p4_q2_acyclic_pi_asdjci}}
\end{figure}

\begin{figure}[t]
\centerline{%
\includegraphics[width=0.24\textwidth]{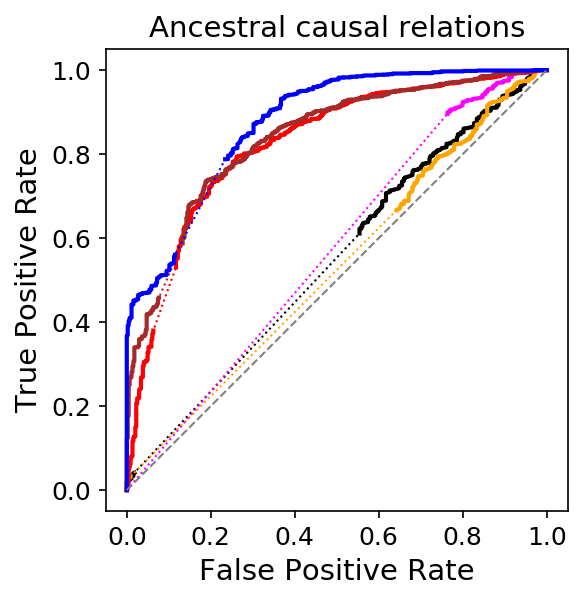}
\includegraphics[width=0.24\textwidth]{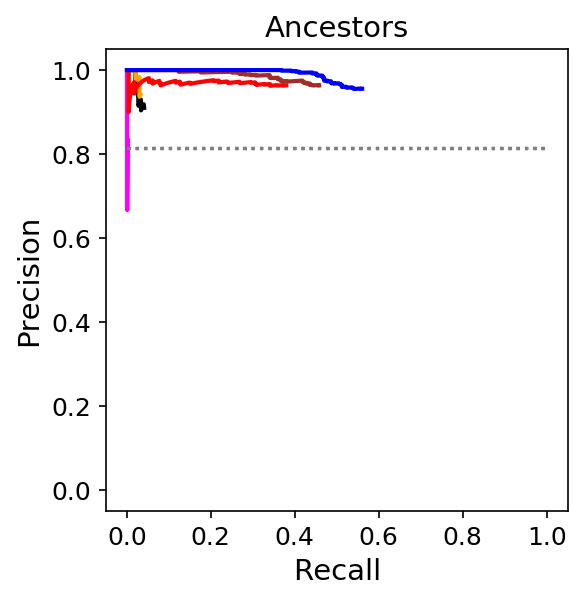}
\includegraphics[width=0.24\textwidth]{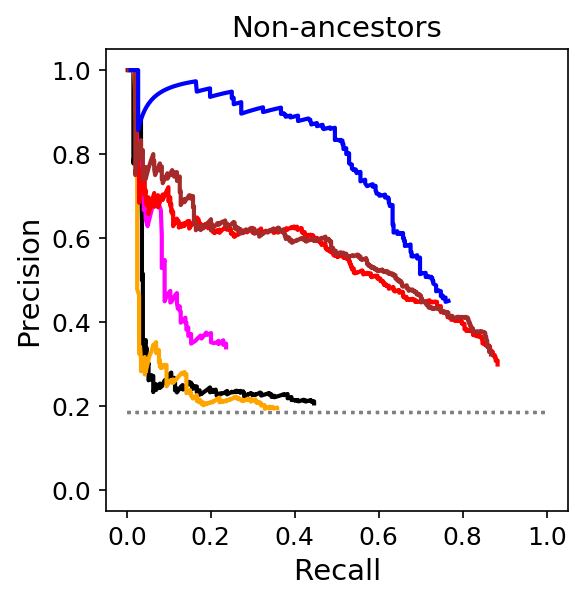}
\includegraphics[width=0.24\textwidth]{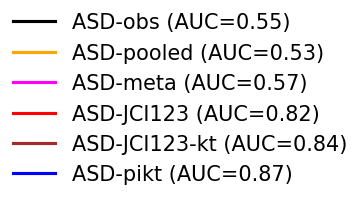}
}
\centerline{%
\includegraphics[width=0.24\textwidth]{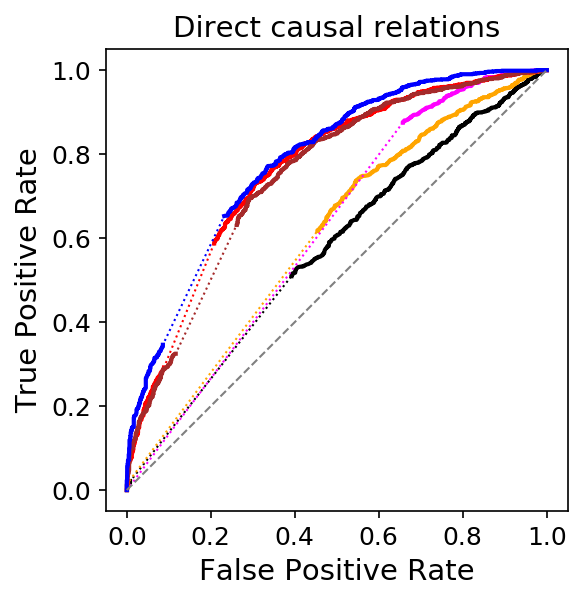}
\includegraphics[width=0.24\textwidth]{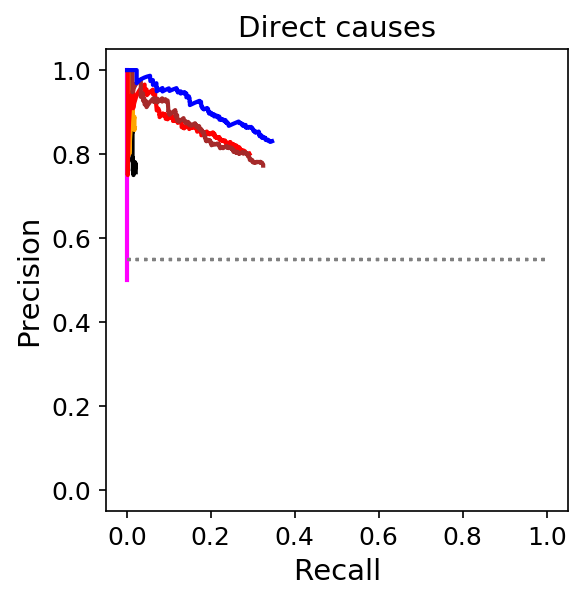}
\includegraphics[width=0.24\textwidth]{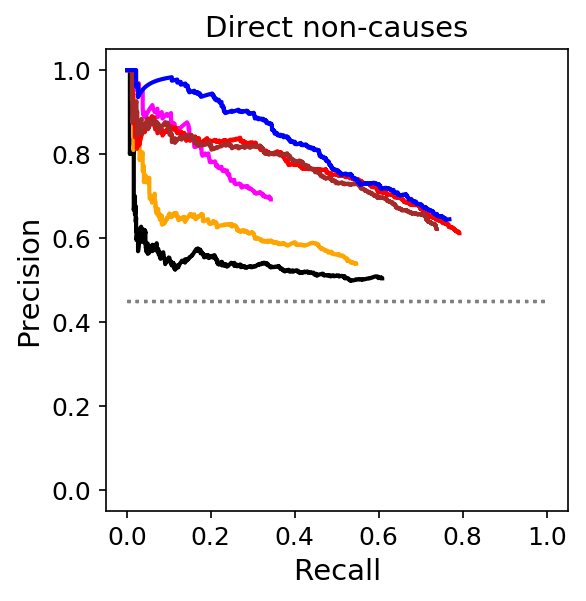}
\includegraphics[width=0.24\textwidth]{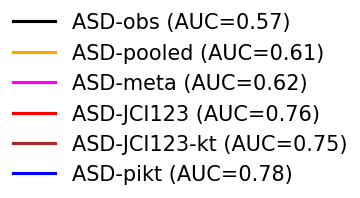}
}
\centerline{%
\includegraphics[width=0.24\textwidth]{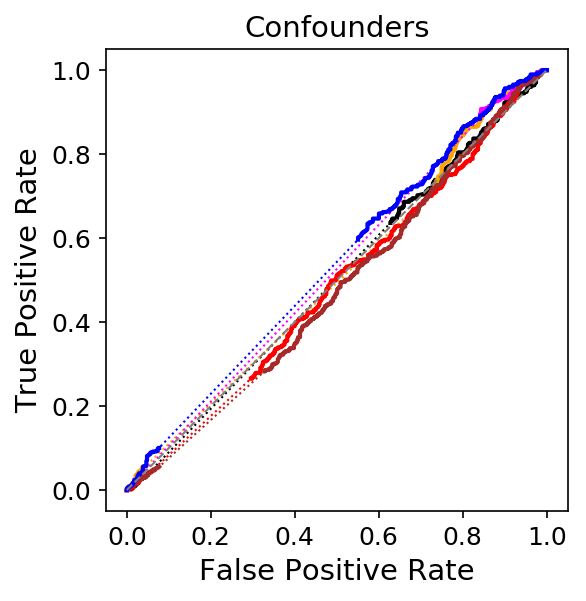}
\includegraphics[width=0.24\textwidth]{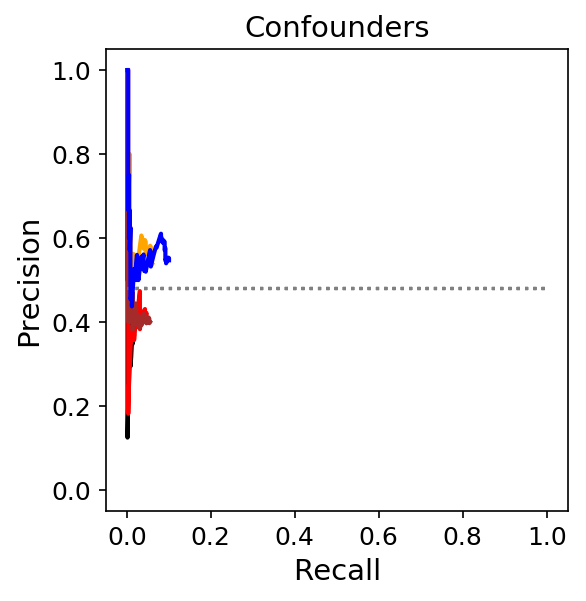}
\includegraphics[width=0.24\textwidth]{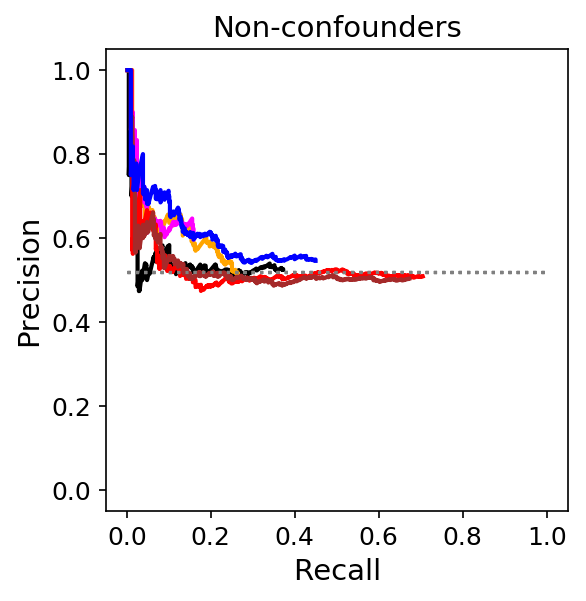}
\includegraphics[width=0.24\textwidth]{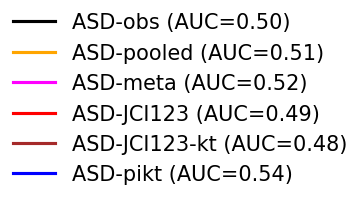}
}
  \caption{\boldcap{Results of some ASD variants (cyclic, perfect interventions)} for small models. We see a similar picture as in
the acyclic case in Figure \ref{fig:simul_p4_q2_acyclic_pi_asdjci}.\label{fig:simul_p4_q2_cyclic_pi_asdjci}}
\end{figure}

Figures \ref{fig:simul_p4_q2_acyclic_pi_asdjci} and \ref{fig:simul_p4_q2_cyclic_pi_asdjci} show 
results for respectively the acyclic and cyclic settings, but now for perfect interventions 
with known targets rather than causal mechanism changes.


In these perfect intervention scenarios, the JCI variants again obtain a much higher precision than any of the baselines, 
with the sole exception of \alg{ASD-pikt}. In this setting, the latter method successfully exploits the assumed perfect
nature of the interventions, thereby outperforming the JCI variants that do not make any assumption
about the nature of the intervention. 
However, a significant disadvantage of \alg{ASD-pikt} in practice is that its assumption of perfect interventions with 
known targets may not be valid. As we already saw in Section~\ref{sec:results_asd-jci_mc}, \alg{ASD-pikt} then breaks down, 
in contrast to the JCI variants.

For the cyclic setting with perfect interventions (Figure~\ref{fig:simul_p4_q2_cyclic_pi_asdjci}), 
we observe that predicting the presence of confounders still seems impossible for all methods, 
but predicting their absence seems at least feasible in principle (although it seems a very challenging task).
\alg{ASD-pikt} again obtains the highest precisions, followed by the JCI variants.

\subsubsection{Influence of JCI Assumptions}

\begin{figure}[t]
\centerline{%
\includegraphics[width=0.24\textwidth]{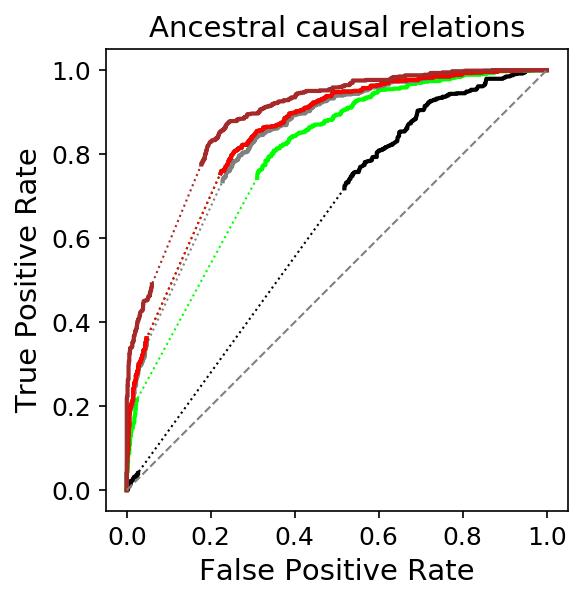}
\includegraphics[width=0.24\textwidth]{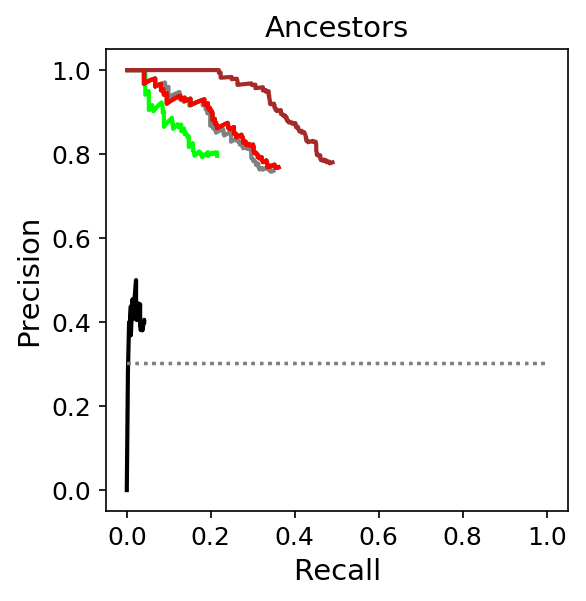}
\includegraphics[width=0.24\textwidth]{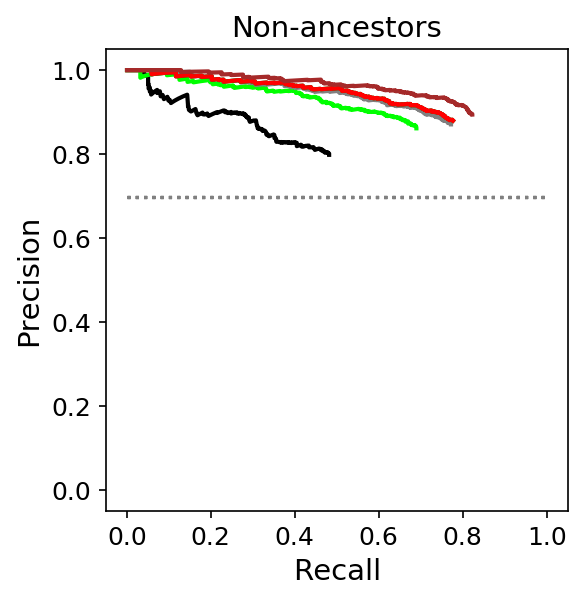}
\includegraphics[width=0.24\textwidth]{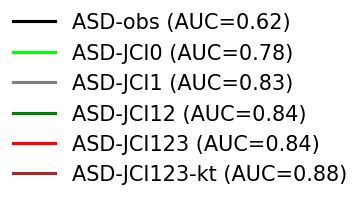}
}
\centerline{%
\includegraphics[width=0.24\textwidth]{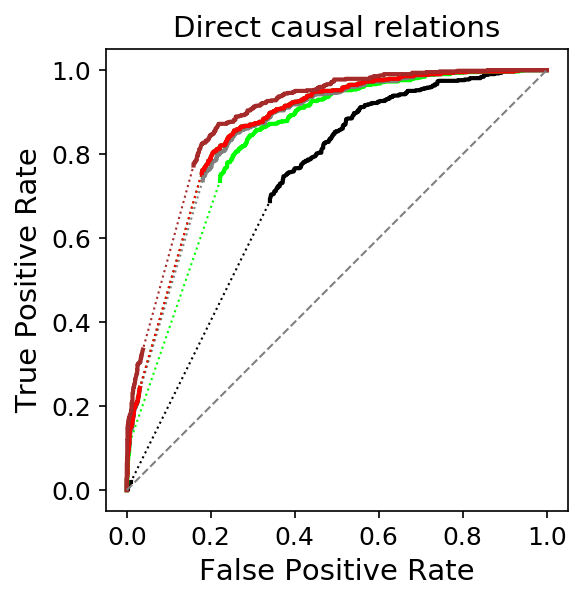}
\includegraphics[width=0.24\textwidth]{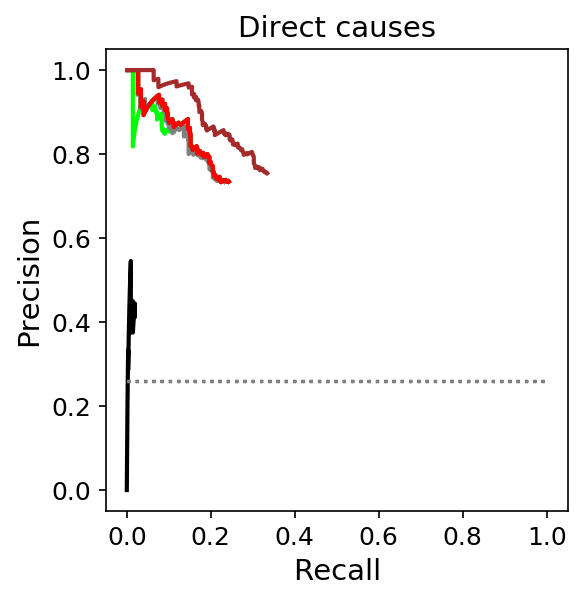}
\includegraphics[width=0.24\textwidth]{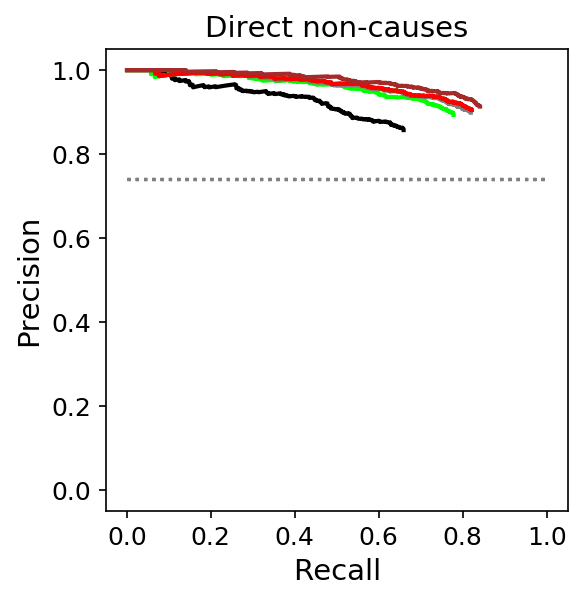}
\includegraphics[width=0.24\textwidth]{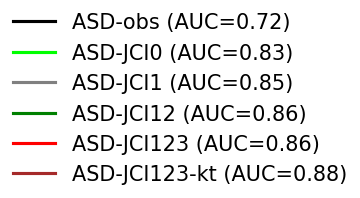}
}
\centerline{%
\includegraphics[width=0.24\textwidth]{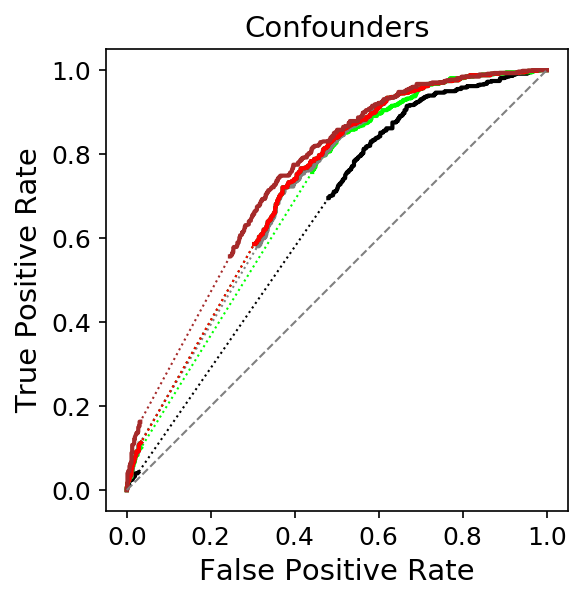}
\includegraphics[width=0.24\textwidth]{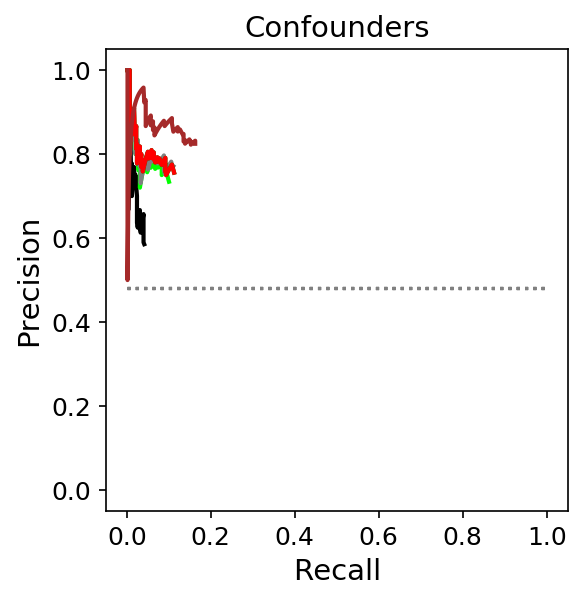}
\includegraphics[width=0.24\textwidth]{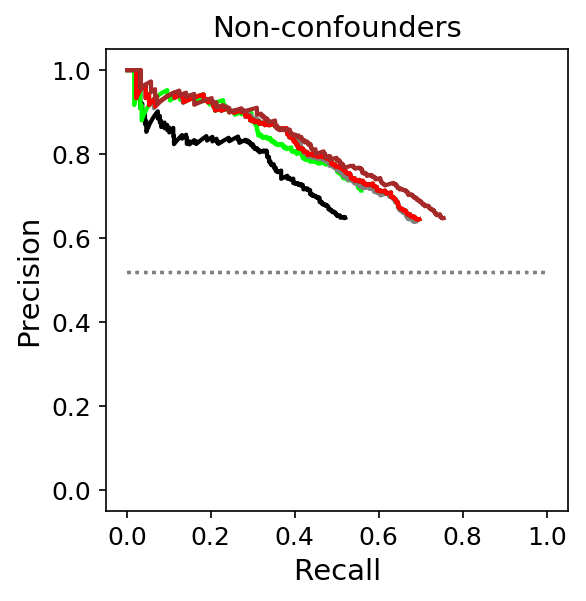}
\includegraphics[width=0.24\textwidth]{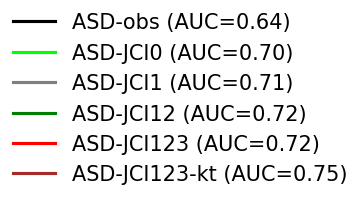}
}
  \caption{\boldcap{Influence of JCI assumptions for ASD-JCI (acyclic, causal mechanism changes)} for small models. Exploiting more prior knowledge is seen to lead to better results. For reference, the non-JCI baseline \alg{ASD-obs} is shown which only uses observational data.\label{fig:simul_p4_q2_acyclic_mc_asdjciN}}
\end{figure}

We now investigate in more detail which JCI assumptions are responsible for the excellent performance of the 
JCI variants of ASD. Figure~\ref{fig:simul_p4_q2_acyclic_mc_asdjciN} shows that, as expected, the more
prior knowledge about the context variables is used, the better the predictions become. 

However, surprisingly, the largest boost in precision with respect to the observational baseline
is due to simply pooling the data and adding
the context variables: \alg{ASD-JCI0} already strongly improves over \alg{ASD-obs}.
Adding more background knowledge regarding the nature of the context variables helps to improve 
the results further. JCI Assumption~\ref{ass:uncaused} yields a marginal improvement. JCI 
Assumptions~\ref{ass:unconfounded} (and \ref{ass:dependences}) do not lead to any 
further improvements for discovering the causal relations between system variables in this setting, though. 
Exploiting knowledge of the intervention targets, on the other hand, turns out to be very 
helpful for getting highly accurate predictions for ancestral relations between system variables, 
and also significantly improves the precision of predicting direct causal relations and confounders.
We have shown here only the results for the acyclic setting with causal
mechanism changes because we obtained qualitatively similar results for the other simulation settings.

We also investigated variants of \alg{ASD-JCI0}, \alg{ASD-JCI1} and
\alg{ASD-JCI12} where we did not perform any independence tests on the context
variables, i.e., using conditional independence testing scheme ``NC'' rather
than ``A''. This is possible because ASD is capable of handling
incomplete inputs. We obtained almost identical results (in all simulation
settings considered) to the standard variants of those methods in which we do
perform conditional independence tests on the context variables themselves (not
shown here).

\subsubsection{Multiple Context Variables vs.\ Single (Merged) Context Variable}

\begin{figure}[t]
\centerline{%
\includegraphics[width=0.24\textwidth]{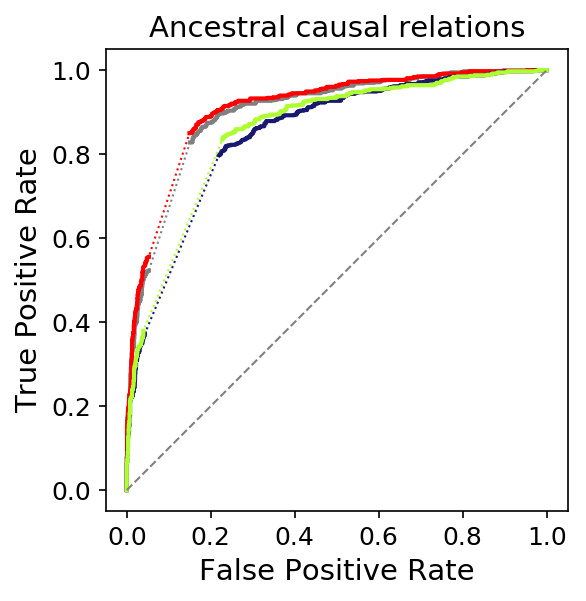}
\includegraphics[width=0.24\textwidth]{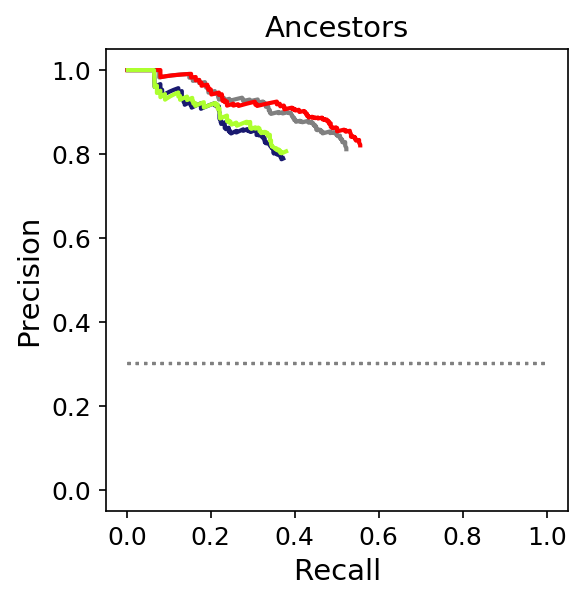}
\includegraphics[width=0.24\textwidth]{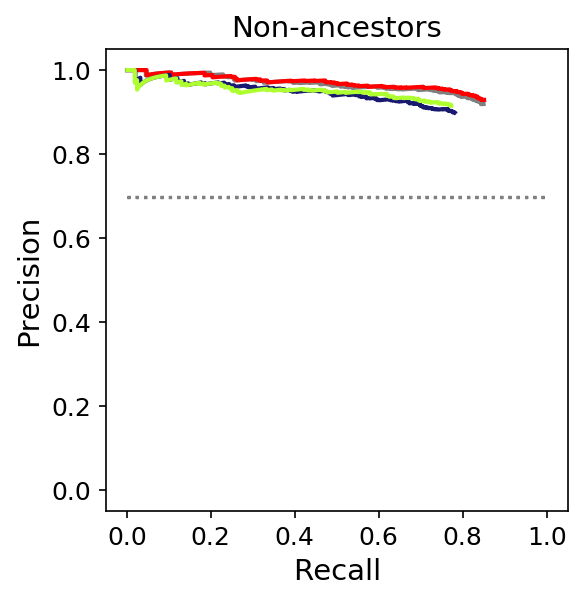}
\includegraphics[width=0.24\textwidth]{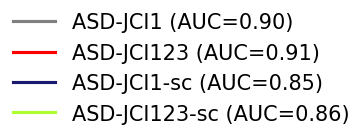}
}
\centerline{%
\includegraphics[width=0.24\textwidth]{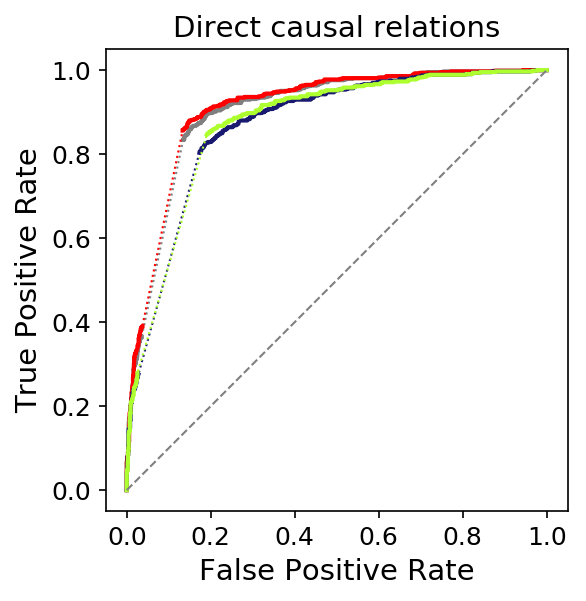}
\includegraphics[width=0.24\textwidth]{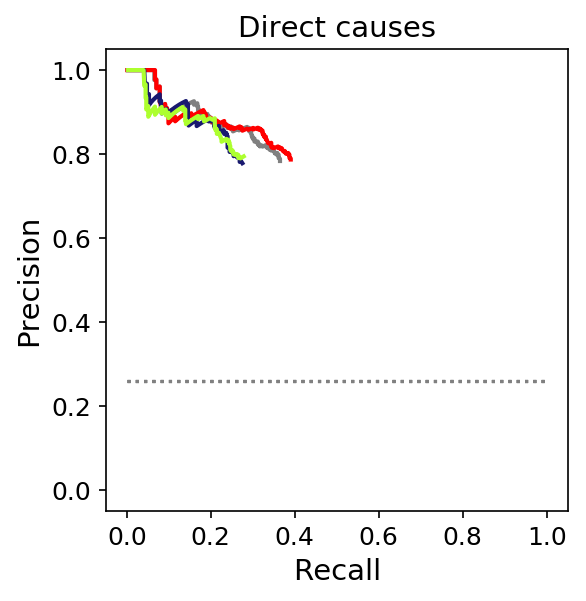}
\includegraphics[width=0.24\textwidth]{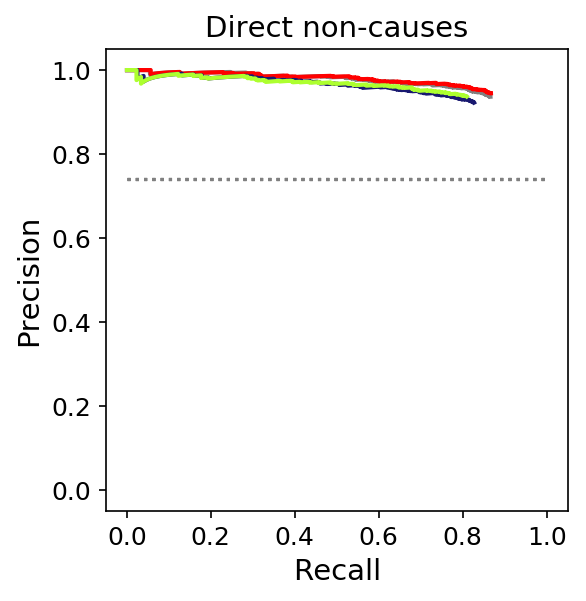}
\includegraphics[width=0.24\textwidth]{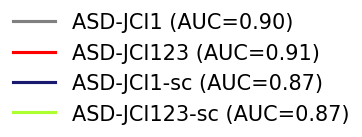}
}
\centerline{%
\includegraphics[width=0.24\textwidth]{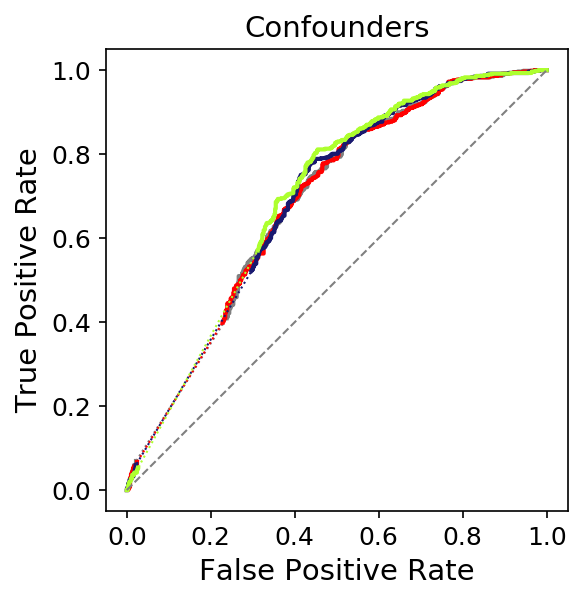}
\includegraphics[width=0.24\textwidth]{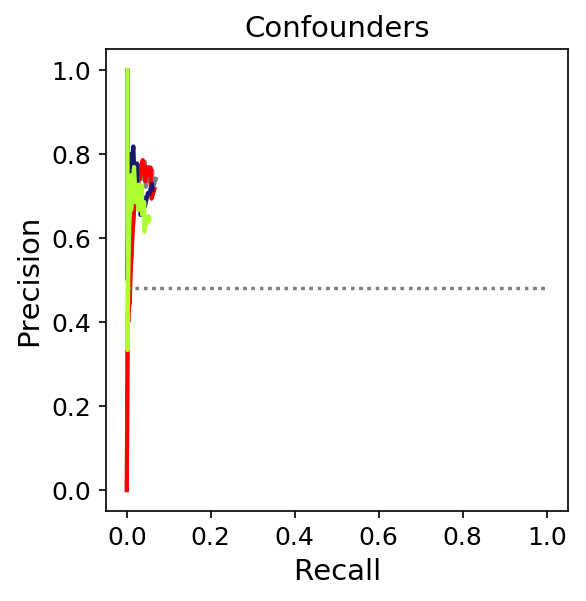}
\includegraphics[width=0.24\textwidth]{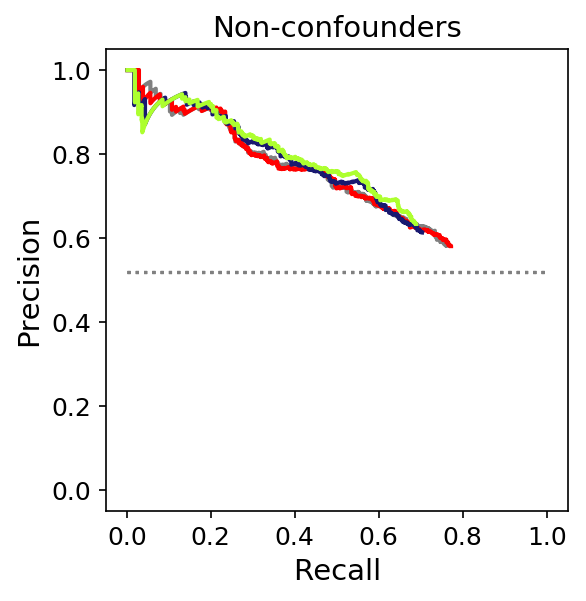}
\includegraphics[width=0.24\textwidth]{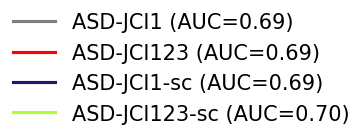}
}
  \caption{\boldcap{Multiple context variables vs.\ single (merged) context variable for ASD-JCI (acyclic, perfect interventions)} for small models. Exploiting multiple context variables leads to better results than using a single (merged) context variable.\label{fig:simul_p4_q2_acyclic_pi_asdjcisc}}
\end{figure}

Figure~\ref{fig:simul_p4_q2_acyclic_pi_asdjcisc} shows that for ASD-JCI variants, exploiting multiple 
context variables leads to better results than using a single (merged) context variable, as expected. Similar
results hold for the cyclic settings and for causal mechanism changes (not shown).

\subsubsection{FCI Variants}

\begin{figure}
\centerline{%
\includegraphics[width=0.24\textwidth]{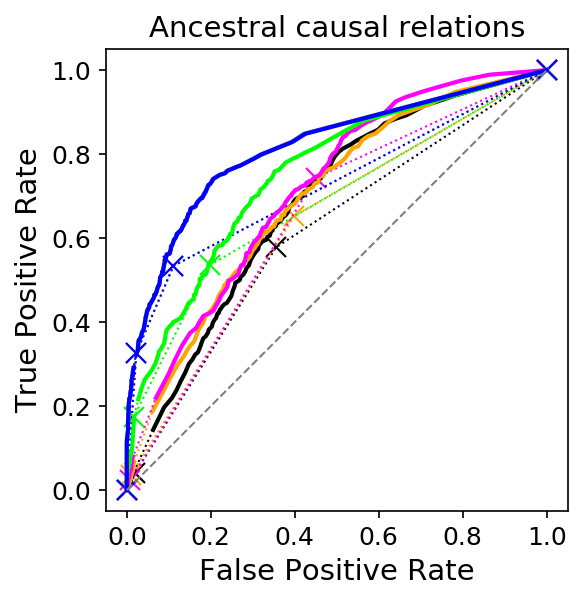}
\includegraphics[width=0.24\textwidth]{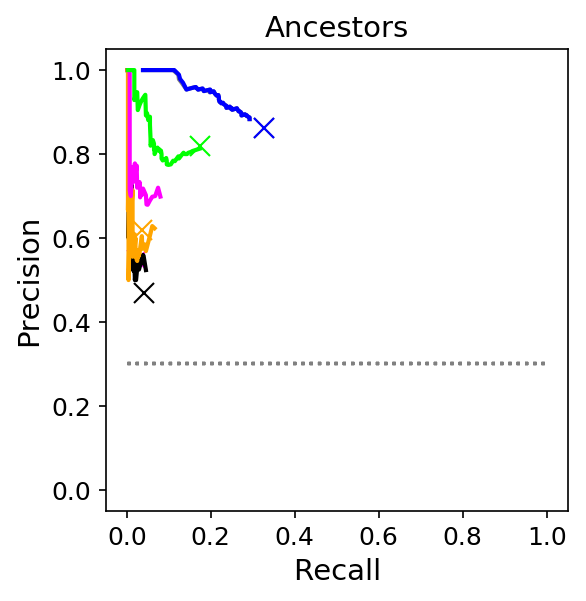}
\includegraphics[width=0.24\textwidth]{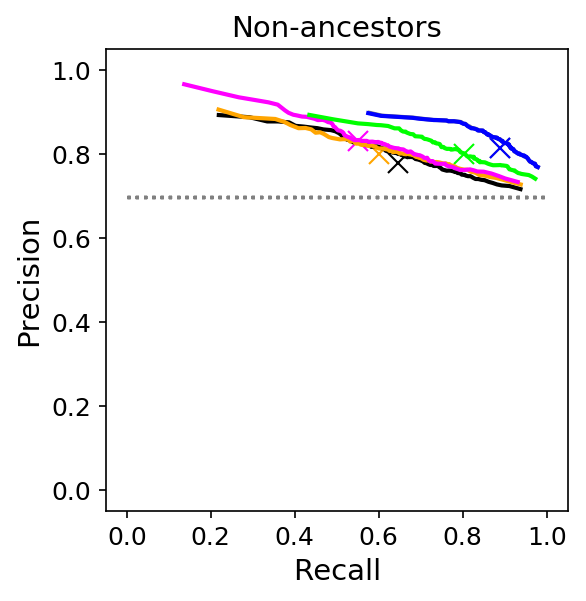}
\includegraphics[width=0.24\textwidth]{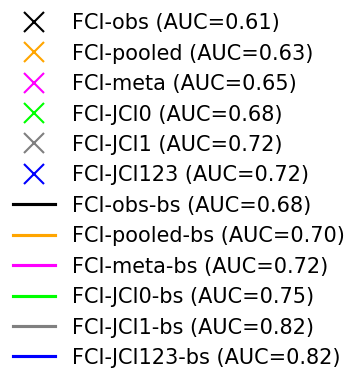}
}
\caption{\boldcap{Results of FCI variants (acyclic, perfect interventions)} for small models.\label{fig:simul_p4_q2_acyclic_pi_FCI}}
\end{figure}

Figure~\ref{fig:simul_p4_q2_acyclic_pi_FCI} shows the results for the various
FCI variants. FCI is seen to be somewhat less accurate 
than ASD, but bootstrapping helps to boost precision for lower recalls. 
Similarly to ASD, we conclude that the JCI variants of FCI substantially
outperform the non-JCI variants. 
\alg{FCI-JCI1} and \alg{FCI-JCI123} seem to yield identical results in this setting. 

Results for causal mechanism changes are very similar to those for perfect
interventions, and therefore are not shown here. The results for the cyclic 
setting are also not shown, because FCI was designed for the acyclic setting.

\subsubsection{LCD and ICP}

\begin{figure}
\centerline{%
\includegraphics[width=0.24\textwidth]{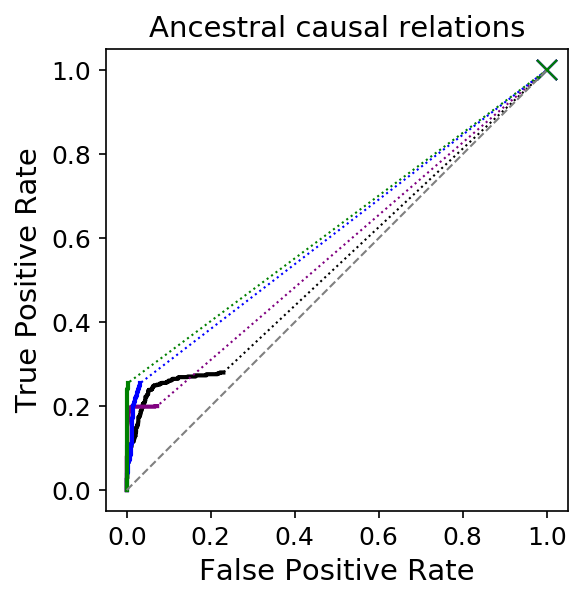}
\includegraphics[width=0.24\textwidth]{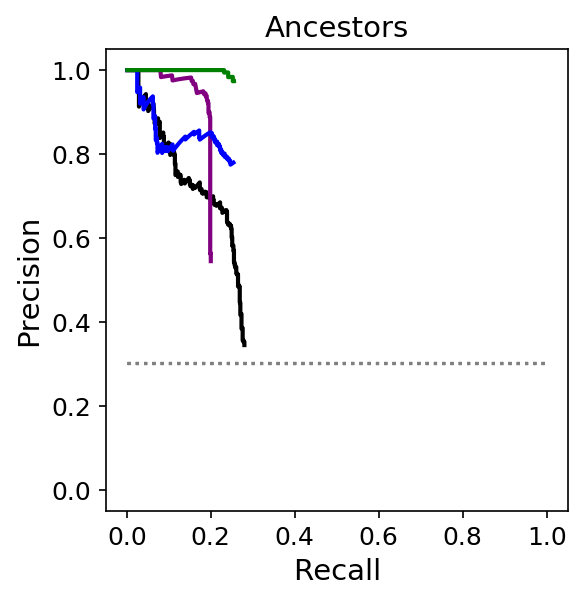}
\includegraphics[width=0.24\textwidth]{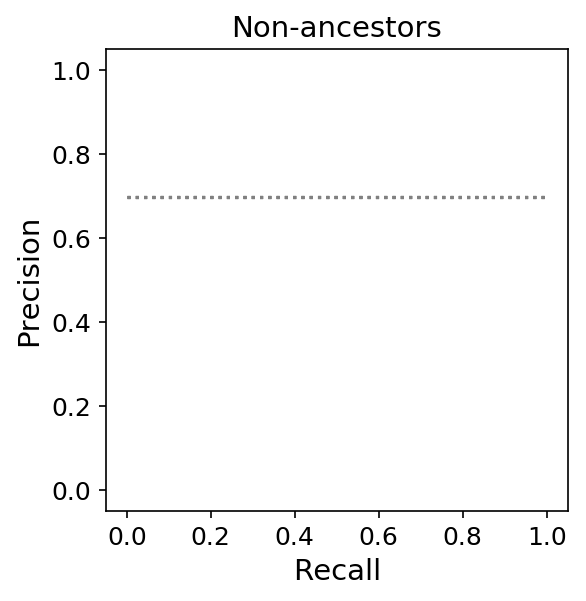}
\includegraphics[width=0.24\textwidth]{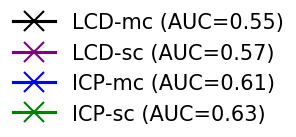}
}
\centerline{%
\includegraphics[width=0.24\textwidth]{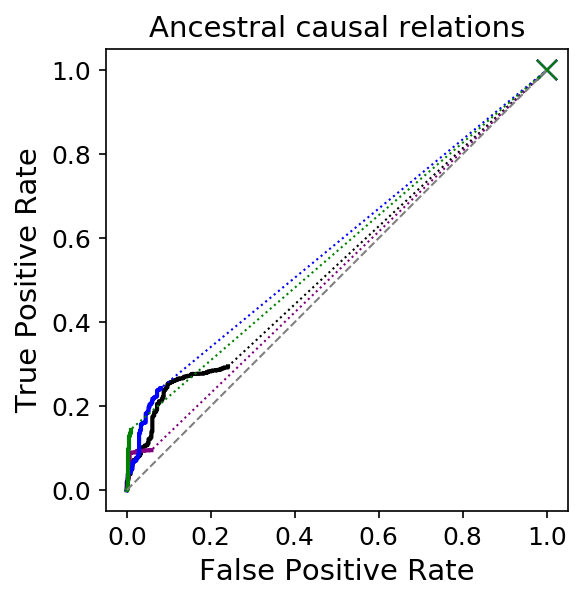}
\includegraphics[width=0.24\textwidth]{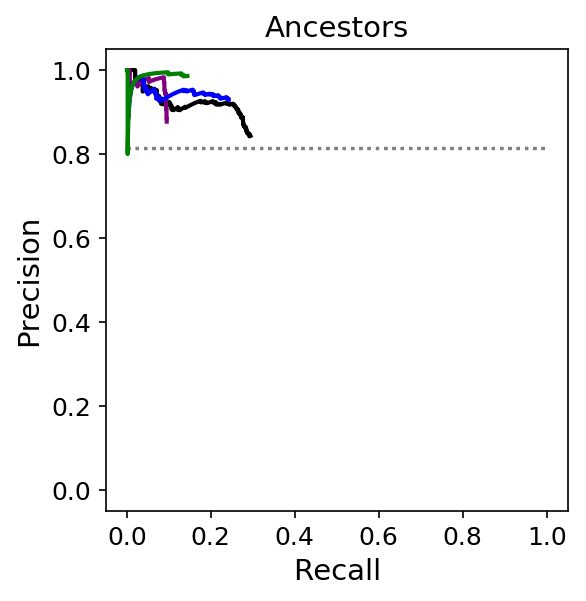}
\includegraphics[width=0.24\textwidth]{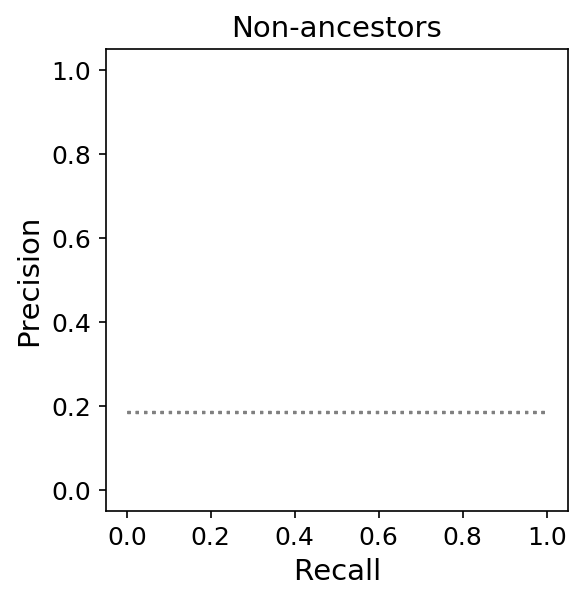}
\includegraphics[width=0.24\textwidth]{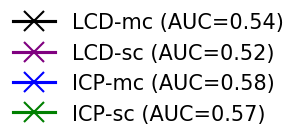}
}
\caption{\boldcap{Multiple context variables vs.\ single (merged) one for \alg{LCD} and \alg{ICP} (perfect interventions)} for small models.
LCD and ICP both benefit from merging the context variables into a single one for the
task of predicting the presence of ancestral relations. Top: acyclic; bottom: cyclic.\label{fig:simul_p4_q2_pi_lcdicp}}
\end{figure}

Figure~\ref{fig:simul_p4_q2_pi_lcdicp} shows the results of the LCD and ICP variants for the task of predicting ancestral relations.
We only show the results for perfect interventions, as the results for causal
mechanism changes are similar. LCD and ICP both only can predict 
the presence of ancestral relations, not their absence. LCD and ICP apparently benefit from merging the context variables into a single one.
A possible explanation of this phenomenon could be that a combination of conditional 
independence tests of the form $C_k \CI X_{i'} \given X_i$ with each $C_k$ binary (and both $X_i$ and 
$X_{i'}$ real-valued) might be less reliable than a single test $\B{C} \CI X_{i'} \given X_i$ where 
$\B{C}$ is categorical with $\gg 2$ states.
Another observation we made is that bootstrapping does lead to only marginal improvements for these methods (not shown).

\subsubsection{Varying the Number of Context Variables}

\begin{figure}[t]
\centerline{%
\includegraphics[width=0.24\textwidth]{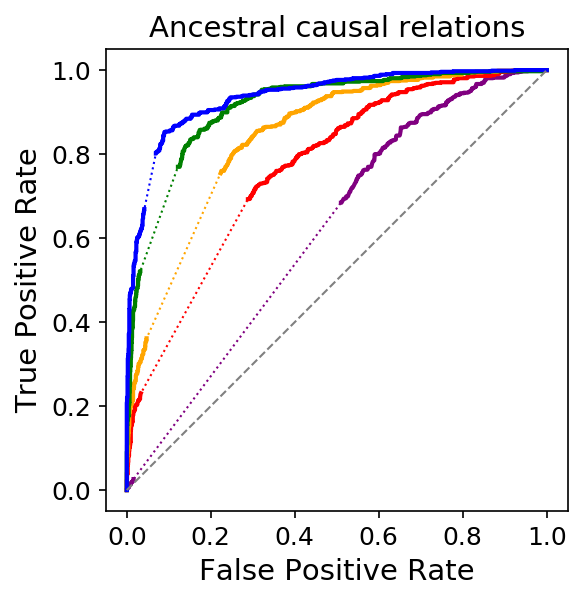}
\includegraphics[width=0.24\textwidth]{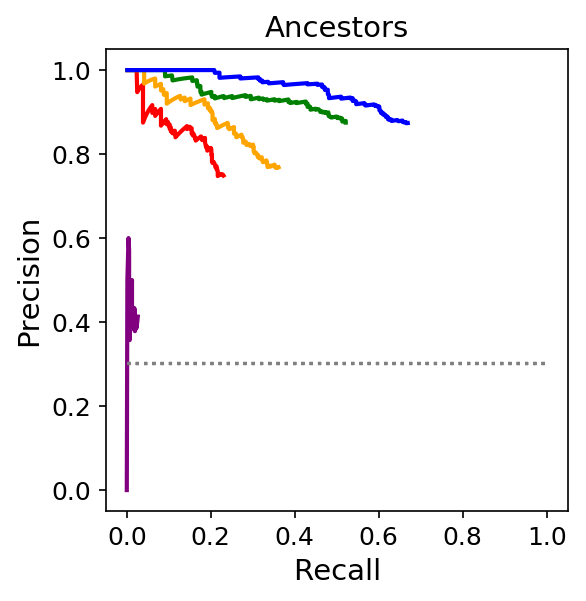}
\includegraphics[width=0.24\textwidth]{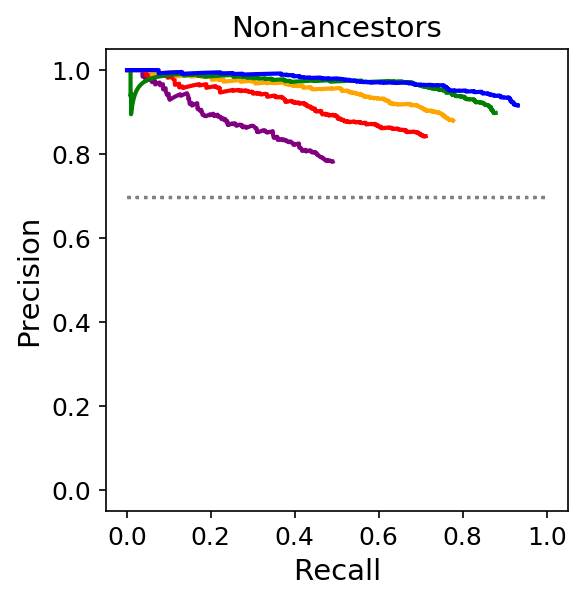}
\includegraphics[width=0.24\textwidth]{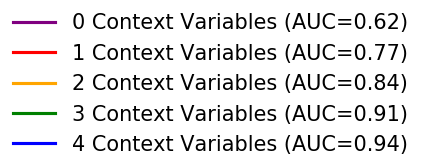}
}
\centerline{%
\includegraphics[width=0.24\textwidth]{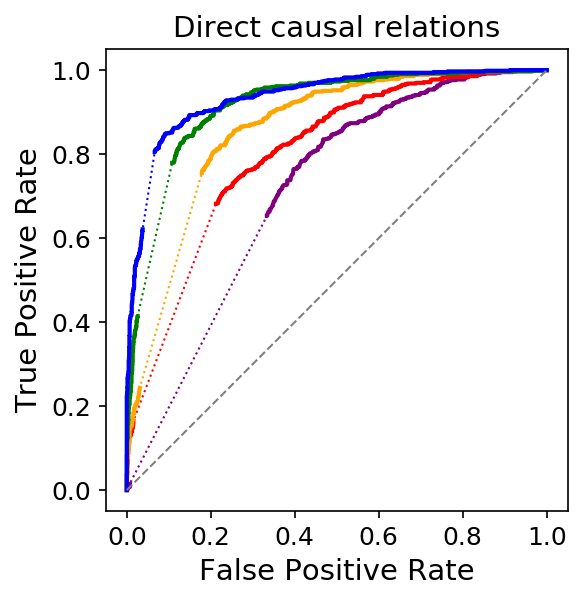}
\includegraphics[width=0.24\textwidth]{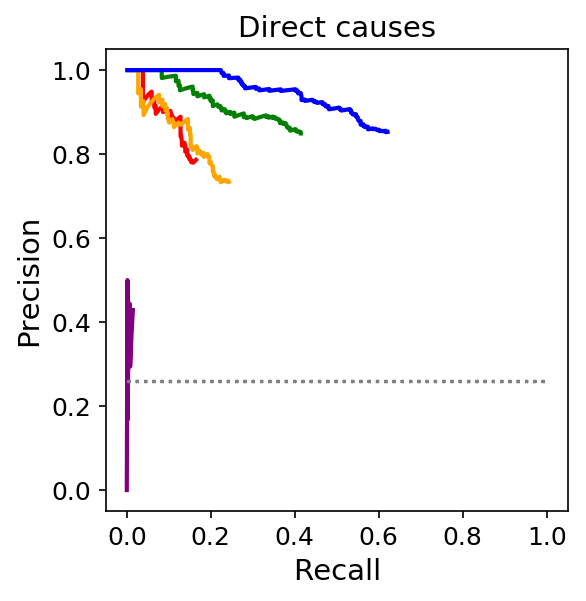}
\includegraphics[width=0.24\textwidth]{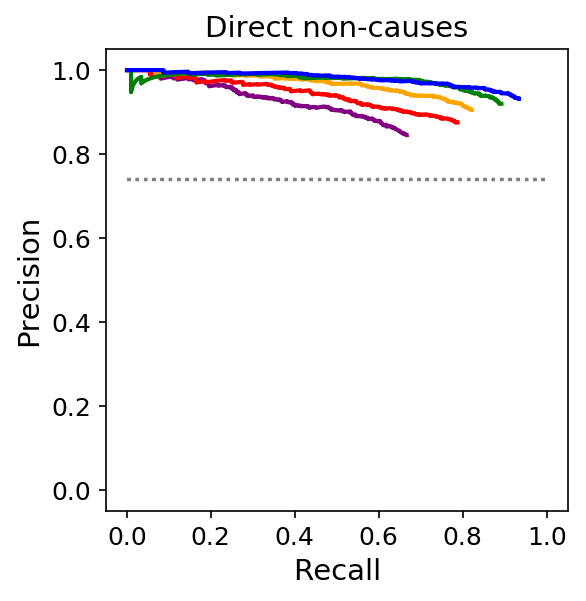}
\includegraphics[width=0.24\textwidth]{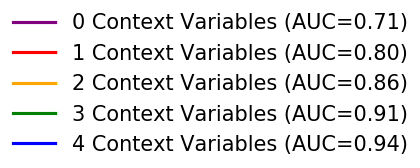}
}
\centerline{%
\includegraphics[width=0.24\textwidth]{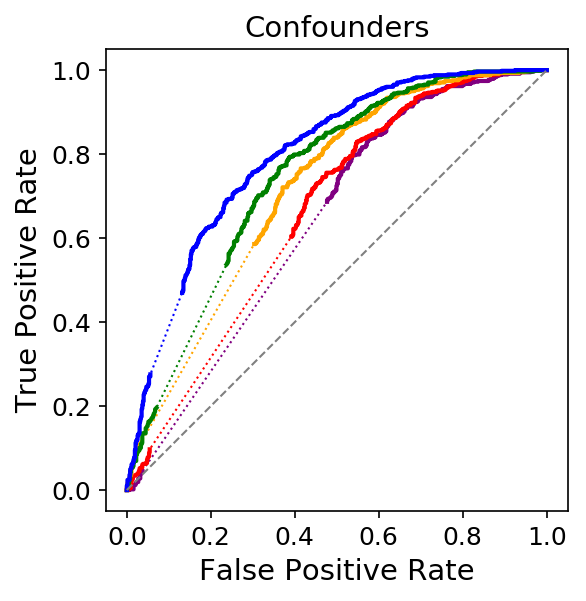}
\includegraphics[width=0.24\textwidth]{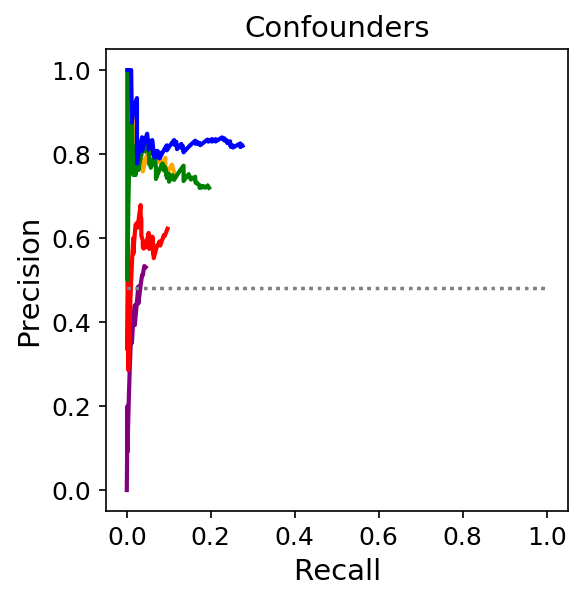}
\includegraphics[width=0.24\textwidth]{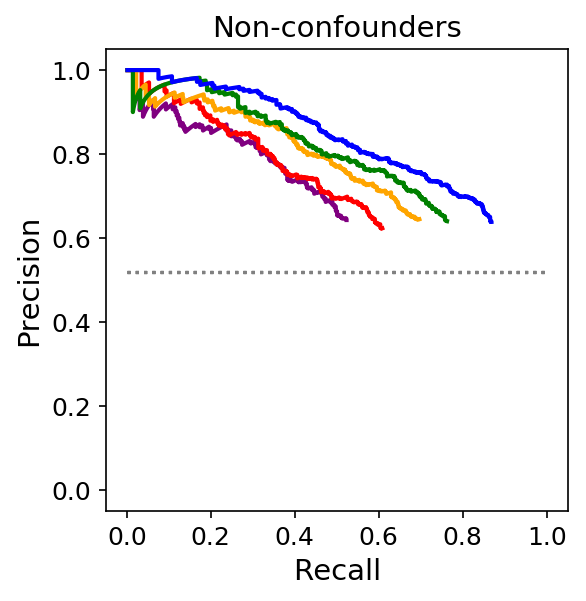}
\includegraphics[width=0.24\textwidth]{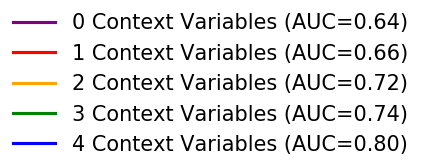}
}
\caption{\boldcap{Results for different numbers of context variables for \alg{ASD-JCI123} (acyclic, causal mechanism changes)} for small models. Taking into account more context variables leads to better predictions.\label{fig:simul_p4_qx_acyclic_mc_asdjci123}}
\end{figure}

As we have seen, discovery of causal relations between system variables can benefit strongly from observing
the system in multiple contexts. As Figure~\ref{fig:simul_p4_qx_acyclic_mc_asdjci123}
shows, the more context variables are taken into account, the better the predictions for \alg{ASD-JCI123} become. 
Although not shown here, the same conclusion holds as well for the other JCI variants 
\alg{ASD-JCI0}, \alg{ASD-JCI1} and \alg{ASD-JCI123-kt}. It does not hold for ASD baselines in general, but
it does for \alg{ASD-pikt} if interventions are perfect. For the JCI variants of FCI (\alg{FCI-JCI0}, \alg{FCI-JCI1} and 
\alg{FCI-JCI123}) we also observed that the more context variables are available, the more accurate the predictions become.
This is in line with our expectation that jointly analyzing data from multiple experiments makes it easier to estimate
the causal structure of the system.

Interestingly, the same conclusion does not hold for \alg{ASD-JCI1-sc} and \alg{ASD-JCI123-sc}.
This suggests that having multiple contexts is mostly beneficial if each context variable targets only a 
small subset of system variables, and only for methods that can explicitly take into account multiple context variables.
For LCD and ICP, precision also does not improve monotonically with the number of context variables.
Although \alg{LCD-mc} and
\alg{ICP-mc} in principle allow for multiple context variables, they suffer from a drop in recall because they focus on
detecting a certain causal pattern that becomes increasingly rare with more context variables. Indeed, for the extreme
case $q=p$, each system variable is targeted by a single context variable in our simulation setting, and hence one would
expect \alg{LCD-mc} and \alg{ICP-mc} to make no predictions at all. Any predictions they make must therefore be false 
positives, resulting in low precision.

\subsubsection{Discovering Indirect Intervention Targets}

\begin{figure}[t]
\centerline{%
\includegraphics[width=0.24\textwidth]{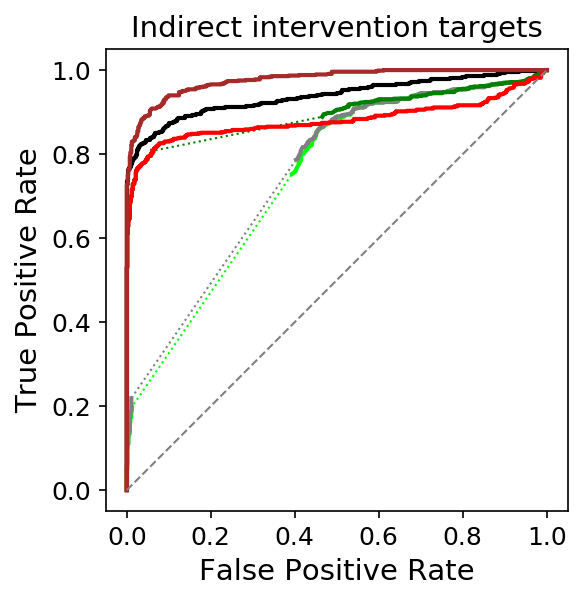}
\includegraphics[width=0.24\textwidth]{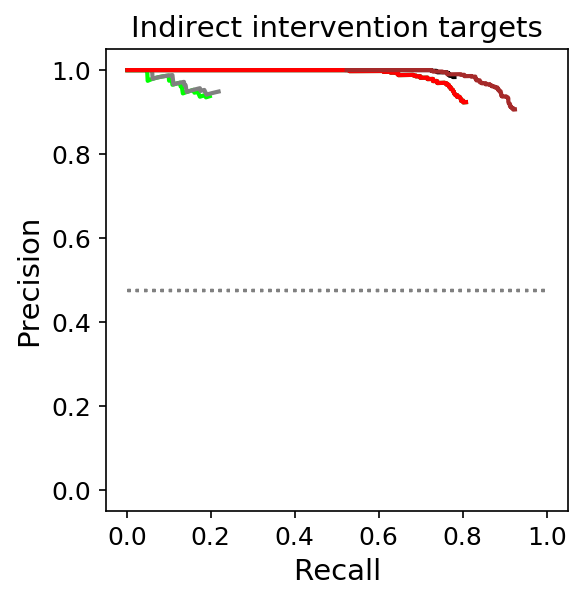}
\includegraphics[width=0.24\textwidth]{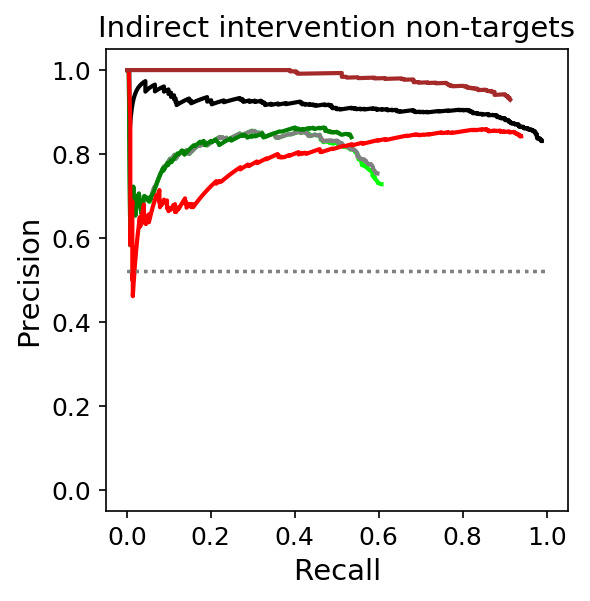}
\includegraphics[width=0.24\textwidth]{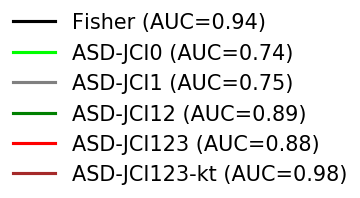}
}
\centerline{%
\includegraphics[width=0.24\textwidth]{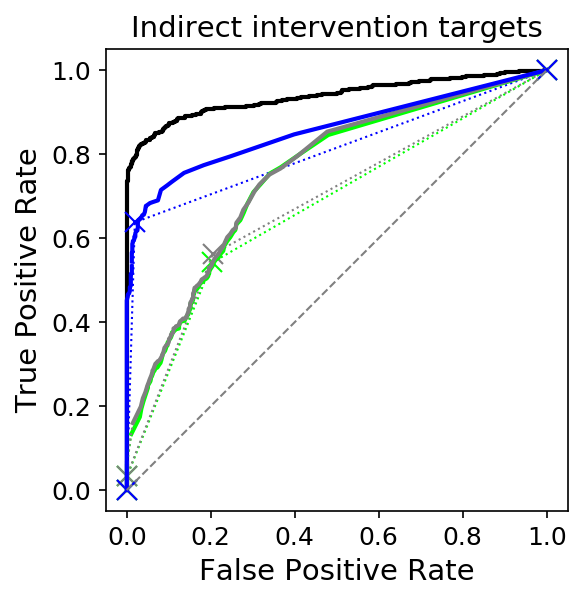}
\includegraphics[width=0.24\textwidth]{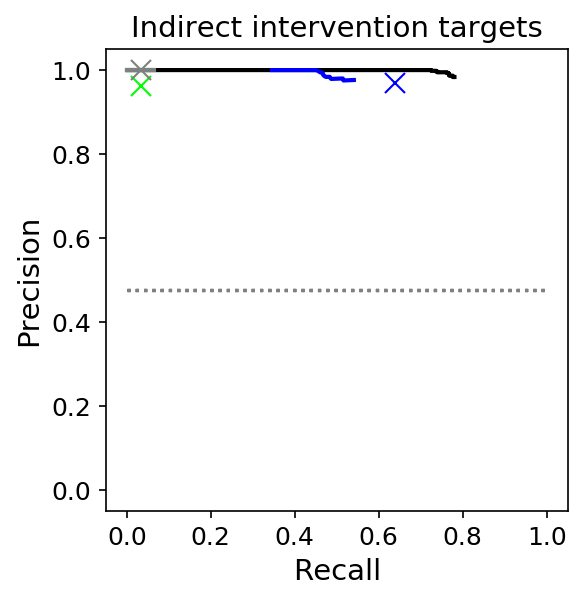}
\includegraphics[width=0.24\textwidth]{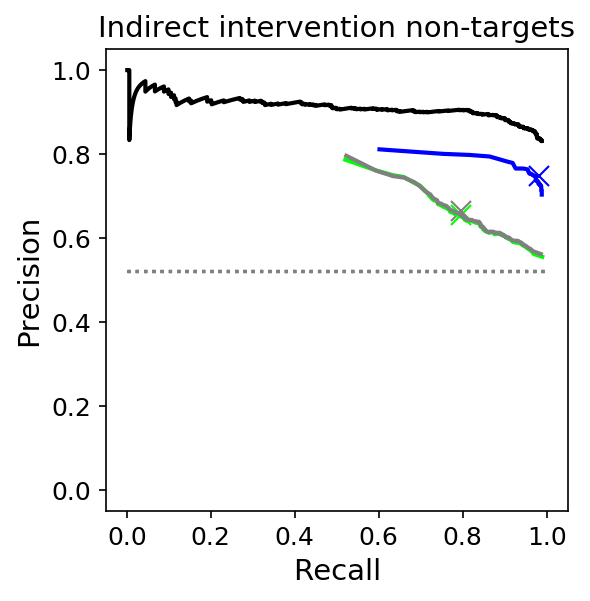}
\includegraphics[width=0.24\textwidth]{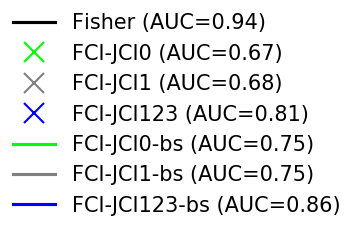}
}
\caption{\boldcap{Discovering indirect intervention targets (acyclic, causal mechanism changes)} on small models. Top: ASD variants; Bottom: FCI variants.\label{fig:simul_p4_q2_acyclic_mc_arel_con2sys}}
\end{figure}
\begin{figure}[t]
\centerline{%
\includegraphics[width=0.24\textwidth]{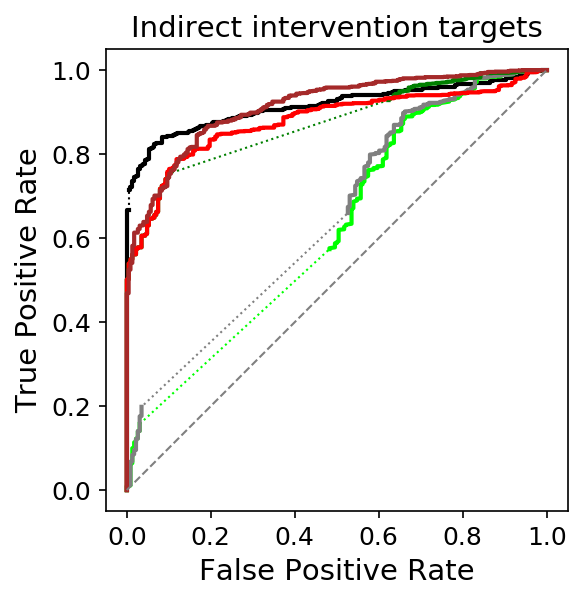}
\includegraphics[width=0.24\textwidth]{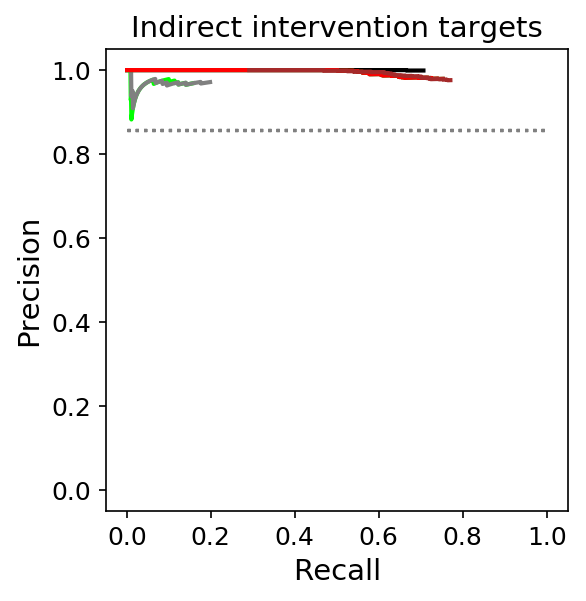}
\includegraphics[width=0.24\textwidth]{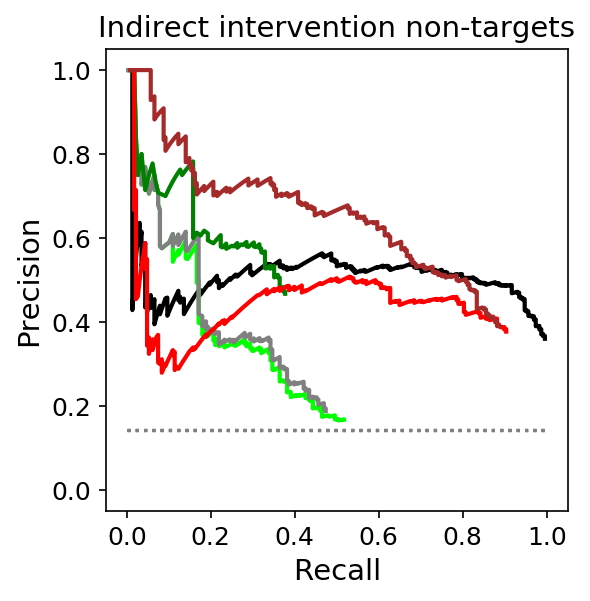}
\includegraphics[width=0.24\textwidth]{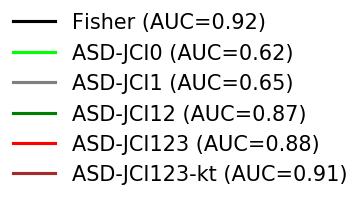}
}
\centerline{%
\includegraphics[width=0.24\textwidth]{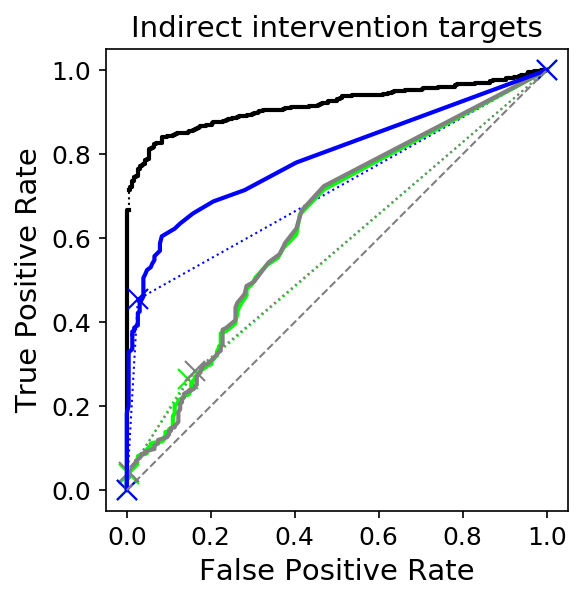}
\includegraphics[width=0.24\textwidth]{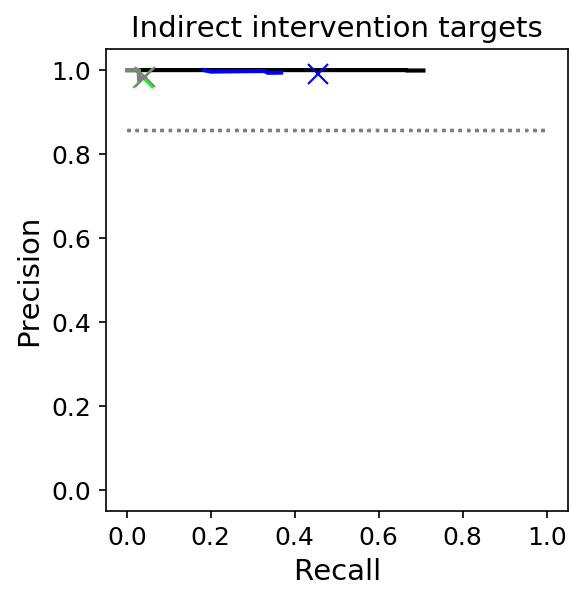}
\includegraphics[width=0.24\textwidth]{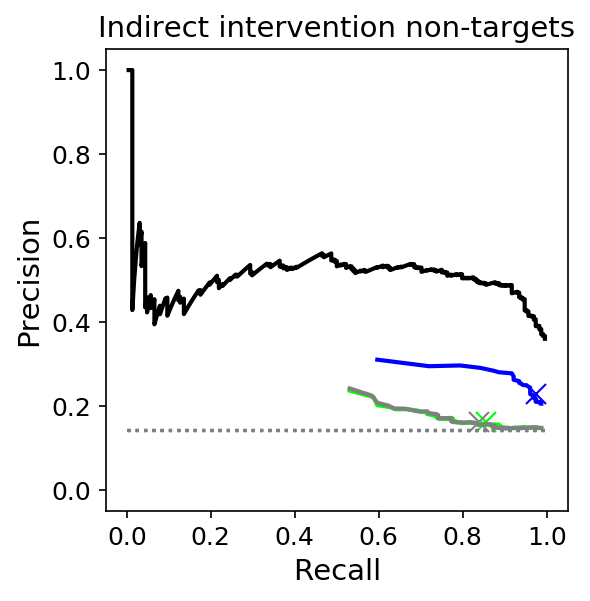}
\includegraphics[width=0.24\textwidth]{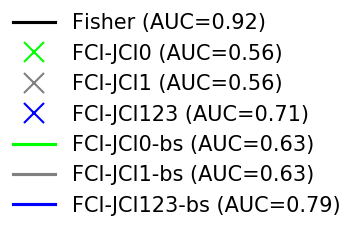}
}
\caption{\boldcap{Discovering indirect intervention targets (cyclic, causal mechanism changes)} on small models. Top: ASD variants; Bottom: FCI variants.\label{fig:simul_p4_q2_cyclic_mc_arel_con2sys}}
\end{figure}

Figure~\ref{fig:simul_p4_q2_acyclic_mc_arel_con2sys} shows for the acyclic setting with causal mechanism changes
how accurately \emph{indirect} intervention targets (i.e., which system variables are descendants of each context
variable) can be discovered by various methods.
Baselines \alg{ASD-obs}, \alg{ASD-pooled}, \alg{ASD-meta} and \alg{ASD-pikt} cannot learn intervention targets
(neither direct ones nor indirect ones), since they do not represent context variables explicitly, and are therefore excluded.\footnote{However, it would be
trivial to extend \alg{ASD-pikt} such that it can predict indirect intervention targets, by combining the known direct 
intervention targets of a certain intervention variable with the descendants of those assumed
targets as predicted by the method as a postprocessing step.}
LCD and ICP also cannot address this task.\footnote{However, when assuming also JCI Assumption~\ref{ass:unconfounded},
both LCD and ICP could be used to learn indirect intervention targets.} Although \alg{ASD-JCI123-kt} makes use of known 
\emph{direct} intervention targets (i.e., which system variables are children of each context variable?), 
this means that there is still a non-trivial task of learning the \emph{indirect} ones.

The task of deciding that a system variable is targeted is an easier one than deciding that a system variable
is \emph{not} targeted by an intervention. 
Although Fisher's test is generally hard to beat when it comes to 
predicting indirect intervention targets, \alg{ASD-JCI123-kt} outperforms it in this setting by exploiting the knowledge about \emph{direct} intervention targets. 
While JCI Assumption~\ref{ass:unconfounded} turned out to be unimportant for 
learning the causal relations between system variables, it is seen to be very useful for this task of 
learning causal relations between context and system variables.

Figure~\ref{fig:simul_p4_q2_cyclic_mc_arel_con2sys} shows a largely similar picture for the cyclic setting with
causal mechanism changes. Surprisingly, FCI variants are also performing quite well in the cyclic setting.
We do not show the results for perfect interventions here, as we observed that this task is easier, and the results are generally 
better, but otherwise mostly similar conclusions are obtained. The only exception is that for perfect
interventions, the best method is Fisher's approach.

\subsubsection{Discovering Direct Intervention Targets}

\begin{figure}[t]
\centerline{%
\includegraphics[width=0.24\textwidth]{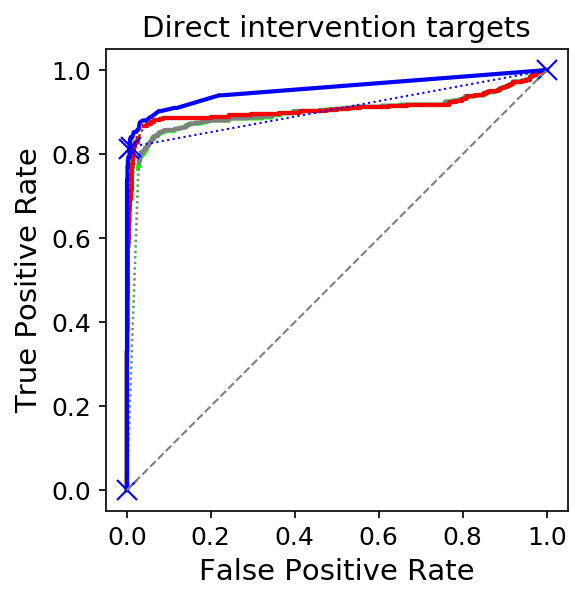}
\includegraphics[width=0.24\textwidth]{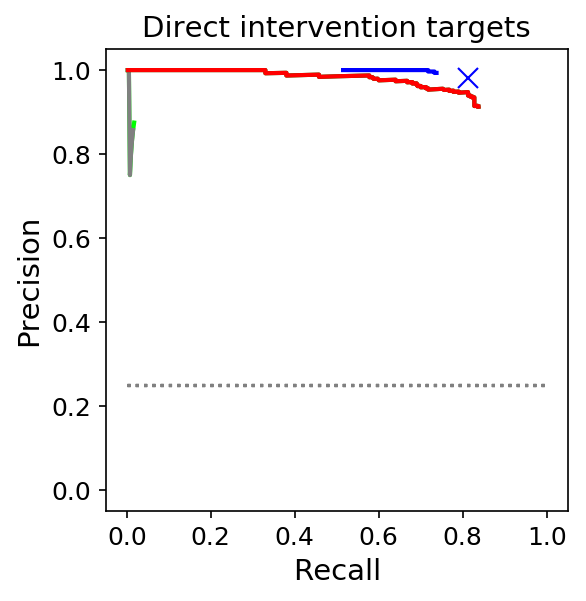}
\includegraphics[width=0.24\textwidth]{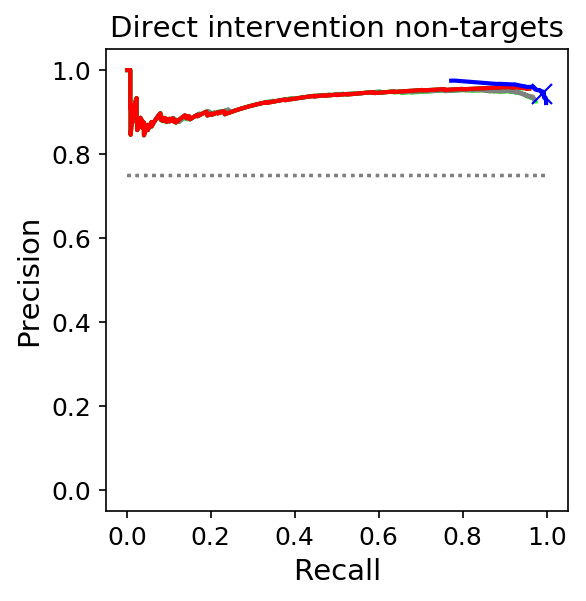}
\includegraphics[width=0.24\textwidth]{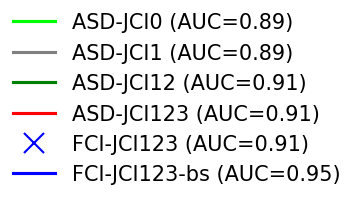}
}
\centerline{%
\includegraphics[width=0.24\textwidth]{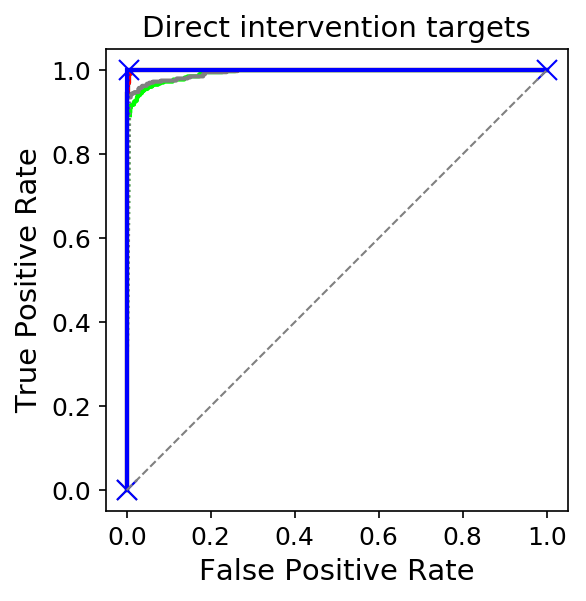}
\includegraphics[width=0.24\textwidth]{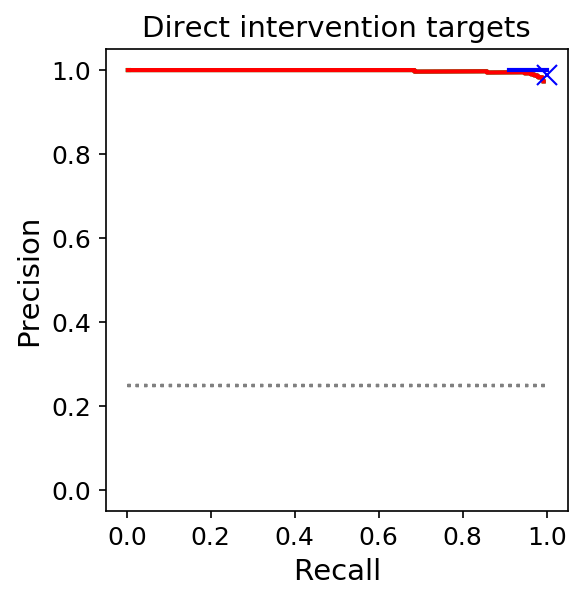}
\includegraphics[width=0.24\textwidth]{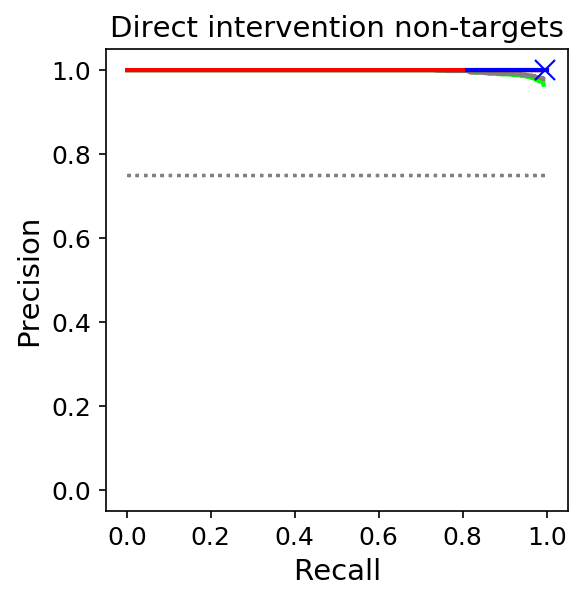}
\includegraphics[width=0.24\textwidth]{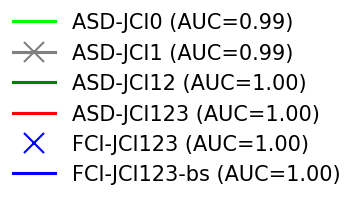}
}
\centerline{%
\includegraphics[width=0.24\textwidth]{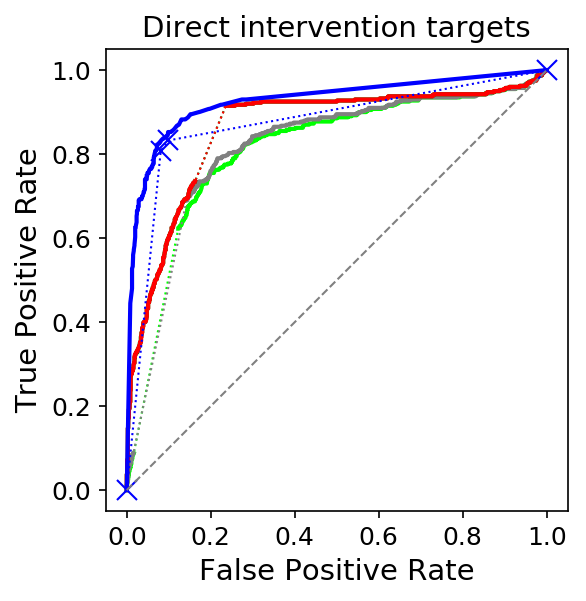}
\includegraphics[width=0.24\textwidth]{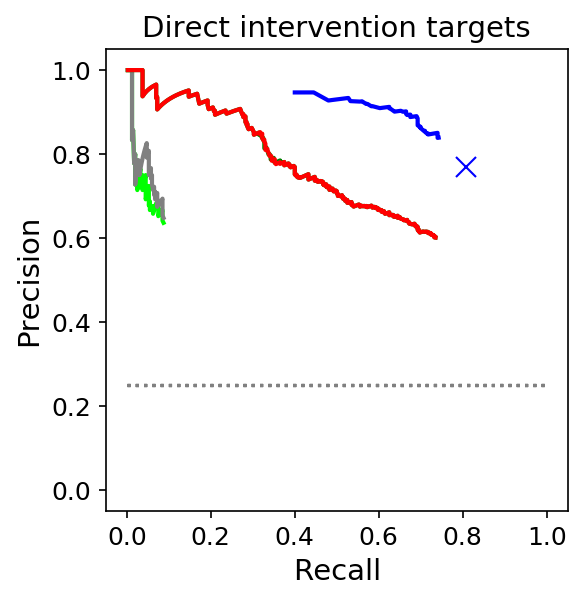}
\includegraphics[width=0.24\textwidth]{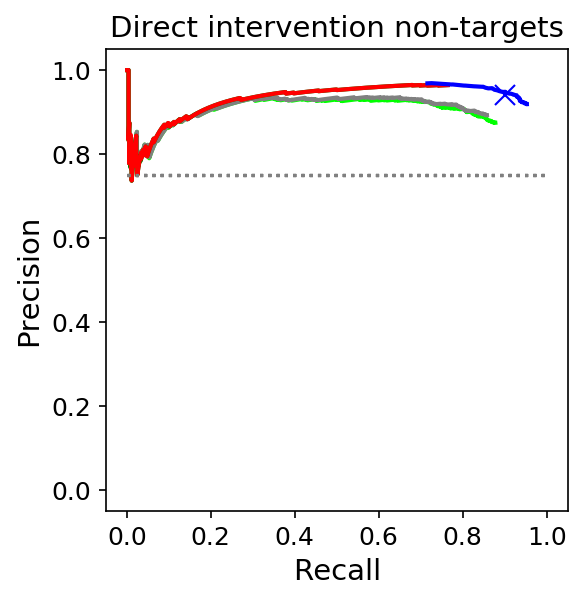}
\includegraphics[width=0.24\textwidth]{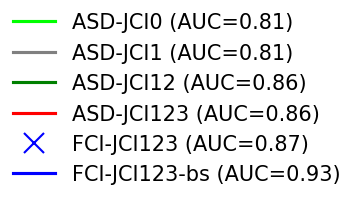}
}
\centerline{%
\includegraphics[width=0.24\textwidth]{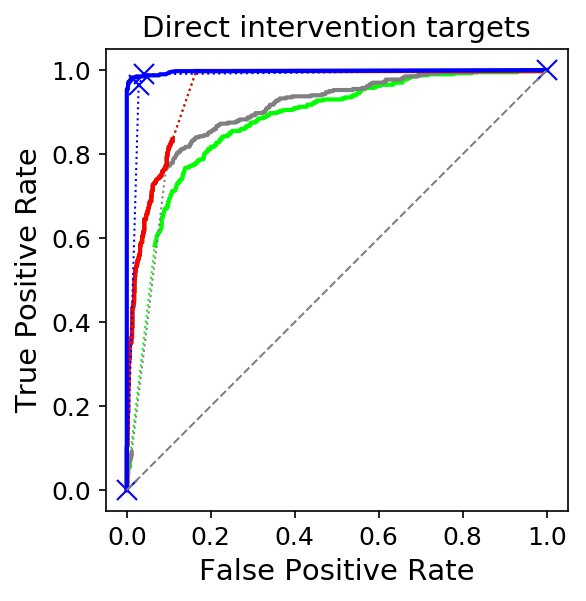}
\includegraphics[width=0.24\textwidth]{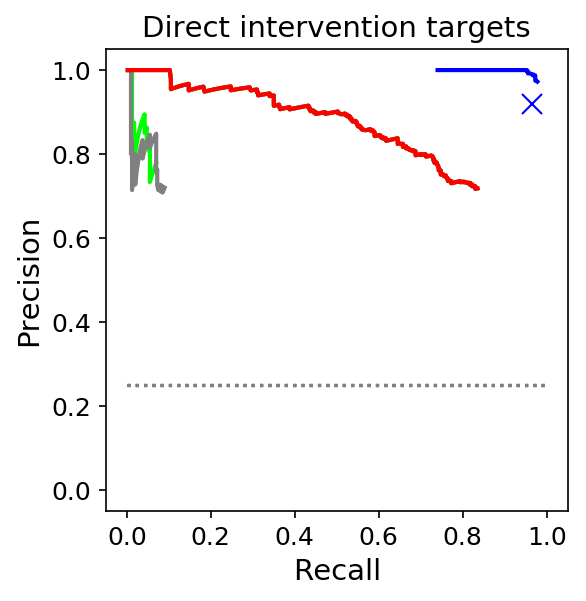}
\includegraphics[width=0.24\textwidth]{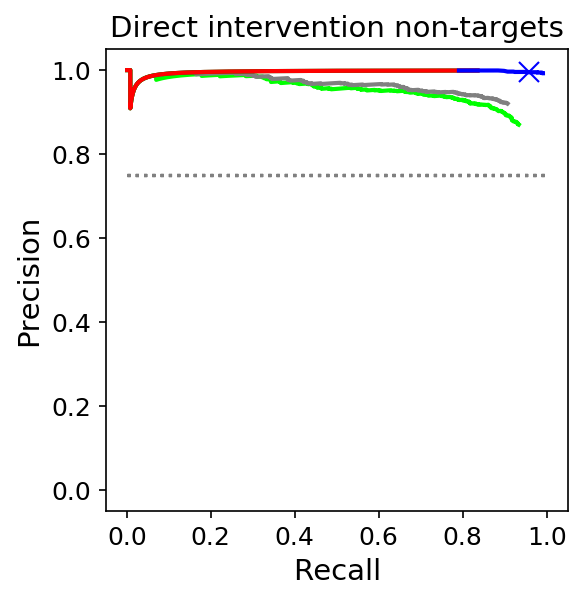}
\includegraphics[width=0.24\textwidth]{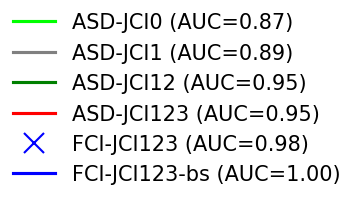}
}
\caption{\boldcap{Discovering direct intervention targets} on small models. From top to bottom: acyclic, causal mechanism changes; acyclic, perfect interventions; cyclic, causal mechanism changes; cyclic, perfect interventions.\label{fig:simul_p4_q2_x_x_edge_con2sys}}
\end{figure}

Fisher's test is not able to predict \emph{direct} intervention targets (i.e., which system variables
are children of each intervention variable?), but the ASD-JCI variants can, as well as \alg{FCI-JCI123}.
Figure~\ref{fig:simul_p4_q2_x_x_edge_con2sys} shows the results for these algorithms in the four 
different simulation settings.
The task is considerably easier in the acyclic setting than in the cyclic setting. Having perfect interventions
makes it slightly easier than with causal mechanism changes. We again notice that
exploiting JCI Assumption~\ref{ass:unconfounded} considerably improves performance on this task.
Surprisingly, \alg{FCI-JCI123} obtains almost perfect precision in all scenarios, outperforming 
\alg{ASD-JCI123} notably in the cyclic cases. We do not understand why this is the case.

\subsubsection{Computation Time}

\begin{figure}
\centerline{\includegraphics[width=0.49\textwidth]{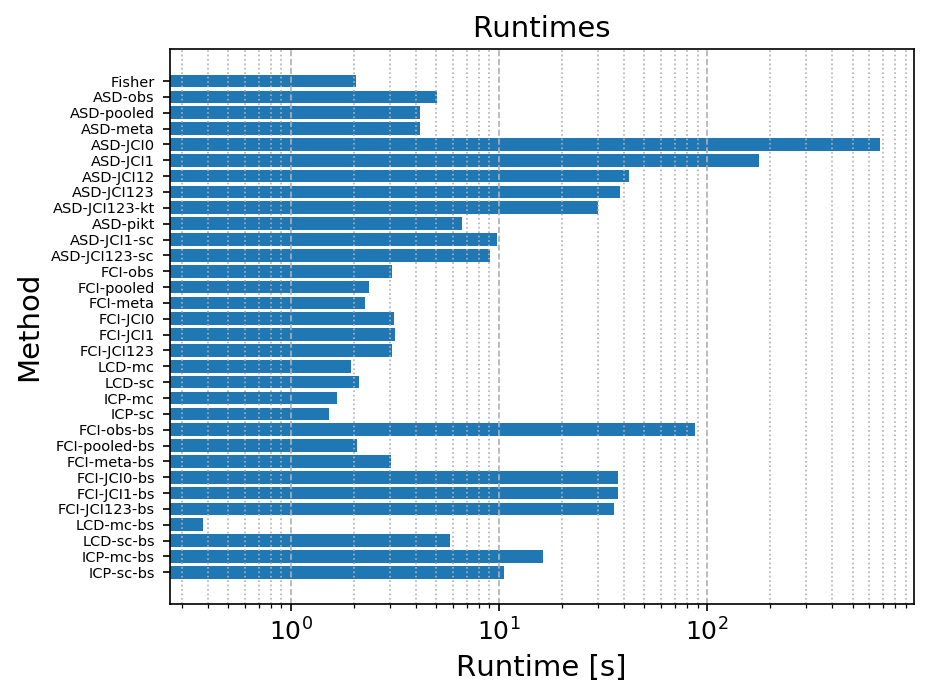}\hfill%
\includegraphics[width=0.49\textwidth]{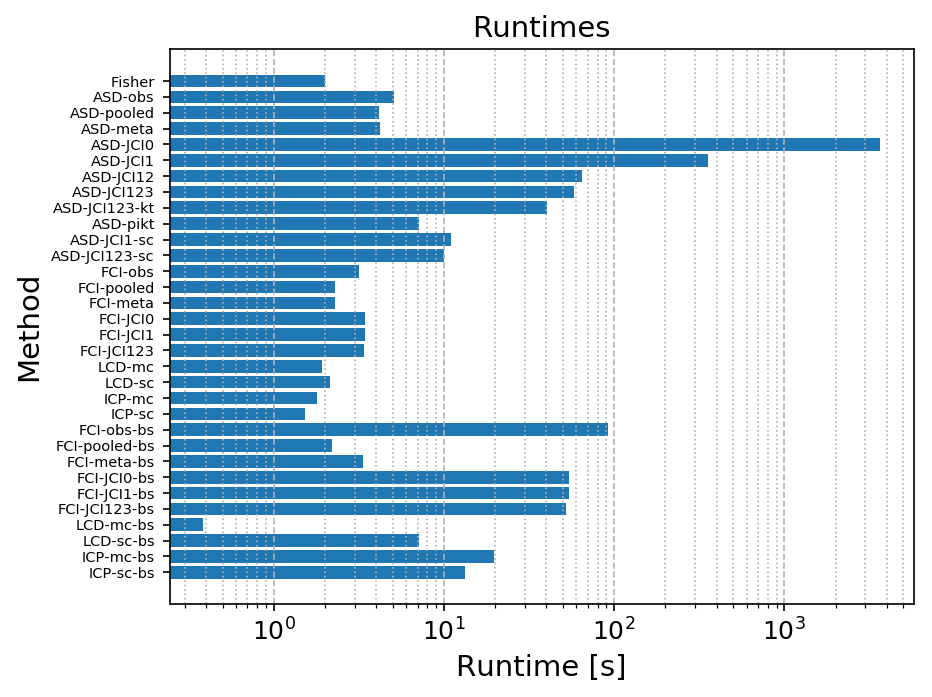}}
  \caption{\boldcap{Runtimes for various methods} on small models. Shown are runtimes for causal mechanism changes; for perfect interventions, runtimes are similar. Left: acyclic; Right: cyclic.\label{fig:simul_p4_q2_runtimes}}
\end{figure}

So far we have only considered the accuracy of the predictions. Another interesting aspect 
is the computation time that various methods need. Figure~\ref{fig:simul_p4_q2_runtimes} shows total computation
time (for all prediction tasks together) for all methods considered thus far. Note the logarithmic scale on the
$x$-axis. We only show runtimes for causal mechanism changes since those for perfect interventions are nearly identical.
On the other hand, we do see that the cyclic setting is more computationally demanding in general than the acyclic one. 

\begin{figure}
\centerline{\includegraphics[width=0.3\textwidth]{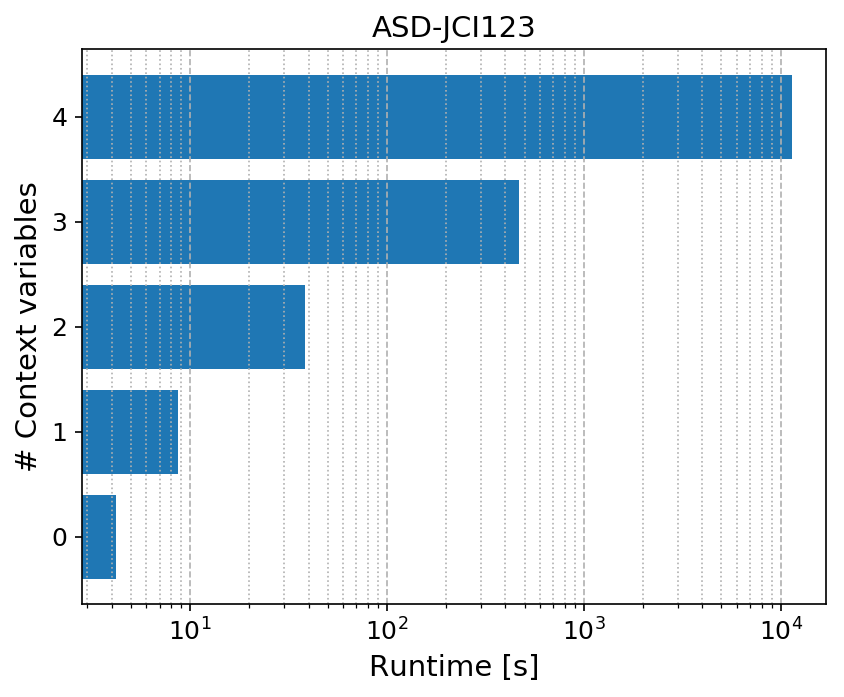}\hfill%
\includegraphics[width=0.3\textwidth]{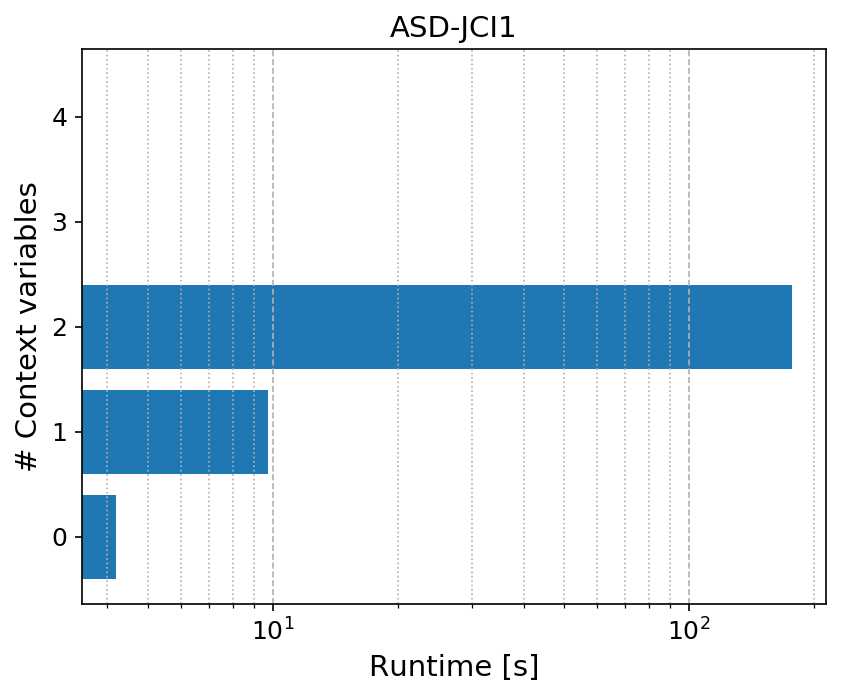}\hfill%
\includegraphics[width=0.3\textwidth]{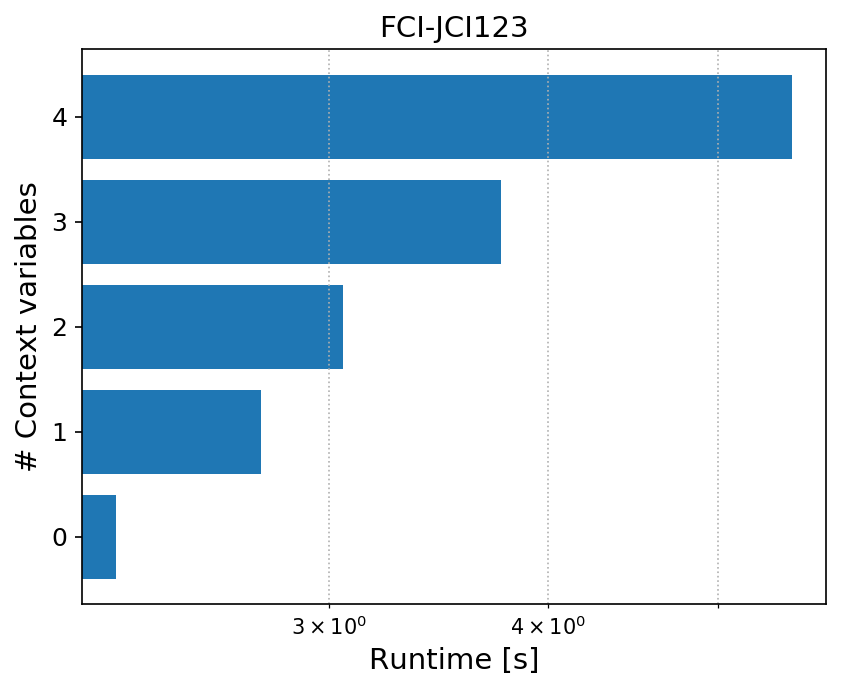}}
\caption{\boldcap{Runtimes for three different algorithms as a function of the number of context variables} on small models. 
  Note the considerably different ranges of the (logarithmic) $x$-axis. 
  Results are omitted if the computation took too long to finish.\label{fig:simul_p4_qx_runtimes}}
\end{figure}

Already for small models of $p+q=4+2=6$ variables, the ASD algorithms become slow because they are
performing an optimization over a large discrete space. The availability of more background knowledge makes the search space
considerably smaller, and hence leads to reduced computation time for the ASD variants. Also, the search space is
considerably larger in the cyclic setting than in the acyclic one. By design, FCI variants are
much faster, but bootstrapping also takes its toll. The fastest methods are Fisher's test, LCD and ICP.
Figure~\ref{fig:simul_p4_qx_runtimes} shows how computation time scales with the number of context variables, for three
JCI implementations (\alg{ASD-JCI123}, \alg{ASD-JCI1} and \alg{FCI-JCI123}).

\subsection{Results: Larger Simulated Models}

We now present results for larger models, with $p=10$ system variables and $q=10$ context variables
(the meaning of the simulation parameters is explained in Section~\ref{sec:simulations}).
We only consider causal mechanism changes here, but we do distinguish the acyclic and cyclic settings.
We again used 500 samples per context. For the acyclic setting, we used $\epsilon=\eta=0.25$, while for
the cyclic setting we used $\epsilon=\eta=0.15$ to get more or less similarly dense graphs in both scenarios.
The motivation for these parameter choices is that they are somewhat comparable to the setting of the real-world data set that we will study in Section~\ref{sec:exp_sachs}.

For these larger models, computation time for bootstrapped FCI methods became prohibitive for the default conditional
independence test (described in Section~\ref{sec:CI}). 
Although we can speed up the implementation of this test that we were using considerably by implementing it 
more efficiently, we here simply replaced it by a standard partial correlation test. This
led to a speedup of about one order of magnitude at no apparent loss of accuracy.

\subsubsection{FCI Variants}

\begin{figure}[t]
\centerline{%
\includegraphics[width=0.24\textwidth]{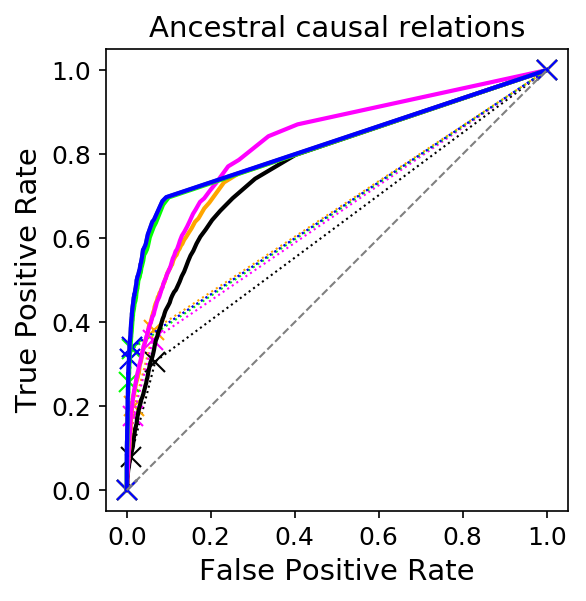}
\includegraphics[width=0.24\textwidth]{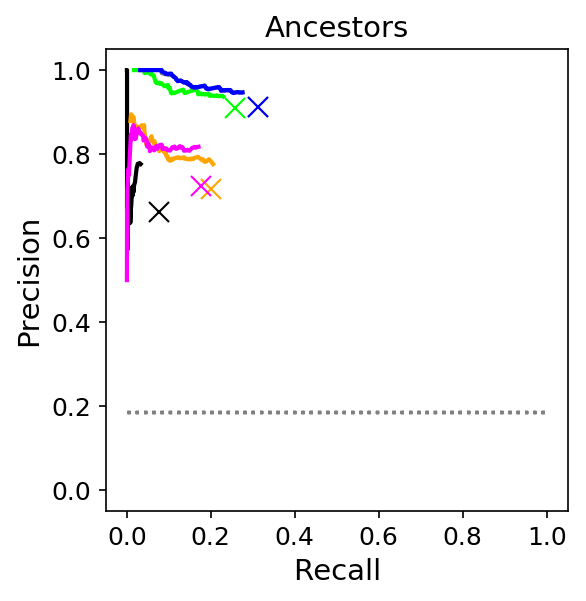}
\includegraphics[width=0.24\textwidth]{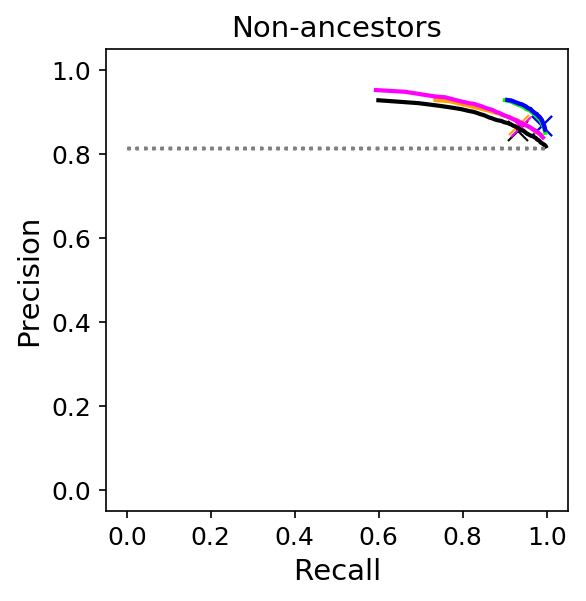}
\includegraphics[width=0.24\textwidth]{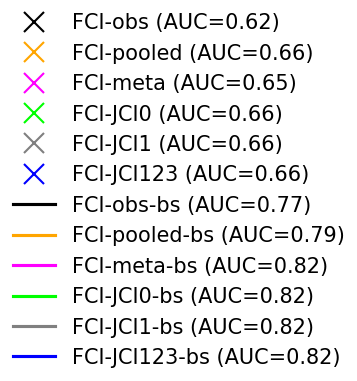}
}
\centerline{%
\includegraphics[width=0.24\textwidth]{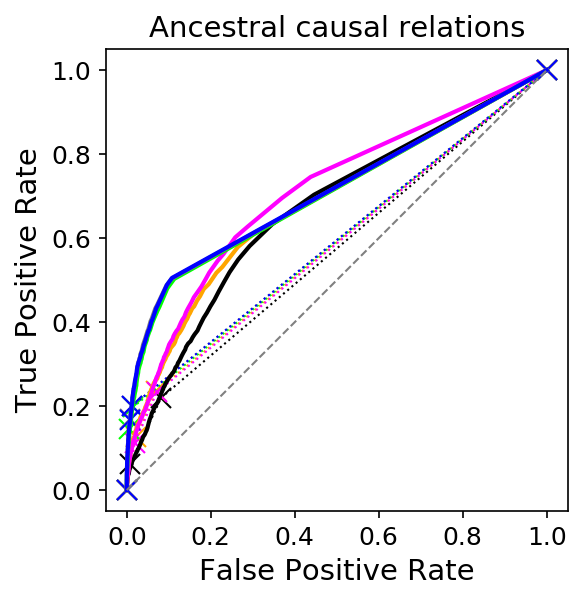}
\includegraphics[width=0.24\textwidth]{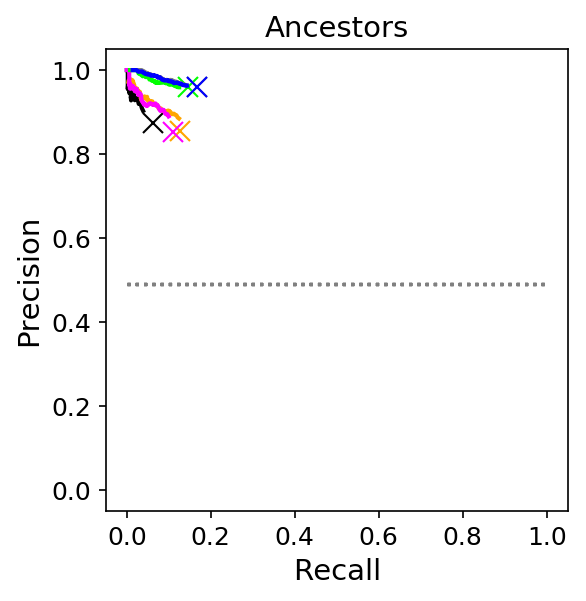}
\includegraphics[width=0.24\textwidth]{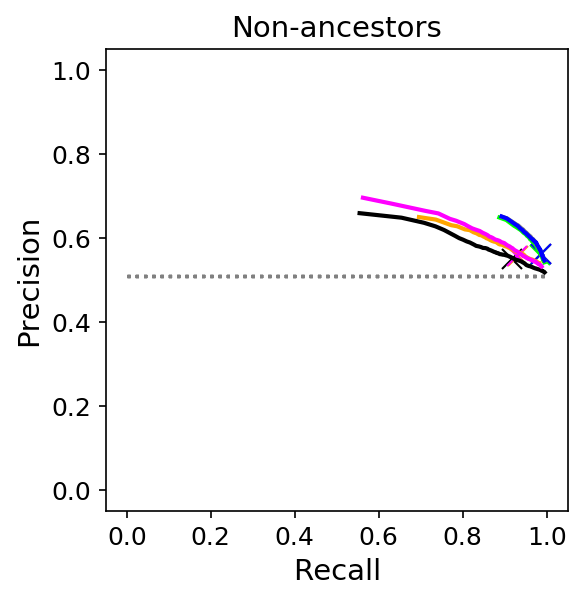}
\includegraphics[width=0.24\textwidth]{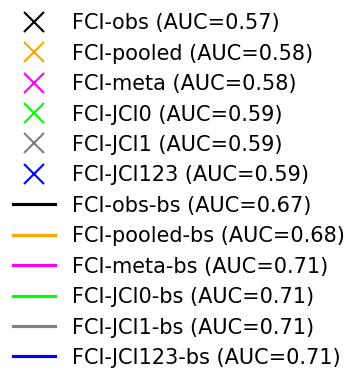}
}
  \caption{\boldcap{FCI results for discovering ancestral causal relations between system variables} in larger models. Top: acyclic; bottom: cyclic.\label{fig:simul_p10_q10_x_mc_arel_fci_sys2sys}}
\end{figure}

Figure~\ref{fig:simul_p10_q10_x_mc_arel_fci_sys2sys} shows the accuracy for the task of predicting ancestral causal relations between system variables for various FCI-JCI variants and for various FCI baselines, in both acyclic and cyclic settings. 
The conclusions are in line with what we already observed for smaller models. 
Again, bootstrapping FCI helps considerably to boost the accuracy of its predictions.
As before, \alg{FCI-obs} (which uses only observational data) performs worst. 
The two baselines \alg{FCI-pooled} and \alg{FCI-meta} (that make use of all data) lead to a moderate improvement.  
The JCI variants (\alg{FCI-JCI0}, \alg{FCI-JCI1} and \alg{FCI-JCI123}) perform the best, delivering almost maximum precision for a considerable recall range on the task of predicting the presence of an ancestral relation.
JCI Assumption \ref{ass:uncaused} does not seem to help much, as the results for \alg{FCI-JCI0} and \alg{FCI-JCI1} seem to be identical.
Assuming in addition JCI Assumption \ref{ass:unconfounded} (and \ref{ass:dependences}) does help to obtain slightly higher precision.
The good performance of FCI variants in the cyclic setting is surprising, since FCI was designed for the acyclic case.

\subsubsection{LCD and ICP}

\begin{figure}[t]
\centerline{%
\includegraphics[width=0.24\textwidth]{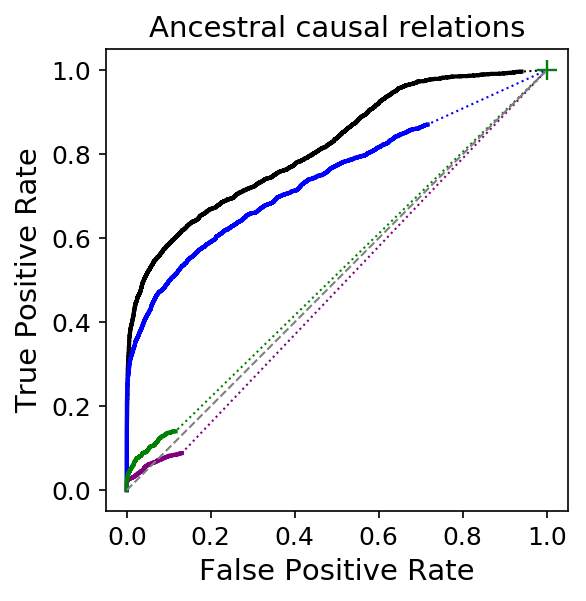}
\includegraphics[width=0.24\textwidth]{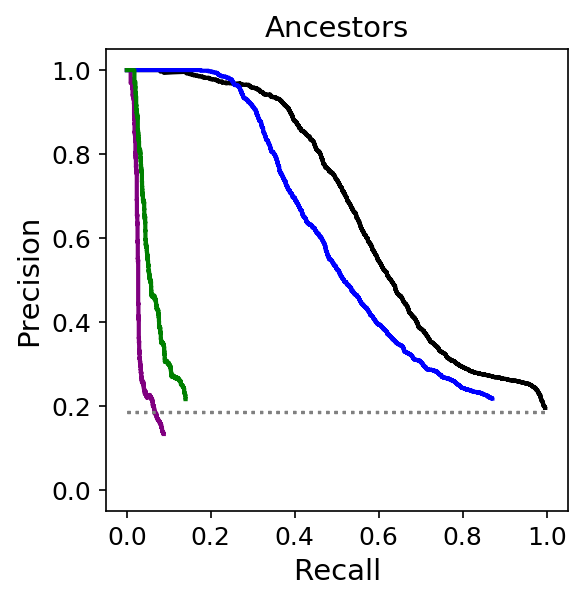}
\includegraphics[width=0.24\textwidth]{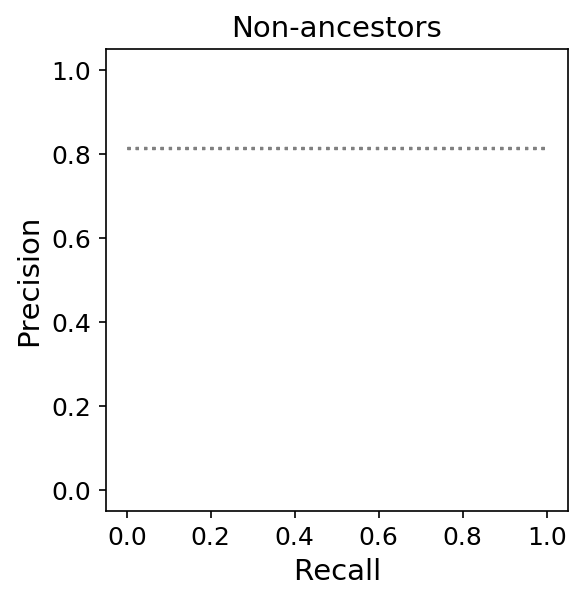}
\includegraphics[width=0.24\textwidth]{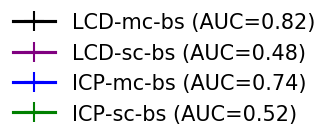}
}
\centerline{%
\includegraphics[width=0.24\textwidth]{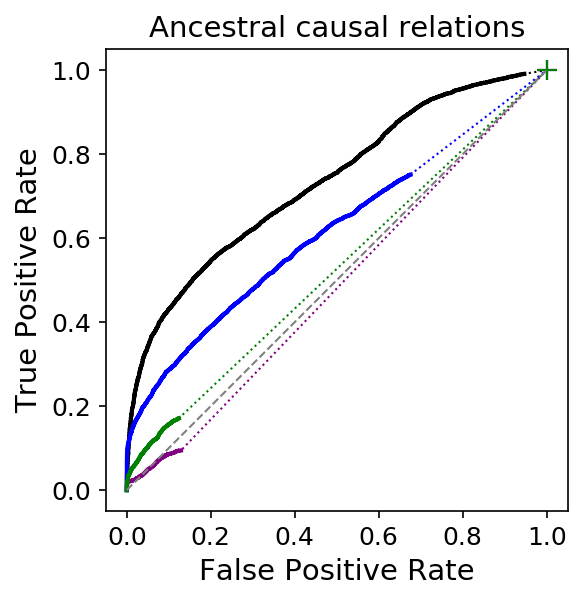}
\includegraphics[width=0.24\textwidth]{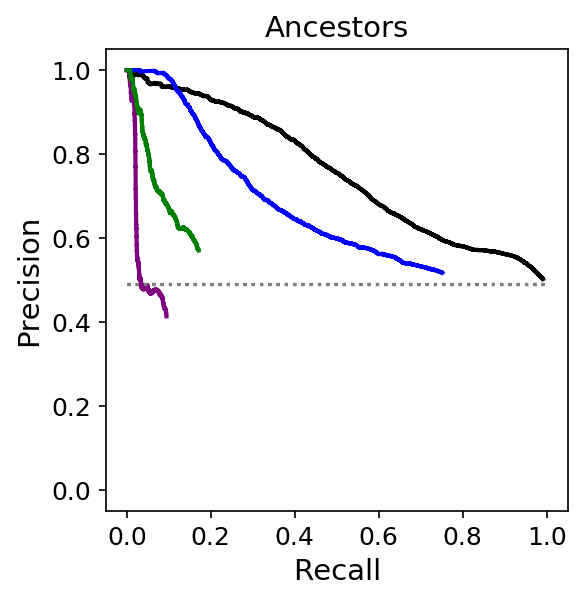}
\includegraphics[width=0.24\textwidth]{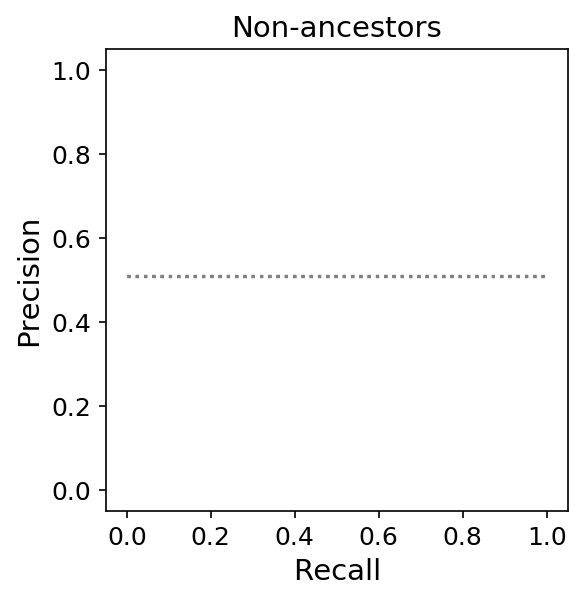}
\includegraphics[width=0.24\textwidth]{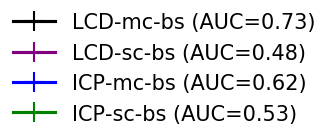}
}
\caption{\boldcap{Bootstrapped LCD and ICP results for discovering ancestral causal relations between system variables}
for larger models. Top: acyclic; bottom: cyclic.\label{fig:simul_p10_q10_x_mc_arel_lcdicp_sys2sys}}
\end{figure}

In Figure~\ref{fig:simul_p10_q10_x_mc_arel_lcdicp_sys2sys}, we show the accuracy of bootstrapped LCD and ICP for the task of predicting ancestral causal relations between system variables, in both the acyclic and the cyclic setting. 
As for FCI, bootstrapping improves the accuracy of LCD and ICP results, and we decided to only show the bootstrapped results here. 
Contrary to what we observed for small models, the ``multiple context'' (``\alg{-mc}'') versions of both algorithms now clearly outperform the versions that use only a single (merged) context (``\alg{-sc}'') in these settings. Interestingly, the accuracy of LCD is quite similar to that of ICP. The additional complexity of ICP apparently does not lead to substantially better results than the LCD algorithm already offers in these settings. Also, the precision of \alg{LCD-mc} is comparable to that of \alg{FCI-JCI123}, the most accurate of the JCI variants of FCI (cf.\ Figure~\ref{fig:simul_p10_q10_x_mc_arel_fci_sys2sys}), on the task of predicting the presence of ancestral relations.

\subsubsection{Discovering Intervention Targets}

\begin{figure}[t]
\centerline{%
\includegraphics[width=0.24\textwidth]{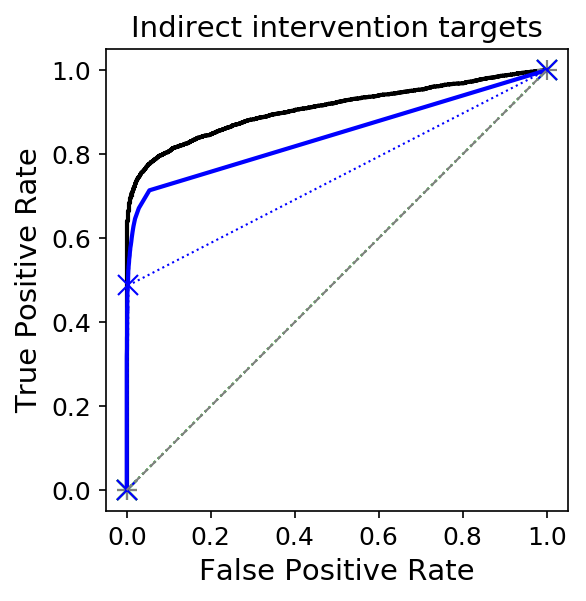}
\includegraphics[width=0.24\textwidth]{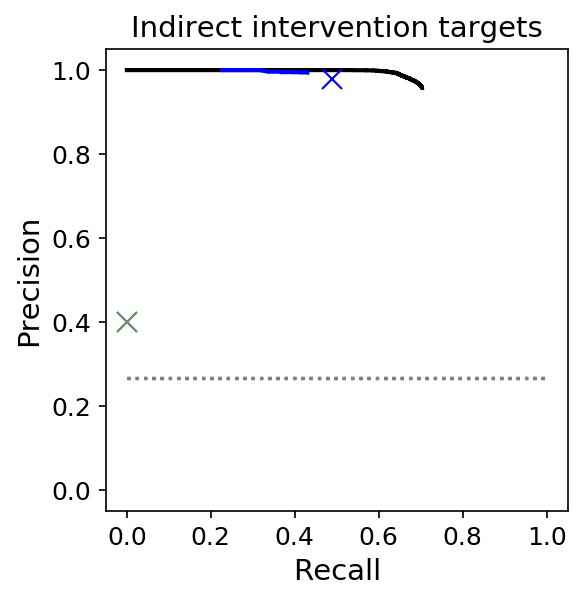}
\includegraphics[width=0.24\textwidth]{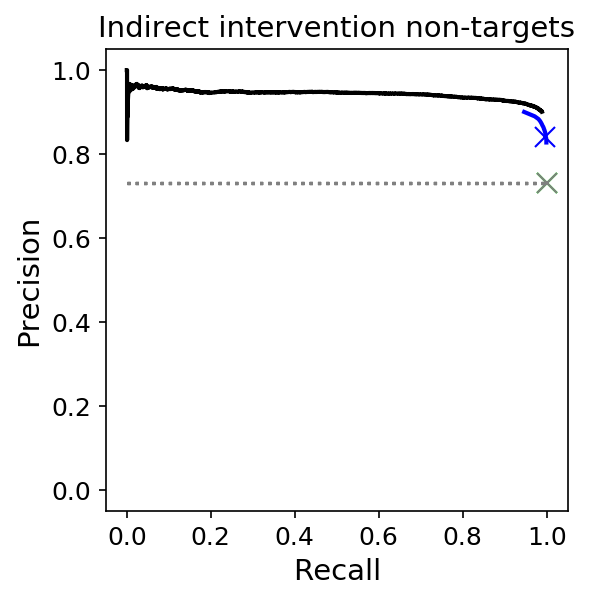}
\includegraphics[width=0.24\textwidth]{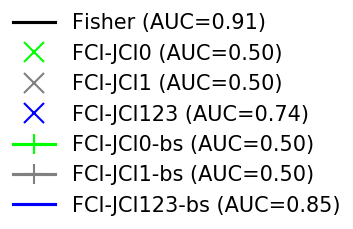}
}
\centerline{%
\includegraphics[width=0.24\textwidth]{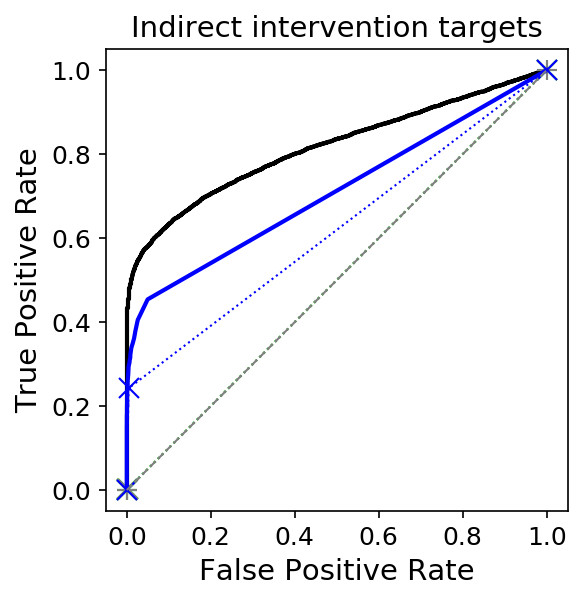}
\includegraphics[width=0.24\textwidth]{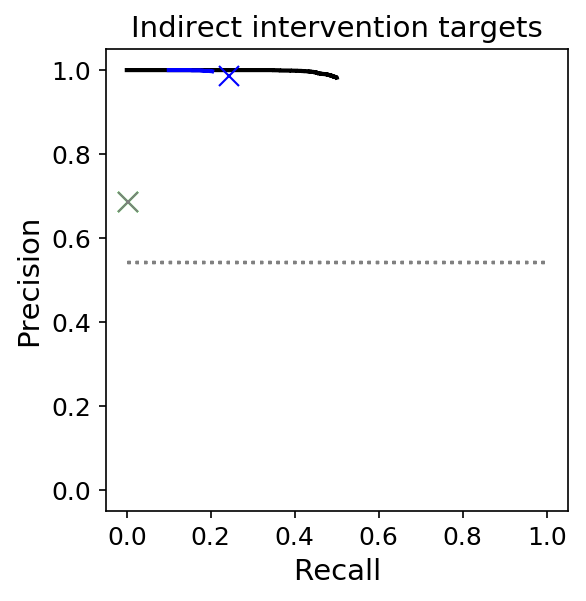}
\includegraphics[width=0.24\textwidth]{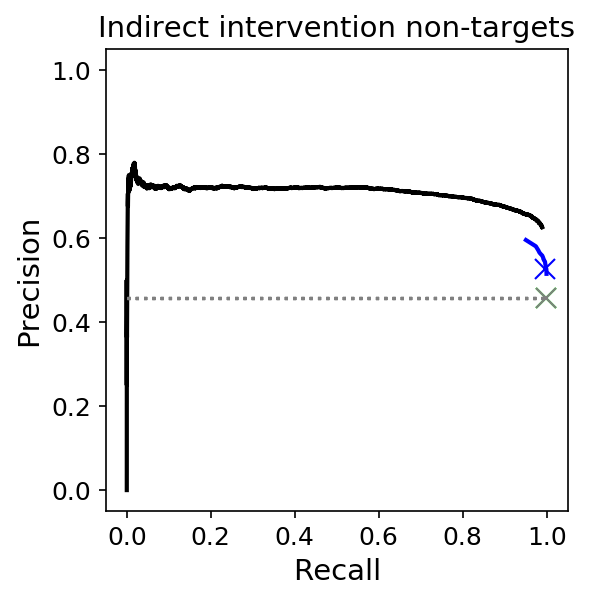}
\includegraphics[width=0.24\textwidth]{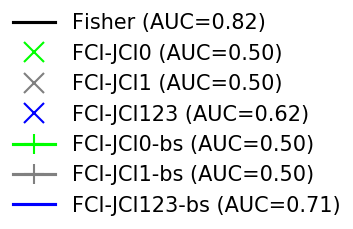}
}
\caption{\boldcap{FCI results for discovering indirect intervention targets} in larger models. Top: acyclic; bottom: cyclic.\label{fig:simul_p10_q10_x_mc_arel_fci_con2sys}}
\end{figure}

\begin{figure}[t]
\centerline{%
\includegraphics[width=0.24\textwidth]{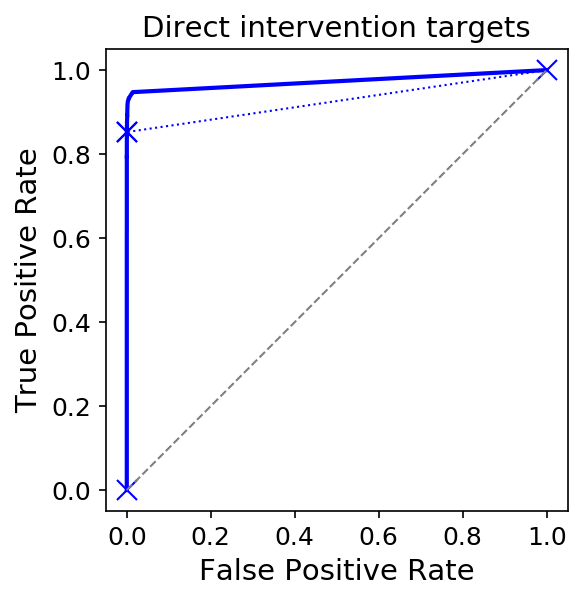}
\includegraphics[width=0.24\textwidth]{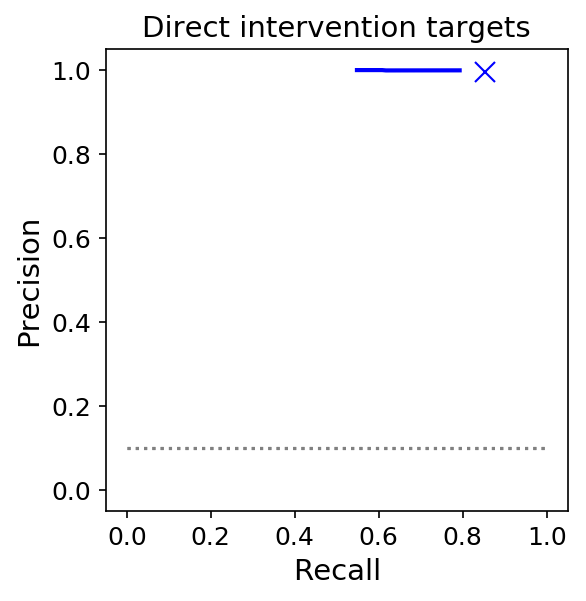}
\includegraphics[width=0.24\textwidth]{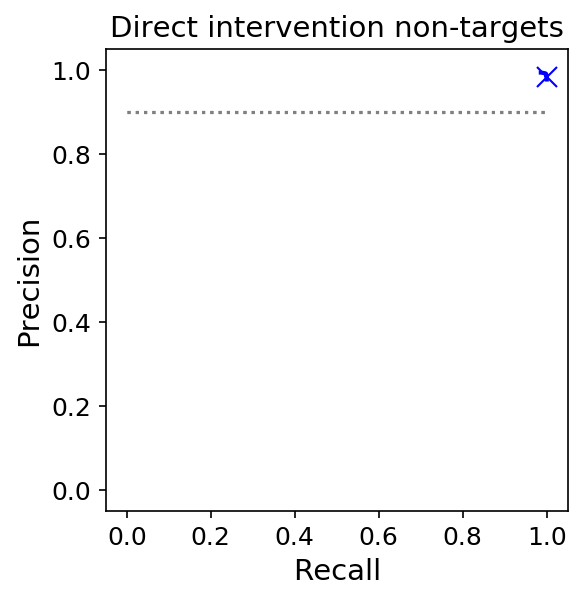}
\includegraphics[width=0.24\textwidth]{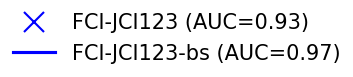}
}
\centerline{%
\includegraphics[width=0.24\textwidth]{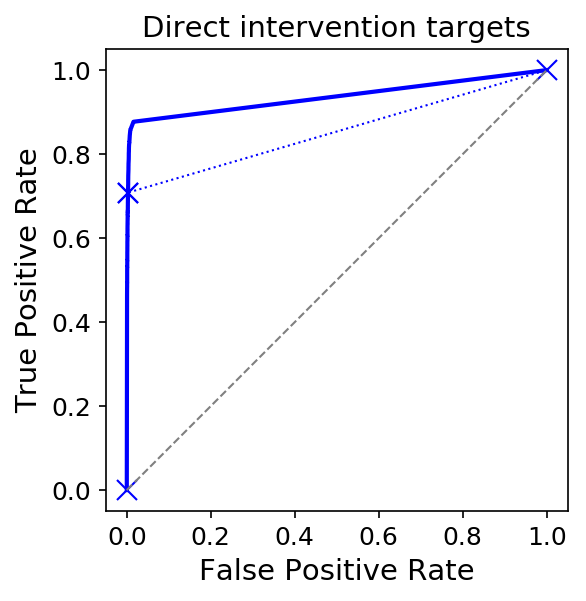}
\includegraphics[width=0.24\textwidth]{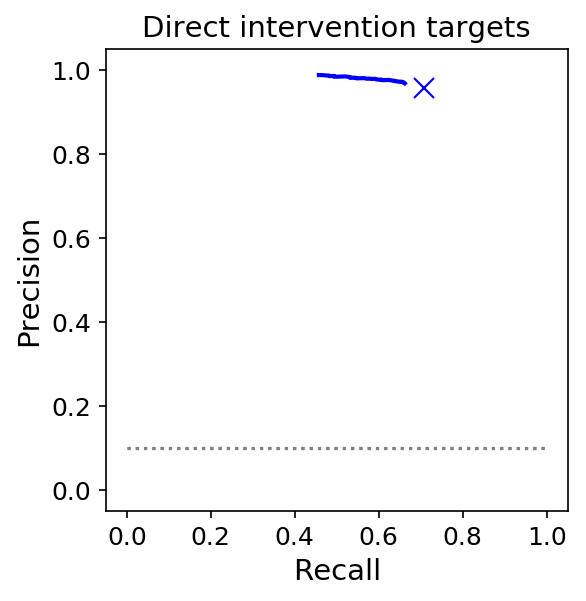}
\includegraphics[width=0.24\textwidth]{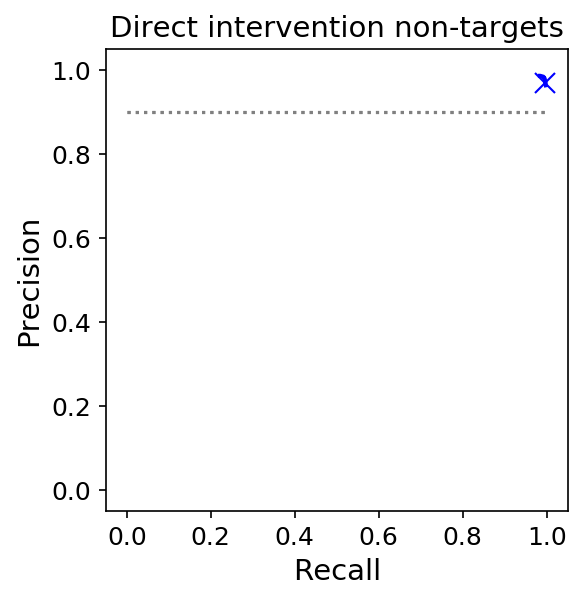}
\includegraphics[width=0.24\textwidth]{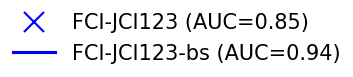}
}
\caption{\boldcap{FCI results for discovering direct intervention targets} in larger models. Top: acyclic; bottom: cyclic.\label{fig:simul_p10_q10_x_mc_edge_con2sys}}
\end{figure}

Figure~\ref{fig:simul_p10_q10_x_mc_arel_fci_con2sys} shows the performance of FCI-JCI variants (and as a baseline,
Fisher's test) on the task of discovering indirect intervention targets, for both the acyclic and cyclic setting.
Interestingly, JCI Assumption~\ref{ass:unconfounded} seems necessary to obtain good results on this task. 
Still, \alg{FCI-JCI123-bs} is outperformed by Fisher's test. 

On the other hand, Fisher's test cannot identify direct intervention targets, whereas \alg{FCI-JCI123} can.
Figure~\ref{fig:simul_p10_q10_x_mc_edge_con2sys} shows that \alg{FCI-JCI123} can identify
the direct intervention targets as well as the non-targets with high precision. Surprisingly, this works also in the cyclic case.

\subsubsection{Computation Time}

\begin{figure}[t]
\centerline{%
\includegraphics[width=0.5\textwidth]{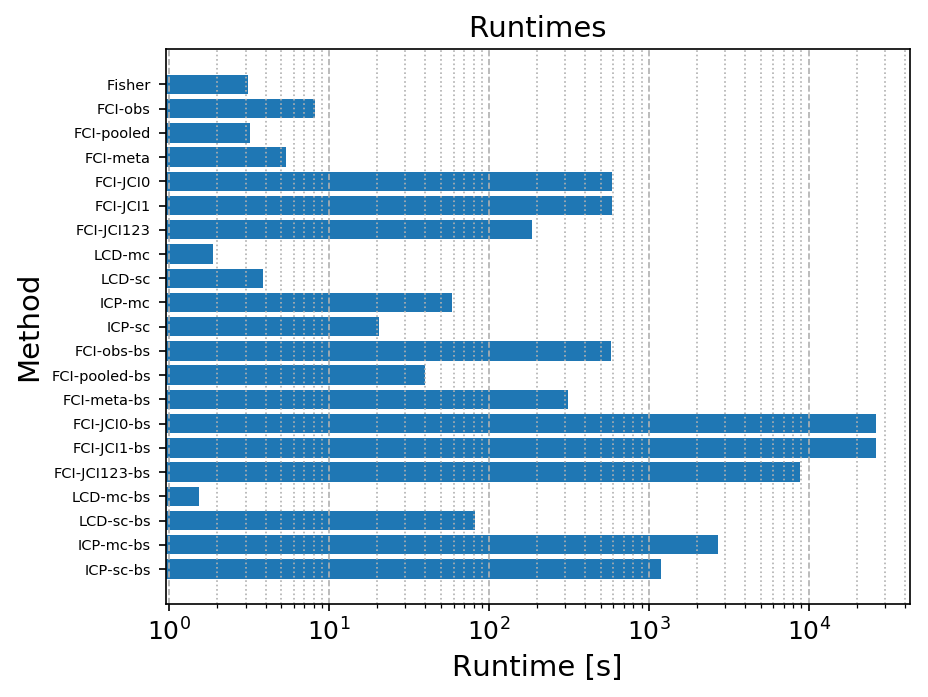}
\includegraphics[width=0.5\textwidth]{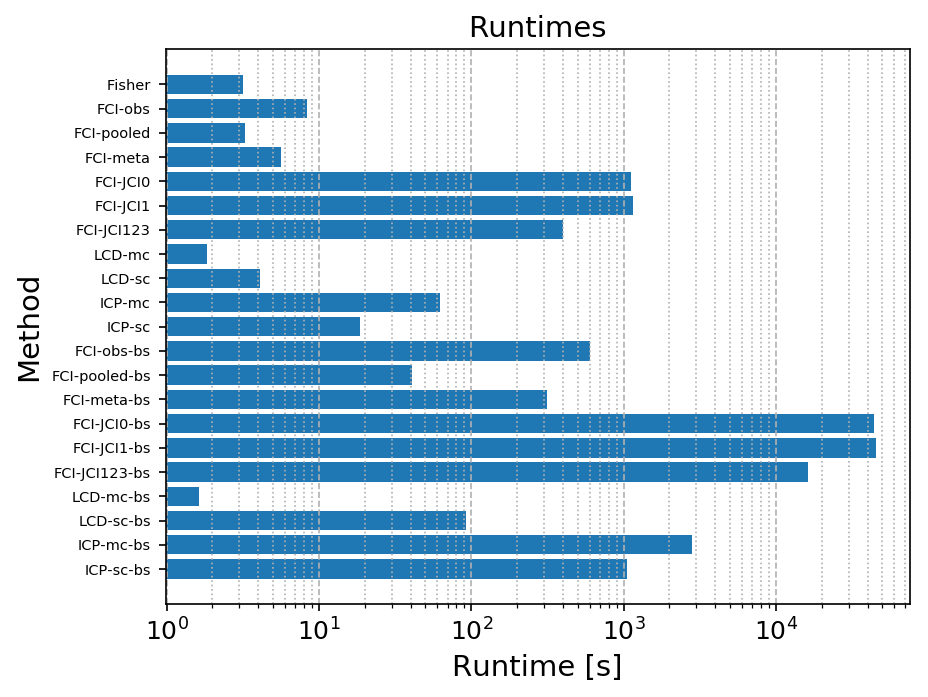}
}
\caption{\boldcap{Runtimes for various methods} on larger models. Left: acyclic; Right: cyclic.\label{fig:simul_p10_q10_runtimes}}
\end{figure}

Figure~\ref{fig:simul_p10_q10_runtimes} shows total runtimes of the methods that we ran on the larger simulated models.
First, we observe no big differences between the runtimes for the cyclic setting with respect to the acyclic one.
However, we do observe huge differences in runtime between various methods.
LCD variants and Fisher's test are by far the fastest. 
ICP variants come second. 
Bootstrapping puts a large toll on computation time for FCI variants.
JCI variants of FCI are much slower than non-JCI variants.
This seems to be mostly due to an exponential increase in the number of conditional independence tests. 
Indeed, we observed that FCI-JCI variants are conditioning on a substantial fraction of all $2^{10}$ subsets of all context variables in the skeleton search phase.
Nevertheless, JCI variants of FCI are still computationally feasible in this setting, even with bootstrapping.

\subsection{Results: Large Simulated Models}

We now present results for large simulated models, with $p=100$ system variables and $q=10$ context variables.
We only consider causal mechanism changes with unknown targets and only the acyclic setting.
We used $\epsilon=\eta=0.02$, which
yields rather sparse graphs, in order to avoid that the computations would take too long
(the meaning of the simulation parameters is explained in Section~\ref{sec:simulations}).
We used only 100 samples per context, because the tasks would become too easy otherwise due to the sparsity of the graphs.
We again used the standard partial correlation test for FCI variants in this setting instead
of the default conditional independence test described in Section~\ref{sec:CI} for computational efficiency reasons.\footnote{It is still possible to use the standard test, and the results are slightly better, but the small gain in precision does not seem to justify the large increase in computation time. Implementing \alg{FCI-JCI123r} as proposed in Section~\ref{sec:fci-jci123r} would probably yield a significant reduction in computation time. Also, alternatives for the skeleton search phase such as FCI+ \citep{ClaassenMooijHeskes_UAI_13} could be employed to gain further speedups. Last but not least, a more efficient implementation of the standard conditional independence test would help considerably.}

\subsubsection{Ancestral Causal Relations between System Variables}

\begin{figure}[t]
\centerline{%
\includegraphics[width=0.24\textwidth]{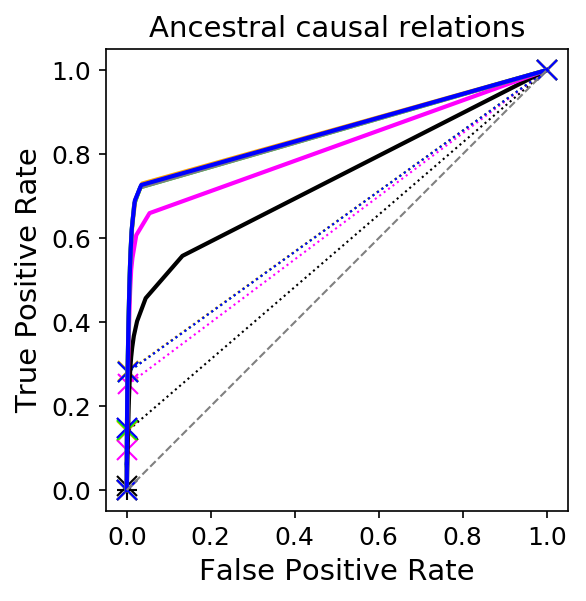}
\includegraphics[width=0.24\textwidth]{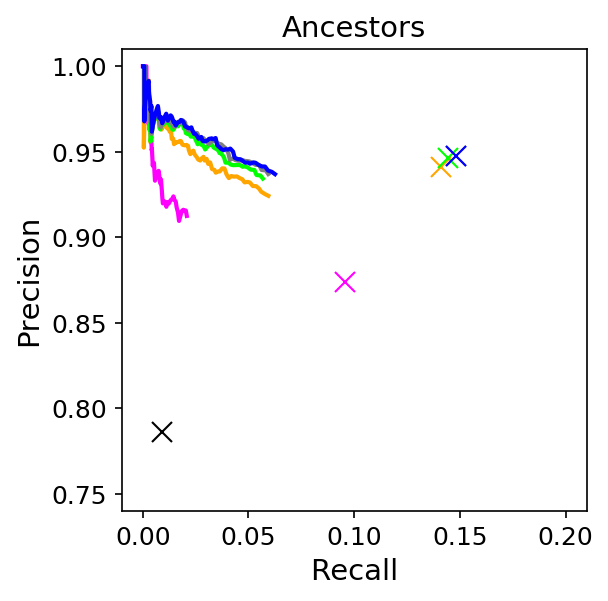}
\includegraphics[width=0.24\textwidth]{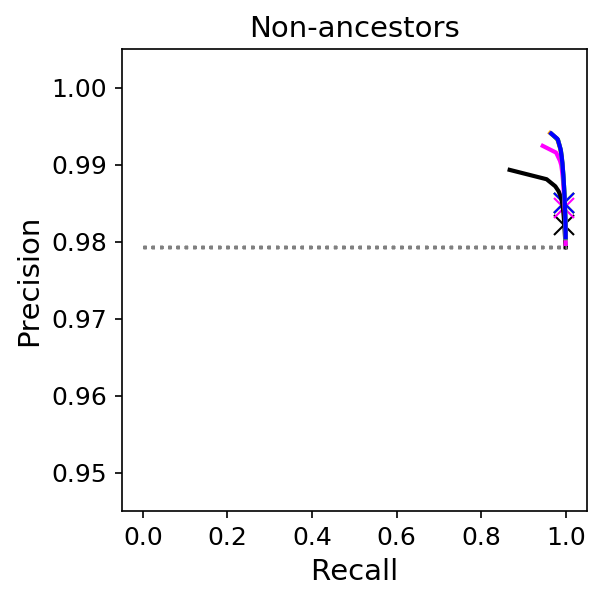}
\includegraphics[width=0.24\textwidth]{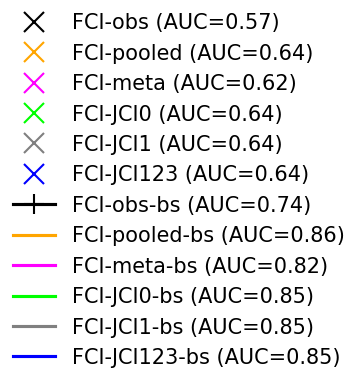}
}
\centerline{%
\includegraphics[width=0.24\textwidth]{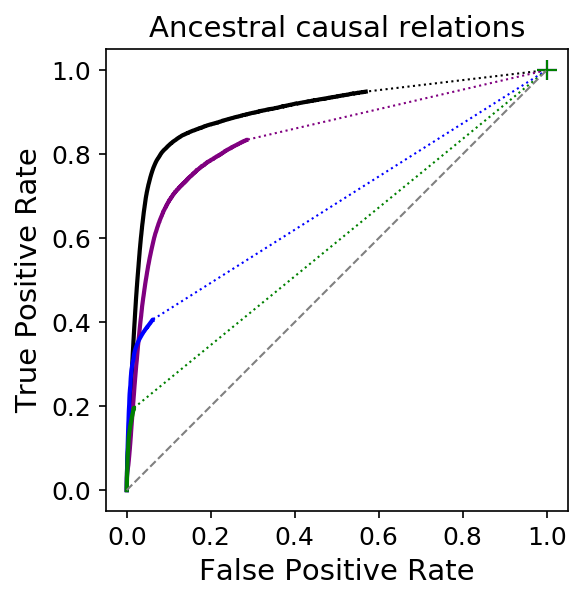}
\includegraphics[width=0.24\textwidth]{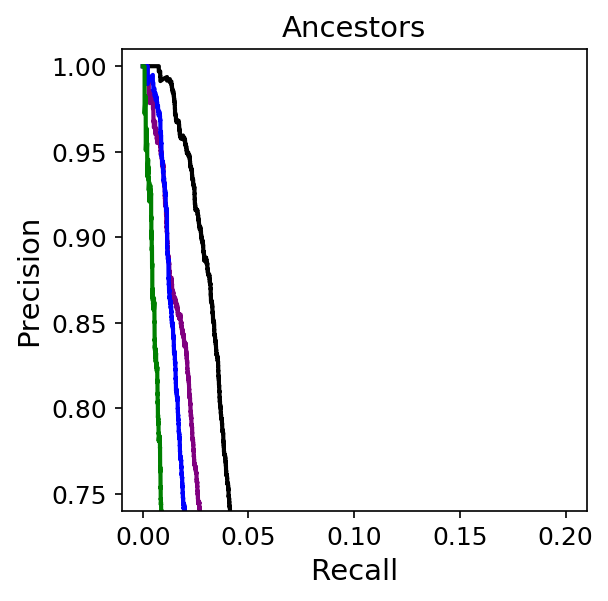}
\includegraphics[width=0.24\textwidth]{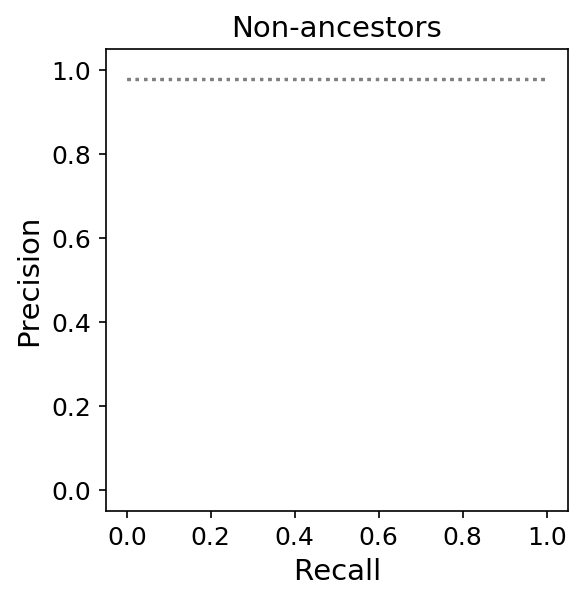}
\includegraphics[width=0.24\textwidth]{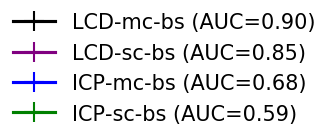}
}
\caption{\boldcap{Discovering ancestral causal relations between system variables} in large models. Top: FCI variants; 
 Bottom: Bootstrapped LCD and ICP variants. Note that we zoomed in on the PR curves.\label{fig:simul_p100_q10_x_mc_arel_sys2sys}}
\end{figure}

Figure~\ref{fig:simul_p100_q10_x_mc_arel_sys2sys} shows the accuracy for the task of discovering ancestral causal relations between system variables for various methods and baselines. 
Overall, we see that JCI variants outperform non-JCI baselines.
On a detailed level, the conclusions are somewhat different than what we saw for smaller models. 

We start by discussing the results for the task of predicting the presence of causal relations.
Like before, we see that bootstrapping FCI helps considerably to increase the precision of the predictions
for the lower recall range. On the other hand, it reduces recall, possibly because only
half of the available data is used and the independence threshold was not adjusted.
As before, \alg{FCI-obs} (which uses only observational data) performs worst.
However, \alg{FCI-obs} outperforms random guessing by a large margin in this setting. 
This is especially noteworthy given that it is only using 100 observational samples. 
Interestingly, the main improvement in this setting is obtained by pooling the data;
whether one includes the context variables (\alg{FCI-JCI0}) or not (\alg{FCI-pooled})
does not seem to make much of a difference.
Also, using more JCI background knowledge yields only small improvements: the differences
between \alg{FCI-JCI0}, \alg{FCI-JCI1} and \alg{FCI-JCI123} are small.

We observe that \alg{LCD-mc-bs} obtains a higher precision at low recall than the FCI-JCI variants. 
On the other hand, the FCI-JCI variants maintain a decent precision over a larger recall range, contrary
to the LCD and ICP variants that only look for very specific patterns, which may explain that their
precision drops off at lower recall than it does for FCI-JCI. The ``multiple context'' (``\alg{-mc}'') 
versions of LCD and ICP outperform the versions that use only a single (merged) context 
(``\alg{-sc}'') in these settings. Interestingly, both variants of LCD outperform the 
corresponding variant of the more complicated ICP algorithm.

For predicting the absence of causal relations, the random guessing baseline already obtains
a high precision, because of the sparsity of the graphs. FCI variants improve on this, roughly
halving the error for the most confident predictions when using the bootstrapped versions. 
FCI-JCI variants again obtain the highest precision on this task, but don't significantly outperform \alg{FCI-pooled}.

\subsubsection{Discovering Intervention Targets}

\begin{figure}[t]
\centerline{%
\includegraphics[width=0.24\textwidth]{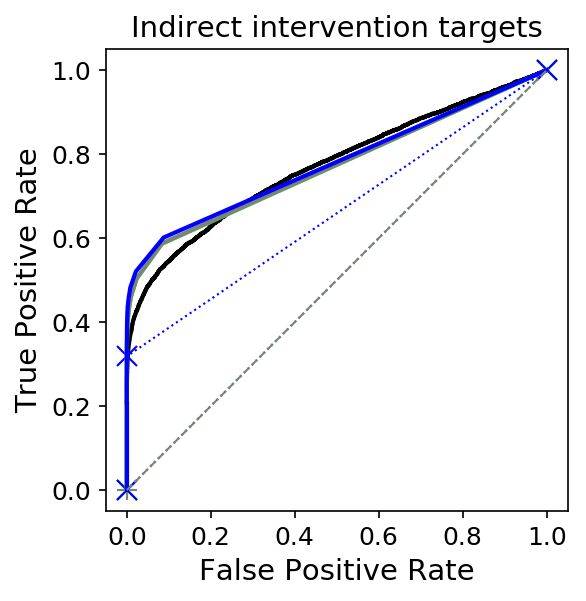}
\includegraphics[width=0.24\textwidth]{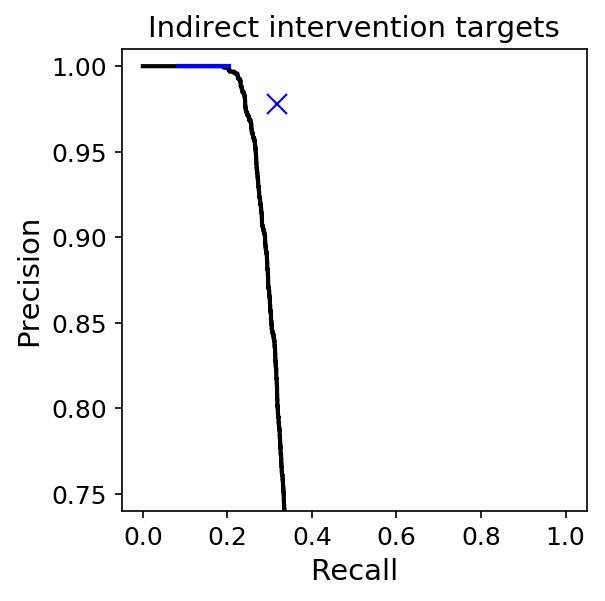}
\includegraphics[width=0.24\textwidth]{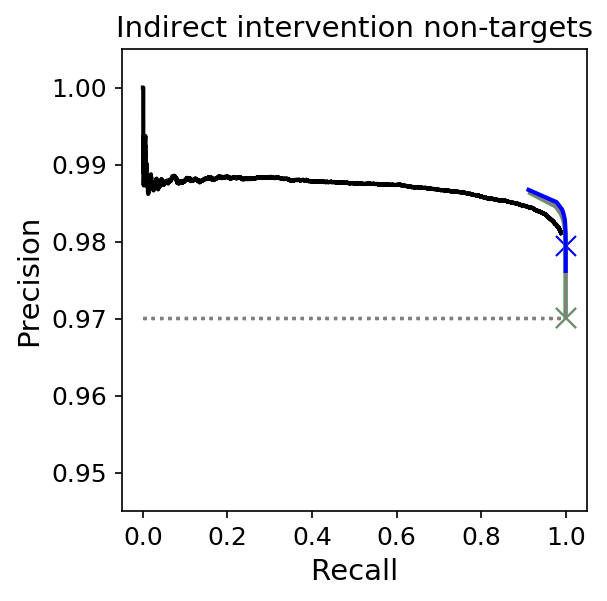}
\includegraphics[width=0.24\textwidth]{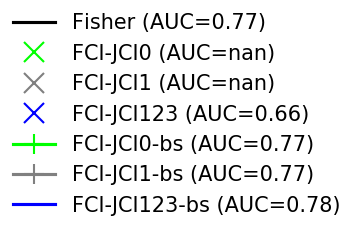}
}
\centerline{%
\includegraphics[width=0.24\textwidth]{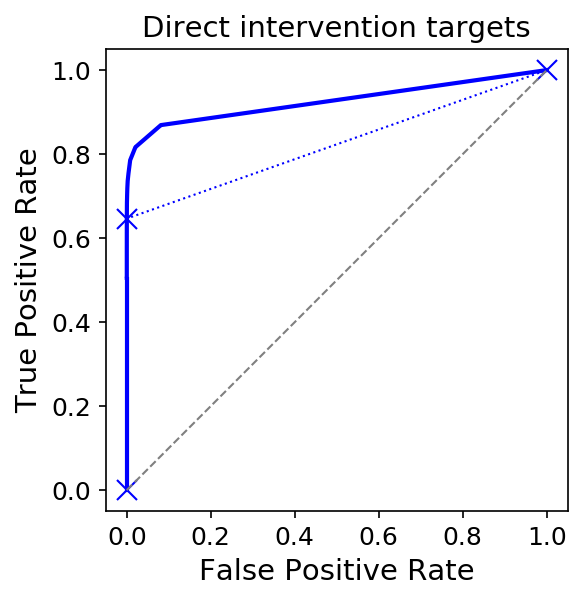}
\includegraphics[width=0.24\textwidth]{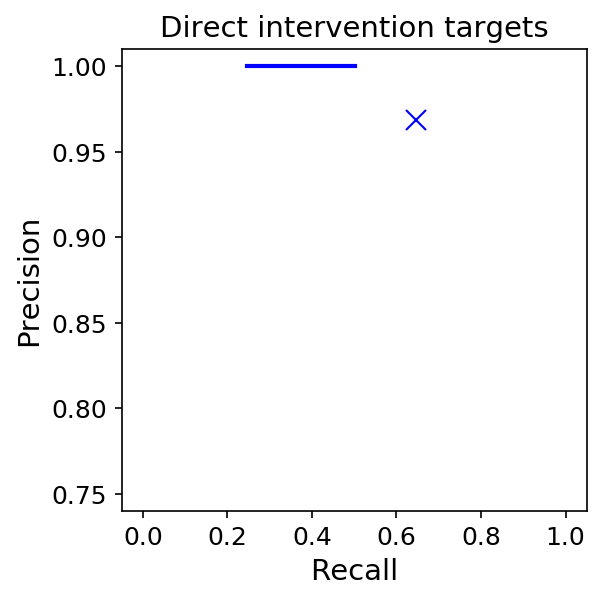}
\includegraphics[width=0.24\textwidth]{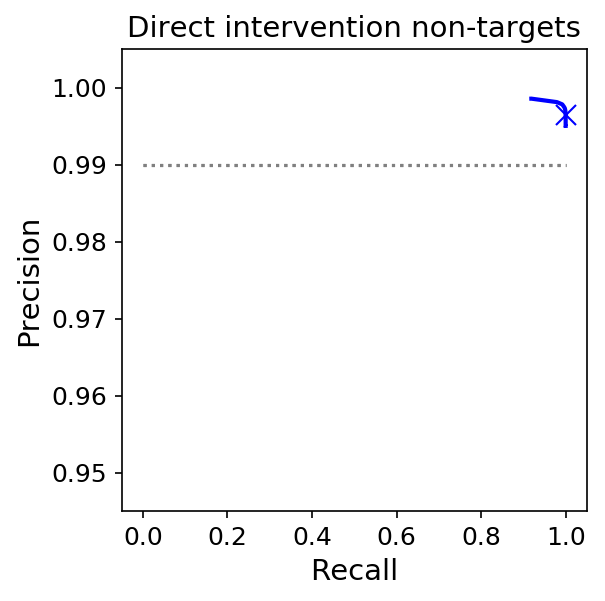}
\includegraphics[width=0.24\textwidth]{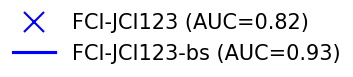}
}
\caption{\boldcap{Results for discovering intervention targets} in large models. Top: indirect intervention targets; Bottom: direct intervention targets. Note that we zoomed in on the PR curves.\label{fig:simul_p100_q10_x_mc_fci_con2sys}}
\end{figure}

Figure~\ref{fig:simul_p100_q10_x_mc_fci_con2sys} shows the performance of FCI-JCI variants (and as a baseline,
Fisher's test) on the task of discovering intervention targets.
For discovering indirect intervention targets, Fisher's test is now slightly outperformed by \alg{FCI-JCI123-bs}. 
Interestingly, JCI Assumption~\ref{ass:unconfounded} seems necessary to obtain any results on that particular task.
Investigating the PAGs shows that the edges between context and system variables are mostly bidirected
for \alg{FCI-JCI1} and \alg{FCI-JCI0}, which explains why these two algorithms yield no predictions at all.
We do not have a good explanation for this behavior, but speculate that the sparse setting with many nodes makes
certain empirical violations of faithfulness quite likely. 
One of the features of \alg{FCI-JCI123} is that it can discover \emph{direct} intervention targets, something
that Fisher's test cannot.
We find that \alg{FCI-JCI123} identifies with high precision direct intervention targets as well as non-targets in this simulation setting.

\subsubsection{Computation Time}

\begin{figure}[t]
\centerline{%
\includegraphics[width=0.5\textwidth]{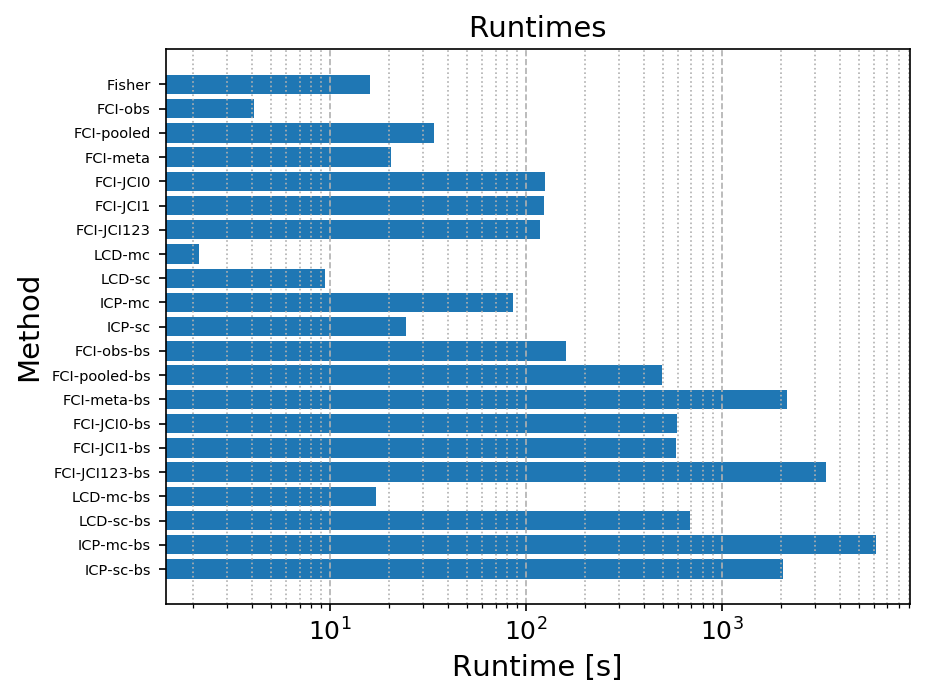}
}
\caption{\boldcap{Runtimes for various methods} on large models.
  \label{fig:simul_p100_q10_runtimes}}
\end{figure}

Figure~\ref{fig:simul_p100_q10_runtimes} shows total runtimes of the methods that we ran on the large simulated models.
For most methods, the largest part of the total running time is spent on performing independence tests. 
In particular, the tests that subdivide data according to context are relatively slow since we have not seriously optimized their implementation.

\subsection{Summary of Results on Simulated Data}

We have seen in our experiments with simulated data that JCI methods typically
outperform non-JCI methods, in some settings by a large margin. For certain
tasks, our newly proposed FCI-JCI algorithms provide the new state-of-the-art.
Interestingly, Fisher's baseline turned out to be hard to beat on the task of
discovering indirect intervention targets. However, there are other tasks for
which Fisher's baseline cannot be applied but for which our newly proposed
methods do apply, such as the task of discovering direct intervention
targets.

As expected, LCD, ICP and ASD variants work in both the acyclic and cyclic
setting. While FCI variants were expected to work only in the acyclic setting,
we were surprised by how well they perform in the cyclic setting.\footnote{These
empirical observations led us to conjecture that FCI does not need to be adapted
for the $\sigma$-separation setting. It was shown very recently that this is indeed
the case: FCI is also sound and complete in that setting \citep{MooijClaassen_2005.00610}.}
Often, but not always, adding more context variables leads to better results.
Having multiple contexts was seen to be mostly beneficial if each context
variable targets only a small subset of system variables, and then only for
methods that can explicitly take into account multiple context variables.
An interesting exception to this are LCD and ICP, which due to their
sensitivity to very specific patterns actually degrade in performance
when too many system variables become directly targeted by context
variables. Exploiting more JCI background knowledge typically led to better
results, but it depends on the task and simulation setting how large the
benefits are. Interestingly, the largest boost in accuracy for discovery
of causal relations between system variables comes already from JCI
Assumption~\ref{ass:simple_scm} (i.e., from pooling the data
and adding the context variables).

As to the relative merits of the various JCI methods that we compared, it is
difficult to state this concisely. Different methods behave differently on
different tasks in different simulation settings, both in terms of precision
and recall as well as in terms of computation
time, and for many methods there is a combination of a task and a simulation
setting in which they do relatively well. Generally speaking, LCD and ICP behave quite
similarly, are relatively fast and obtain high precision but have low recall;
ASD-JCI variants are among the most accurate and have highest recall, but
computation time explodes for more than a handful of variables; performance of
FCI-JCI methods turns out to be somewhere in between, both in terms of accuracy
and in terms of computation time.
When implemented properly, their scalability is comparable to that of 
standard FCI, where the total number of variables $|\C{I}| + |\C{K}|$ and the 
sparsity of the underlying MAGs are important factors in the computation time.

Some aspects that should be kept in mind when interpreting these results is
that all simulations have been done under JCI Assumptions~\JCIABC, and
there was no model misspecification. From that perspective, it was to be
expected that JCI methods would work well. Also, we make no claims as to how
our conclusions would generalize to different settings, for example, with 
non-Gaussian noise distributions, discrete variables, or continuous variables 
with non-linear interactions. For those settings, the choice of the conditional
independence tests could have a large influence on the results, for example. 
A detailed study of that is beyond the scope of this paper.

\subsection{Results: Real-world (Flow Cytometry) Data\label{sec:exp_sachs}}

In this subsection, we present an application of the Joint Causal Inference framework on 
real-world data: the flow cytometry data of \citet{SPP05}. 
The data consists of a collection of data sets, where each data set corresponds with a different experimental condition in which a system was perturbed and subsequently measured. 
The system consists of an individual cell, drawn randomly from a collection of primary human immune system cells.
The system variables measure the abundances of several phosphorylated protein and phospholipid components in an individual cell using a measurement technology known as flow cytometry.
Performing the measurement destroys the cell, and hence, it is not possible to obtain multiple measurements over time from the same cell. 
Instead, snap-shot measurements of thousands of individual cells are available, obtained in different experimental conditions in which the cells were perturbed with molecular interventions, performed by administering certain reagents to the cells.
Most of these interventions are not perfect, but rather change the activity of some component by an unknown amount.
There is prior knowledge about the targets of these interventions (see Table~\ref{tab:sachs_contexts}), but it is not clear whether interventions are as specific as claimed.
Many existing causal discovery approaches assume that the true causal graph is acyclic and that the system variables are causally sufficient. 
However, it is known that these cellular signaling networks contain strong feedback loops, and it is quite likely that some of the variables may be subject to latent confounding.
Thus, this type of experimental data constitutes a compelling motivation for the Joint Causal Inference framework.

\begin{figure}
  \centerline{\includegraphics[width=0.4\textwidth]{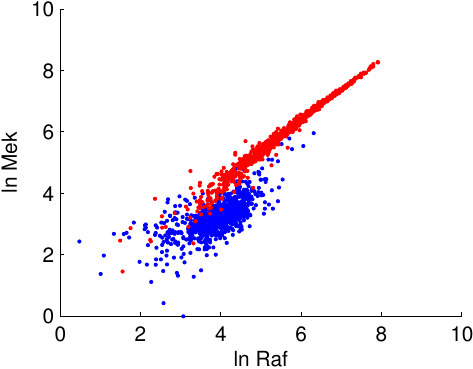}\hfill\includegraphics[width=0.4\textwidth]{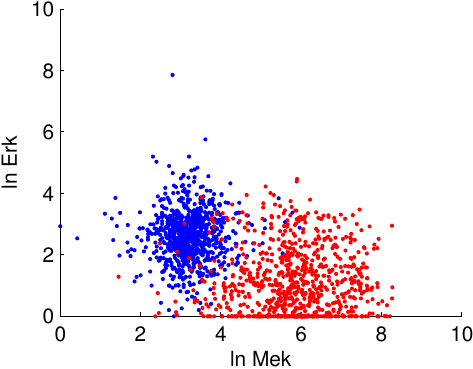}}
  \caption{Log-abundances of Mek vs.\ Raf (left) and of Erk vs.\ Mek (right). Blue: observational baseline ($C_\alpha=1, \B{C}_{\setminus \alpha}=\B{0}$); Red: reagent U0126 added ($C_\zeta=1$). We observe: (i) the measurement noise is quite small; (ii) Raf and
  Mek are highly correlated (``consensus network'': Raf is a direct cause of Mek); 
  (iii) strong evidence for feedback (intervening on Mek changes Raf abundance) if we assume that U0126 directly
  targets Mek but not Raf; (iv) the Mek \emph{inhibitor} U0126 \emph{increases} Mek abundance (so modeling this as a perfect
  intervention would not be realistic); (v) Mek and Erk are independent in both contexts
  (even though Mek is a direct cause of Erk according to the ``consensus network''), an apparent violation of
  faithfulness.\label{fig:raf_mek_erk}}
\end{figure}

\begin{figure}
  \centering
  \subfigure[``Consensus Network'']{\label{fig:Sachs_PAGs_consensus}\includegraphics[width=0.6\textwidth]{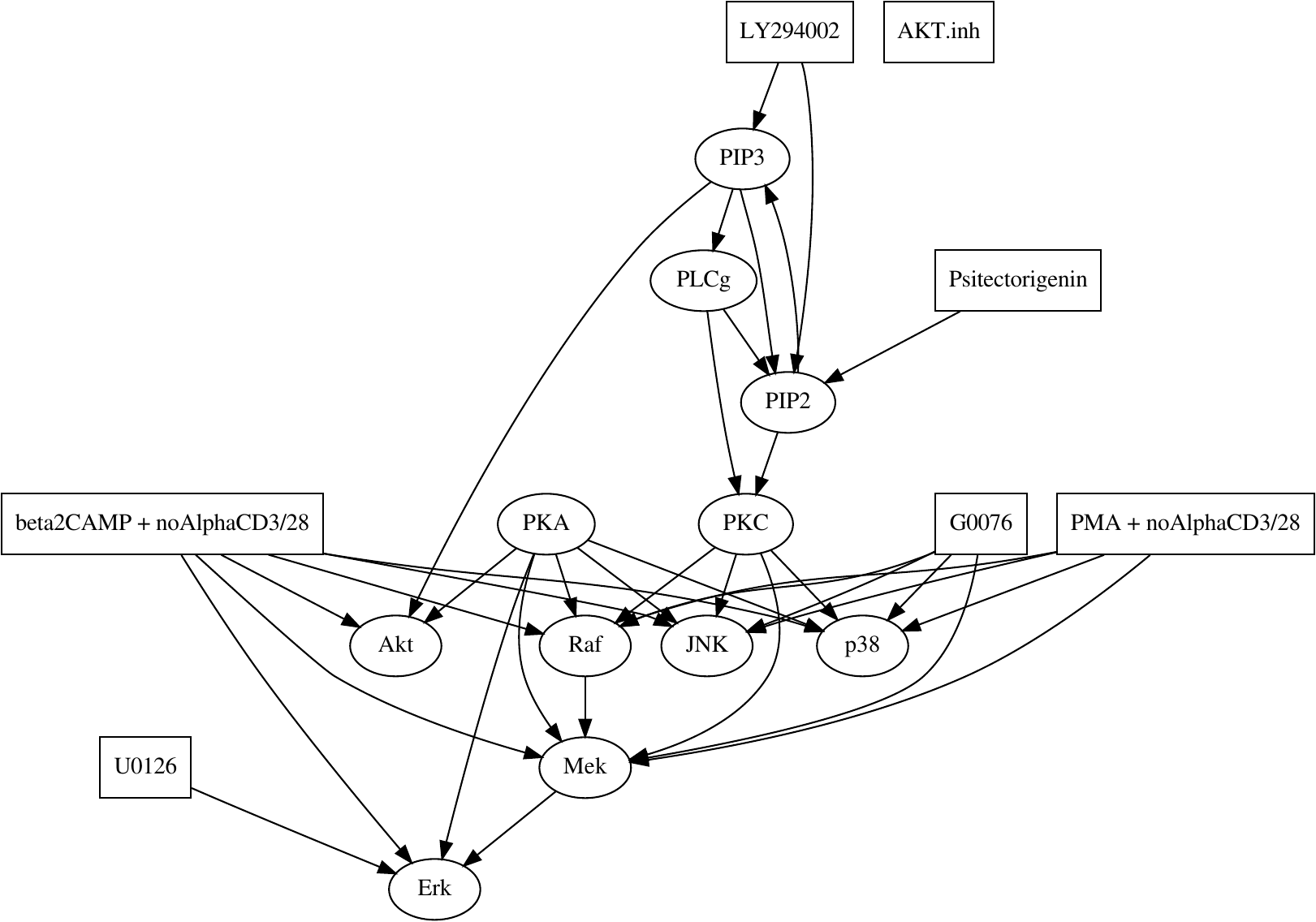}}\hfill
  \subfigure[\alg{FCI-pooled}]{\label{fig:Sachs_PAGs_FCI-pooled}\includegraphics[width=0.25\textwidth]{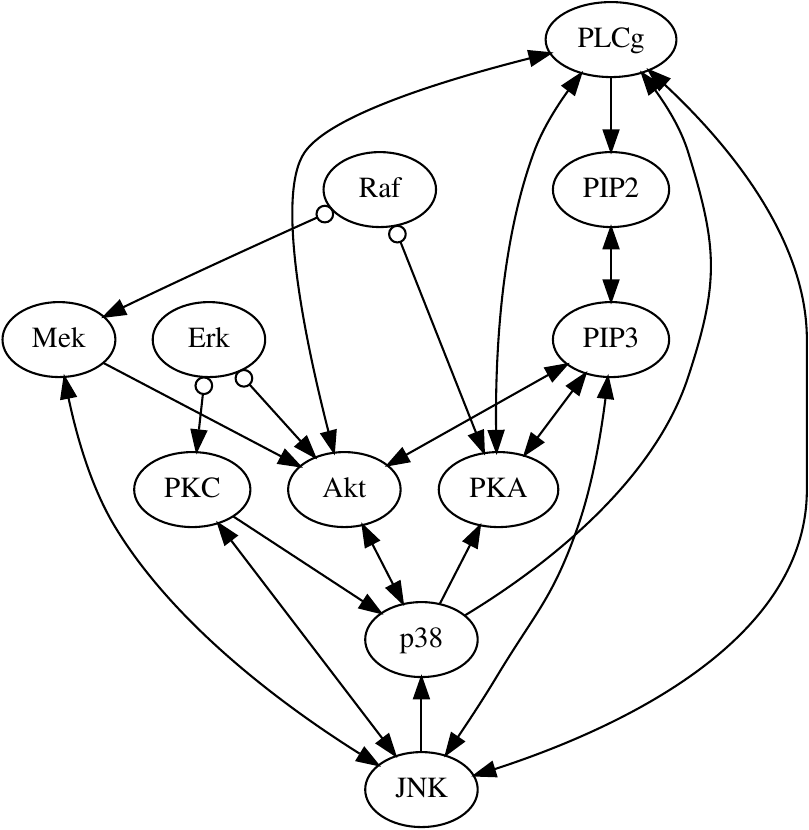}}\\
  \subfigure[\alg{FCI-obs}]{\label{fig:Sachs_PAGs_FCI-obs}\includegraphics[width=0.4\textwidth]{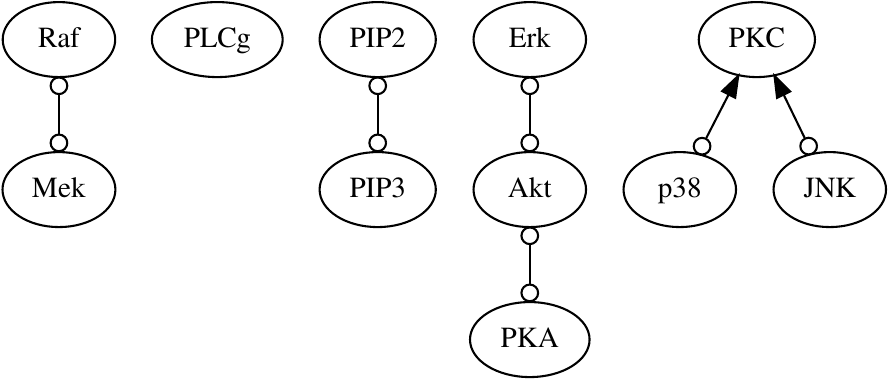}}\hfill
  \subfigure[\alg{FCI-meta}]{\label{fig:Sachs_PAGs_FCI-meta}\includegraphics[width=0.4\textwidth]{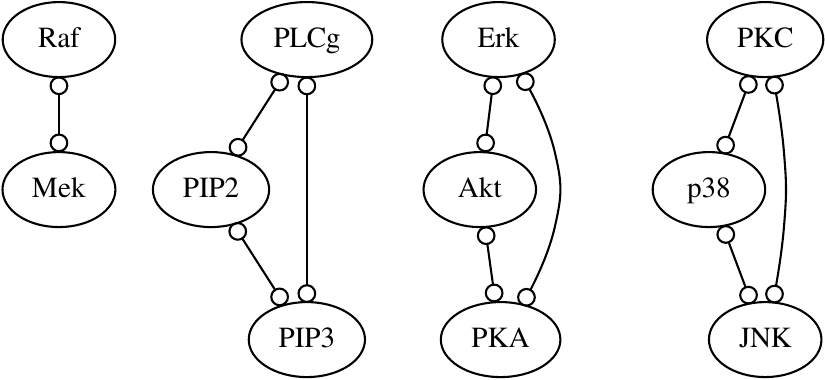}}
  \caption{\label{fig:Sachs_PAGs_baselines}\boldcap{PAGs resulting from various FCI baselines on the flow cytometry data of \citet{SPP05}.}
  Intervention variables are denoted with rectangles, system variables with ellipses.
  From top to bottom:
  (a) ``Consensus network'' according to \citet{SPP05}; 
  (b) FCI on pooled data (without adding the context variables);
  (c) FCI on the first (``observational'') data set in which only global activators $\alpha$-CD3 and $\alpha$-CD28 have been administered;
  (d) FCI with as input the result of Fisher's method for combining conditional independence test results from all data sets.}
\end{figure}

Over the years, this particular flow cytometry data set has become a ``benchmark'' in causal discovery
(see e.g.\ \citet{RamseyAndrews2018} for some references).
Many causal discovery methods have been applied to this data, and in many
cases, the ``consensus network'' in \citet{SPP05}, visualized in Figure~\ref{fig:Sachs_PAGs_baselines}(a), 
was used as a ground truth
to evaluate the results of the causal discovery procedure. However, we would like
to point out that there are good reasons to be skeptical about the assumption that the ``consensus network'' 
represents the true causal graph of the system (as acknowledged by several domain experts we spoke).
Indeed, by inspecting the data one can find many examples where the data is 
incompatible with the hypothesis that the ``consensus network'' is a realistic and complete
description of the underlying system. For example, according to the ``consensus network'', an intervention that
inhibits the activity of Mek should have no effect on Raf (because Raf is
a direct cause of Mek and there is no feedback loop from Mek to Raf). 
However, in the data we see an increase of more
than one order of magnitude in the abundance of Raf when U0126 (a reagent assumed
to inhibit Mek activity) is added to the cells (see Figure~\ref{fig:raf_mek_erk}). 
So either U0126 also directly targets Raf, or Mek must be a cause of Raf, in both cases contradicting the
``consensus network''. In the literature regarding this signaling
pathway (which has been studied in great detail since it plays an important role in many human cancers) it is often suggested that there 
should be a feedback loop back from Erk to Raf
(whose molecular mechanism is still unknown).
This would be in line with our observations from the data in Figure~\ref{fig:raf_mek_erk}, and also imply that the ``consensus
network'' is incomplete.
This is just one example illustrating that the data is not entirely compatible with
the ``consensus network'', and it is easy to find more of these examples via visual 
inspection of the data.
Therefore, we will \emph{not} make use of the ``consensus network'' as a ground truth 
to compare with when evaluating the output of the causal discovery algorithms.

\begin{table}\centering
{\small\begin{tabular}{|cccccc|c|l|}
\hline
  $C_\gamma$ & $C_\delta$ & $C_\epsilon$ & $C_\zeta$ & $C_\eta$ & $(C_\alpha,C_\theta,C_\iota)$ & $N_{\B{C}}$ & Reagents added \\
\hline
  0 & 0 & 0 & 0 & 0 & (1,0,0) & 853 & $\alpha$-CD3, $\alpha$-CD28\\
  1 & 0 & 0 & 0 & 0 & (1,0,0) & 911 & $\alpha$-CD3, $\alpha$-CD28, AKT inhibitor\\
  0 & 1 & 0 & 0 & 0 & (1,0,0) & 723 & $\alpha$-CD3, $\alpha$-CD28, G0076\\
  0 & 0 & 1 & 0 & 0 & (1,0,0) & 810 & $\alpha$-CD3, $\alpha$-CD28, Psitectorigenin\\
  0 & 0 & 0 & 1 & 0 & (1,0,0) & 799 & $\alpha$-CD3, $\alpha$-CD28, U0126\\
  0 & 0 & 0 & 0 & 1 & (1,0,0) & 848 & $\alpha$-CD3, $\alpha$-CD28, LY294002\\
  0 & 0 & 0 & 0 & 0 & (0,1,0) & 913 & PMA\\
  0 & 0 & 0 & 0 & 0 & (0,0,1) & 707 & $\beta$2CAMP\\
\hline
\end{tabular}}
\caption{Experimental design for part of the \citet{SPP05} flow cytometry data used in our experiments.\label{tab:sachs_experimental_design_used}}
\end{table}

\citet{SPP05} use an MCMC method to estimate the structure of a
causal Bayesian network from the combined observational and interventional data, making use of the modified
BDe score proposed by \citet{CooperYoo1999}. \citet{EatonMurphy07} later used a dynamic programming
algorithm to solve the estimation problem exactly.
Like the original analysis by \citet{SPP05}, many causal discovery methods
that have been applied on this data rely on the background knowledge about the intervention
types and targets, for which we provide (our interpretation) in Table~\ref{tab:sachs_contexts}. A notable exception
is \citet{EatonMurphy07}, who were the first to estimate the intervention targets directly from the data. 
Exploiting the background knowledge on intervention targets and types simplifies the causal 
discovery problem considerably.
However, the accuracy of this background knowledge is not universally accepted. In
particular, many biologists that we spoke with were skeptical about the assumed specificity of the interventions
(i.e., the interventions may have additional direct effects that are not listed in the table).

We ran various FCI-JCI variants on a subset of the flow cytometry data.\footnote{As preprocessing, we simply took the logarithm of the raw values.}
The experimental design of the original data is described in Table~\ref{tab:sachs_experimental_design} (left), p.\ \pageref{tab:sachs_experimental_design}. 
In order to avoid deterministic relations between context variables, we merged
context variables $C_\alpha$ ($\alpha$-CD3/CD28), $C_\theta$ (PMA) and $C_\iota$ ($\beta$2CAMP),
as discussed in Section~\ref{sec:faithfulness}, leading to the experimental design in Table~\ref{tab:sachs_experimental_design} (right).
This means that we must interpret the merged context variable as referring to the addition of PMA or $\beta$2CAMP, combined with the \emph{omission} of $\alpha$-CD3/CD28. 
However, note that there are still approximate conditional independences in this experimental design of the form 
$C_\beta \CI C_k \given (C_\alpha,C_\theta,C_\iota)$ for $k \in \{\gamma,\delta,\epsilon,\zeta,\eta\}$.
This could lead to problems with JCI Assumption~\ref{ass:dependences}.
For that reason, but also in order to enable comparisons with other results reported in the literature, we only used the 8 (out of 14) experimental conditions in the data set in which no ICAM.2 had been administered (i.e., with $C_\beta=0$), and ignored the others.

Similarly to \citet{EatonMurphy07}, we do not use the background knowledge regarding intervention types or targets. 
We only assume that the experimental setting is captured by the JCI framework.
JCI Assumption~\ref{ass:uncaused} should be true because the intervention is performed some time (approximately 20 minutes) before the measurements are done.
We have already discussed the validity of JCI Assumption~\ref{ass:unconfounded} for this particular experimental setting in Section~\ref{sec:JCI_assumptions}. Assuming that the context variables provide a complete causal description of the context (in particular, that there are no unintended batch effects), JCI Assumption~\ref{ass:unconfounded} applies.
JCI Assumption~\ref{ass:dependences} then also applies since there are no conditional independences in the context distribution (after merging $C_\alpha$, $C_\theta$ and $C_\iota$ and leaving out all contexts with $C_\beta = 1$). 
When using \alg{FCI-JCI123}, we can then learn the intervention targets from the data itself, without making use of the background knowledge on intervention types and targets. 

\begin{figure}\centering
\includegraphics[width=0.49\textwidth]{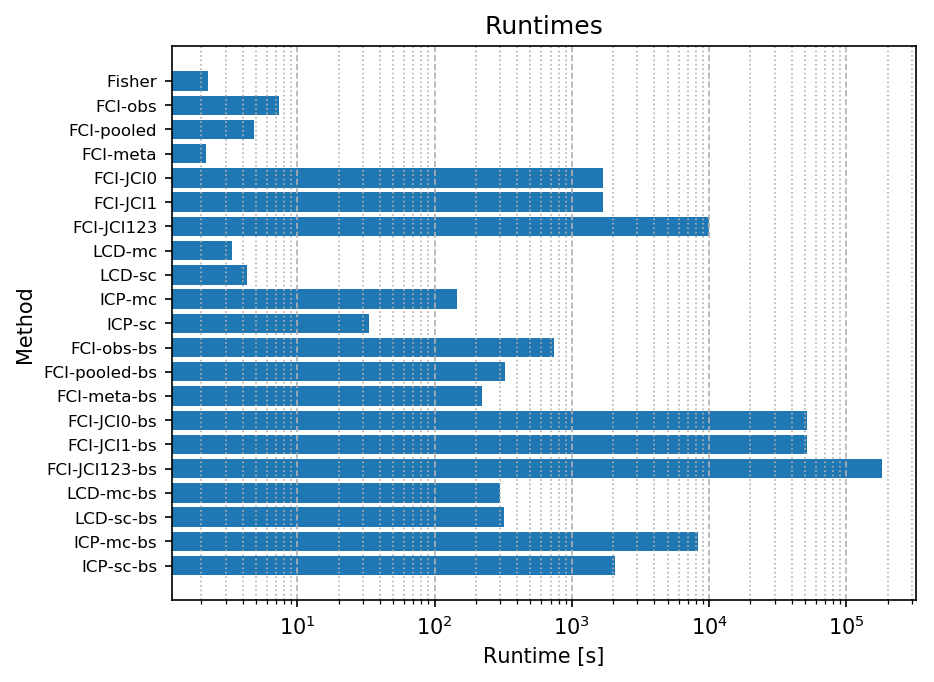}
\caption{\boldcap{Runtimes for various methods on the flow cytometry data of \citet{SPP05}.}\label{fig:sachs_runtimes}}
\end{figure}

For comparison, we also ran \alg{FCI-obs}, i.e., standard FCI using only the observational data set
(i.e., the one in which only global activators $\alpha$-CD3 and $\alpha$-CD28 have been administered). 
We also ran \alg{FCI-meta}, which uses Fisher's method to combine $p$-values of conditional independence tests in the 8 separate experimental conditions, which are then used as input for standard FCI. Finally, we ran \alg{FCI-pooled}, i.e., standard FCI on the 8 experimental conditions pooled together (but excluding context variables). In all those FCI variants, we assumed that no selection bias would be present.
We additionally compare with other JCI implementations, in particular, multiple variants of LCD and ICP.
Computation times of various implementations are reported in Figure~\ref{fig:sachs_runtimes}.

\begin{figure}
  \centering
  \subfigure[\alg{FCI-JCI123}]{\label{fig:Sachs_PAGs_FCI_JCI123}\includegraphics[width=0.5\textwidth]{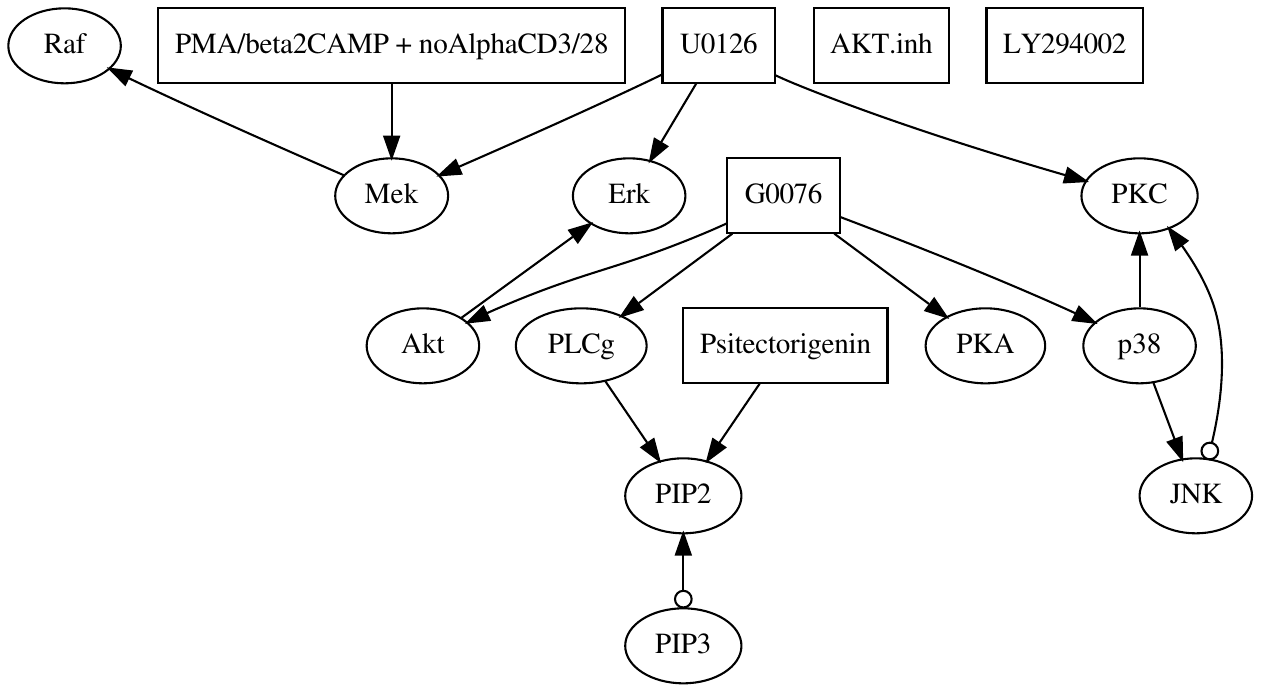}}\\
  \subfigure[\alg{FCI-JCI1}]{\label{fig:Sachs_PAGs_FCI_JCI1}\includegraphics[width=0.5\textwidth]{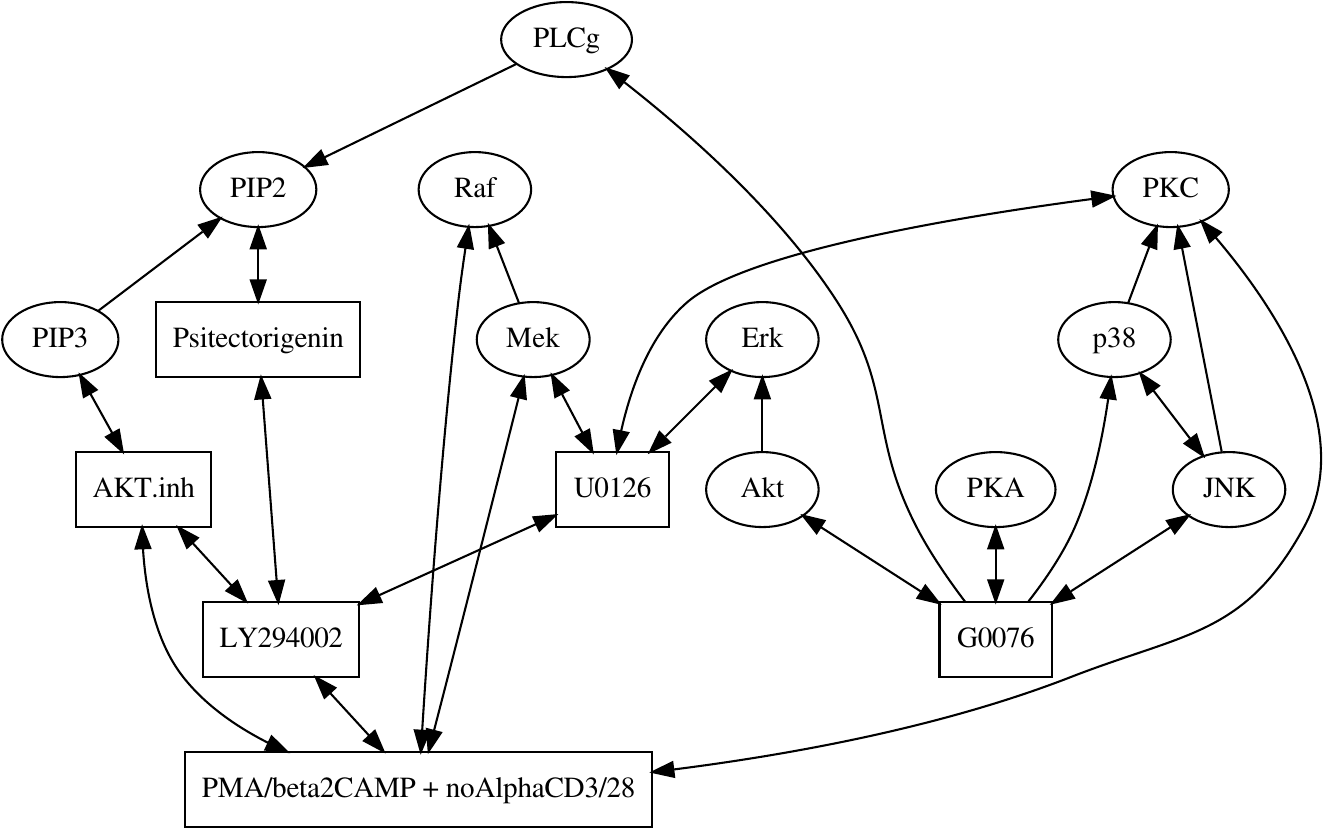}}
  \subfigure[\alg{FCI-JCI0}]{\label{fig:Sachs_PAGs_FCI_JCI0}\includegraphics[width=0.5\textwidth]{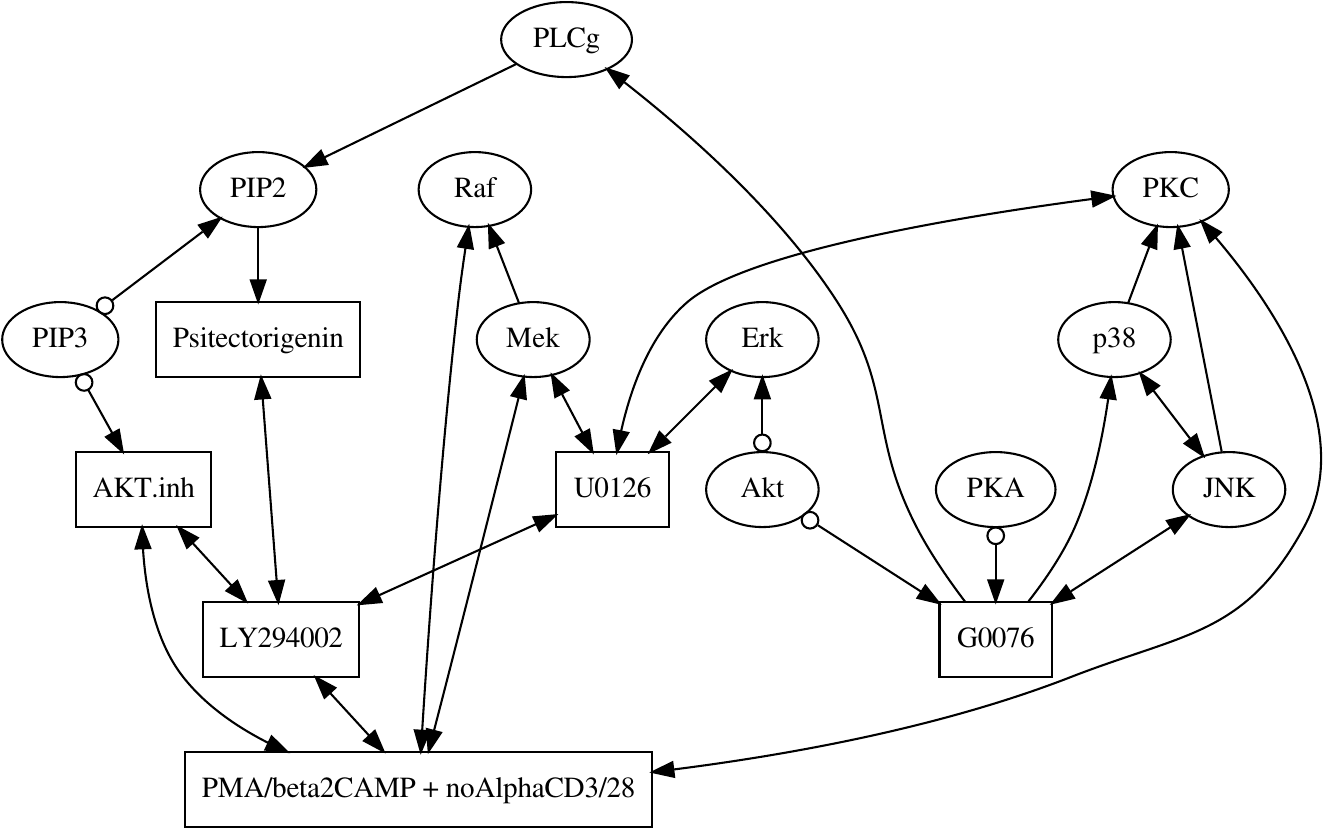}}
  \caption{\label{fig:Sachs_PAGs_JCI}\boldcap{PAGs resulting from various FCI-JCI variants on the flow cytometry data of \citet{SPP05}.}
  These causal discovery methods do not make use of the biological prior knowledge regarding intervention types and targets,
  but learn the intervention targets from the data.
  Intervention variables are denoted with rectangles, system variables with ellipses. 
  From top to bottom, less JCI Assumptions are made. Note that these are individual PAGs that have not been bootstrapped.
  To get an idea of the robustness, Figure~\ref{fig:comparisonSachs} shows also the corresponding bootstrap estimates for certain features of the PAGs.}
\end{figure}

\begin{figure}\centering
\includegraphics[height=17cm]{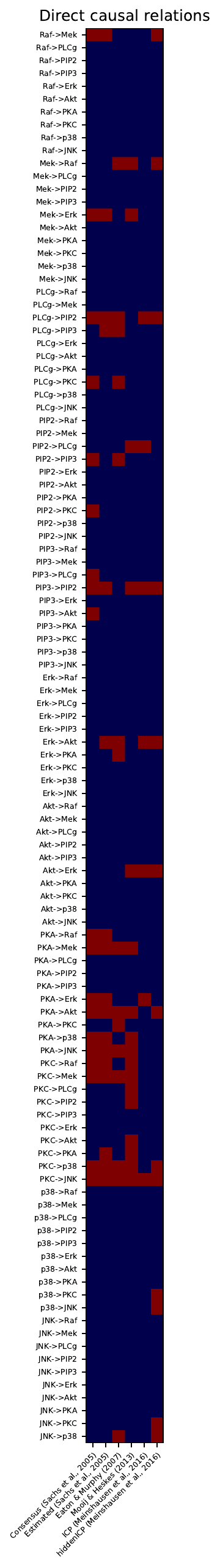}
\quad
\includegraphics[height=17cm]{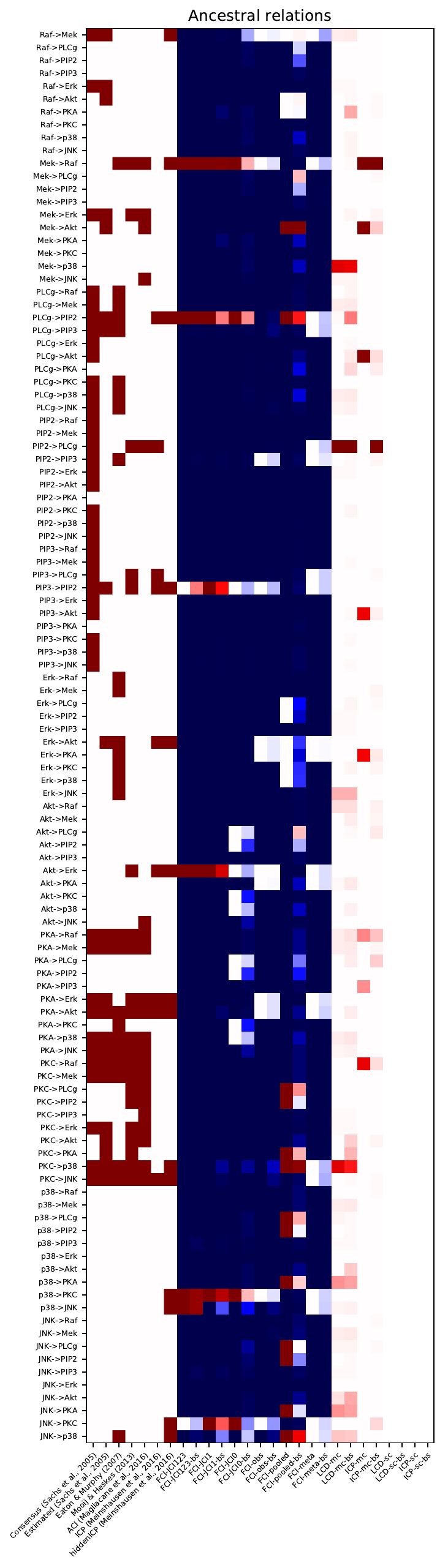}\\
\includegraphics[width=0.5\textwidth]{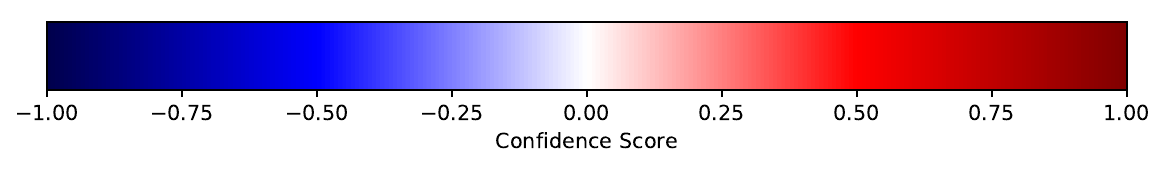}
  \caption{\boldcap{Causal relationships between the biochemical agents in the flow cytometry data of \citet{SPP05}}, according to different causal discovery methods and the ``consensus network'' according to \citet{SPP05} (which we do not consider as a reliable complete ground truth). We also included results of causal discovery methods reported in other works.
  \label{fig:comparisonSachs}}
\end{figure}

The ``consensus network'' and the PAGs obtained by the FCI baselines are shown in Figure~\ref{fig:Sachs_PAGs_baselines}. 
Figure~\ref{fig:Sachs_PAGs_JCI} shows PAGs obtained by the FCI-JCI variants. Note that we show the PAGs obtained without bootstrapping, although these are not necessarily stable. Therefore, we show in Figure~\ref{fig:comparisonSachs} also the bootstrapped results for the learned ancestral causal relations between system variables, and the learned intervention targets. One can see here that most, but not all, of these features are stably predicted by the FCI-JCI variants. In particular, the causal relations\footnote{We write $i \causes j$ if $i$ is a cause of $j$.} Mek$\causes$Raf, PLCg$\causes$PIP2, Akt$\causes$Erk, P38$\causes$PKC are predicted by both \alg{FCI-JCI123} and \alg{FCI-JCI1} with high confidence.

Regarding the learned indirect intervention targets, \alg{FCI-JCI123} has an advantage over \alg{FCI-JCI1} because it can exclude bidirected edges between context and system variables. Nonetheless, \alg{FCI-JCI1} predicts (in accordance with \alg{FCI-JCI123}) that G0076 affects PLCg, PIP2, PKC and P38.
For discovering indirect intervention targets, \alg{Fisher}'s test for causality is simple and powerful. However, it is not able to learn the \emph{direct} intervention targets, like \alg{FCI-JCI123} can. Although the direct intervention targets learned by \alg{FCI-JCI123} do not all correspond with the ``consensus network'', they do agree for example on Psitectorigenin directly targeting PIP2. Most direct intervention targets learned by \alg{FCI-JCI123} were also found by \citet{EatonMurphy07}. It is interesting to see that \alg{FCI-JCI123} considers both Mek and Erk as direct intervention targets of U0126, while Raf is identified as an indirect target. 

In the absence of a reliable ground truth, we draw the following conclusions from these results. 
First, the reference methods that exploit knowledge on intervention types and targets obtain rather consistent results.
Second, the JCI methods that do not assume knowledge on intervention types and targets (and the approach by \citet{EatonMurphy07}) show less consistent results.
This could indicate that the data contain not enough signal in order to solve this more ambitious task reliably.
In that case, some model misspecification (for example, strong non-linearities or deviations from Gaussianity, which makes a simple partial correlation based conditional independence test inadequate) could lead to inconsistencies between methods.
Nonetheless, the performance of the FCI-JCI variants appears to be a considerable improvement over the simple FCI baselines
\alg{FCI-obs,} \alg{FCI-meta} and \alg{FCI-pooled}.
Remarkably, \alg{FCI-JCI123} manages to orient most of the edges.
The output also resembles the consensus network, although some of the edges seem to be reversed. 
Considering that we have not taken into account the available background knowledge on the intervention types and targets (Table~\ref{tab:sachs_contexts}), and that we have not used any tuning of the parameters we consider this still an impressive and encouraging result that illustrates the potential that JCI has for analyzing complex scientific data sets.

\begin{figure}\centering
  \includegraphics[height=17cm]{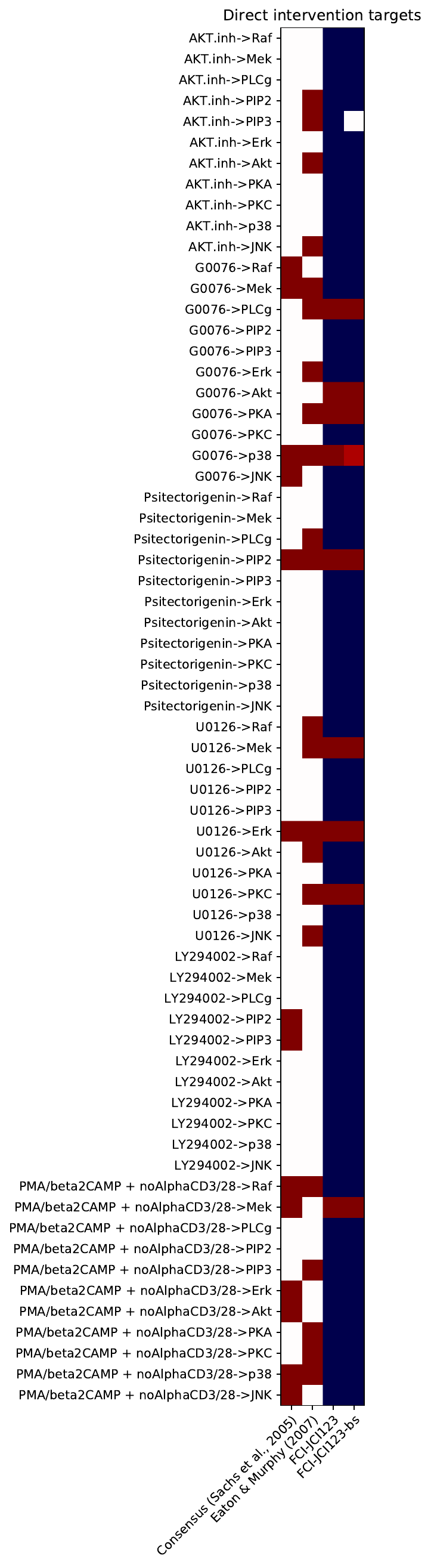}
  \includegraphics[height=17cm]{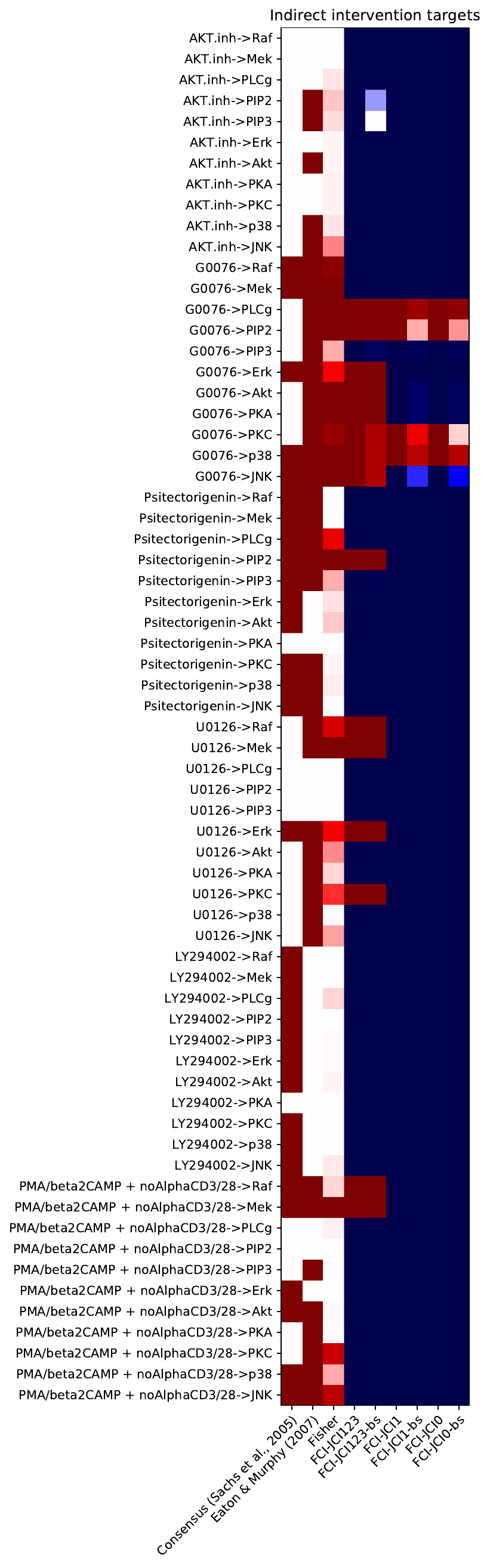}\\
  \includegraphics[width=0.5\textwidth]{sachs_noICAM_LEG.pdf}
  \caption{\boldcap{Intervention effects on biochemical agents in the flow cytometry data of \citet{SPP05}}, according to different causal discovery methods and the ``consensus network'' according to \citet{SPP05} (which we do not consider as a reliable complete ground truth). We also included results of causal discovery methods reported in other works.
  \label{fig:comparisonSachsInterventions}}
\end{figure}

\section{Conclusions and Discussion}\label{sec:conclusion}

In this work, we proposed Joint Causal Inference (JCI), a powerful and elegant
framework for causal discovery from data sets from multiple contexts. JCI
generalizes the ideas of causal discovery based on experimentation (as in
randomized controlled trials and A/B-testing) to multiple context and system
variables. Seen from another perspective, it also generalizes the ideas of
causal discovery from purely observational data to the setting of data sets from
multiple contexts---for example, different interventional regimes---by reducing
the latter to a special case of the former, with additional background
knowledge on the causal relationships involving the context variables.
We proposed different flavours of JCI that differ in the amount of background knowledge that is
assumed, some being more conservative than others.
JCI can
be implemented with any causal discovery method that can take into account the 
background knowledge. Surprisingly, we saw that one can even apply an
off-the-shelf causal discovery algorithm for purely observational data on the
pooled data (with context variables included), completely ignoring the background knowledge, and thereby
already obtain significant improvements in the accuracy of the discovered
causal relations.

We have seen how JCI deals with different 
types of interventions in a unified fashion, how it reduces learning intervention targets
to learning the causal relations between context and system variables, and that it
allows one to fully exploit all the information in the joint distribution on system and context variables.
JCI was partially inspired by the approach by \citet{EatonMurphy07}, but is much more generally applicable,
as it allows for latent confounders and cycles, which are both important in many application domains. 
Especially noteworthy is that more conservative flavours of JCI allow for confounders between system 
and context variables, which cannot always be excluded, for example because the relevant aspects of the
system's context were only partially observed. 

 We have investigated various
implementations of JCI, amongst which some existing algorithms (LCD, ICP, and standard estimators for the
presence of a causal effect in a randomized controlled trial), and also proposed novel implementations that
are adaptations of algorithms for causal discovery from purely observational data to the JCI setting. In particular, we
proposed ASD-JCI, an adaptation of the method of \citet{HEJ2014} combined with ideas from \citet{MagliacaneClaassenMooij_NIPS_16}, which is
very flexible and accurate.
By replacing d-separation with $\sigma$-separation \citep{ForreMooij_UAI_18}, ASD-JCI can also be used in 
general nonlinear cyclic settings. A major disadvantage of ASD-JCI is that 
it becomes computationally extremely expensive already for as few as about $7$ variables.
We also proposed FCI-JCI, an adaptation of the FCI algorithm that enables it to 
exploit the applicable JCI background knowledge. This algorithm is less accurate than ASD-JCI, but much faster.

We evaluated different implementations of the JCI approach on synthetic data.
We saw that JCI implementations outperform other state-of-the-art causal
discovery algorithms in most settings. In some cases, the gains were quite extreme;
for example, while purely observational causal discovery methods did not 
perform better than random guessing on small models, JCI variants were able to discover with
almost perfect precision ancestral causal relations between system variables.
The only case in which all JCI implementations
were outperformed by another causal discovery algorithm that combines data from
different contexts, was the setting in which the contexts correspond with 
perfect interventions with known targets. The reason is that none of the
JCI implementations exploited the perfect nature of the interventions. 
However, we also saw that if interventions are not perfect (for example, in the case
of causal mechanism changes), JCI implementations still perform very well,
while algorithms relying on the perfect nature of interventions may suffer from model
misspecification.
Another interesting observation we made in the experiments on synthetic data is that for the
task of discovering indirect (ancestral) causal relations, the classic (and very
simple and fast) LCD algorithm can be competitive with more sophisticated algorithms,
like ICP and bootstrapped FCI-JCI.

We further illustrated the use of JCI by analyzing flow cytometry protein expression data
\citep{SPP05}, a famous ``benchmark'' in the field of causal discovery. Unfortunately, 
applying ASD-JCI on the 11 system
and 6 context variables would take excessive amounts of computation time, 
so we had to resort to FCI-JCI instead for causal discovery on a global scale. 
We compared with LCD and ICP variants that do causal discovery locally.
The results of various methods differ considerably, but show also some consistent
patterns. This suggests that there is indeed a strong causal signal in the data, but
it seems hard to conclude which of the various methods is best equipped to extract 
this signal most reliably, because the ground truth is only partially known.
In future work, we plan to analyze more recent 
cytometry data sets that will allow for a more principled validation.
Because often the true causal structure is not known, while interventional
data is available, this requires to extend JCI-based causal discovery with causal 
prediction techniques, enabling one to predict the results of a particular 
intervention \citep{Magliacane++_NeurIPS_18}.

JCI offers increased flexibility 
when it comes to designing experiments for the purpose of causal discovery, 
as the JCI framework
facilitates analysis of data from almost arbitrary experimental designs. 
This allows researchers to
trade off the number and complexity of experiments to be done with the
reliability of the analysis of the data for the purpose of causal discovery.
Compared with existing methods, the framework offered by JCI is the most
generally applicable, handling various intervention types and other context
changes in a unified and non-parametric way, allowing for latent variables and
cycles, and also applies when intervention types and targets are unknown, a
common situation in causal discovery for complex systems.

As future work, we plan to (i) weaken the faithfulness assumption of JCI with respect
to the context variables to allow for even more general experimental designs, (ii)
address the problem of learning from data sets with non-identical (but overlapping)
sets of observed variables, (iii) address selection bias, (iv) develop algorithms
that need less computation time for delivering reliable results, (v) work on more 
applications on real-world data.

\acks{We thank Thijs van Ommen for useful discussions and the reviewers and editor for their constructive comments.
SM, JMM and TC were supported by NWO, the Netherlands Organization for Scientific Research (VIDI grant 639.072.410). SM was also supported by the Dutch programme COMMIT/ under the Data2Semantics project.
TC was supported by EU-FP7 grant agreement n.603016 (MATRICS).}

\bookmarksetup{startatroot}
\appendix

\section{Proofs}\label{sec:app:proofs}

In this appendix we provide the proofs that were omitted from the main text.

\subsection{JCI Foundations}

\begin{customthm}{\ref{thm:replace_context}}
\cmdTheoremReplaceContext
\end{customthm}
\begin{proof}
Let $\C{M}$ be an SCM of the form \eref{eq:SCM_JCI_ass}. Under JCI Assumption~\ref{ass:uncaused}, the structural equations for the context variables do not depend on the system variables:
$$C_k = f_k(\B{C}_{\pasub{\C{H}}{k} \cap \C{K}}, \B{E}_{\pasub{\C{H}}{k} \cap \C{J}}), \qquad k \in \C{K}.$$
Because of JCI Assumption \ref{ass:unconfounded}, $\pasub{\C{H}}{\C{K}} \cap \pasub{\C{H}}{\C{I}} \cap \C{J} = \emptyset$, i.e., the context variables do not share any exogenous variable with the system variables.
This means that in $\C{G}(\C{M})$, any edge between a context variable and a system variable must be a directed edge pointing
from context to system variable, i.e., of the form $k \to i$ with $k \in \C{K}$, $i \in \C{I}$.

Since the structural equations for the system variables of $\tilde{\C{M}}$ coincide with
those of $\C{M}$, their solutions (in terms of the context and exogenous variables) also
coincide, even after any perfect intervention on a subset of the system variables.
Since $\B{C}$ is independent of $\B{E}_{\pasub{\C{H}}{\C{I}}}$ (both for $\C{M}$ as well as for $\tilde{\C{M}}$),
and since $\Prb_{\C{E}} = \Prb_{\tilde{\C{E}}_{\C{J}}}$ by assumption,
this implies that the interventional distributions of $\C{M}$ and $\tilde{\C{M}}$ coincide
for any perfect intervention on a subset of system variables if $\Prb_{\C{M}}(\B{C}) = \Prb_{\tilde{\C{M}}}(\B{C})$.

Assume now that $\C{G}(\tilde{\C{M}})_{\C{K}}$ and $\C{G}(\C{M})_{\C{K}}$ induce the same separations.
In the remainder of this proof, ``open'' can be read either consistently as ``$\sigma$-open'' or as ``$d$-open''.
Note that by assumption, $\C{G}(\C{M})_{\intervene(\C{K})}
= \C{G}(\tilde{\C{M}})_{\intervene(\C{K})}$, and that the edges in $\C{G}(\C{M})_{\intervene(\C{K})}$
are a subset of those in $\C{G}(\C{M})$, and of those in $\C{G}(\tilde{\C{M}})$.
We will prove that $\C{G}(\tilde{\C{M}})$ and $\C{G}(\C{M})$ induce the same separations by first
showing that for any two context nodes connected by a path $\pi$ in $\C{G}(\C{M})_{\C{K}}$ such that $\pi$ is 
$A$-open in $\C{G}(\C{M})$ for some $A \subseteq \C{I} \cup \C{K}$, we can find a path $\pi'$
in $\C{G}(\C{M})_{\C{K}}$ between the two nodes that is $A'$-open in $\C{G}(\C{M})$ where
$A' = A \cap \C{K} \cup B$ with $B \subseteq \C{K} \cap \ansub{\C{G}(\C{M})_{\intervene(\C{K})}}{A \setminus \C{K}}$. 
For $\pi$ to be $A$-open in $\C{G}(\C{M})$, any collider on $\pi$ that is not a $\C{G}(\C{M})$-ancestor of $A \cap \C{K}$ must
be a $\C{G}(\C{M})$-ancestor of $A \setminus \C{K}$. Since the latter does not necessarily imply that the collider must also be 
$\C{G}(\tilde{\C{M}})$-ancestor of $A \setminus \C{K}$, the idea will be to replace the variables from $A \setminus \C{K}$ in the conditioning set by variables in $\C{K} \cap \ansub{\C{G}(\C{M})_{\intervene(\C{K})}}{A \setminus \C{K}}$
(i.e., context nodes that are guaranteed to be both $\C{G}(\C{M})$-ancestors and $\C{G}(\tilde{\C{M}})$-ancestors of $A \setminus \C{K}$)
that are $\C{G}(\C{M})$-descendants of those colliders that are not already $\C{G}(\C{M})$-ancestors of $A \cap \C{K}$.
It will turn out that this is not always possible to achieve for $\pi$, but that we can construct another path $\pi'$
for which this can be done. 

Consider a path $\pi$ in $\C{G}(\C{M})_{\C{K}}$ between $k_0 \in \C{K}$ and $k_n \in \C{K}$
that is $A$-open in $\C{G}(\C{M})$ for some $A \subseteq \C{I} \cup \C{K}$. We will 
iteratively construct a walk in $\C{G}(\C{M})_{\C{K}}$ between the same two nodes $k_0$ and $k_n$ that is both
$A$-open in $\C{G}(\C{M})$ and $(A \cap \C{K}) \cup B$-open in $\C{G}(\C{M})$, where
$B \subseteq \C{K} \cap \ansub{\C{G}(\C{M})_{\intervene(\C{K})}}{A \setminus \C{K}}$.
We will proceed by induction. Suppose a walk $\pi_m$ in
$\C{G}(\C{M})_{\C{K}}$ between $k_0$ and $k_n$ is $A$-open in $\C{G}(\C{M})$. Then it is
$A \cup B_m$-open in $\C{G}(\C{M})$ where 
$B_m = (\C{K} \cap \ansub{\C{G}(\C{M})_{\intervene(\C{K})}}{A \setminus \C{K}}) \setminus \ncol{\pi_m}$.
Consider the ``problematic'' colliders $\col{\pi_m} \setminus \ansub{\C{G}(\C{M})}{A \cap \C{K} \cup B_m}$
on $\pi_m$, i.e., the ones that are not ancestors of $A \cap \C{K} \cup B_m$. If there
are any, choose one such problematic collider $c \in \C{K}$ on $\pi_m$. 
Since $c$ is not $\C{G}(\C{M})$-ancestor of $A \cap \C{K} \cup B_m$,
but $\pi_m$ is $A$-open, it has to be $\C{G}(\C{M})$-ancestor of $A \setminus \C{K}$.
This means that there is a directed path in $\C{G}(\C{M})$ that starts at
$c$, passes through zero or more context nodes, none of which lie in $A \cap \C{K} \cup B_m$ by assumption, and then 
through zero or more system nodes, until it ends at a system node in $A \setminus \C{K}$.
Let $k_c \in \C{K}$ be the last context node on this directed path before the path crosses
the context-system boundary. By assumption, 
$k_c$ must exist as a non-collider on $\pi_m$ (otherwise it would be in $B_m$ and $c$ would be
$\C{G}(\C{M})$-ancestor of $B_m$), hence we can make a shortcut
by replacing the subwalk of $\pi_m$ between $c$ and $k_c$ by a directed path 
$c \to \dots \to k_c$ in $\C{G}(\C{M})_{\C{K}}$, which 
necessarily entirely consists of context nodes that are not in $A$. 
If $k_c$ occurs more than once on this new walk, remove the entire subwalk between the
two outermost occurrences of $k_c$, such that $k_c$ only occurs once.
This new walk $\pi_{m+1}$ must be $A$-open: $c$ (if still present) is now a non-collider 
that is not in $A$, none of the (non-collider) nodes on the directed path (if still present)
between $c$ and $k_c$ are in $A$, and $k_c$ itself is not in $A$ and is a $\C{G}(\C{M})$-ancestor of $A$, 
so it does not matter whether it is a collider or non-collider. 
The number of problematic colliders on $\pi_{m+1}$ is at least one less than on $\pi_m$:
$c$ is no longer a collider, and if $k_c$ became a collider on $\pi_{m+1}$, it won't be problematic
(as it is itself in $(\C{K} \cap \ansub{\C{G}(\C{M})_{\intervene(\C{K})}}{A \setminus \C{K}})$
and cannot also occur as non-collider on $\pi_{m+1}$),
We repeat this procedure until no problematic colliders are present anymore. This yields a walk 
$\pi_M$ that is both $A$-open and $A'$-open, with $A' = (A \cap \C{K}) \cup B$ where
$B = B_M = (\C{K} \cap \ansub{\C{G}(\C{M})_{\intervene(\C{K})}}{A \setminus \C{K}}) \setminus \ncol{\pi_M}$.
We now shorten this $A'$-open walk $\pi_M$ in $\C{G}(\C{M})_{\C{K}}$ into an $A'$-open path $\pi'$ in $\C{G}(\C{M})_{\C{K}}$.

This implies that there must be an $A'$-open path $\tilde{\pi}'$ in $\C{G}(\tilde{\C{M}})_{\C{K}}$
connecting $k_0$ and $k_n$, by assumption. Every collider on $\tilde{\pi}'$ is a $\C{G}(\tilde{\C{M}})_{\C{K}}$-ancestor
of $A \cap \C{K} \cup B$, and hence $\C{G}(\tilde{\C{M}})$-ancestor of $A \cap \C{K} \cup B$,
and hence $\C{G}(\tilde{\C{M}})$-ancestor of $A$.
Therefore, $\tilde{\pi}'$ is also $A'$-open in $\C{G}(\tilde{\C{M}})$.
But then it must also be $A$-open, as we can add $A \setminus \C{K}$ to the conditioning set
without blocking any non-collider on $\tilde{\pi}'$, and then remove $B \setminus (A \setminus \C{K})$ 
from the conditioning set as all colliders are
still kept open due to either being $\C{G}(\tilde{\C{M}})$-ancestor of $A \cap \C{K}$ or of $A \setminus \C{K}$.

Consider now any path in $\C{G}(\C{M})$ that is $A$-open,
for $A \subseteq \C{I} \cup \C{K}$. Any edge on the path between a system node and a context node must
be of the form $i \ot k$ (with $i \in \C{I}$, $k \in \C{K}$) or $k \to i$, where $i$ is in another
strongly-connected component than $k$ and $k$ cannot be in $A$ (because the path was assumed to be $A$-open).
Replacing each longest subpath consisting entirely of context nodes $k_0 \dots k_n$ (with all $k_0, \dots, k_n \in \C{K}$) 
by a corresponding $A$-open path in $\C{G}(\tilde{\C{M}})_{\C{K}}$ between $k_0$ and $k_n$ 
gives a walk in $\C{G}(\tilde{\C{M}})$ that by construction is also $A$-open in $\C{G}(\tilde{\C{M}})$. 
Any system collider on this walk must be a collider on the original path, 
and therefore $\C{G}(\C{M})$-ancestor of $A$, and therefore also $\C{G}(\tilde{\C{M}})$-ancestor of $A$.
Any system non-collider on this walk is also a system non-collider on the original path and therefore not in $A$
or, in case of $\sigma$-separation, pointing only to nodes in the same strongly-connected component of $\C{G}(\C{M})$, 
and hence of $\C{G}(\tilde{\C{M}})$. 
Any context non-collider on this walk cannot be in $A$,
or, in case of $\sigma$-separation, points to the same strongly-connected component in $\C{G}(\tilde{\C{M}})$,
since the replacing path in $\C{G}(\tilde{\C{M}})_{\C{K}}$
was $A$-open by construction. Any context collider on this walk that is a $\C{G}(\C{M})_{\C{K}}$-ancestor of  
$(A \cap \C{K}) \cup B$, and therefore must be $\C{G}(\tilde{\C{M}})$-ancestor of $A$.
The walk can be shortened into an $A$-open path in $\C{G}(\tilde{\C{M}})$. 

Similarly, one can show that any path in $\C{G}(\tilde{\C{M}})$ that is $A$-open, there must be a
corresponding path in $\C{G}(\C{M})$ that is $A$-open.
\end{proof}

\begin{customcor}{\ref{coro:JCI_ass_3}}
\cmdCorollaryJCIAssC
\end{customcor}
\begin{proof}
Let $\C{M}$ be an SCM of the form \eref{eq:SCM_JCI_ass}. Under JCI Assumption~\ref{ass:uncaused}, the structural equations for the context variables do not depend on the system variables:
$$C_k = f_k(\B{C}_{\pasub{\C{H}}{k} \cap \C{K}}, \B{E}_{\pasub{\C{H}}{k} \cap \C{J}}), \qquad k \in \C{K}.$$
Because of JCI Assumption \ref{ass:unconfounded}, $\pasub{\C{H}}{\C{K}} \cap \pasub{\C{H}}{\C{I}} \cap \C{J} = \emptyset$, i.e., the context variables do not share any exogenous variable with the system variables.

Consider now the modified SCM $\tilde{\C{M}}$ of the form:
\begin{equation*}
  \tilde{\C{M}}:
  \begin{cases}
    C_k = g_k(\B{E}_C), & \qquad  k \in \C{K}\\
    X_i = f_i(\B{X}_{\pasub{\C{H}}{i} \cap \C{I}}, \B{C}_{\pasub{\C{H}}{i} \cap \C{K}}, \B{E}_{\pasub{\C{H}}{i} \cap \C{J}}), & \qquad i \in \C{I},\\
    \Prb(\B{E}) = \prod_{j\in\tilde{\C{J}}} \Prb(E_j),
  \end{cases}
\end{equation*}
where $\tilde{\C{J}} = \C{J} \cup \{C\}$ contains an additional exogenous variable
$\B{E}_C \in \prod_{k\in\C{K}} \C{E}_k$ with components $(\B{E}_C)_k \in \C{C}_k$
with distribution $\Prb(\B{E}_C) = \Prb_{\C{M}}(\B{C})$
and $g_k$ the projection on the $k^{\mathrm{th}}$ component $g_k : \B{E}_C \mapsto (\B{E}_C)_k$.
By construction, this SCM $\tilde{\C{M}}$ satisfies JCI Assumptions~\ref{ass:uncaused} and \ref{ass:unconfounded}.
The only aspect that requires some work is to prove that $\tilde{\C{M}}$ as constructed above is simple (Definition~\ref{def:simple_scm}).

Take $\C{O} \subseteq \C{I}$ and consider the solution function for $\C{O}$ according to $\C{M}$:
$$\B{g}_{\C{O}} : \BC{X}_{(\pasub{\C{H}}{\C{O}}\setminus\C{O})\cap\C{I}} \times \BC{C}_{(\pasub{\C{H}}{\C{O}}\setminus\C{O})\cap\C{K}} \times \BC{E}_{\pasub{\C{H}}{\C{O}} \cap \C{J}} \to \BC{X}_{\C{I} \setminus \C{O}}.$$
This solves the structural equations for $\C{O} \setminus \C{I}$, and since these are the same for
$\tilde{\C{M}}$ as for $\C{M}$, the same solution function works also for $\tilde{\C{M}}$. 
Now take $\C{Q} \subseteq \C{K}$ and consider the solution function $\B{g}_{\C{Q}} : \C{E}_C \to \BC{C}_Q$ with
components $\C{E}_C \to \C{C}_k : \B{e}_C \mapsto g_k(\B{e}_C)$, $k \in \C{Q}$. Any other solution function can be
obtained by composition. We conclude that $\tilde{\C{M}}$ is simple.
$\tilde{\C{M}}$ also induces the same context distribution $\Prb_{\tilde{\C{M}}}(\B{E}) = \Prb_{\C{M}}(\B{E})$ and
satisfies JCI Assumption~\ref{ass:dependences} by construction. The other statements now follow by applying 
Theorem~\ref{thm:replace_context}, where the only thing left to show is that
$\C{G}(\tilde{\C{M}})_{\C{K}}$ and $\C{G}(\C{M})_{\C{K}}$ induce the same $\sigma$-separations if
the context distribution $\Prb_{\C{M}}(\B{C})$ contains no conditional independences, and the
same $d$-separations if in addition the Directed Global Markov Property holds for $\C{M}$.

Marginalizing out the system variables (both in $\C{M}$ as well as in $\tilde{\C{M}}$) yields
$\C{M}_{\setminus\C{I}}$ and $\tilde{\C{M}}_{\setminus\C{I}}$, with graphs
$\C{G}(\C{M}_{\setminus\C{I}}) = \C{G}(\C{M})_{\C{K}}$ and 
$\C{G}(\tilde{\C{M}}_{\setminus\C{I}}) = \C{G}(\tilde{\C{M}})_{\C{K}}$, respectively. 
By the Generalized Directed Global Markov property, since 
$\Prb_{\C{M}}(\B{C})$ has no conditional independences, there must be a $K$-$\sigma$-open path in 
$\C{G}(\C{M})_{\C{K}}$ between any
two context nodes $k \ne k' \in \C{K}$, for any $K\subseteq \C{K}$ with $\{k,k'\} \cap K = \emptyset$. 
If the Directed Global Markov property holds for $\C{M}$, then it holds for $\C{G}(\C{M}_{\setminus\C{I}})$, 
and hence there must even be a $K$-$d$-open path in $\C{G}(\C{M})_{\C{K}}$ between any two context nodes 
$k \ne k' \in \C{K}$, for any $K\subseteq \C{K}$ with $\{k,k'\} \cap K = \emptyset$. 
Since by construction $\C{G}(\tilde{\C{M}})_{\C{K}}$ contains all bidirected edges $k \oto k'$,
there is a $K$-$d$-open path in $\tilde{\C{M}}_{\setminus\C{I}}$ between any two context nodes
$k \ne k' \in \C{K}$, for any $K\subseteq \C{K}$ with $\{k,k'\} \cap K = \emptyset$.
\end{proof}

The following Lemma and Corollary extend these fundamental results further, which enables one to state
a precise relationship between our JCI approach of jointly modeling system and context with
alternative approaches based on modeling the system conditional on its context (e.g., \citet{YangKatcoffUhler2018}).
\begin{lemma}
  Let $\C{M}$ be an SCM that satisfies JCI Assumptions~\ref{ass:simple_scm}, \JCIAB. 
  Then the same \emph{restricted} separations hold in $\C{G}(\C{M})$ as in the conditional system graph
  $\C{G}(\C{M})_{\intervene(\C{K})}$, i.e.,
    $$\sep{X}{Y}{Z}{\C{G}(\C{M})_{\intervene(\C{K})}} \iff \sep{X}{Y}{Z}{\C{G}(\C{M})}$$
  whenever $X,Y,Z \subseteq \C{I} \cup \C{K}$ with $X \cap \C{K} = \emptyset$ and $\C{K} \subseteq Y \cup Z$ 
  (where ``separations'' can refer to either $d$-separations or $\sigma$-separations).
\end{lemma}
\begin{proof}
Let $X,Y,Z \subseteq \C{I}\cup\C{K}$ be such that $X \cap \C{K} = \emptyset$ and $\C{K} \subseteq Y \cup Z$. 
Let $\C{G}_1, \C{G}_2$ be two graphs in $\{${}$\C{G}(\C{M})$, $\C{G}(\C{M})_{\intervene(\C{K})}\}$. 
Let $\pi$ be a path in $\C{G}_1$ between a node in $X$ and a node in $Y$ that is open in $\C{G}_1$ and that does not
contain any non-endpoint nodes in $X \cup Y$. It cannot have non-endpoint nodes in $\C{K}$, because those would be either 
in $Y$ (a contradiction), or in $Z$ (and since they would be non-colliders with an outgoing directed edge pointing to another strongly-connected component, they would block the path, another contradiction). But then the same path $\pi$ must be 
present in $\C{G}_2$ as well. It is easy to see that it must also be open in $\C{G}_2$, since for each $i \in \C{I}$,
$\desub{\C{G}_1}{i} = \desub{\C{G}_2}{i}$ and $\sccsub{\C{G}_1}{i} = \sccsub{\C{G}_2}{i}$.
\end{proof}

We can now formulate the following slightly adapted version of Corollary~\ref{coro:JCI_ass_3}:
\begin{corollary}\label{coro:JCI_ass_3_WLOG}
Assume that JCI Assumptions \ref{ass:simple_scm}, \ref{ass:uncaused} and \ref{ass:unconfounded} hold for SCM $\C{M}$.
Then there exists an SCM $\tilde{\C{M}}$ that satisfies JCI Assumptions \ref{ass:simple_scm}, \ref{ass:uncaused} and \ref{ass:unconfounded} and \ref{ass:dependences}, such that
  \begin{enumerate}[(i)]
    \item the conditional system graphs coincide: $\C{G}(\C{M})_{\intervene(\C{K})} = \C{G}(\tilde{\C{M}})_{\intervene(\C{K})}$;
    \item as a consequence, the same \emph{restricted} separations hold in $\C{G}(\tilde{\C{M}})$ as in $\C{G}(\C{M})$ and in their corresponding conditional system graphs, i.e.,
      $$\sep{X}{Y}{Z}{\C{G}(\tilde{\C{M}})} \iff \sep{X}{Y}{Z}{\C{G}(\tilde{\C{M}})_{\intervene(\C{K})}} \iff \sep{X}{Y}{Z}{\C{G}(\C{M})_{\intervene(\C{K})}} \iff \sep{X}{Y}{Z}{\C{G}(\C{M})}$$
      whenever $X,Y,Z \subseteq \C{I} \cup \C{K}$ with $X \cap \C{K} = \emptyset$ and $\C{K} \subseteq Y \cup Z$ (where ``separations'' can refer to either $d$-separations or $\sigma$-separations);
    \item for any perfect intervention on the system variables $\intervene(I,\B{\xi}_I)$ with $I \subseteq \C{I}$ (including the non-intervention $I = \emptyset$), and any perfect intervention on all context variables $\intervene(\C{K},\B{c})$:
      $$\Prb_{\tilde{\C{M}}}(\B{X} \given \intervene(\C{K},\B{c}), \intervene(I,\B{\xi}_I)) = \Prb_{\C{M}}(\B{X} \given \intervene(\C{K},\B{c}), \intervene(I,\B{\xi}_I));$$
    \item as a consequence, $\Prb_{\C{M}}(\B{X} \given \B{C}) = \Prb_{\tilde{\C{M}}}(\B{X} \given \B{C})$, and in particular, the same restricted conditional independences hold, i.e.,
      $$\indep{X}{Y}{Z}{\C{G}(\tilde{\C{M}})} \iff \indep{X}{Y}{Z}{\C{G}(\tilde{\C{M}})_{\intervene(\C{K})}} \iff \indep{X}{Y}{Z}{\C{G}(\C{M})_{\intervene(\C{K})}} \iff \indep{X}{Y}{Z}{\C{G}(\C{M})}$$
      whenever $X,Y,Z \subseteq \C{I} \cup \C{K}$ with $X \cap \C{K} = \emptyset$ and $\C{K} \subseteq Y \cup Z$;
    \item the context distribution $\Prb_{\tilde{\C{M}}}(\B{C})$ contains no conditional or marginal independences.
  \end{enumerate}
\end{corollary}
\begin{proof}
  The same SCM $\tilde{\C{M}}$ as constructed in the proof of Corollary~\ref{coro:JCI_ass_3},
  but with a generic distribution of $\Prb(\B{E}_C)$ that contains no conditional or marginal independences,
  is easily seen to fulfill all requirements.
\end{proof}

\subsection{Minimal Conditional (In)Dependencies}\label{app:mci}

In this section we generalize two useful Lemmas from \citet{ClaassenHeskes2011} to the cyclic setting.
\begin{definition}
Let $X,Y,Z,S \subseteq \C{V}$ be sets of nodes in a DMG $\C{G} = \langle \C{V},\C{E},\C{F} \rangle$. 
Let $\SEP$ denote a DMG-separation property,
  e.g., $d$-separation ($\SEP^d$) or $\sigma$-separation ($\SEP^\sigma$). We say that the \emph{minimal separation}
  $$\sep{X}{Y}{S \cup [Z]}{\C{G}}$$
holds if and only if
  $$\sep{X}{Y}{S \cup Z}{\C{G}} \quad\land\quad \forall Q \subsetneq Z: \con{X}{Y}{S \cup Q}{\C{G}}.$$
In words: all nodes in $Z$ are required (in the context of the nodes in $S$) to separate $X$ from $Y$.
Similarly: we say that the \emph{minimal connection}
  $$\con{X}{Y}{S \cup [Z]}{\C{G}}$$
holds if and only if
  $$\con{X}{Y}{S \cup Z}{\C{G}}\quad\land\quad \forall Q \subsetneq Z: \sep{X}{Y}{S \cup Q}{\C{G}}.$$
In words: all nodes in $Z$ are required (in the context of the nodes in $S$) to connect $X$ with $Y$.
\end{definition}
Note that despite the notation, a minimal connection is not the logical negation of a minimal separation.

Minimal connections imply the absence of certain ancestral relations:
\begin{lemma}\label{lemm:min_con}
Let $\{X\},\{Y\},S,\{Z\} \subseteq \C{V}$ be mutually disjoint sets of nodes in a DMG $\C{G} = \langle \C{V},\C{E},\C{F} \rangle$.
For both $d$-separation ($\SEP^d$) and $\sigma$-separation ($\SEP^\sigma$), we have that:
$$\con{X}{Y}{S \cup [\{Z\}]}{\C{G}} \implies Z \notin \ansub{\C{G}}{\{X, Y\} \cup S}.$$
\end{lemma}
\begin{proof}
The minimal connection means that all paths between $X$ and
$Y$ are closed when conditioning on $S$ and there exists at least one
path between $X$ and $Y$ that is open when conditioning on $S \cup \{Z\}$.
For $d$-separation, this means that such a path (i) contains a collider not in $\ansub{\C{G}}{S}$, 
  (ii) every collider is in $\ansub{\C{G}}{S \cup \{Z\}}$,
(iii) every non-collider is not in $S \cup \{Z\}$. 
For $\sigma$-separation, this means that such a path 
  (i) contains a collider not in $\ansub{\C{G}}{S}$, (ii) every collider is in 
  $\ansub{\C{G}}{S \cup \{Z\}}$,
(iii) every non-collider is either not in $S \cup \{Z\}$, or if it is,
it points to neighboring nodes in the same strongly-connected component only.

Thus there exists a path between $X$ and $Y$ that contains a collider in $\ansub{\C{G}}{\{Z\}}$ that 
is not in $\ansub{\C{G}}{S}$. If $Z \in \ansub{\C{G}}{S}$ this would be a contradiction.
If $Z \in \ansub{\C{G}}{X}$, then we can consider the walk between $X$ and $Y$ obtained
from composing the subpath of the original path between $Y$ and the first
collider (starting from $Y$) in $\ansub{\C{G}}{\{Z\}} \setminus \ansub{\C{G}}{S}$ with a directed 
path to $Z$ and then on to $X$, without passing through nodes in $S$.
This walk between $X$ and $Y$ must be open when conditioning on $S$, and hence there exists a path
between $X$ and $Y$ that is open when conditioning on $S$, a contradiction.
Similarly we obtain a contradiction if $Z \in \ansub{\C{G}}{Y}$.
\end{proof}

On the other hand, minimal separations imply the presence of certain ancestral relations:
\begin{lemma}\label{lemm:min_sep}
Let $\{X\},\{Y\},S,Z \subseteq \C{V}$ be mutually disjoint sets of nodes in a DMG $\C{G} = \langle \C{V},\C{E},\C{F} \rangle$.
For both $d$-separation ($\SEP^d$) and $\sigma$-separation ($\SEP^\sigma$), we have that:
$$\sep{X}{Y}{S \cup [Z]}{\C{G}} \implies Z \subseteq \ansub{\C{G}}{\{X, Y\} \cup S}.$$
\end{lemma}
\begin{proof}
Let $Q \subsetneq Z$.
Consider a path between $X$ and $Y$ that is open when conditioning on $S \cup Q$, but becomes blocked when conditioning on $S \cup Z$. 
  For $d$-separation, this means that (i) every collider on the path is in $\ansub{\C{G}}{S \cup Q}$, (ii) every non-collider is not in $S \cup Q$, and (iii) it contains a non-collider in $S\cup Z$.
  For $\sigma$-separation, this means that (i) every collider on the path is in $\ansub{\C{G}}{S \cup Q}$, (ii) every non-collider is either not in $S \cup Q$ or if it is, it points to a neighboring node on the path in another strongly-connected component, and (iii) it contains a non-collider in $S\cup Z$ that points to a neighboring node on the path in another strongly-connected component. 
  In both cases, we have that (i) every collider on the path is in $\ansub{\C{G}}{S \cup Q}$ and (ii) it contains a non-collider in $Z \setminus Q$.
Consider a maximal directed subpath of the path starting at a non-collider $U$ in $Z \setminus Q$ and stopping at a collider or at an end node.
  Then $U \in \ansub{\C{G}}{\{X,Y\} \cup S \cup Q}$. 

  So, for each $Q \subsetneq Z$, there exists a $U \in Z\setminus Q$ with $U \in \ansub{\C{G}}{\{X,Y\} \cup S \cup Q}$.
Thus for every $Z_i \in Z$, we either obtain (taking $Q = Z \setminus \{Z_i\}$) an ancestral relation of the form $Z_i \in \ansub{\C{G}}{\{X,Y\} \cup S}$, or, otherwise, at least $Z_i \in \ansub{\C{G}}{Z_j}$ for some $Z_j \in Z \setminus \{Z_i\}$. Define a directed graph $\C{A}$ with nodes $Z \cup \{\omega\}$ (where $\omega$ represents $\{X,Y\}\cup S$) and add an edge $Z_i \to \omega$ whenever our construction yields an ancestral relation of the form $Z_i \in \ansub{\C{G}}{\{X,Y\} \cup S}$, or otherwise, an edge $Z_i \to Z_j$ if our construction yields $Z_i \in \ansub{\C{G}}{Z_j}$.
 
  Then, taking the transitive closure of the constructed directed graph $\C{A}$ and using transitivity of ancestral relations, for any $Z_i \in Z$ we either obtain $Z_i \in \ansub{\C{G}}{\{X,Y\} \cup S}$, or $Z_i$ is in some strongly-connected component $C \subseteq Z$ in $\C{A}$. In the latter case, we can apply the reasoning above (taking now $Q = Z \setminus C$) to conclude that there exists a $Z_j \in C$ with $Z_j \in \ansub{\C{G}}{\{X,Y\} \cup S}$ or $Z_j \in \ansub{\C{G}}{C'}$ where $C' \subseteq Z$ is another strongly-connected component. Since the strongly-connected components of $Z$ form an acyclic structure, repeating this reasoning a finite number of times, we ultimately conclude that $Z_i \in \ansub{\C{G}}{\{X,Y\} \cup S}$. 
\end{proof}

An implication of this is that the intersection of all sets that separate a node $X$ from a node $Y$ can only consist of ancestors of $X$ or $Y$:
\begin{proposition}\label{prop:intersection_sepsets}
Let $X,Y \in \C{V}$ be different nodes in a DMG $\C{G} = \langle \C{V},\C{E},\C{F} \rangle$.
For $d$-separation ($\SEP^d$) or $\sigma$-separation ($\SEP^\sigma$), consider
  $Z^* := \bigcap \{Z \subseteq \C{V} : X \notin Z, Y \notin Z, \sep{X}{Y}{Z}{\C{G}}\}$. 
Then $Z^* \subseteq \ansub{\C{G}}{\{X,Y\}}$.
\end{proposition}
\begin{proof}
First, note that
  $Z^* = \bigcap \{Z \subseteq \C{V} : X \notin Z, Y \notin Z, \sep{X}{Y}{[Z]}{\C{G}}\}$. From Lemma~\ref{lemm:min_sep}, $\sep{X}{Y}{[Z]}{\C{G}}$ implies $Z \subseteq \ansub{\C{G}}{\{X,Y\}}$. Hence $Z^* \subseteq \ansub{\C{G}}{\{X,Y\}}$.
\end{proof}

\section{Soundness, Consistency and Completeness Properties of FCI-JCI}\label{sec:app:fcijci}

In this appendix we will formulate and prove various results concerning soundness and completeness of FCI-JCI variants.

\subsection{Preliminaries on MAGs and PAGs}\label{sec:app:preliminaries}

We start by summarizing the basic definitions and results from the theory of maximal ancestral graphs and partial ancestral graphs \citep{SGS2000,RichardsonSpirtes02,Zhang2006,Zhang2008_AI,Zhang2008_JMLR} that we will need.

\subsubsection{Directed Maximal Ancestral Graphs}\label{sec:app:dmag}

\citet{RichardsonSpirtes02} introduced a class of graphs known as \emph{maximal ancestral graphs (MAGs)}.
The general formulation of MAGs allows for undirected edges which are useful when modeling selection bias, but here we will only use \emph{directed} maximal ancestral graphs without undirected edges (sometimes abbreviated as DMAGs in the literature) as we assume for simplicity that there is no selection bias.
In order to define a directed maximal ancestral graph, we need the notion of inducing path.
\begin{definition}
Let $\C{G} = \langle \C{V}, \C{E}, \C{F} \rangle$ be an acyclic directed mixed graph (ADMG). 
An \emph{inducing path between two nodes $u,v \in \C{V}$} is a path in $\C{G}$ between $u$ and $v$ on which every node 
(except for the end nodes) is a collider on the path and an ancestor in $\C{G}$ of an end node of the path.
\end{definition}
We can now state:
\begin{definition}
  A directed mixed graph $\C{G} = \langle \C{V}, \C{E}, \C{F} \rangle$ is called a \emph{directed maximal ancestral graph (DMAG)} if all of the following conditions hold: 
  \begin{compactenum}
    \item Between any two different nodes there is at most one edge, and there are no self-cycles;
    \item The graph contains no directed or almost directed cycles (``ancestral'');
    \item There is no inducing path between any two non-adjacent nodes (``maximal'').
  \end{compactenum}
\end{definition}
Given an ADMG, we can define a corresponding DMAG \citep{RichardsonSpirtes02}:
\begin{definition}
  Let $\C{G} = \langle \C{V}, \C{E}, \C{F} \rangle$ be an ADMG. 
  The \emph{directed maximal ancestral graph induced by $\C{G}$} is denoted $\DMAG(\C{G})$ and is defined as
  $\DMAG(\C{G}) = \langle \tilde{\C{V}}, \tilde{\C{E}}, \tilde{\C{F}} \rangle$ such that $\tilde{\C{V}} = \C{V}$ and
  for each pair $u,v \in \C{V}$ with $u \ne v$, there is an edge in $\DMAG(\C{G})$ between $u$ and $v$ if and only if 
  there is an inducing path between $u$ and $v$ in $\C{G}$, and in that case the edge in $\DMAG(\C{G})$ connecting $u$ and $v$ is:
    $$\begin{cases}
      \text{$u \to  v$} & \text{if $u \in \ansub{\C{G}}{v}$}, \\
      \text{$u \ot  v$} & \text{if $v \in \ansub{\C{G}}{u}$}, \\
      \text{$u \oto v$} & \text{if $u \not\in \ansub{\C{G}}{v}$ and $v \not\in \ansub{\C{G}}{u}$.}
    \end{cases}$$
\end{definition}
An important property of the induced DMAG is that it preserves all ancestral and non-ancestral relations.
More precisely, for two nodes $u,v$ in ADMG $\C{G}$: $u \in \ansub{\C{G}}{v}$ if and only if $u \in \ansub{\DMAG(\C{G})}{v}$.
Another important property of the induced DMAG is that it preserves all d-separations.
Indeed, $\dsep{A}{B}{C}{\DMAG(\C{G})} \iff \dsep{A}{B}{C}{\C{G}}$ for all $A, B, C \subseteq \C{V}$.
We sometimes identify a DMAG $\C{H}$ with the set of ADMGs $\C{G}$ such that $\DMAG(\C{G}) = \C{H}$.
For an acyclic SCM $\C{M}$, we will define its induced DMAG as $\DMAG(\C{M}) := \DMAG(\C{G}(\C{M}))$. 

For a directed maximal ancestral graph $\C{H}$, define its independence model to be
$$\IM(\C{H}) := \{ \langle A, B, C \rangle : A, B, C \subseteq \C{V}, \dsep{A}{B}{C}{\C{H}} \},$$
i.e., the set of all d-separations entailed by the DMAG. 
For a simple SCM $\C{M}$ with endogenous index set $\C{I}$ and distribution $\Prb_{\C{M}}(\B{X})$, 
we define its independence model to be
$$\IM(\C{M}) := \{ \langle A, B, C \rangle : A, B, C \subseteq \C{I}, \indep{\B{X}_A}{\B{X}_B}{\B{X}_C}{\Prb_{\C{M}}} \},$$
i.e., the set of all (conditional) independences that hold in its (observational) distribution.
If $\C{M}$ is acyclic, then by the Markov property, $\IM(\C{M}) \supseteq \IM(\DMAG(\C{M}))$; 
the faithfulness assumption then means that $\IM(\C{M}) \subseteq \IM(\DMAG(\C{M}))$.

\subsubsection{Directed Partial Ancestral Graphs}\label{sec:app:dpag}

Since in many cases, the true DMAG is unknown, it is often convenient when performing causal reasoning to be able to represent a set of hypothetical DMAGs in a compact way. 
For this purpose, \emph{partial ancestral graphs (PAGs)} have been introduced \citep{Zhang2006}.
Again, since we are assuming no selection bias for simplicity, we will only discuss \emph{directed} PAGs (that is, PAGs without undirected or circle-tail edges, i.e., edges of the form $\{\ttt,\ttc,\ctt\}$).
\begin{definition}
  We call a mixed graph $\C{G} = \langle\C{V},\C{E}\rangle$ with nodes $\C{V}$ and edges $\C{E}$ of the types $\{\to,\ot,\otc,\oto,\ctc,\cto\}$ a \emph{directed partial ancestral graph (DPAG)} if:
  \begin{compactenum}
    \item Between any two different nodes there is at most one edge, and there are no self-cycles;
    \item The graph contains no directed or almost directed cycles (``ancestral'');
    \item There is no inducing path between any two non-adjacent nodes (``maximal'').
  \end{compactenum}
\end{definition}
Given a DMAG or DPAG, its induced \emph{skeleton} is an undirected graph with the same nodes and with an edge between any pair of nodes if and only if the two nodes are adjacent in the DMAG or DPAG.
We often identify a DPAG with the set of all DMAGs that have the same skeleton as the DPAG, have an arrowhead (tail) on each edge mark for which the DPAG has an arrowhead (tail) at that corresponding edge mark, and for each circle in the DPAG, have either an arrowhead or a tail at the corresponding edge mark. 
Hence, the circles in a DPAG can be thought of as to represent either an arrowhead or a tail.

We extend the definitions of (directed) walks, (directed) paths and colliders for directed mixed graphs to apply also to DPAGs. 
Edges of the form $i \ot j, i \otc j, i \oto j$ are called \emph{into $i$}, and similarly, edges of the form $i \to j, i \cto j, i \oto j$ are called \emph{into $j$}. Edges of the form $i \to j$ and $j \ot i$ are called \emph{out of $i$}. In addition, we define:
\begin{definition}
  A path $v_0,e_1,v_1,\dots,v_n$ between nodes $v_0$ and $v_n$ in a DPAG $\C{G} = \langle\C{V},\C{E}\rangle$ is called a \emph{possibly directed path from $v_0$ to $v_n$} if for each $i=1,\dots,n$, the edge $e_i$ between $v_{i-1}$ and $v_i$ is not into $v_{i-1}$ (i.e., is of the form $v_{i-1} \ctc v_i$, $v_{i-1} \cto v_i$, or $v_{i-1} \to v_i$). 
  The path is called \emph{uncovered} if every subsequent triple is unshielded, i.e., $v_i$ and $v_{i-2}$ are not adjacent in $\C{G}$ for $i=2,\dots,n$.
\end{definition}

If $\C{H}_1$ and $\C{H}_2$ are DMAGs, then we call them \emph{Markov equivalent} if $\IM(\C{H}_1) = \IM(\C{H}_2)$. 
One can show that this implies that $\C{H}_1$ and $\C{H}_2$ must have the same skeleton and the same unshielded colliders.
The FCI algorithm maps the independence model $\IM(\C{H})$ of a DMAG $\C{H}$ to a DPAG $\C{P}$. 
\citet{Zhang2008_AI} showed that FCI is \emph{sound} and \emph{complete}, which means that
\begin{itemize}
  \item $\C{P}$ has the same skeleton as $\C{H}$;
  \item As a set of DMAGs, $\C{P}$ contains $\C{H}$ and all Markov equivalent DMAGs;
  \item For every circle edge mark in $\C{P}$, there exists a DMAG in $\C{P}$ Markov equivalent to $\C{H}$ that has a tail at the corresponding edge mark, and there exists a DMAG in $\C{P}$ Markov equivalent to $\C{H}$ that has an arrowhead at the corresponding edge mark.
\end{itemize}
We will denote the completely oriented directed partial ancestral graph that contains the Markov equivalence class of a DMAG $\C{H}$ by $\CDPAG(\C{H})$.
For an acyclic SCM $\C{M}$ we will denote its corresponding CDPAG representation as $\CDPAG(\C{M}) := \CDPAG(\DMAG(\C{G}(\C{M})))$. 

We will make use of the notion of \emph{(in)visible} edges in a DMAG \citep{Zhang2008_JMLR}:
\begin{definition}
A directed edge $i \to j$ in a DMAG is said to be \emph{visible} if there is a node
$k$ not adjacent to $j$, such that either there is an edge between $k$ and $i$ that is into $i$, or there is a
collider path between $k$ and $i$ that is into $i$ and every collider on the path is a parent of $j$. Otherwise
$i \to j$ is said to be \emph{invisible}.
\end{definition}
We will use the same notion in a DPAG, but call it \emph{definitely visible} (and its negation \emph{possibly invisible}).
If a directed edge in a DPAG is definitely visible, it must be visible in all DMAGs in the DPAG.

\subsection{Soundness and Consistency of FCI-JCI}\label{sec:app:fcijci_sound}

We are now equipped to prove the soundness (and for some cases, completeness) of FCI-JCI, the adaptation of FCI that 
incorporates the JCI background knowledge that we introduced in Section~\ref{sec:FCI-JCI}.

First, the soundness of FCI-JCI is easy to prove by checking that the soundness of the FCI orientation rules is not invalidated by the JCI background knowledge.
\begin{theorem}\label{theo:fcijci_sound}
  Let $\C{M}$ be an acyclic SCM that satisfies JCI Assumption~\ref{ass:simple_scm} and a subset of JCI Assumptions~\JCIABC. 
  Suppose that its distribution $\Prb_{\C{M}}(\B{X},\B{C})$ is faithful w.r.t.\ the graph $\C{G}(\C{M})$. 
  With input $\IM(\C{M})$, and with the right JCI Assumptions, FCI-JCI outputs a DPAG that contains $\DMAG(\C{M})$.
\end{theorem}
\begin{proof}
  First note that the skeleton obtained by FCI-JCI must coincide with that of $\DMAG(\C{M})$ (as it would for standard FCI). 
  Indeed, if JCI Assumption~\ref{ass:dependences} is made, the context nodes are all adjacent in $\DMAG(\C{M})$ by assumption. 
  For all other edges, and also for edges between context nodes if JCI Assumption~\ref{ass:dependences} is not made: the edge is in the skeleton found by FCI-JCI if and only if it is in the skeleton of $\DMAG(\C{M})$, for the same reason as for standard FCI.

  Furthermore, one can easily see that FCI rule $\C{R}$0 is still sound and will not conflict with the application of the background knowledge stemming from the JCI Assumptions. 

  The extra orientation rules to incorporate the JCI background knowledge are easily seen to be sound themselves, as they just impose additional features on the DPAG that are satisfied by $\DMAG(\C{M})$ by assumption.

  By checking the soundness proofs of each of the standard FCI orientation rules $\C{R}$1-$\C{R}$4 and $\C{R}$8-$\C{R}$10 in \citet{Zhang2006}, it is obvious that all these rules are sound when applied on any DPAG as long as (i) it contains the true DMAG and (ii) FCI rule $\C{R}0$ has been completely applied, i.e., all unshielded colliders have been oriented as such. Rules $\C{R}$5-$\C{R}$7 are not needed since we assumed no selection bias.

  Hence all the rules applied by FCI-JCI are sound (as long as they are applied in the prescribed ordering), and hence the final DPAG must contain the true $\DMAG(\C{M})$.
\end{proof}

In general, soundness of a constraint-based causal discovery algorithm implies consistency of the algorithm when using appropriate conditional independence tests.
\begin{lemma}\label{lemm:consistency_constraint_based}
  If a conditional independence test, including the choice of the sample-size dependent threshold (to decide between the null and alternative hypothesis), is consistent, then any sound constraint-based causal discovery algorithm based on the test is asymptotically consistent.
\end{lemma}
\begin{proof}
  Consistency of the conditional independence test means that for any distribution, the probability of a Type I or Type II error converges to 0 as sample size $N \to \infty$. Since the number of tests is finite for a fixed number of variables, and the number of possible predictions made by the algorithm is finite, the probability of \emph{any} error then converges to 0. Any constraint-based algorithm that is sound (i.e., that would return correct answers when using an independence oracle, including the possible answer ``unknown'') is therefore asymptotically consistent if it makes use of that conditional independence test.
\end{proof}
As an example of a consistent test, \citet{KalischBuehlmann2007} provide a choice of the threshold for the standard partial correlation test that ensures
asymptotic consistency of the test under the assumption that the distribution is multivariate Gaussian. Another example of a
distribution-free and even strongly-consistent conditional independence test is proposed by \citet{GyorfiWalk2012}.
In general, when using the $p$-value as a test statistic, one should
choose the sample-size dependent threshold $\alpha_N$ (where the $p$-value of the test result is used to decide ``dependence''
if $p \le \alpha_N$ and ``independence'' otherwise) in such a way that $\alpha_N \to 0$ as $N \to \infty$ at a suitable rate. 

\subsection{Completeness of FCI-JCI}\label{sec:app:fcijci_complete}

Regarding completeness, we currently only know how to prove the completeness of the variants \alg{FCI-JCI0} and \alg{FCI-JCI123}.
In particular, we do not know whether \alg{FCI-JCI1} is complete.
The completeness of \alg{FCI-JCI0} is obvious, because \alg{FCI-JCI0} reduces to the standard FCI algorithm without additional background knowledge.
\begin{theorem}\label{theo:fcijci0_complete}
  Let $\C{M}$ be an acyclic SCM that satisfies JCI Assumption~\ref{ass:simple_scm}. 
  Suppose that its distribution $\Prb_{\C{M}}(\B{X},\B{C})$ is faithful w.r.t.\ the graph $\C{G}(\C{M})$. 
  With input $\IM(\C{M})$, the output of \alg{FCI-JCI0} is a CDPAG in which all edge marks that can possibly be identified from $\IM(\C{M})$ have been oriented.
\end{theorem}
\begin{proof}
  Follows immediately from the completeness of FCI \citep{Zhang2008_AI} under the additional assumption of no selection bias.
\end{proof}
Proving the completeness of \alg{FCI-JCI123} is more work. 
The proof strategy is to introduce additional variables that mimic the JCI background knowledge.
We can then apply the completeness results for the standard FCI algorithm \citep{Zhang2008_AI}.
\begin{theorem}\label{theo:fcijci123_complete}
  Let $\C{M}$ be an acyclic SCM that satisfies JCI Assumptions~\ref{ass:simple_scm}, \JCIABC. 
  Suppose that its distribution $\Prb_{\C{M}}(\B{X},\B{C})$ is faithful w.r.t.\ the graph $\C{G}(\C{M})$. 
  With input $\IM(\C{M})$, the output of \alg{FCI-JCI123} is a DPAG in which all edge marks that can possibly be identified from $\IM(\C{M})$ and the JCI background knowledge have been oriented.
\end{theorem}
\begin{proof}
  The adjacency phase (skeleton search) of \alg{FCI-JCI123} and the orientation of unshielded triples by applying FCI rule $\C{R}$0 are both sound, as we have seen in Theorem~\ref{theo:fcijci_sound}.
  Furthermore, the skeleton and unshielded colliders found by \alg{FCI-JCI123} will be the same as found by standard FCI (in particular, note that FCI would not orient any unshielded colliders on a context node, since the true $\DMAG(\C{M})$ does not have these).\footnote{If we would use the results of statistical conditional independence tests on a finite data sample, then there could be differences between the DPAGs constructed by FCI and \alg{FCI-JCI123} after these stages.}

  Before continuing with the FCI orientation rules $\C{R}$1--$\C{R}$4 and $\C{R}$8--$\C{R}$10, \alg{FCI-JCI123} now uses the JCI background knowledge to orient the following edges:
  \begin{itemize}
    \item $k \oto k'$ for all $k \ne k' \in \C{K}$;
    \item $k \to i$ for $k \in \C{K}$, $i \in \C{I}$ if $k$ and $i$ are adjacent.
  \end{itemize}
  After this background orientation step, the only edges in the skeleton that remain to be (further) oriented are the ones connecting two system variables. 
  Denote the DPAG identified by \alg{FCI-JCI123} so far by $\C{P}$.

  Each DMAG $\C{H}$ with $\IM(\C{H}) = \IM(\C{M})$ and that satisfies the JCI Assumptions~\ref{ass:simple_scm}, \JCIABC\ must be contained in $\C{P}$.
  Consider any such DMAG $\C{H}$. 
  We can extend it to a DMAG $\C{H}^*$, defined over an extended set of variables $\C{I} \dot\cup \C{K} \dot\cup \{r\} \dot\cup \bar{\C{K}}$ where $\bar{\C{K}} := \{ \bar{k} : k \in \C{K} \}$ is a copy of $\C{K}$, by adding edges $r \to k$ for all $k \in \C{K}$, adding edges $\bar{k} \to k$ for all $k \in \C{K}$, and removing all bidirected edges $k \oto k'$ for all $k \ne k' \in \C{K}$ (see also Figure~\ref{fig:FCI_extended_MAG}).
  By construction, the marginal DMAG of $\C{H}^*$ on $\C{I} \cup \C{K}$ is $\C{H}$.\footnote{For the notion of marginal DMAG, see \citet{RichardsonSpirtes02}.}

  If we run FCI on $\IM(\C{H}^*)$ then the first stages of the algorithm yield a DPAG $\C{P}^*$ with:
  \begin{itemize}
    \item the skeleton of $\C{H}^*$, which equals the skeleton of $\C{P}$ together with the additionally constructed edges ($r \sts k$ for all $k \in \C{K}$, and $\bar{k} \sts k$ for all $k \in \C{K}$) but without the edges between the context variables ($k \sts k'$ for $k \ne k' \in \C{K}$);
    \item the unshielded colliders in $\C{P}$ plus the additionally constructed unshielded colliders ($r \sto k \ots \bar{k}$
      for all $k \in \C{K}$), which are all identified by rule $\C{R}$0;
    \item the arrowheads identified by rule $\C{R}$1 that could also be found in $\C{P}$, plus the edge orientations $k \to i$ for $k \in \C{K}$, $i \in \C{I} \cap \chsub{\C{G}(\C{M})}{k}$ that are obtained from rule $\C{R}$1;
  \end{itemize}
  This means that the subgraph of DPAG $\C{P}$ (obtained by \alg{FCI-JCI123} so far) induced on the system nodes $\C{I}$ is identical to the subgraph of DPAG $\C{P}^*$ (obtained by FCI so far) induced on the system nodes $\C{I}$. 
  In addition, all pairs of a system node $i \in \C{I}$ and a context node $k \in \C{K}$ are identically connected in the two DPAGs.

  By examining rules $\C{R}$1-$\C{R}$4 and $\C{R}$8-$\C{R}$10 of FCI (which are used without modifications in \alg{FCI-JCI123}) in detail, one can check that they will perform exactly the same edge mark orientations on $\C{P}$ as on $\C{P}^*$.
  For rules $\C{R}$2, $\C{R}$3 and $\C{R}$8-$\C{R}$10 this is obvious because the only subsets of nodes that play a role in those rules necessarily must be in $\C{I}$. 
  For rules $\C{R}$1 and $\C{R}$4 the situation is only slightly more complicated: a single node appearing in those rules can be in $\C{I} \cup \C{K}$, while all others must be in $\C{I}$. 
  Hence each of these rules is applicable to some tuple of nodes in $\C{P}$ if and only if it is applicable to the same tuple of nodes in $\C{P}^*$. 
  Hence, the final DPAG obtained by \alg{FCI-JCI123} from $\IM(\C{M})$ and the final DPAG obtained by FCI from $\IM(\C{H}^*)$ induce identical subgraphs on the system nodes $\C{I}$.
  
  Thus, if all DMAGs $\C{H}$ in $\IM(\C{M})$ that satisfy JCI Assumptions~\ref{ass:simple_scm}, \JCIABC\ have a certain invariant edge mark on an 
  edge between two system variables, then all extended DMAGs $\C{H}^*$ must have the same invariant edge mark. All these extended
  DMAGs $\C{H}^*$ must be Markov equivalent, since FCI arrives at the same CDPAG for all $\IM(\C{H}^*)$.
  Now suppose \alg{FCI-JCI123} left an edge mark on an edge between two system variables unoriented. Then FCI must also
  leave the corresponding edge mark unoriented when it is run on $\IM(\DMAG(\C{M})^*)$. This means that there must exist DMAGs that 
  are Markov equivalent to $\DMAG(\C{M})^*$ that have an arrowhead at that spot, but also DMAGs that are Markov equivalent to 
  $\DMAG(\C{M})^*$ that have a tail at that spot. Marginalizing those DMAGs down to $\C{I} \cup \C{K}$ gives DMAGs that are
  Markov equivalent to $\DMAG(\C{M})$, satisfy JCI Assumptions~\ref{ass:simple_scm}, \JCIABC\, and have an arrowhead respectively a tail at that spot. 
  This means that all edge marks between system variables that could possibly be oriented, have been oriented by \alg{FCI-JCI123}.
  This completes the proof of arrowhead and tail completeness of \alg{FCI-JCI123}.
\end{proof}

\begin{figure}\centering
  \begin{tikzpicture}
    \begin{scope}
      \node at (-2.5,3) {(a)};
      \draw (-3,0.6) edge[dotted] (3,0.6);
      \node[var] (C0) at (-2,1.2) {$C_\alpha$};
      \node[var] (C1) at (0,1.2) {$C_\beta$};
      \node[var] (C2) at (2,1.2) {$C_\gamma$};
      \draw[biarr] (C0) edge (C1);
      \draw[biarr] (C1) edge (C2);
      \draw[biarr,bend left] (C0) edge (C2);
      \node[var] (X0) at (-2,0) {$X_0$};
      \node[var] (X1) at (0,0) {$X_1$};
      \node[var] (X2) at (2,0) {$X_2$};
      \node[var] (X3) at (-1,-1.2) {$X_3$};
      \node[var] (X4) at (1,-1.2) {$X_4$};
      \draw[arr] (C0) edge (X0);
      \draw[arr] (C1) edge (X1);
      \draw[arr] (C2) edge (X2);
      \draw[arr] (C2) edge (X1);
      \draw[arr] (X0) edge (X3);
      \draw[arr] (X3) edge (X4);
      \draw[biarr] (X4) edge (X1);
      \draw[arr] (X2) to (X1);
    \end{scope}
    \begin{scope}[xshift=7cm]
      \node at (-2.5,3) {(b)};
      \draw (-3,0.6) edge[dotted] (3,0.6);
      \node[var,fill=blue!20!white] (R) at (0,2.4) {$C_r$};
      \node[var] (C0) at (-2,1.2) {$C_\alpha$};
      \node[var] (C1) at (0,1.2) {$C_\beta$};
      \node[var] (C2) at (2,1.2) {$C_\gamma$};
      \node[var,fill=blue!20!white] (U0) at (-1,1.2) {$C_{\bar{\alpha}}$};
      \node[var,fill=blue!20!white] (U1) at (1,1.2) {$C_{\bar{\beta}}$};
      \node[var,fill=blue!20!white] (U2) at (3,1.2) {$C_{\bar{\gamma}}$};
      \draw[arr] (R) -- (C0);
      \draw[arr] (R) -- (C1);
      \draw[arr] (R) -- (C2);
      \draw[arr] (U0) -- (C0);
      \draw[arr] (U1) -- (C1);
      \draw[arr] (U2) -- (C2);
      \node[var] (X0) at (-2,0) {$X_0$};
      \node[var] (X1) at (0,0) {$X_1$};
      \node[var] (X2) at (2,0) {$X_2$};
      \node[var] (X3) at (-1,-1.2) {$X_3$};
      \node[var] (X4) at (1,-1.2) {$X_4$};
      \draw[arr] (C0) edge (X0);
      \draw[arr] (C1) edge (X1);
      \draw[arr] (C2) edge (X2);
      \draw[arr] (C2) edge (X1);
      \draw[arr] (X0) edge (X3);
      \draw[arr] (X3) edge (X4);
      \draw[biarr] (X4) edge (X1);
      \draw[arr] (X2) to (X1);
    \end{scope}
  \end{tikzpicture}
  \caption{(a) Example DMAG satisfying JCI Assumptions~\JCIABC\ and (b) corresponding extended DMAG with additional variables as used in the proof of Theorem~\ref{theo:fcijci123_complete}.\label{fig:FCI_extended_MAG}}
\end{figure}

\subsection{Reading off Definite (Non-)Ancestors From a DPAG}\label{sec:app:dpag_ancestors}

\citet{Zhang2006} conjectured the soundness and completeness of a criterion to read off definite ancestral relations from a CDPAG.
\citet{Roumpelaki++_UAIWS_16} proved soundness of this criterion.\footnote{\citet{Roumpelaki++_UAIWS_16} also claim to have proved completeness, but their proof is flawed: the last part of the proof that aims to prove that $u,v$ are non-adjacent appears to be incomplete.}
We will need a slightly stronger result (with a similar proof) for DPAGs:
\begin{proposition}\label{prop:dpag_ancestors}
  Let $\C{M}$ be an acyclic SCM. Let $\C{P}$ be a DPAG that contains $\DMAG(\C{M})$, and in which all unshielded colliders in $\DMAG(\C{M})$ have been oriented. For two nodes $i,j \in \C{P}$: If 
  \begin{itemize}
    \item there is a directed path from $i$ to $j$ in $\C{P}$, or 
    \item there exist uncovered possibly directed paths from $i$ to $j$ in $\C{P}$ of the form $i,u,\dots,j$ and $i,v,\dots,j$ such that $u,v$ are non-adjacent nodes in $\C{P}$, 
  \end{itemize}
  then $i$ causes $j$ according to $\C{M}$, i.e., $i \in \ansub{\C{G}(\C{M})}{j}$.
\end{proposition}
\begin{proof}
  First, if there is a directed path from $i$ to $j$ in $\C{P}$, it must be in any DMAG in $\C{P}$, hence there must be a directed path from $i$ to $j$ in $\DMAG(\C{M})$ as well.
  Therefore $i \in \ansub{\C{G}(\C{M})}{j}$. 

  Second, assume that there exist uncovered possibly directed paths from $i$ to $j$ in $\C{P}$ of the form $i,u,\dots,j$ and $i,v,\dots,j$ such that $u,v$ are non-adjacent in $\C{P}$. 
  If $\DMAG(\C{M})$ has $i \to u$, the path $i,u,\dots,j$ must actually correspond to a directed path in $\DMAG(\C{M})$ because otherwise it would contain unshielded colliders that were not oriented, contradicting the assumptions. 
  If $\DMAG(\C{M})$ has $i \ots u$ instead, it must have $i \to v$ to avoid an unshielded collider $u \sto i \ots v$ that was not oriented, and hence must have a directed path $i,v,\dots,j$. 
  In both cases, $\DMAG(\C{M})$ must have a directed path from $i$ to $j$, and hence $i \in \ansub{\C{G}(\C{M})}{j}$.
\end{proof}

\citet[p.\ 137]{Zhang2006} provides a sound and complete criterion to read off definite non-ancestors from a CDPAG. 
It is easy to prove the soundness of the criterion also for (arbitrary) DPAGs:
\begin{proposition}\label{prop:dpag_nonancestors}
  Let $\C{M}$ be an acyclic SCM. 
  Let $\C{P}$ be a DPAG that contains $\DMAG(\C{M})$.
  For two nodes $i,j \in \C{P}$:
  if there is no possibly-directed path from $i$ to $j$ in $\C{P}$ then $i \notin \ansub{\C{G}(\C{M})}{j}$.
\end{proposition}
\begin{proof}
  If $i \in \ansub{\C{G}(\C{M})}{j}$ then there is a directed path from $i$ to $j$ in $\DMAG(\C{M})$. 
  Since $\C{P}$ contains $\DMAG(\C{M})$, this must correspond with a possibly-directed path from $i$ to $j$ in $\C{P}$.
\end{proof}

These two propositions allow us to read off (a subset of the) ancestral and non-ancestral relations that are identifiable
from the conditional independences in the joint distribution and the JCI background knowledge (if applicable) from the DPAGs 
output by the various FCI variants (\alg{FCI-JCI123}, \alg{FCI-JCI1}, \alg{FCI-JCI0}, FCI).

\subsection{Discovering Direct Intervention Targets with \alg{FCI-JCI123}}\label{sec:app:fcijci123_intervention_targets}

\begin{figure}\centering
  \begin{tikzpicture}
    \begin{scope}
      \node at (-2.7,1.7) {(a)};
      \draw (-3,0.6) edge[dotted] (0.5,0.6);
      \node[var] (C0) at (-2,1.2) {$C$};
      \node[var] (X0) at (-2,0) {$X_1$};
      \node[var] (X1) at (-0.5,0) {$X_2$};
      \draw[arr] (C0) edge (X0);
      \draw[arr] (X0) edge (X1);
      \draw[biarr,bend left] (X0) edge (X1);
    \end{scope}
    \begin{scope}[xshift=4cm]
      \node at (-2.7,1.7) {(b)};
      \draw (-3,0.6) edge[dotted] (0.5,0.6);
      \node[var] (C0) at (-2,1.2) {$C$};
      \node[var] (X0) at (-2,0) {$X_1$};
      \node[var] (X1) at (-0.5,0) {$X_2$};
      \draw[arr] (C0) edge (X0);
      \draw[arr] (C0) edge (X1);
      \draw[arr] (X0) edge (X1);
    \end{scope}
    \begin{scope}[xshift=8cm]
      \node at (-2.7,1.7) {(c)};
      \draw (-3,0.6) edge[dotted] (0.5,0.6);
      \node[var] (C0) at (-2,1.2) {$C$};
      \node[var] (X0) at (-2,0) {$X_1$};
      \node[var] (X1) at (-0.5,0) {$X_2$};
      \draw[arr] (C0) edge (X0);
      \draw[arr] (C0) edge (X1);
      \draw[carc] (X0) edge (X1);
    \end{scope}
  \end{tikzpicture}
  \caption{Example to illustrate that directed edges in the DPAG obtained by \alg{FCI-JCI123} do not necessarily correspond with a direct cause. (a) Graph $\C{G}(\C{M})$, satisfying JCI Assumptions~\JCIABC; (b) corresponding $\DMAG(\C{M})$; (c) corresponding DPAG $\C{P}$ output by \alg{FCI-JCI123}. The directed edge $C \to X_2$ in $\C{P}$ identified by \alg{FCI-JCI123} does not correspond with a direct causal effect of $C$ on $X_2$ (note that there is no directed edge $C \to X_2$ in $\C{G}(\C{M})$).\label{fig:spurious_direct_edge}}
\end{figure}

One of the features of \alg{FCI-JCI123} is that it allows one to read off direct intervention targets from the DPAG that it outputs. Na\"ively interpreting a directed edge $k \to i$ from context node $k \in \C{K}$ to system node $i \in \C{I}$ in the DPAG output by \alg{FCI-JCI123} as meaning that $k$ directly targets $i$ is incorrect, as can be seen from the example in Figure~\ref{fig:spurious_direct_edge}. Here we propose a provably correct (but possibly incomplete) procedure.

We will first consider how to read direct intervention targets from DMAGs before applying this to DPAGs.
\begin{lemma}\label{lemm:fci_jci123_mag}
  Let $\C{M}$ be an acyclic SCM that satisfies JCI Assumptions~\ref{ass:simple_scm}, \JCIABC.
  For $k \in \C{K}$, $i \in \C{I}$: if
  \begin{itemize}
    \item $k \to i$ in $\DMAG(\C{M})$, and 
    \item for all nodes $j$ in $\DMAG(\C{M})$ s.t.\ $k \to j \to i$ in $\DMAG(\C{M})$, $j \to i$ is visible,
  \end{itemize}
  then $k$ is a direct cause of $i$ according to $\C{M}$, i.e., $k \to i \in \C{G}(\C{M})$.
\end{lemma}
\begin{proof}
  Because the edge $k \to i$ is present in $\DMAG(\C{M})$, there must be an inducing path between $k$ and $i$ in $\C{G}(\C{M})$.
  This must be a collider path into $i$ where each collider is ancestor of $i$.
  First suppose that the path consists of more than a single edge. 
  Denote its first collider (the one adjacent to $k$) by $j$.
  If the first edge on the path would be into $k$, then it must be $k \oto j$ and $j$ must be a context node (because of JCI Assumptions~\JCIAB).
  Similarly, all subsequent nodes on the inducing path (except for the final node $i$) must be collider nodes and hence in $\C{K}$.  
  But then the final edge is between a context node and system node $i$ and into the context node, contradicting JCI Assumption~\ref{ass:uncaused} or \ref{ass:unconfounded}.
  Hence the first edge on the inducing path must be $k \to j$.
  The same edge $k \to j$ must then occur in $\DMAG(\C{M})$.
  The remainder of the inducing path is actually an inducing path between $j$ and $i$ that is into $j$. 
  By \citet[Lemma 9]{Zhang2008_JMLR}, $j \to i$ is in $\DMAG(\C{M})$ and it is invisible.
  This contradicts the assumption.
  Therefore the inducing path in $\C{G}(\C{M})$ between $k$ and $i$ must consist of a single edge.
  This must be out of $k$ because of JCI Assumptions~\JCIAB, and is thus necessarily of the form $k \to i$.
  Hence $k \to i$ is in $\C{G}(\C{M})$.
\end{proof}
The following result enables us to read off direct intervention targets from the DPAG output by \alg{FCI-JCI123}.
\begin{proposition}
Let $\C{M}$ be an acyclic SCM that satisfies JCI Assumptions~\ref{ass:simple_scm}, \JCIABC.
Suppose that its distribution $\Prb_{\C{M}}(\B{X},\B{C})$ is faithful w.r.t.\ the graph $\C{G}(\C{M})$. 
Let $\C{P}$ be the DPAG output by \alg{FCI-JCI123} with input $\IM(\C{M})$.
Let $k \in \C{K}$, $i \in \C{I}$. 
  \begin{itemize}
    \item If $k$ is not adjacent to $i$ in $\C{P}$, $k$ is not a direct cause of $i$ according to $\C{M}$, i.e., $k \to i \not\in \C{G}(\C{M})$. 
    \item If:
      \begin{compactenum}
        \item $k \to i$ in $\C{P}$, and 
        \item for all system nodes $j \in \C{I}$ s.t.\ $k \to j$ in $\C{P}$ and $j \ctc i$ or $j \cto i$ or $j \to i$ in $\C{P}$, the edge $j \to i$ is definitely visible in the DPAG obtained from $\C{P}$ by replacing the edge between $j$ and $i$ by $j \to i$,
      \end{compactenum}
    then $k$ is a direct cause of $i$ according to $\C{M}$, i.e., $k \to i \in \C{G}(\C{M})$.
  \end{itemize}
\end{proposition}
\begin{proof}
  Because \alg{FCI-JCI123} is sound (Theorem~\ref{theo:fcijci123_complete}), $\C{P}$ contains $\DMAG(\C{M})$. 

For the first statement: if $k$ is not adjacent to $i$ in $\C{P}$, then the two nodes are not adjacent in any DMAG in $\C{P}$, and in particular, in $\DMAG(\C{M})$. 
  This means that  $k \to  i \not\in \C{G}(\C{M})$, because otherwise, $k \to i$ would be in $\DMAG(\C{M})$, a contradiction.

The second statement follows from Lemma~\ref{lemm:fci_jci123_mag} and from the JCI Assumptions~\JCIABC.
\end{proof}
If the context variables represent interventions, then this allows us to learn (a subset of) the direct targets and non-targets of each intervention. 
While it is easy to see that this criterion is sound, we do not know whether it is complete.

\section{ASD: Accounting for Strong Dependencies}\label{sec:app:asd}

The causal discovery and reasoning algorithm that we refer to as ASD (Accounting for Strong Dependencies) used in this
work is based on an algorithm proposed by \citet{HEJ2014} and extensions
proposed by \citet{MagliacaneClaassenMooij_NIPS_16} and \citet{ForreMooij_UAI_18}. To
make this paper more self-contained, we give here a short description of this algorithm,
referring the reader for more details to the original publications.

\citet{HEJ2014} formulate causal discovery as an optimization problem where a
loss function is minimized over possible causal graphs. Intuitively, the loss
function can be thought of as measuring the amount of evidence \emph{against} the
hypothesis that the data was generated by an SCM with a particular graph.
The loss function depends on the hypothetical causal graph and on a list of input statements. 
For the purely observational case, the input consists of a list $S = \big((a_j,b_j,Z_j,\lambda_j)\big)_{j=1}^n$ of 
weighted conditional independence statements. Here, the weighted statement $(a_j,b_j,Z_j,\lambda_j)$
with $\{a_j\},\{b_j\},Z_j$ disjoint sets of endogenous variable indices and 
$\lambda_j \in \bar{\RN} := \RN \cup \{-\infty,+\infty\}$
encodes that the conditional independence $X_{a_j} \CI X_{b_j} \given \B{X}_{Z_j}$ holds
with ``confidence'' $\lambda_j$, where a finite value of $\lambda_j$ gives a ``soft constraint''
and a value of $\lambda_j=\pm\infty$ imposes a ``hard constraint''. Positive weights 
encode that we have empirical support \emph{in favor} of the independence, whereas negative weights
encode empirical support \emph{against} the independence (in other words, in favor of
\emph{dependence}).
The loss function simply sums the absolute weights of all the input statements that would be violated 
if the true causal graph would consist of the hypothetical one:
\begin{equation*}
  \mathcal{L}(\C{G}, S) := \sum_{(a_j,b_j,Z_j,\lambda_j)\in S} \lambda_j (\I_{\lambda_j > 0} - \I_{a_j \SEP_{\C{G}} b_j \given Z_j}),
\end{equation*}
where $\I$ is the indicator function.
While the original implementation by \citet{HEJ2014} makes use of $d$-separation, \citet{ForreMooij_UAI_18}
show how this can be modified for $\sigma$-separation.
Causal discovery can now be formulated as the optimization problem:
\begin{equation}\label{eq:argmin}
  \C{G}^* = \argmin_{\C{G}\in\mathbb{G}(\C{I})} \mathcal{L}(\C{G}, S),
\end{equation}
where $\mathbb{G}(\C{I})$ denotes the set of all possible causal graphs with nodes $\C{I}$
(ADMGs in the acyclic case, and DMGs in the cyclic case).

The optimization problem (\ref{eq:argmin}) may have multiple minima, for example because the underlying causal graph is not identifiable from the inputs. Nonetheless, some of the features of the causal graph (e.g., the presence or absence of a certain
directed edge) may still be identifiable. 
Let $f : \mathbb{G}(\C{I}) \to \{0,1\}$ be a feature, i.e., a Boolean function of the causal graph $\C{G}$.
We employ the method proposed by \citet{MagliacaneClaassenMooij_NIPS_16} for scoring
the confidence that feature $f$ is present in the causal graph by calculating the difference between the optimal losses under the 
additional hard constraints that the feature $f$ is present vs.\ that the feature $f$ is absent in the causal graph:
\begin{equation}\label{eq:confidence_feature}
  C(f,S) := \min_{\C{G} \in \mathbb{G}(\C{I}): \lnot f(\C{G})} \mathcal{L}(\C{G},S) - \min_{\C{G} \in \mathbb{G}(\C{I}): f(\C{G})} \mathcal{L}(\C{G},S).
\end{equation}
This confidence is positive if there is less evidence against its presence than against its absence, negative if there is less evidence against its absence than its presence, and vanishes if there is as much evidence against its presence as there is against its absence.
As features, we can consider for example the presence of a direct causal relation, the presence of
an (ancestral) causal relation, and the presence of a latent confounder.

\citet{MagliacaneClaassenMooij_NIPS_16} showed that this scoring method is sound for oracle inputs.
\begin{theorem}\label{theo:ASD_sound_complete}
For any feature $f : \mathbb{G}(\C{I}) \to \{0,1\}$, the ASD confidence score $C(f,S)$ of \eref{eq:confidence_feature} is sound and complete for oracle inputs with infinite weights. In other words,
$C(f,S)=\infty$ if $f$ is identifiable from the inputs,
$C(f,S)=-\infty$ if $\lnot f$ is identifiable from the inputs, and $C(f,S)=0$ if $f$ is unidentifiable from the inputs.
\end{theorem}
Additionally, \citet{MagliacaneClaassenMooij_NIPS_16} showed that the scoring method is asymptotically consistent under a consistency condition on the weights that encode the confidence of conditional (in)dependence.
\begin{theorem}\label{theo:ASD_consistent}
Assume that the weights are asymptotically consistent, meaning that
\begin{equation}\label{eq:weightFreqConsistency}
\lambda_j \xto{P} \begin{cases}
  -\infty & X_{a_j} \nCI X_{b_j} \given \B{X}_{Z_j} \\
  +\infty & X_{a_j} \CI X_{b_j} \given \B{X}_{Z_j},
\end{cases}
\end{equation}
(where $\xto{P}$ means convergence in probability) as the number of samples $N \to \infty$.
Then for any feature $f : \mathbb{G}(\C{I}) \to \{0,1\}$, the ASD confidence score $C(f,S)$ of \eref{eq:confidence_feature} is asymptotically consistent, i.e., $C(f,S) \xto{P} \infty$ if $f$ is identifiably true, $C(f,S) \xto{P} -\infty$ if $f$ is identifiably false, and $C(f,S) \xto{P} 0$ otherwise.
\end{theorem}

In our experiments, we used the weights proposed in \citet{MagliacaneClaassenMooij_NIPS_16}:
$\lambda_j = \log p_j - \log \alpha$, where $p_j$ is the p-value of a statistical test with independence
as null hypothesis, and $\alpha$ is a significance level (e.g., 1\%).
These weights have the desirable property that independences typically get a smaller absolute weight 
than strong dependencies. This leads to the strong dependencies dominating the loss function, which explains
the acronym ASD (Accounting for Strong Dependencies) that we use here to describe this method.
By choosing a sample-size dependent threshold $\alpha_N$ such that $\alpha_N \to 0$ as $N \to \infty$ at a suitable rate, 
these weights may become asymptotically consistent. \citet{KalischBuehlmann2007} provide a choice of $\alpha_N$ for 
partial correlation tests that ensures
asymptotic consistency under the assumption that the distribution is multivariate Gaussian.
Another possibility for obtaining consistent weights would be to base them on the 
distribution-free and strongly-consistent conditional independence test proposed by \citet{GyorfiWalk2012}.


\end{document}